\documentclass{article}

\usepackage{arxiv-a4}

\usepackage{balance} 

\usepackage{etoolbox}
\usepackage{graphicx}
\graphicspath{{./images/}}
\usepackage{latexsym}
\usepackage{amsmath}
\usepackage{pifont} \usepackage{amsfonts}
\usepackage[inline]{enumitem}

\def\cS{{\mathcal{S}}}
\def\cSt{{\mathcal{S}_{\mathcal{T}}}}
\def\cA{{\mathcal{A}}}

\def\temp{\tau} \def\typeOf{\phi} \def\mdp{\text{\sc mdp}}
\def\mdpb{\text{\sc mdp-b}}
\def\mdps{\text{\sc mdp-s}}
\def\ie{{\em i.e.}}
\def\eg{{\em e.g.}}

\def\pred{\text{pred}}
\mathchardef\mhyphen="2D

\def\RApred{R^\cA_\pred}
\def\RSpred{R^\cS_\pred}
\def\piApred{\pi^\cA_\pred}
\def\piSpred{\pi^\cS_\pred}

\def\myDiscount{0.99}

 \usepackage{amsthm}

\newtheorem{proposition}{Proposition}

\newcommand{\eqdef}     {\stackrel{{\textrm{\rm\tiny def}}}{=}}

\DeclareMathOperator*{\argmax}{arg\,max}

\usepackage{tikz}
\usetikzlibrary{positioning,angles,quotes,automata}

\makeatletter
\newtoks\@tabtoks
\newcommand\addtabtoks[1]{\global\@tabtoks\expandafter{\the\@tabtoks#1}}
\newcommand\eaddtabtoks[1]{\edef\mytmp{#1}\expandafter\addtabtoks\expandafter{\mytmp}}
\newcommand*\resettabtoks{\global\@tabtoks{}}
\newcommand*\printtabtoks{\the\@tabtoks}
\makeatother

\usepackage{fontawesome5}

\usetikzlibrary{arrows.meta}
\usepackage[normalem]{ulem} \usepackage{cancel} \usepackage[inline]{enumitem}

\usepackage{subcaption}

\usepackage{booktabs} \usepackage{siunitx} \usepackage{adjustbox} \usepackage{pgfplots}

\usepackage[noabbrev]{cleveref}

\usepackage{xcolor}
\newcommand{\persComment}[3]{
	\ifmmode
	\text{\textcolor{#3}{[#2] #1}}
	\else
	\textcolor{#3}{[#2] \em #1}
	\fi
}

\title{How to Exhibit More Predictable Behaviors}

\renewcommand{\headeright}{Preprint - \today}
\author{
  Salomé Lepers$^1$ \\
  Vincent Thomas$^1$
  \and
  Sophie Lemonnier$^{1,2}$ \\
  Olivier Buffet$^1$
\and
  \begin{minipage}{.99\linewidth}
  \small \centering
  $^{(1)}$Université de Lorraine, CNRS, Inria, LORIA, F-54000 Nancy, France\\
  $^{(2)}$Université de Lorraine, PErSEUs, F-57045 Metz, France
  \end{minipage}
}

\begin{document}

\maketitle

\begin{abstract}
This paper looks at predictability problems, \ie, wherein an agent must choose its strategy in order to optimize the predictions that an external observer could make.
We address these problems while taking into account uncertainties on the environment dynamics and on the observed agent's policy.
To that end, we assume that the observer
\begin{enumerate*}
	\item seeks to predict the agent's future action or state at each time step, and
	\item models the agent using a stochastic policy computed from a known underlying problem,
\end{enumerate*}
and we leverage on the framework of observer-aware Markov decision processes (OAMDPs).
We propose action and state predictability performance criteria through reward functions built on the observer's belief about the agent policy; show that these induced {\em predictable} OAMDPs can be represented by goal-oriented or discounted MDPs; and analyze the properties of the proposed reward functions both theoretically and empirically on two types of grid-world problems.
\end{abstract}

\section{Introduction}

\label{sec|introduction}
\begin{figure}[ht!]
	\centering
	\includegraphics[width=0.55\columnwidth]{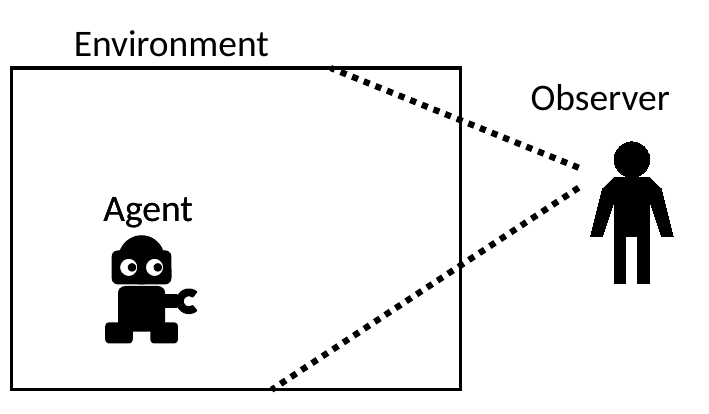}
	\caption{Agent in its environment and a passive observer}
	\label{fig|conscient}
\end{figure}

In a human-agent collaboration scenario, some properties of the agent behavior can be useful for the human and sometimes allow a better collaboration.
Recent papers suggest ways of obtaining such behaviors.
In particular, when an agent is aware that it is being observed by a passive human, as in \Cref{fig|conscient},
 it can control the information disclosed to the observer through its behavior.

Chakraborti et al. \cite{DBLP:conf/aips/ChakrabortiKSSK19} build on previous work to derive a taxonomy of these concepts.
In particular, they distinguish between
\begin{enumerate*}
	\item transmitting information, with properties such as
	{\em legibility} (legible behaviors convey intentions, \ie, actual task at hand, via action choices),
	{\em explicability} (explicable behaviors conform to observers’ expectations, \ie, they appear to have some purpose), and
	{\em predictability} (a behavior is predictable if it is easy to guess the end of an on-going trajectory);

	or
	\item hiding information, as through
{\em obfuscation}, when the agent tries to hide its real goal.
\end{enumerate*}
They propose a general framework for such problems under the hypothesis that transitions are deterministic, and work mostly with plans (a sequence of actions inducing a state sequence).
In their approach, the human is modeled by the robot as having a model of the robot+environment system (including the robot's possible tasks), and is thus able to predict the robot behavior and adapt to it.

Each of the properties they discuss can be relevant in some situations.
They convey different kinds of information to the observer, and can be mutually exclusive.
Chakraborti et al. \cite{DBLP:conf/aips/ChakrabortiKSSK19} point out that an explicable plan can be unpredictable, \eg, when  multiple explicable plans exist.
Similarly, Fisac et al. \cite{FisacEtAl-AFR20} suggest that, if an agent acts legibly, then one can infer its goal but not necessarily how it is going to achieve this goal.
{\em Predictability} is meant to ensure that the agent's behavior conveys this information.

Schadenberg  et al. \cite{10.1145/3461534} explain that {\em Predictability} has a real interest when considering human-robot interaction.
Their work mainly focuses on how human observers react to a {\em hand-coded} social robot behavior depending on whether the cause of responsive actions is visible or not.
As we do in our experiments, they distinguish the participants' performance in predicting the robot's behavior, called the {\em behavioral predictability}, and their perception of the predictability of the robot behavior, called the {\em attributed predictability}.
They observe that both predictabilities are not necessarily aligned, and point out that, depending on the scenario, one may want to optimize either the behavioral predictability, for instance with industrial robots, or the behavioral predictability, for instance with social robots.
Unlike them, we are interested in {\em automatically deriving} predictable behaviors, and only consider fully observable settings.

\begin{figure}[ht!]
	\centering
	\includegraphics[width=.55\columnwidth]{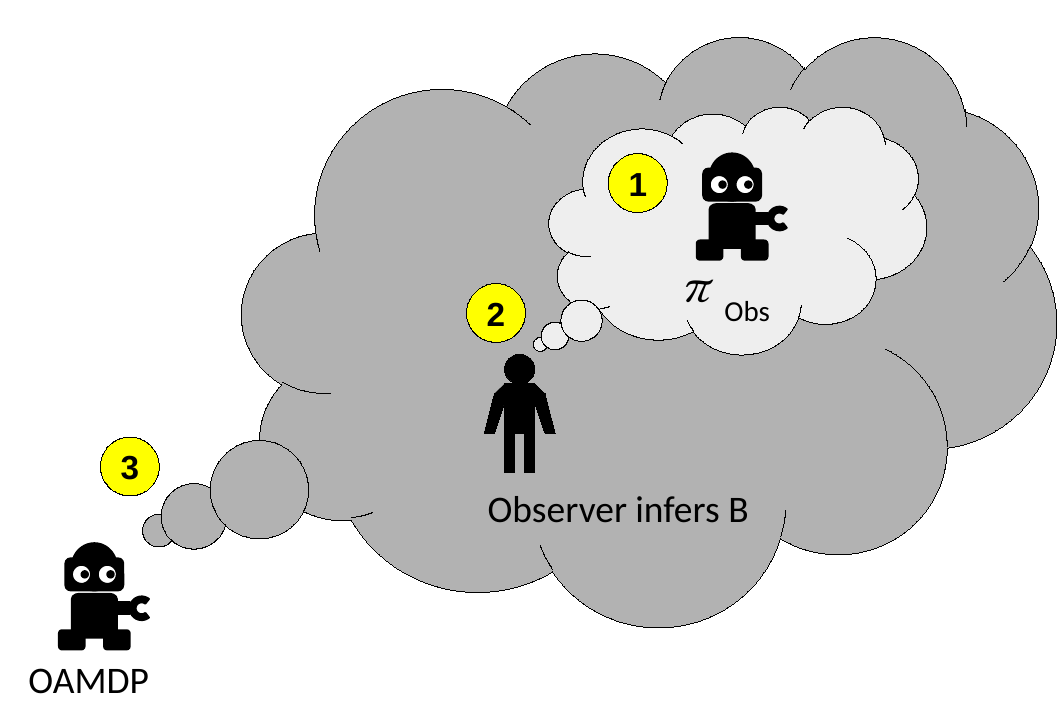}
	\caption{An OAMDP agent (3) assumes that the observer's expectation (2) is that the agent behaves so as to achieve some task (1).}
\label{fig|OAMDP}
\end{figure}

Miura and Zilberstein \cite{pmlr-v161-miura21a} build a unifying framework while assuming stochastic transitions, namely {\em observer-aware  Markov decision processes} (OAMDPs), adopting a similar approach as Chakraborti et al., as illustrated in \Cref{fig|OAMDP}.
Among other things, they work also on legibility, explicability, and predictability.
Yet, as we will further discuss in \Cref{sec|background},
the two OAMDP approaches to predictability they consider are not fully satisfying: one amounts to returning an optimal policy for the low-level MDP, and the other reasons on full trajectories, which does not seem appropriate in a stochastic environment (and turns out to be prohibitive).

Our objective in this paper is to propose a more satisfying approach to predictability by working not with complete trajectories, but with actions or states at each time step.
This implies reasoning on dynamic variables, which requires introducing a variant of the OAMDP formalism.
Moreover, we also consider not only discounted problems, but also stochastic shortest-path (\ie, goal-oriented) problems.

\Cref{sec|background} provides background on Markov decision processes and observer-aware MDPs.
Our approach to action and state predictability, through dedicated reward functions, is described in \Cref{sec|contribution}, along with proofs that well-defined problems are induced.
Experiments are then presented in \Cref{sec|XP}, where we generate and interprete policies on two types of grid-world problems, comparing with standard MDP solution policies, and then, in \Cref{sec|XPinVivo}, where human observers are confronted with these policies on some of these problems.

\section{Background}
\label{sec|background}

\subsection{Markov Decision Processes}
A Markov decision process (MDP) \cite{Bellman-jmm57}
 is specified through a tuple  {$\langle \cS, \cA, T, R, \gamma, \cSt \rangle$} where:
\begin{itemize}
\item $\cS$ is a set of states; \item $\cA$ is a set of actions;
\item $T: \cS \times  \cA \times \cS \to [0;1]$, the transition function, gives the probability
  $T(s,a,s')$ that action $a$ performed in state $s$ will lead to state $s'$;
\item $R : \cS \times \cA \times \cS \to \mathbb{R}$, the reward function, gives the immediate reward $R(s,a,s')$ received upon transition $(s,a,s')$.
\item $\gamma \in [0,1]$ is a discount factor; and
\item $\cSt \subset \cS$ is a set of terminal states:
for all $s,a \in \cS\times \cA$, $T(s,a,s)=1$ and $R(s,a,s)=0$.
\end{itemize}
Then, a (stochastic) {\em policy} $\pi: \cS \to \Delta(\cA)$ maps states to distributions over actions, $\pi(a|s)$ denoting the probability to perform $a$ when in $s$.
When a policy is deterministic, $\pi(s)$ denotes the only possible action in $s$.
Assuming $\gamma<1$, the value of a policy $\pi$ is the sum of discounted rewards on an infinite horizon:
\begin{align*}
	V^{\pi}(s)
	& \eqdef \mathbb{E}_\pi \left[ \sum_{t=0}^\infty \gamma^{t}R(S_t, A_t) | S_0=s  \right],
\end{align*}
and an optimal policy $\pi^*$ is such that, for all $s$, $V^{\pi*}(s) = \max_{\pi} V^{\pi}(s)$.
The {\em value iteration} (VI) algorithm \cite{Bellman-jmm57} approximates $V^*$, the value function common to all optimal policies, by iterating the following computation (where $k$ is the current iteration):
\begin{align*}
  V_{k+1}(s)
  & \gets \max_a \sum_{s'} T(s,a,s') \cdot \left( R(s,a,s') + \gamma V_k(s') \right).
  \intertext{
    Calculations stop when the {\em Bellman residual} is below a threshold: }
  & \underbrace{\max_s |V_{k+1}(s) - V_k(s)|}_{\text{Bellman residual}} \leq \frac{1-\gamma}{\gamma} \epsilon.
\intertext{Then, an $\epsilon$-optimal policy is obtained by acting greedily with respect to the solution value function $V_{k}$, \ie, using}
  \pi_k(s)
  & \gets \argmax_a \sum_{s'} T(s,a,s') \cdot \left( R(s,a,s') + \gamma V_k(s') \right).
\end{align*}

The same dynamic programming operator and $\epsilon$-greedy selection apply when $\gamma=1$ if $\cSt$ is not empty.\footnote{
  No stopping criterion provides guarantees about the solution quality in general SSPs (cf. \cite{Hansen-mmor17}).
  Here, we simply stop the algorithm when the Bellman residual is below some threshold $\eta\ll\epsilon$ and assume that $V$ is $\epsilon$-close to $V^*$.
}
Such problems are called {\em Shortest Stochastic Path} problems (SSPs) \cite{Bertsekas-dpoc05,HanZil-aij01}. SSPs are more general than MDPs because any MDP can be turned into an SSP with, at any time step, a $1-\gamma$ probability to transition to a terminal state \cite[Sec.~7.3]{Bertsekas-dpoc05}.

Let us call {\em proper} a policy $\pi$ that reaches $\cSt$ with probability $1$ from any state.
We will from now on make the assumptions that, in our SSPs:
\begin{description}
\item[(A1)] for any policy $\pi$ and any state $s$, $\pi$ reaches $\cSt$ with probability 1 from $s$ iff $V^\pi(s) > -\infty$; and
\item[(A2)] at least one proper policy $\pi$ exists (\ie, $\forall s$, $V^\pi(s) > -\infty$).
\end{description}
In particular, the first assumption holds if, for all $(s,a,s') \in (\cS \setminus \cSt) \times \cA \times (\cS \setminus \cSt)$, $R(s,a,s')<0$.

\subsection{Observer-Aware Markov Decision Processes}

As introduced by Miura and Zilberstein, an observer-aware MDP (OAMDP) \cite{pmlr-v161-miura21a} models a situation wherein an agent attempts to maximize an observer's information regarding some target random variable, called {\em type}, under some model of the observer's evolving belief about this type.
Formally, an OAMDP is described by an 8-tuple $\langle \cS, \cA, T, \gamma, \cSt, \Theta, B, R \rangle$, where:
\begin{itemize}
	\item $\langle \cS, \cA, T, \gamma, \cSt \rangle$
	is a reward-less discounted MDP ($\gamma<1$);
	\item $\Theta$
	is a finite set of {\em types} representing a characteristic of the agent such as possible goals, intentions or capabilities;
	\item $B: H^* \to \Delta^{|\Theta|}$
	gives the assumed belief of the observer given a history ($H=\cS \times \cA$); \item $R: \cS \times \cA \times \Delta^{|\Theta|} \to \mathbb{R}$ is the reward function.
\end{itemize}
In most of the cases they consider, Miura and Zilberstein derive $B$ by relying on Baker et al.'s ``BST''\footnote{The acronym stands for the authors' initial letters.}
Bayesian belief update rule \cite{BakSaxTen-cog09}, \ie, considering that, again from the agent's viewpoint, the observer models the agent's behavior for a given type through an MDP by
\begin{enumerate}
\item using a corresponding reward function $R^\theta_\mdp$;
\item solving the discounted MDP $\langle \cS, \cA, T, R^\theta_\mdp, \gamma, \cSt \rangle$ (where all components but $R^\theta_\mdp$ come from the OAMDP definition) to obtain $V^{\theta,*}_\mdp$;
\item building a stochastic ``softmax'' policy such that, $\forall (s,a)$,
  \begin{align*}
		\pi^\theta_\mdp(a|s) & =
		\frac{
			e^{\frac{1}{\temp}Q^{\theta,*}_\mdp(s,a)}
		}{
			\sum_{a'} e^{\frac{1}{\temp}Q^{\theta,*}_\mdp(s,a')}
		},
		\text{ where} \\
		Q^{\theta,*}_\mdp(s,a) & = \sum_{s'} T(s,a,s') \cdot \left( r(s,a,s') + \gamma V^{\theta,*}_\mdp(s') \right),
  \end{align*}
  and temperature $\temp > 0$ allows tuning the policy's optimality (thus the agent's assumed rationality for the observer).
\end{enumerate}
With $\pi_\mdp\equiv(\pi^\theta_\mdp)_{\theta \in \Theta}$ in hand, the observer's belief function about the type can then be obtained through Bayesian inference.

Miura and Zilberstein \cite{pmlr-v161-miura21a} use the OAMDP framework to formalize various observer-aware problems from the literature, including legibility, explainability, and predictability.
For predictability, which we now focus on, they mention two approaches.
The first one builds on Dragan et al.'s idea to ``model the predictability of a trajectory as simply proportional to the value (negative cost) of a trajectory'' \cite{DraLeeSri-hri13}, which, in the OAMDP setting, translates into
\begin{enumerate*}
\item having a single type $\theta^0$, and
\item optimizing the underlying reward function $R^{\theta^0}_\mdp$, \ie, acting greedily wrt $Q^{\theta^0,*}_\mdp$ (rather than following $\pi^{\theta^0}_\mdp$).
\end{enumerate*}
The second approach builds on Fisac et al.'s $t$-predictability \cite{FisacEtAl-AFR20}, which maximizes $Pr(a_{t+1}, \dots , a_T |a_1, \dots , a_t)$ in deterministic settings, by using a type for each possible trajectory---\ie,
exponentially many types---over a finite temporal horizon.

In the following, we propose an alternative approach to predictability and discuss its properties.

\section{Contribution}
\label{sec|contribution}

As a preliminary contribution, while Miura and Zilberstein consider only discounted OAMDPs, we introduce OASSPs (thus, using $\gamma=1$). This mainly raises the question: Under which conditions do proper policies exist in the induced SSP?
We will discuss this issue in the context of predictability.

\subsection{Predictable Observer-Aware MDPs}

Both approaches to predictability mentioned by Miura and Zilberstein are inspired by work in deterministic settings, reasoning on trajectories.
Because both the softmax policy $\pi_\mdp$ and the dynamics of the system can be stochastic, we instead propose to try predicting either actions or states, both alternatives ({\em action} and {\em state predictability}) possibly leading to different solutions.
Yet, OAMDP types $\theta$ are static variables (as types in Bayesian games \cite{HarsanyiBG-I-ms67,FudTir-gt91}), while actions and states are dynamic.
This leads us to introduce pOAMDPs (predictable OAMDPs), where we instead talk of a (dynamic) {\em target variable}, also noted $\theta_t$, which is now a function of the current transition: $\theta_t=\typeOf(s_t,a_t,s_{t+1})$.
This
\begin{enumerate*}
\item does not allow encoding problems where the target variable is static and hidden (\ie, is a type), \eg, legibility or explicability, but
\item still allows
  \begin{enumerate*}
  \item defining and solving the observer's MDP (because the type does not influence the system dynamics), and
  \item using the BST belief update (because of the Markovian nature of the target variables).
  \end{enumerate*}
\end{enumerate*}

The following sections describe respectively, for both the action and state predictabilities:
\begin{enumerate*}
	\item how to derive $B$ and solve the pOAMDP given a reward function $R$, and
	\item the reward functions proposed to formalize predictability, along with properties of the resulting decision problems.
\end{enumerate*}

\subsection{Belief Function and Properties of pOAMDPs}

For action predictability, $\Theta=\cA$, $\typeOf(s,a,s')=a$, and $B$ is
\begin{align*}
	B : &
	\begin{array}{ccc}
		H^* & \to & \Delta^{|\cA|}, \\
		(s_0, a_0, \dots, s_t) & \mapsto & \pi_\mdp(A_t | s_t).
	\end{array}
\intertext{For state predictability, $\Theta=\cS$, $\typeOf(s,a,s')=s'$, and $B$ is}
	B : &
	\begin{array}{ccc}
		H^* & \to & \Delta^{|\cS|}, \\
		(s_0, a_0, \dots, s_t) & \mapsto & \sum_{a'}\pi_\mdp(a' | s_t)\cdot T(s_t,a',S_{t+1}).
	\end{array}
\end{align*}
In both cases, since $B$ depends only on the current state, $s_t$, we can denote the belief about target variable $\theta$ under $s_t$ as $B(s_t)=b_{s_t}(\theta)$ and redefine the pOAMDP reward function (not the observer's one) as $R'(s_t,a_t) \eqdef R(s_t,a_t,b_{s_t}(\theta))$ instead of $R(s_t,a_t,B(s_0,a_0,\dots,s_t))$.

The agent's sequential decision-making problem can then be expressed as an MDP $\langle \cS, \cA, T, R', \gamma, \cSt \rangle$ solvable with an algorithm such as value iteration.
The solving complexity is thus the complexity of solving both the observer MDP and the MDP induced by the pOAMDP.
In contrast, in the case of OAMDPs \cite{pmlr-v161-miura21a}, one generally cannot obtain such an MDP, and solving the pOAMDP requires specific algorithms in which the action choice is linked to the whole state-action history (so that the tree of possible futures that needs to be accounted for grows exponentially).

\subsection{pOAMDP Reward Function}
\label{sec|fonction_recompenses}

\paragraph{Reward Definition}
When in state $s$, to predict the next target variable's value (action or state) as well as possible, the observer should pick one of the most likely values according to her model of the agent's behavior.
This means picking an action in $\argmax_{a\in \cA} b_s(a)$ (or a state in $\argmax_{s'\in \cS} b_s(s')$).
We will assume that the observer samples her prediction uniformly from this set, and thus define $\pred(\theta|s) \eqdef \frac{1}{|\argmax_{\theta\in \Theta} b_s(\theta)|}$ if $\theta \in \argmax_{\theta\in \Theta} b_s(\theta)$, and $0$ otherwise. Note: From now on, we focus on action predictability, only highlighting some points for state predictability.

Then, considering an SSP (thus with $\gamma=1$), we would like to minimize the expected number of prediction errors made by the observer along a trajectory.
For a single transition $(s,a,s')$, assuming the above model of observer prediction, the probability of a bad action prediction is $1 - \pred(a|s)$.
Because we are in a maximization rather than a minimization setting, and generalizing the formula to both action and state predictabilities, this leads to defining the reward function as:
\begin{align*}
  R^\Theta_\pred(s,a,s')
  & \eqdef \pred(\typeOf(s,a,s')|s) - 1.
\end{align*}
Then, in any state $s$, $-V^*(s)$ gives the expected number of future prediction errors.

\paragraph{Valid SSPs?}
An important question is whether this reward function induces a valid SSP, which requires ensuring that assumptions (A1) and (A2) are satisfied.

\begin{proposition}
  Let us assume that
  \begin{enumerate*}[label=(\roman*)]
  \item $\gamma=1$,
  \item the MDP considered by the observer is a valid SSP, and
  \item $\RApred$ is the pOAMDP reward function.
  \end{enumerate*}
Then the pOAMDP is a well-defined problem as its induced SSP satisfies assumptions (A1) and (A2).
\end{proposition}

\begin{proof}

  (A1) Let $\pi$ be a policy, and (if it exists) $\cS' \subseteq (\cS \setminus \cSt)$ be a connex subset of states under $\pi$, \ie, once reached, all states are visited infinitely often.
Let $s'\in\cS'$ be a state in which an optimal policy $\pi^*_\mdp$ of the observer SSP would leave $\cS'$.
Then, $\pi^*_\mdp(s')\neq\pi(s')$, so that $\pred(\pi(s')|s')<1$ and $\RApred(s',\pi(s'),s'')<0$ for any $s''$.
As a consequence, states in $\cS'$ being visited infinitely often, for any $s\in\cS'$, $V^\pi(s)=-\infty$.
On the other hand, if, for some state $s\in\cS$, $\pi$ reaches $\cSt$ with probability $1$, then, trivially, $V^\pi(s)>-\infty$.
This proves that (A1) holds.

  (A2) Let us point out that whether a policy is proper or not depends on the reachability of terminal states, not on the reward function.
Since the observer SSP satisfies assumption (A2) and only differs from the pOASSP in its rewards function, the induced SSP also satisfies assumption (A2).
\end{proof}

This result does not hold for state predictability.

\begin{proposition}
  Let us assume that
  \begin{enumerate*}[label=(\roman*)]
  \item $\gamma=1$,
  \item the MDP considered by the observer is a valid SSP, and
  \item $\RSpred$ is the pOAMDP reward function.
  \end{enumerate*}
Then the pOAMDP may be an ill-defined problem as its induced SSP satisfies assumption (A2), but may not satisfy assumption (A1).
\end{proposition}

\begin{proof}
  The proof that assumption (A2) holds is the same as for action predictability.
  
  To prove that assumption (A1) may not hold, let us consider an OAMDP with:
  \begin{itemize}
  \item $\cS = \{s_0, s_G\}$, with $s_0$ initial and $s_G$ terminal;
  \item $\cA = \{a_1, a_2\}$;
  \item the transition function described in \Cref{fig|counterExample|sPred};
  \item $r_{\mdp}(s_0,a,s')=1$ for all $a$ and $s'$; and
  \item the state-predictability reward function.
  \end{itemize}

  \begin{figure}
    \centerline{
      \begin{tikzpicture}[auto,
        node distance=15mm,
        >=latex]

        \tikzstyle{round}=[thick,draw=black,circle]
        
        \node[round] (s0) {$s_0$};
        \node[rectangle,above of= s0] (a1) {$a_1$};
        \node[rectangle,right of= s0] (a2) {$a_2$};
        \node[round,accepting,right of= a2] (sg) {$s_G$};
        
        \draw[->] (s0) to [out=60,in=-45] (a1);
        \draw[->] (s0) to [out=30,in=150] (a2);
        \draw[->] (a1) to [out=-135,in=120] node [right,font=\scriptsize] {$1$} (s0);
        \draw[->] (a2) to [out=-150,in=-30] node [font=\scriptsize] {$0.9$} (s0);
        \draw[->] (a2) to [out=0,in=180] node [font=\scriptsize] {$0.1$} (sg);
      \end{tikzpicture}
    }
    
    \caption{Transition function of an ill-defined p-OASSP for state predictability, with transition probabilities as edge labels.}
    \label{fig|counterExample|sPred}
  \end{figure}

  In this setting, $\pi_\mdp(s_0)$ is deterministic since there is a single optimal action when in $s_0$: $a_2$.
Then, due to the transition function, when applying $\pi_\mdp$ in $s_0$, the most probable next state is $s_0$, meaning that the observer should bet on $s_0$.
As a consequence, we can evaluate the policies $\pi_{a_1}$ and $\pi_{a_2}$ that respectively always pick $a_1$ and $a_2$ when in $s_0$:
  \begin{align}
    V^{\pi_{a_1}}(s_0)
    & = T(s_0,a_1,s_0) \cdot \left( \RSpred(s_0,a_1,s_0) + V^{\pi_{a_1}}(s_0) \right) \\
    & = 1 \cdot \left( 0 +  V^{\pi_{a_1}}(s_0) \right) \\
    & = 0,^{\text{[One could argue that the value is undefined.]}}
    \intertext{and}
    V^{\pi_{a_2}}(s_0)
    & = T(s_0,a_2,s_0) \cdot \left( \RSpred(s_0,a_2,s_0) + V^{\pi_{a_2}}(s_0) \right) \\
    & \quad + T(s_0,a_2,s_G) \cdot \left( \RSpred(s_0,a_2,s_G) + V^{\pi_{a_2}}(s_G) \right)
    \\
    & = 0.9 \cdot \left( 0 + V^{\pi_{a_2}}(s_0) \right)
    + 0.1 \cdot \left( -1 + 0 \right)
    \\
    & = 0.9 \cdot V^{\pi_{a_2}}(s_0) -0.1
    \\
    & = -1.
  \end{align}
  $\pi_{a_1}$ is thus the only optimal policy for state-predictability, although it never reaches the terminal state, which breaks assumption (A1).
\end{proof}

This negative result does not prevent from using $\RSpred$ in OA-SSPs, in particular:
\begin{itemize}
\item if linearly combined with another reward function that necessarily induces a valid SSP, \eg, $\RApred$ or a non-positive reward function (such as $R(s,a,s')=-1$ for any $(s,a,s')$); and
\item in case of deterministic dynamics; indeed, if a policy $\pi$ induces an absorbing subset of states $\cS' \subseteq \cS \setminus \cSt$, then $T(s,\pi(s)) \neq T(s,\pi_{\mdp}(s))$ for some $s\in \cS'$, so that $\RSpred(s,\pi(s),T(s,\pi(s)))<0$, meaning that $\pi$ has a negative infinite value at least in $s$ (so that assumption (A1) holds).
\end{itemize}

\medskip

In the case of (discounted) MDPs, we will rely on the same reward definition.
The interpretation of $-V^*(s)$ is similar if one sees the problem as an equivalent SSP with a $1-\gamma$ termination probability at each time step.

\medskip

The next two sections study this approach to action and state predictability on simple examples
\begin{enumerate*}
\item {\em in silico}, observing and analyzing the policies obtained through our approach, and compared with simple MDP policies; and
\item {\em in vivo}, \ie, confronting actual human observers with several policies.
\end{enumerate*}

\section{Generating and Interpreting Policies} \label{sec|XP}

These first experiments aim at illustrating and better understanding the policies induced by the proposed reward function, and in particular at determining whether they can be considered as predictable.
The code will be made available under an open license.

\subsection{Protocol}

To describe the two types of pOAMDPs considered in our experiments, let us just detail the corresponding MDPs, both set in 4-connected grid worlds, that the observer will take into account:
\begin{itemize}
\item an SSP, named \emph{maze}, in which the agent wants to reach a terminal goal state; and
\item a discounted MDP (with no terminal state), named \emph{firefighter}, in which the agent uses water sources to extinguish fires.
\end{itemize}
To facilitate the analysis, most problems have deterministic dynamics.

\paragraph{Maze problem}
A \emph{maze} (cf. \Cref{fig|labyS}) contains walls (in dark grey), normal cells (in white), slippery cells (in cyan), and terminal cells (pink disks).
The starting cell is marked by a circle. More formally, in this SSP:
\begin{itemize}
\item each state $s$ in $\cS$ indicates the $(x,y)$ coordinates of the agent in a normal, slippery, or terminal cell;
\item $\cSt$ is a non-empty (but also possibly non-singleton) subset of $\cS$;
\item $\cA=\{up, down, left, right \}$;
\item $T(s,a,s')$ encodes the agent's moves: an agent in a normal cell moves in the direction indicated by its action if no wall prevents it; in a slippery cell, the agent has a probability $p$ ($0.5$ in our experiments) of making a 2-cell rather than 1-cell move (if possible); in a terminal cell, the agent does not move;
\item $R_\mdp$, the reward function, returns
  \begin{itemize}\item a default penalty of $-0.04$ for each move,
  \item $-1$ when the agent hits a wall,
\item $+1$ upon reaching a terminal state $s_f$, and
  \item $0$ when the agent stays in the terminal state.
  \end{itemize}
\end{itemize}

This SSP trivially satisfies assumptions (A1) and (A2).

\paragraph{Firefighter problem}
Similar grids are used for the \emph{firefighter} problem, but with terminal cells replaced by fires and water sources (cf. \Cref{fig|pompier1}).
The agent now has a water tank,
which is emptied upon reaching a (never extinguished) fire, and filled upon reaching a (never emptied) water source.
More formally, in this $\gamma = \myDiscount$-discounted MDP:
\begin{itemize}
\item each state $s$ in $\cS$ is represented by a triplet  $(x,y,w)$ with $(x,y)$ the agent's coordinates and $w$ a boolean encoding whether its water tank is full;
\item $\cA =\{up, down, left, right \}$;
\item $T(s,a,s')$ is similar to the \emph{maze} problem, except that $w$ becomes false upon reaching a fire, and true upon reaching a water source;
\item $R_\mdp$, the reward function, returns
  \begin{itemize}\item a default penalty of $-0.04$ for each move,
  \item $-1$ when the agent hits a wall, and
  \item $+1$ when the agent reaches a fire while carrying water ($w=$true).
  \end{itemize}
\end{itemize}
Optimal MDP policies consist in endlessly going back and forth between a water source and a fire.

\paragraph{Baseline Policies}
The pOAMDP solution policy, denoted $\pi^\Theta_\pred$, will be compared with near-optimal solutions of the observer MDP obtained as follows.
We solve the observer MDP until convergence to an $\epsilon$-optimal value function.
Then, in each state $s$, let $\psi(s)\eqdef\{a\in\cA\ |\ Q^*(s,a) \leq V^*(s,a)-2\epsilon\}$.
This set necessarily contains all optimal actions.
With this, we can first define $\pi_\mdps$, a {\em stochastic} policy that, in each state $s$, samples actions uniformly from $\psi(s)$.

In practice, algorithms will often be biased, having a preference order over actions.
We thus also consider the policies that, in each state $s$, deterministically pick the preferred action given a predefined order.
These {\em biased} (and deterministic) policies are denoted $\pi_\mdpb$, not distinguishing them from each other.

\paragraph{pOAMDP Model}
For both types of problems and for each grid environment, a pOAMDP is derived using the previously proposed reward function for predictability $R^\Theta_\pred$.
The baseline policy $\pi_\mdps$ described above serves to identify the observer's possible predictions.
Since each pOAMDP can be considered as an MDP, pOAMDPs are solved by using again the value iteration algorithm with an appropriate discount factor (details in the next section), resulting in the $R^\Theta_\pred$ policy.
Note that our approach does not make use of the softmax policy, thus making its temperature parameter $\temp$ irrelevant.

\subsection{Results}

The figures present both stochastic MDP policies $\pi_\mdps$ (which also ``cover'' all deterministic policies $\pi_\mdpb$), and pOAMDP policies $\pi^\Theta_\pred$, the arrows indicating all $\epsilon$-optimal actions.

\def\mpwidth{.23\textwidth}
\def\aboxwidth{.99\textwidth}

\begin{figure*}[ht!]

  \newcommand{\twoMazes}[2]{
    \begin{minipage}{\mpwidth}
      \centering
      \adjustbox{width=\aboxwidth}{
        \input{expFinalVers/#1_MDP_policyStoch.tikz}
      }
    \end{minipage}
    \hspace{-.10cm}{\small $M_{#2}$}\hspace{-.15cm}
    \begin{minipage}{\mpwidth}
      \centering
      \adjustbox{width=\aboxwidth}{
        \input{expFinalVers/#1_OAMDP_Action_Pred.tikz} }
    \end{minipage}
  }

    \begin{minipage}{\mpwidth}
      \centering
      \adjustbox{width=\aboxwidth}{
        \begin{tikzpicture}
\draw (0,0) grid (11,12);
\draw[fill=gray] (0,0) rectangle (1,1);
\draw[fill=gray] (1,0) rectangle (2,1);
\draw[fill=gray] (2,0) rectangle (3,1);
\draw[fill=gray] (3,0) rectangle (4,1);
\draw[fill=gray] (4,0) rectangle (5,1);
\draw[fill=gray] (5,0) rectangle (6,1);
\draw[fill=gray] (6,0) rectangle (7,1);
\draw[fill=gray] (7,0) rectangle (8,1);
\draw[fill=gray] (8,0) rectangle (9,1);
\draw[fill=gray] (9,0) rectangle (10,1);
\draw[fill=gray] (10,0) rectangle (11,1);
\draw[fill=gray] (0,1) rectangle (1,2);
\draw[fill=gray] (1,1) rectangle (2,2);
\draw[line width=1mm] (2.500000,1.500000) circle (0.300000);
\draw[fill=gray] (3,1) rectangle (4,2);
\draw[fill=gray] (4,1) rectangle (5,2);
\draw[fill=gray] (5,1) rectangle (6,2);
\draw[fill=gray] (6,1) rectangle (7,2);
\draw[fill=gray] (7,1) rectangle (8,2);
\draw[fill=gray] (8,1) rectangle (9,2);
\draw[fill=gray] (9,1) rectangle (10,2);
\draw[fill=gray] (10,1) rectangle (11,2);
\draw[fill=gray] (0,2) rectangle (1,3);
\draw[fill=gray] (1,2) rectangle (2,3);
\draw[fill=gray] (3,2) rectangle (4,3);
\draw[fill=gray] (4,2) rectangle (5,3);
\draw[fill=gray] (5,2) rectangle (6,3);
\draw[fill=gray] (6,2) rectangle (7,3);
\draw[fill=gray] (7,2) rectangle (8,3);
\draw[fill=gray] (8,2) rectangle (9,3);
\draw[fill=gray] (9,2) rectangle (10,3);
\draw[fill=gray] (10,2) rectangle (11,3);
\draw[fill=gray] (0,3) rectangle (1,4);
\draw[fill=gray] (6,3) rectangle (7,4);
\draw[fill=gray] (7,3) rectangle (8,4);
\draw[fill=gray] (8,3) rectangle (9,4);
\draw[fill=gray] (9,3) rectangle (10,4);
\draw[fill=gray] (10,3) rectangle (11,4);
\draw[fill=gray] (0,4) rectangle (1,5);
\draw[fill=gray] (2,4) rectangle (3,5);
\draw[fill=gray] (3,4) rectangle (4,5);
\draw[fill=gray] (4,4) rectangle (5,5);
\draw[fill=gray] (10,4) rectangle (11,5);
\draw[fill=gray] (0,5) rectangle (1,6);
\draw[fill=gray] (2,5) rectangle (3,6);
\draw[fill=gray] (3,5) rectangle (4,6);
\draw[fill=gray] (4,5) rectangle (5,6);
\draw[fill=gray] (10,5) rectangle (11,6);
\draw[fill=gray] (0,6) rectangle (1,7);
\draw[fill=gray] (2,6) rectangle (3,7);
\draw[fill=gray] (3,6) rectangle (4,7);
\draw[fill=gray] (4,6) rectangle (5,7);
\draw[fill=gray] (10,6) rectangle (11,7);
\draw[fill=gray] (0,7) rectangle (1,8);
\draw[fill=gray] (2,7) rectangle (3,8);
\draw[fill=gray] (3,7) rectangle (4,8);
\draw[fill=gray] (4,7) rectangle (5,8);
\draw[fill=gray] (10,7) rectangle (11,8);
\draw[fill=gray] (0,8) rectangle (1,9);
\draw[fill=gray] (2,8) rectangle (3,9);
\draw[fill=gray] (3,8) rectangle (4,9);
\draw[fill=gray] (4,8) rectangle (5,9);
\draw[fill=gray] (10,8) rectangle (11,9);
\draw[fill=gray] (0,9) rectangle (1,10);
\draw[fill=gray] (2,9) rectangle (3,10);
\draw[fill=gray] (3,9) rectangle (4,10);
\draw[fill=gray] (4,9) rectangle (5,10);
\draw[fill=gray] (5,9) rectangle (6,10);
\draw[fill=gray] (6,9) rectangle (7,10);
\draw[fill=gray] (7,9) rectangle (8,10);
\draw[fill=gray] (8,9) rectangle (9,10);
\draw[fill=gray] (10,9) rectangle (11,10);
\draw[fill=gray] (0,10) rectangle (1,11);
\draw[fill=pink] (9.500000,10.500000) circle (0.500000);
\draw[fill=gray] (10,10) rectangle (11,11);
\draw[fill=gray] (0,11) rectangle (1,12);
\draw[fill=gray] (1,11) rectangle (2,12);
\draw[fill=gray] (2,11) rectangle (3,12);
\draw[fill=gray] (3,11) rectangle (4,12);
\draw[fill=gray] (4,11) rectangle (5,12);
\draw[fill=gray] (5,11) rectangle (6,12);
\draw[fill=gray] (6,11) rectangle (7,12);
\draw[fill=gray] (7,11) rectangle (8,12);
\draw[fill=gray] (8,11) rectangle (9,12);
\draw[fill=gray] (9,11) rectangle (10,12);
\draw[fill=gray] (10,11) rectangle (11,12);
\node [align=center] at(-0.500000,0.500000) {0};
\node [align=center] at(-0.500000,1.500000) {1};
\node [align=center] at(-0.500000,2.500000) {2};
\node [align=center] at(-0.500000,3.500000) {3};
\node [align=center] at(-0.500000,4.500000) {4};
\node [align=center] at(-0.500000,5.500000) {5};
\node [align=center] at(-0.500000,6.500000) {6};
\node [align=center] at(-0.500000,7.500000) {7};
\node [align=center] at(-0.500000,8.500000) {8};
\node [align=center] at(-0.500000,9.500000) {9};
\node [align=center] at(-0.500000,10.500000) {10};
\node [align=center] at(-0.500000,11.500000) {11};
\node [align=center] at(0.500000,-0.500000) {A};
\node [align=center] at(1.500000,-0.500000) {B};
\node [align=center] at(2.500000,-0.500000) {C};
\node [align=center] at(3.500000,-0.500000) {D};
\node [align=center] at(4.500000,-0.500000) {E};
\node [align=center] at(5.500000,-0.500000) {F};
\node [align=center] at(6.500000,-0.500000) {G};
\node [align=center] at(7.500000,-0.500000) {H};
\node [align=center] at(8.500000,-0.500000) {I};
\node [align=center] at(9.500000,-0.500000) {J};
\node [align=center] at(10.500000,-0.500000) {K};
\draw[-Latex,draw=black!100] (1.500000,10.500000)--(1.900000,10.500000);
\draw[-Latex,draw=black!100] (2.500000,10.500000)--(2.900000,10.500000);
\draw[-Latex,draw=black!100] (3.500000,10.500000)--(3.900000,10.500000);
\draw[-Latex,draw=black!100] (4.500000,10.500000)--(4.900000,10.500000);
\draw[-Latex,draw=black!100] (5.500000,10.500000)--(5.900000,10.500000);
\draw[-Latex,draw=black!100] (6.500000,10.500000)--(6.900000,10.500000);
\draw[-Latex,draw=black!100] (7.500000,10.500000)--(7.900000,10.500000);
\draw[-Latex,draw=black!100] (8.500000,10.500000)--(8.900000,10.500000);
\draw[-Latex,draw=black!100] (9.500000,10.500000)--(9.100000,10.500000);
\draw[-Latex,draw=black!100] (9.500000,10.500000)--(9.500000,10.900000);
\draw[-Latex,draw=black!100] (9.500000,10.500000)--(9.900000,10.500000);
\draw[-Latex,draw=black!100] (9.500000,10.500000)--(9.500000,10.100000);
\draw[-Latex,draw=black!100] (1.500000,9.500000)--(1.500000,9.900000);
\draw[-Latex,draw=black!100] (9.500000,9.500000)--(9.500000,9.900000);
\draw[-Latex,draw=black!100] (1.500000,8.500000)--(1.500000,8.900000);
\draw[-Latex,draw=black!100] (5.500000,8.500000)--(5.900000,8.500000);
\draw[-Latex,draw=black!100] (6.500000,8.500000)--(6.900000,8.500000);
\draw[-Latex,draw=black!100] (7.500000,8.500000)--(7.900000,8.500000);
\draw[-Latex,draw=black!100] (8.500000,8.500000)--(8.900000,8.500000);
\draw[-Latex,draw=black!100] (9.500000,8.500000)--(9.500000,8.900000);
\draw[-Latex,draw=black!100] (1.500000,7.500000)--(1.500000,7.900000);
\draw[-Latex,draw=black!100] (5.500000,7.500000)--(5.500000,7.900000);
\draw[-Latex,draw=black!100] (5.500000,7.500000)--(5.900000,7.500000);
\draw[-Latex,draw=black!100] (6.500000,7.500000)--(6.500000,7.900000);
\draw[-Latex,draw=black!100] (6.500000,7.500000)--(6.900000,7.500000);
\draw[-Latex,draw=black!100] (7.500000,7.500000)--(7.500000,7.900000);
\draw[-Latex,draw=black!100] (7.500000,7.500000)--(7.900000,7.500000);
\draw[-Latex,draw=black!100] (8.500000,7.500000)--(8.500000,7.900000);
\draw[-Latex,draw=black!100] (8.500000,7.500000)--(8.900000,7.500000);
\draw[-Latex,draw=black!100] (9.500000,7.500000)--(9.500000,7.900000);
\draw[-Latex,draw=black!100] (1.500000,6.500000)--(1.500000,6.900000);
\draw[-Latex,draw=black!100] (5.500000,6.500000)--(5.500000,6.900000);
\draw[-Latex,draw=black!100] (5.500000,6.500000)--(5.900000,6.500000);
\draw[-Latex,draw=black!100] (6.500000,6.500000)--(6.500000,6.900000);
\draw[-Latex,draw=black!100] (6.500000,6.500000)--(6.900000,6.500000);
\draw[-Latex,draw=black!100] (7.500000,6.500000)--(7.500000,6.900000);
\draw[-Latex,draw=black!100] (7.500000,6.500000)--(7.900000,6.500000);
\draw[-Latex,draw=black!100] (8.500000,6.500000)--(8.500000,6.900000);
\draw[-Latex,draw=black!100] (8.500000,6.500000)--(8.900000,6.500000);
\draw[-Latex,draw=black!100] (9.500000,6.500000)--(9.500000,6.900000);
\draw[-Latex,draw=black!100] (1.500000,5.500000)--(1.500000,5.900000);
\draw[-Latex,draw=black!100] (5.500000,5.500000)--(5.500000,5.900000);
\draw[-Latex,draw=black!100] (5.500000,5.500000)--(5.900000,5.500000);
\draw[-Latex,draw=black!100] (6.500000,5.500000)--(6.500000,5.900000);
\draw[-Latex,draw=black!100] (6.500000,5.500000)--(6.900000,5.500000);
\draw[-Latex,draw=black!100] (7.500000,5.500000)--(7.500000,5.900000);
\draw[-Latex,draw=black!100] (7.500000,5.500000)--(7.900000,5.500000);
\draw[-Latex,draw=black!100] (8.500000,5.500000)--(8.500000,5.900000);
\draw[-Latex,draw=black!100] (8.500000,5.500000)--(8.900000,5.500000);
\draw[-Latex,draw=black!100] (9.500000,5.500000)--(9.500000,5.900000);
\draw[-Latex,draw=black!100] (1.500000,4.500000)--(1.500000,4.900000);
\draw[-Latex,draw=black!100] (5.500000,4.500000)--(5.500000,4.900000);
\draw[-Latex,draw=black!100] (5.500000,4.500000)--(5.900000,4.500000);
\draw[-Latex,draw=black!100] (6.500000,4.500000)--(6.500000,4.900000);
\draw[-Latex,draw=black!100] (6.500000,4.500000)--(6.900000,4.500000);
\draw[-Latex,draw=black!100] (7.500000,4.500000)--(7.500000,4.900000);
\draw[-Latex,draw=black!100] (7.500000,4.500000)--(7.900000,4.500000);
\draw[-Latex,draw=black!100] (8.500000,4.500000)--(8.500000,4.900000);
\draw[-Latex,draw=black!100] (8.500000,4.500000)--(8.900000,4.500000);
\draw[-Latex,draw=black!100] (9.500000,4.500000)--(9.500000,4.900000);
\draw[-Latex,draw=black!100] (1.500000,3.500000)--(1.500000,3.900000);
\draw[-Latex,draw=black!100] (1.500000,3.500000)--(1.900000,3.500000);
\draw[-Latex,draw=black!100] (2.500000,3.500000)--(2.900000,3.500000);
\draw[-Latex,draw=black!100] (3.500000,3.500000)--(3.900000,3.500000);
\draw[-Latex,draw=black!100] (4.500000,3.500000)--(4.900000,3.500000);
\draw[-Latex,draw=black!100] (5.500000,3.500000)--(5.500000,3.900000);
\draw[-Latex,draw=black!100] (2.500000,2.500000)--(2.500000,2.900000);
\draw[-Latex,draw=black!100] (2.500000,1.500000)--(2.500000,1.900000);
\end{tikzpicture}
      }
    \end{minipage}
    \hspace{-.10cm}{\small $M_{1}$}\hspace{-.15cm}
    \begin{minipage}{\mpwidth}
      \centering
      \adjustbox{width=\aboxwidth}{
        \begin{tikzpicture}
\draw (0,0) grid (11,12);
\draw[fill=gray] (0,0) rectangle (1,1);
\draw[fill=gray] (1,0) rectangle (2,1);
\draw[fill=gray] (2,0) rectangle (3,1);
\draw[fill=gray] (3,0) rectangle (4,1);
\draw[fill=gray] (4,0) rectangle (5,1);
\draw[fill=gray] (5,0) rectangle (6,1);
\draw[fill=gray] (6,0) rectangle (7,1);
\draw[fill=gray] (7,0) rectangle (8,1);
\draw[fill=gray] (8,0) rectangle (9,1);
\draw[fill=gray] (9,0) rectangle (10,1);
\draw[fill=gray] (10,0) rectangle (11,1);
\draw[fill=gray] (0,1) rectangle (1,2);
\draw[fill=gray] (1,1) rectangle (2,2);
\draw[line width=1mm] (2.500000,1.500000) circle (0.300000);
\draw[fill=gray] (3,1) rectangle (4,2);
\draw[fill=gray] (4,1) rectangle (5,2);
\draw[fill=gray] (5,1) rectangle (6,2);
\draw[fill=gray] (6,1) rectangle (7,2);
\draw[fill=gray] (7,1) rectangle (8,2);
\draw[fill=gray] (8,1) rectangle (9,2);
\draw[fill=gray] (9,1) rectangle (10,2);
\draw[fill=gray] (10,1) rectangle (11,2);
\draw[fill=gray] (0,2) rectangle (1,3);
\draw[fill=gray] (1,2) rectangle (2,3);
\draw[fill=gray] (3,2) rectangle (4,3);
\draw[fill=gray] (4,2) rectangle (5,3);
\draw[fill=gray] (5,2) rectangle (6,3);
\draw[fill=gray] (6,2) rectangle (7,3);
\draw[fill=gray] (7,2) rectangle (8,3);
\draw[fill=gray] (8,2) rectangle (9,3);
\draw[fill=gray] (9,2) rectangle (10,3);
\draw[fill=gray] (10,2) rectangle (11,3);
\draw[fill=gray] (0,3) rectangle (1,4);
\draw[fill=gray] (6,3) rectangle (7,4);
\draw[fill=gray] (7,3) rectangle (8,4);
\draw[fill=gray] (8,3) rectangle (9,4);
\draw[fill=gray] (9,3) rectangle (10,4);
\draw[fill=gray] (10,3) rectangle (11,4);
\draw[fill=gray] (0,4) rectangle (1,5);
\draw[fill=gray] (2,4) rectangle (3,5);
\draw[fill=gray] (3,4) rectangle (4,5);
\draw[fill=gray] (4,4) rectangle (5,5);
\draw[fill=gray] (10,4) rectangle (11,5);
\draw[fill=gray] (0,5) rectangle (1,6);
\draw[fill=gray] (2,5) rectangle (3,6);
\draw[fill=gray] (3,5) rectangle (4,6);
\draw[fill=gray] (4,5) rectangle (5,6);
\draw[fill=gray] (10,5) rectangle (11,6);
\draw[fill=gray] (0,6) rectangle (1,7);
\draw[fill=gray] (2,6) rectangle (3,7);
\draw[fill=gray] (3,6) rectangle (4,7);
\draw[fill=gray] (4,6) rectangle (5,7);
\draw[fill=gray] (10,6) rectangle (11,7);
\draw[fill=gray] (0,7) rectangle (1,8);
\draw[fill=gray] (2,7) rectangle (3,8);
\draw[fill=gray] (3,7) rectangle (4,8);
\draw[fill=gray] (4,7) rectangle (5,8);
\draw[fill=gray] (10,7) rectangle (11,8);
\draw[fill=gray] (0,8) rectangle (1,9);
\draw[fill=gray] (2,8) rectangle (3,9);
\draw[fill=gray] (3,8) rectangle (4,9);
\draw[fill=gray] (4,8) rectangle (5,9);
\draw[fill=gray] (10,8) rectangle (11,9);
\draw[fill=gray] (0,9) rectangle (1,10);
\draw[fill=gray] (2,9) rectangle (3,10);
\draw[fill=gray] (3,9) rectangle (4,10);
\draw[fill=gray] (4,9) rectangle (5,10);
\draw[fill=gray] (5,9) rectangle (6,10);
\draw[fill=gray] (6,9) rectangle (7,10);
\draw[fill=gray] (7,9) rectangle (8,10);
\draw[fill=gray] (8,9) rectangle (9,10);
\draw[fill=gray] (10,9) rectangle (11,10);
\draw[fill=gray] (0,10) rectangle (1,11);
\draw[fill=pink] (9.500000,10.500000) circle (0.500000);
\draw[fill=gray] (10,10) rectangle (11,11);
\draw[fill=gray] (0,11) rectangle (1,12);
\draw[fill=gray] (1,11) rectangle (2,12);
\draw[fill=gray] (2,11) rectangle (3,12);
\draw[fill=gray] (3,11) rectangle (4,12);
\draw[fill=gray] (4,11) rectangle (5,12);
\draw[fill=gray] (5,11) rectangle (6,12);
\draw[fill=gray] (6,11) rectangle (7,12);
\draw[fill=gray] (7,11) rectangle (8,12);
\draw[fill=gray] (8,11) rectangle (9,12);
\draw[fill=gray] (9,11) rectangle (10,12);
\draw[fill=gray] (10,11) rectangle (11,12);
\node [align=center] at(-0.500000,0.500000) {0};
\node [align=center] at(-0.500000,1.500000) {1};
\node [align=center] at(-0.500000,2.500000) {2};
\node [align=center] at(-0.500000,3.500000) {3};
\node [align=center] at(-0.500000,4.500000) {4};
\node [align=center] at(-0.500000,5.500000) {5};
\node [align=center] at(-0.500000,6.500000) {6};
\node [align=center] at(-0.500000,7.500000) {7};
\node [align=center] at(-0.500000,8.500000) {8};
\node [align=center] at(-0.500000,9.500000) {9};
\node [align=center] at(-0.500000,10.500000) {10};
\node [align=center] at(-0.500000,11.500000) {11};
\node [align=center] at(0.500000,-0.500000) {A};
\node [align=center] at(1.500000,-0.500000) {B};
\node [align=center] at(2.500000,-0.500000) {C};
\node [align=center] at(3.500000,-0.500000) {D};
\node [align=center] at(4.500000,-0.500000) {E};
\node [align=center] at(5.500000,-0.500000) {F};
\node [align=center] at(6.500000,-0.500000) {G};
\node [align=center] at(7.500000,-0.500000) {H};
\node [align=center] at(8.500000,-0.500000) {I};
\node [align=center] at(9.500000,-0.500000) {J};
\node [align=center] at(10.500000,-0.500000) {K};
\draw[-Latex,draw=black!100] (1.500000,10.500000)--(1.900000,10.500000);
\draw[-Latex,draw=black!100] (2.500000,10.500000)--(2.900000,10.500000);
\draw[-Latex,draw=black!100] (3.500000,10.500000)--(3.900000,10.500000);
\draw[-Latex,draw=black!100] (4.500000,10.500000)--(4.900000,10.500000);
\draw[-Latex,draw=black!100] (5.500000,10.500000)--(5.900000,10.500000);
\draw[-Latex,draw=black!100] (6.500000,10.500000)--(6.900000,10.500000);
\draw[-Latex,draw=black!100] (7.500000,10.500000)--(7.900000,10.500000);
\draw[-Latex,draw=black!100] (8.500000,10.500000)--(8.900000,10.500000);
\draw[-Latex,draw=black!100] (9.500000,10.500000)--(9.100000,10.500000);
\draw[-Latex,draw=black!100] (9.500000,10.500000)--(9.500000,10.900000);
\draw[-Latex,draw=black!100] (9.500000,10.500000)--(9.900000,10.500000);
\draw[-Latex,draw=black!100] (9.500000,10.500000)--(9.500000,10.100000);
\draw[-Latex,draw=black!100] (1.500000,9.500000)--(1.500000,9.900000);
\draw[-Latex,draw=black!100] (9.500000,9.500000)--(9.500000,9.900000);
\draw[-Latex,draw=black!100] (1.500000,8.500000)--(1.500000,8.900000);
\draw[-Latex,draw=black!100] (5.500000,8.500000)--(5.900000,8.500000);
\draw[-Latex,draw=black!100] (6.500000,8.500000)--(6.900000,8.500000);
\draw[-Latex,draw=black!100] (7.500000,8.500000)--(7.900000,8.500000);
\draw[-Latex,draw=black!100] (8.500000,8.500000)--(8.900000,8.500000);
\draw[-Latex,draw=black!100] (9.500000,8.500000)--(9.500000,8.900000);
\draw[-Latex,draw=black!100] (1.500000,7.500000)--(1.500000,7.900000);
\draw[-Latex,draw=black!100] (5.500000,7.500000)--(5.500000,7.900000);
\draw[-Latex,draw=black!100] (6.500000,7.500000)--(6.500000,7.900000);
\draw[-Latex,draw=black!100] (7.500000,7.500000)--(7.500000,7.900000);
\draw[-Latex,draw=black!100] (8.500000,7.500000)--(8.500000,7.900000);
\draw[-Latex,draw=black!100] (8.500000,7.500000)--(8.900000,7.500000);
\draw[-Latex,draw=black!100] (9.500000,7.500000)--(9.500000,7.900000);
\draw[-Latex,draw=black!100] (1.500000,6.500000)--(1.500000,6.900000);
\draw[-Latex,draw=black!100] (5.500000,6.500000)--(5.500000,6.900000);
\draw[-Latex,draw=black!100] (6.500000,6.500000)--(6.500000,6.900000);
\draw[-Latex,draw=black!100] (7.500000,6.500000)--(7.500000,6.900000);
\draw[-Latex,draw=black!100] (7.500000,6.500000)--(7.900000,6.500000);
\draw[-Latex,draw=black!100] (8.500000,6.500000)--(8.900000,6.500000);
\draw[-Latex,draw=black!100] (9.500000,6.500000)--(9.500000,6.900000);
\draw[-Latex,draw=black!100] (1.500000,5.500000)--(1.500000,5.900000);
\draw[-Latex,draw=black!100] (5.500000,5.500000)--(5.500000,5.900000);
\draw[-Latex,draw=black!100] (6.500000,5.500000)--(6.500000,5.900000);
\draw[-Latex,draw=black!100] (6.500000,5.500000)--(6.900000,5.500000);
\draw[-Latex,draw=black!100] (7.500000,5.500000)--(7.900000,5.500000);
\draw[-Latex,draw=black!100] (8.500000,5.500000)--(8.900000,5.500000);
\draw[-Latex,draw=black!100] (9.500000,5.500000)--(9.500000,5.900000);
\draw[-Latex,draw=black!100] (1.500000,4.500000)--(1.500000,4.900000);
\draw[-Latex,draw=black!100] (5.500000,4.500000)--(5.500000,4.900000);
\draw[-Latex,draw=black!100] (5.500000,4.500000)--(5.900000,4.500000);
\draw[-Latex,draw=black!100] (6.500000,4.500000)--(6.900000,4.500000);
\draw[-Latex,draw=black!100] (7.500000,4.500000)--(7.900000,4.500000);
\draw[-Latex,draw=black!100] (8.500000,4.500000)--(8.900000,4.500000);
\draw[-Latex,draw=black!100] (9.500000,4.500000)--(9.500000,4.900000);
\draw[-Latex,draw=black!100] (1.500000,3.500000)--(1.500000,3.900000);
\draw[-Latex,draw=black!100] (2.500000,3.500000)--(2.100000,3.500000);
\draw[-Latex,draw=black!100] (3.500000,3.500000)--(3.900000,3.500000);
\draw[-Latex,draw=black!100] (4.500000,3.500000)--(4.900000,3.500000);
\draw[-Latex,draw=black!100] (5.500000,3.500000)--(5.500000,3.900000);
\draw[-Latex,draw=black!100] (2.500000,2.500000)--(2.500000,2.900000);
\draw[-Latex,draw=black!100] (2.500000,1.500000)--(2.500000,1.900000);
\end{tikzpicture} }
    \end{minipage}
  
  \hfill
  
    \begin{minipage}{\mpwidth}
      \centering
      \adjustbox{width=\aboxwidth}{
        \input{expFinalVers/5_MDP_policyStoch.tikz}
      }
    \end{minipage}
    \hspace{-.10cm}{\small $M_{2}$}\hspace{-.15cm}
    \begin{minipage}{\mpwidth}
      \centering
      \adjustbox{width=\aboxwidth}{
        \begin{tikzpicture}
\draw (0,0) grid (11,13);
\draw[fill=gray] (0,0) rectangle (1,1);
\draw[fill=gray] (1,0) rectangle (2,1);
\draw[fill=gray] (2,0) rectangle (3,1);
\draw[fill=gray] (3,0) rectangle (4,1);
\draw[fill=gray] (4,0) rectangle (5,1);
\draw[fill=gray] (5,0) rectangle (6,1);
\draw[fill=gray] (6,0) rectangle (7,1);
\draw[fill=gray] (7,0) rectangle (8,1);
\draw[fill=gray] (8,0) rectangle (9,1);
\draw[fill=gray] (9,0) rectangle (10,1);
\draw[fill=gray] (10,0) rectangle (11,1);
\draw[fill=gray] (0,1) rectangle (1,2);
\draw[fill=pink] (9.500000,1.500000) circle (0.500000);
\draw[fill=gray] (10,1) rectangle (11,2);
\draw[fill=gray] (0,2) rectangle (1,3);
\draw[fill=gray] (2,2) rectangle (3,3);
\draw[fill=gray] (3,2) rectangle (4,3);
\draw[fill=gray] (4,2) rectangle (5,3);
\draw[fill=gray] (5,2) rectangle (6,3);
\draw[fill=gray] (6,2) rectangle (7,3);
\draw[fill=gray] (7,2) rectangle (8,3);
\draw[fill=gray] (8,2) rectangle (9,3);
\draw[fill=gray] (10,2) rectangle (11,3);
\draw[fill=gray] (0,3) rectangle (1,4);
\draw[fill=gray] (2,3) rectangle (3,4);
\draw[fill=gray] (3,3) rectangle (4,4);
\draw[fill=gray] (4,3) rectangle (5,4);
\draw[fill=gray] (10,3) rectangle (11,4);
\draw[fill=gray] (0,4) rectangle (1,5);
\draw[fill=gray] (2,4) rectangle (3,5);
\draw[fill=gray] (3,4) rectangle (4,5);
\draw[fill=gray] (4,4) rectangle (5,5);
\draw[fill=gray] (10,4) rectangle (11,5);
\draw[fill=gray] (0,5) rectangle (1,6);
\draw[fill=gray] (2,5) rectangle (3,6);
\draw[fill=gray] (3,5) rectangle (4,6);
\draw[fill=gray] (4,5) rectangle (5,6);
\draw[fill=gray] (10,5) rectangle (11,6);
\draw[fill=gray] (0,6) rectangle (1,7);
\draw[fill=gray] (2,6) rectangle (3,7);
\draw[fill=gray] (3,6) rectangle (4,7);
\draw[fill=gray] (4,6) rectangle (5,7);
\draw[fill=gray] (10,6) rectangle (11,7);
\draw[fill=gray] (0,7) rectangle (1,8);
\draw[fill=gray] (2,7) rectangle (3,8);
\draw[fill=gray] (3,7) rectangle (4,8);
\draw[fill=gray] (4,7) rectangle (5,8);
\draw[fill=gray] (10,7) rectangle (11,8);
\draw[fill=gray] (0,8) rectangle (1,9);
\draw[fill=gray] (2,8) rectangle (3,9);
\draw[fill=gray] (3,8) rectangle (4,9);
\draw[fill=gray] (4,8) rectangle (5,9);
\draw[fill=gray] (10,8) rectangle (11,9);
\draw[fill=gray] (0,9) rectangle (1,10);
\draw[fill=gray] (6,9) rectangle (7,10);
\draw[fill=gray] (7,9) rectangle (8,10);
\draw[fill=gray] (8,9) rectangle (9,10);
\draw[fill=gray] (9,9) rectangle (10,10);
\draw[fill=gray] (10,9) rectangle (11,10);
\draw[fill=gray] (0,10) rectangle (1,11);
\draw[fill=gray] (1,10) rectangle (2,11);
\draw[fill=gray] (2,10) rectangle (3,11);
\draw[fill=gray] (3,10) rectangle (4,11);
\draw[fill=gray] (4,10) rectangle (5,11);
\draw[fill=gray] (6,10) rectangle (7,11);
\draw[fill=gray] (7,10) rectangle (8,11);
\draw[fill=gray] (8,10) rectangle (9,11);
\draw[fill=gray] (9,10) rectangle (10,11);
\draw[fill=gray] (10,10) rectangle (11,11);
\draw[fill=gray] (0,11) rectangle (1,12);
\draw[fill=gray] (1,11) rectangle (2,12);
\draw[fill=gray] (2,11) rectangle (3,12);
\draw[fill=gray] (3,11) rectangle (4,12);
\draw[fill=gray] (4,11) rectangle (5,12);
\draw[line width=1mm] (5.500000,11.500000) circle (0.300000);
\draw[fill=gray] (6,11) rectangle (7,12);
\draw[fill=gray] (7,11) rectangle (8,12);
\draw[fill=gray] (8,11) rectangle (9,12);
\draw[fill=gray] (9,11) rectangle (10,12);
\draw[fill=gray] (10,11) rectangle (11,12);
\draw[fill=gray] (0,12) rectangle (1,13);
\draw[fill=gray] (1,12) rectangle (2,13);
\draw[fill=gray] (2,12) rectangle (3,13);
\draw[fill=gray] (3,12) rectangle (4,13);
\draw[fill=gray] (4,12) rectangle (5,13);
\draw[fill=gray] (5,12) rectangle (6,13);
\draw[fill=gray] (6,12) rectangle (7,13);
\draw[fill=gray] (7,12) rectangle (8,13);
\draw[fill=gray] (8,12) rectangle (9,13);
\draw[fill=gray] (9,12) rectangle (10,13);
\draw[fill=gray] (10,12) rectangle (11,13);
\node [align=center] at(-0.500000,0.500000) {0};
\node [align=center] at(-0.500000,1.500000) {1};
\node [align=center] at(-0.500000,2.500000) {2};
\node [align=center] at(-0.500000,3.500000) {3};
\node [align=center] at(-0.500000,4.500000) {4};
\node [align=center] at(-0.500000,5.500000) {5};
\node [align=center] at(-0.500000,6.500000) {6};
\node [align=center] at(-0.500000,7.500000) {7};
\node [align=center] at(-0.500000,8.500000) {8};
\node [align=center] at(-0.500000,9.500000) {9};
\node [align=center] at(-0.500000,10.500000) {10};
\node [align=center] at(-0.500000,11.500000) {11};
\node [align=center] at(-0.500000,12.500000) {12};
\node [align=center] at(0.500000,-0.500000) {A};
\node [align=center] at(1.500000,-0.500000) {B};
\node [align=center] at(2.500000,-0.500000) {C};
\node [align=center] at(3.500000,-0.500000) {D};
\node [align=center] at(4.500000,-0.500000) {E};
\node [align=center] at(5.500000,-0.500000) {F};
\node [align=center] at(6.500000,-0.500000) {G};
\node [align=center] at(7.500000,-0.500000) {H};
\node [align=center] at(8.500000,-0.500000) {I};
\node [align=center] at(9.500000,-0.500000) {J};
\node [align=center] at(10.500000,-0.500000) {K};
\draw[-Latex,draw=black!100] (5.500000,11.500000)--(5.500000,11.100000);
\draw[-Latex,draw=black!100] (5.500000,10.500000)--(5.500000,10.100000);
\draw[-Latex,draw=black!100] (1.500000,9.500000)--(1.500000,9.100000);
\draw[-Latex,draw=black!100] (2.500000,9.500000)--(2.100000,9.500000);
\draw[-Latex,draw=black!100] (3.500000,9.500000)--(3.900000,9.500000);
\draw[-Latex,draw=black!100] (4.500000,9.500000)--(4.900000,9.500000);
\draw[-Latex,draw=black!100] (5.500000,9.500000)--(5.500000,9.100000);
\draw[-Latex,draw=black!100] (1.500000,8.500000)--(1.500000,8.100000);
\draw[-Latex,draw=black!100] (5.500000,8.500000)--(5.900000,8.500000);
\draw[-Latex,draw=black!100] (6.500000,8.500000)--(6.900000,8.500000);
\draw[-Latex,draw=black!100] (7.500000,8.500000)--(7.900000,8.500000);
\draw[-Latex,draw=black!100] (8.500000,8.500000)--(8.900000,8.500000);
\draw[-Latex,draw=black!100] (9.500000,8.500000)--(9.500000,8.100000);
\draw[-Latex,draw=black!100] (1.500000,7.500000)--(1.500000,7.100000);
\draw[-Latex,draw=black!100] (5.500000,7.500000)--(5.900000,7.500000);
\draw[-Latex,draw=black!100] (5.500000,7.500000)--(5.500000,7.100000);
\draw[-Latex,draw=black!100] (6.500000,7.500000)--(6.900000,7.500000);
\draw[-Latex,draw=black!100] (7.500000,7.500000)--(7.900000,7.500000);
\draw[-Latex,draw=black!100] (8.500000,7.500000)--(8.900000,7.500000);
\draw[-Latex,draw=black!100] (9.500000,7.500000)--(9.500000,7.100000);
\draw[-Latex,draw=black!100] (1.500000,6.500000)--(1.500000,6.100000);
\draw[-Latex,draw=black!100] (5.500000,6.500000)--(5.500000,6.100000);
\draw[-Latex,draw=black!100] (6.500000,6.500000)--(6.900000,6.500000);
\draw[-Latex,draw=black!100] (6.500000,6.500000)--(6.500000,6.100000);
\draw[-Latex,draw=black!100] (7.500000,6.500000)--(7.900000,6.500000);
\draw[-Latex,draw=black!100] (8.500000,6.500000)--(8.900000,6.500000);
\draw[-Latex,draw=black!100] (9.500000,6.500000)--(9.500000,6.100000);
\draw[-Latex,draw=black!100] (1.500000,5.500000)--(1.500000,5.100000);
\draw[-Latex,draw=black!100] (5.500000,5.500000)--(5.500000,5.100000);
\draw[-Latex,draw=black!100] (6.500000,5.500000)--(6.500000,5.100000);
\draw[-Latex,draw=black!100] (7.500000,5.500000)--(7.900000,5.500000);
\draw[-Latex,draw=black!100] (7.500000,5.500000)--(7.500000,5.100000);
\draw[-Latex,draw=black!100] (8.500000,5.500000)--(8.900000,5.500000);
\draw[-Latex,draw=black!100] (9.500000,5.500000)--(9.500000,5.100000);
\draw[-Latex,draw=black!100] (1.500000,4.500000)--(1.500000,4.100000);
\draw[-Latex,draw=black!100] (5.500000,4.500000)--(5.500000,4.100000);
\draw[-Latex,draw=black!100] (6.500000,4.500000)--(6.500000,4.100000);
\draw[-Latex,draw=black!100] (7.500000,4.500000)--(7.500000,4.100000);
\draw[-Latex,draw=black!100] (8.500000,4.500000)--(8.900000,4.500000);
\draw[-Latex,draw=black!100] (8.500000,4.500000)--(8.500000,4.100000);
\draw[-Latex,draw=black!100] (9.500000,4.500000)--(9.500000,4.100000);
\draw[-Latex,draw=black!100] (1.500000,3.500000)--(1.500000,3.100000);
\draw[-Latex,draw=black!100] (5.500000,3.500000)--(5.900000,3.500000);
\draw[-Latex,draw=black!100] (6.500000,3.500000)--(6.900000,3.500000);
\draw[-Latex,draw=black!100] (7.500000,3.500000)--(7.900000,3.500000);
\draw[-Latex,draw=black!100] (8.500000,3.500000)--(8.900000,3.500000);
\draw[-Latex,draw=black!100] (9.500000,3.500000)--(9.500000,3.100000);
\draw[-Latex,draw=black!100] (1.500000,2.500000)--(1.500000,2.100000);
\draw[-Latex,draw=black!100] (9.500000,2.500000)--(9.500000,2.100000);
\draw[-Latex,draw=black!100] (1.500000,1.500000)--(1.900000,1.500000);
\draw[-Latex,draw=black!100] (2.500000,1.500000)--(2.900000,1.500000);
\draw[-Latex,draw=black!100] (3.500000,1.500000)--(3.900000,1.500000);
\draw[-Latex,draw=black!100] (4.500000,1.500000)--(4.900000,1.500000);
\draw[-Latex,draw=black!100] (5.500000,1.500000)--(5.900000,1.500000);
\draw[-Latex,draw=black!100] (6.500000,1.500000)--(6.900000,1.500000);
\draw[-Latex,draw=black!100] (7.500000,1.500000)--(7.900000,1.500000);
\draw[-Latex,draw=black!100] (8.500000,1.500000)--(8.900000,1.500000);
\draw[-Latex,draw=black!100] (9.500000,1.500000)--(9.100000,1.500000);
\draw[-Latex,draw=black!100] (9.500000,1.500000)--(9.500000,1.900000);
\draw[-Latex,draw=black!100] (9.500000,1.500000)--(9.900000,1.500000);
\draw[-Latex,draw=black!100] (9.500000,1.500000)--(9.500000,1.100000);
\end{tikzpicture} }
    \end{minipage}

    \begin{minipage}{\mpwidth}
      \centering
      \adjustbox{width=\aboxwidth}{
        \begin{tikzpicture}
\draw (0,0) grid (11,11);
\draw[fill=gray] (0,0) rectangle (1,1);
\draw[fill=gray] (1,0) rectangle (2,1);
\draw[fill=gray] (2,0) rectangle (3,1);
\draw[fill=gray] (3,0) rectangle (4,1);
\draw[fill=gray] (4,0) rectangle (5,1);
\draw[fill=gray] (5,0) rectangle (6,1);
\draw[fill=gray] (6,0) rectangle (7,1);
\draw[fill=gray] (7,0) rectangle (8,1);
\draw[fill=gray] (8,0) rectangle (9,1);
\draw[fill=gray] (9,0) rectangle (10,1);
\draw[fill=gray] (10,0) rectangle (11,1);
\draw[fill=gray] (0,1) rectangle (1,2);
\draw[fill=pink] (9.500000,1.500000) circle (0.500000);
\draw[fill=gray] (10,1) rectangle (11,2);
\draw[fill=gray] (0,2) rectangle (1,3);
\draw[fill=gray] (5,2) rectangle (6,3);
\draw[fill=gray] (6,2) rectangle (7,3);
\draw[fill=gray] (7,2) rectangle (8,3);
\draw[fill=gray] (8,2) rectangle (9,3);
\draw[fill=gray] (10,2) rectangle (11,3);
\draw[fill=gray] (0,3) rectangle (1,4);
\draw[fill=gray] (5,3) rectangle (6,4);
\draw[fill=gray] (10,3) rectangle (11,4);
\draw[fill=gray] (0,4) rectangle (1,5);
\draw[fill=gray] (5,4) rectangle (6,5);
\draw[fill=gray] (10,4) rectangle (11,5);
\draw[fill=gray] (0,5) rectangle (1,6);
\draw[fill=gray] (5,5) rectangle (6,6);
\draw[fill=gray] (10,5) rectangle (11,6);
\draw[fill=gray] (0,6) rectangle (1,7);
\draw[fill=gray] (2,6) rectangle (3,7);
\draw[fill=gray] (3,6) rectangle (4,7);
\draw[fill=gray] (4,6) rectangle (5,7);
\draw[fill=gray] (5,6) rectangle (6,7);
\draw[fill=gray] (10,6) rectangle (11,7);
\draw[fill=gray] (0,7) rectangle (1,8);
\draw[fill=gray] (10,7) rectangle (11,8);
\draw[fill=gray] (0,8) rectangle (1,9);
\draw[fill=gray] (2,8) rectangle (3,9);
\draw[fill=gray] (3,8) rectangle (4,9);
\draw[fill=gray] (4,8) rectangle (5,9);
\draw[fill=gray] (5,8) rectangle (6,9);
\draw[fill=gray] (6,8) rectangle (7,9);
\draw[fill=gray] (7,8) rectangle (8,9);
\draw[fill=gray] (8,8) rectangle (9,9);
\draw[fill=gray] (9,8) rectangle (10,9);
\draw[fill=gray] (10,8) rectangle (11,9);
\draw[fill=gray] (0,9) rectangle (1,10);
\draw[line width=1mm] (1.500000,9.500000) circle (0.300000);
\draw[fill=gray] (2,9) rectangle (3,10);
\draw[fill=gray] (3,9) rectangle (4,10);
\draw[fill=gray] (4,9) rectangle (5,10);
\draw[fill=gray] (5,9) rectangle (6,10);
\draw[fill=gray] (6,9) rectangle (7,10);
\draw[fill=gray] (7,9) rectangle (8,10);
\draw[fill=gray] (8,9) rectangle (9,10);
\draw[fill=gray] (9,9) rectangle (10,10);
\draw[fill=gray] (10,9) rectangle (11,10);
\draw[fill=gray] (0,10) rectangle (1,11);
\draw[fill=gray] (1,10) rectangle (2,11);
\draw[fill=gray] (2,10) rectangle (3,11);
\draw[fill=gray] (3,10) rectangle (4,11);
\draw[fill=gray] (4,10) rectangle (5,11);
\draw[fill=gray] (5,10) rectangle (6,11);
\draw[fill=gray] (6,10) rectangle (7,11);
\draw[fill=gray] (7,10) rectangle (8,11);
\draw[fill=gray] (8,10) rectangle (9,11);
\draw[fill=gray] (9,10) rectangle (10,11);
\draw[fill=gray] (10,10) rectangle (11,11);
\node [align=center] at(-0.500000,0.500000) {0};
\node [align=center] at(-0.500000,1.500000) {1};
\node [align=center] at(-0.500000,2.500000) {2};
\node [align=center] at(-0.500000,3.500000) {3};
\node [align=center] at(-0.500000,4.500000) {4};
\node [align=center] at(-0.500000,5.500000) {5};
\node [align=center] at(-0.500000,6.500000) {6};
\node [align=center] at(-0.500000,7.500000) {7};
\node [align=center] at(-0.500000,8.500000) {8};
\node [align=center] at(-0.500000,9.500000) {9};
\node [align=center] at(-0.500000,10.500000) {10};
\node [align=center] at(0.500000,-0.500000) {A};
\node [align=center] at(1.500000,-0.500000) {B};
\node [align=center] at(2.500000,-0.500000) {C};
\node [align=center] at(3.500000,-0.500000) {D};
\node [align=center] at(4.500000,-0.500000) {E};
\node [align=center] at(5.500000,-0.500000) {F};
\node [align=center] at(6.500000,-0.500000) {G};
\node [align=center] at(7.500000,-0.500000) {H};
\node [align=center] at(8.500000,-0.500000) {I};
\node [align=center] at(9.500000,-0.500000) {J};
\node [align=center] at(10.500000,-0.500000) {K};
\draw[-Latex,draw=black!100] (1.500000,9.500000)--(1.500000,9.100000);
\draw[-Latex,draw=black!100] (1.500000,8.500000)--(1.500000,8.100000);
\draw[-Latex,draw=black!100] (1.500000,7.500000)--(1.900000,7.500000);
\draw[-Latex,draw=black!100] (1.500000,7.500000)--(1.500000,7.100000);
\draw[-Latex,draw=black!100] (2.500000,7.500000)--(2.900000,7.500000);
\draw[-Latex,draw=black!100] (3.500000,7.500000)--(3.900000,7.500000);
\draw[-Latex,draw=black!100] (4.500000,7.500000)--(4.900000,7.500000);
\draw[-Latex,draw=black!100] (5.500000,7.500000)--(5.900000,7.500000);
\draw[-Latex,draw=black!100] (6.500000,7.500000)--(6.900000,7.500000);
\draw[-Latex,draw=black!100] (6.500000,7.500000)--(6.500000,7.100000);
\draw[-Latex,draw=black!100] (7.500000,7.500000)--(7.900000,7.500000);
\draw[-Latex,draw=black!100] (7.500000,7.500000)--(7.500000,7.100000);
\draw[-Latex,draw=black!100] (8.500000,7.500000)--(8.900000,7.500000);
\draw[-Latex,draw=black!100] (8.500000,7.500000)--(8.500000,7.100000);
\draw[-Latex,draw=black!100] (9.500000,7.500000)--(9.500000,7.100000);
\draw[-Latex,draw=black!100] (1.500000,6.500000)--(1.500000,6.100000);
\draw[-Latex,draw=black!100] (6.500000,6.500000)--(6.900000,6.500000);
\draw[-Latex,draw=black!100] (6.500000,6.500000)--(6.500000,6.100000);
\draw[-Latex,draw=black!100] (7.500000,6.500000)--(7.900000,6.500000);
\draw[-Latex,draw=black!100] (7.500000,6.500000)--(7.500000,6.100000);
\draw[-Latex,draw=black!100] (8.500000,6.500000)--(8.900000,6.500000);
\draw[-Latex,draw=black!100] (8.500000,6.500000)--(8.500000,6.100000);
\draw[-Latex,draw=black!100] (9.500000,6.500000)--(9.500000,6.100000);
\draw[-Latex,draw=black!100] (1.500000,5.500000)--(1.900000,5.500000);
\draw[-Latex,draw=black!100] (1.500000,5.500000)--(1.500000,5.100000);
\draw[-Latex,draw=black!100] (2.500000,5.500000)--(2.900000,5.500000);
\draw[-Latex,draw=black!100] (2.500000,5.500000)--(2.500000,5.100000);
\draw[-Latex,draw=black!100] (3.500000,5.500000)--(3.900000,5.500000);
\draw[-Latex,draw=black!100] (3.500000,5.500000)--(3.500000,5.100000);
\draw[-Latex,draw=black!100] (4.500000,5.500000)--(4.500000,5.100000);
\draw[-Latex,draw=black!100] (6.500000,5.500000)--(6.900000,5.500000);
\draw[-Latex,draw=black!100] (6.500000,5.500000)--(6.500000,5.100000);
\draw[-Latex,draw=black!100] (7.500000,5.500000)--(7.900000,5.500000);
\draw[-Latex,draw=black!100] (7.500000,5.500000)--(7.500000,5.100000);
\draw[-Latex,draw=black!100] (8.500000,5.500000)--(8.900000,5.500000);
\draw[-Latex,draw=black!100] (8.500000,5.500000)--(8.500000,5.100000);
\draw[-Latex,draw=black!100] (9.500000,5.500000)--(9.500000,5.100000);
\draw[-Latex,draw=black!100] (1.500000,4.500000)--(1.900000,4.500000);
\draw[-Latex,draw=black!100] (1.500000,4.500000)--(1.500000,4.100000);
\draw[-Latex,draw=black!100] (2.500000,4.500000)--(2.900000,4.500000);
\draw[-Latex,draw=black!100] (2.500000,4.500000)--(2.500000,4.100000);
\draw[-Latex,draw=black!100] (3.500000,4.500000)--(3.900000,4.500000);
\draw[-Latex,draw=black!100] (3.500000,4.500000)--(3.500000,4.100000);
\draw[-Latex,draw=black!100] (4.500000,4.500000)--(4.500000,4.100000);
\draw[-Latex,draw=black!100] (6.500000,4.500000)--(6.900000,4.500000);
\draw[-Latex,draw=black!100] (6.500000,4.500000)--(6.500000,4.100000);
\draw[-Latex,draw=black!100] (7.500000,4.500000)--(7.900000,4.500000);
\draw[-Latex,draw=black!100] (7.500000,4.500000)--(7.500000,4.100000);
\draw[-Latex,draw=black!100] (8.500000,4.500000)--(8.900000,4.500000);
\draw[-Latex,draw=black!100] (8.500000,4.500000)--(8.500000,4.100000);
\draw[-Latex,draw=black!100] (9.500000,4.500000)--(9.500000,4.100000);
\draw[-Latex,draw=black!100] (1.500000,3.500000)--(1.900000,3.500000);
\draw[-Latex,draw=black!100] (1.500000,3.500000)--(1.500000,3.100000);
\draw[-Latex,draw=black!100] (2.500000,3.500000)--(2.900000,3.500000);
\draw[-Latex,draw=black!100] (2.500000,3.500000)--(2.500000,3.100000);
\draw[-Latex,draw=black!100] (3.500000,3.500000)--(3.900000,3.500000);
\draw[-Latex,draw=black!100] (3.500000,3.500000)--(3.500000,3.100000);
\draw[-Latex,draw=black!100] (4.500000,3.500000)--(4.500000,3.100000);
\draw[-Latex,draw=black!100] (6.500000,3.500000)--(6.900000,3.500000);
\draw[-Latex,draw=black!100] (7.500000,3.500000)--(7.900000,3.500000);
\draw[-Latex,draw=black!100] (8.500000,3.500000)--(8.900000,3.500000);
\draw[-Latex,draw=black!100] (9.500000,3.500000)--(9.500000,3.100000);
\draw[-Latex,draw=black!100] (1.500000,2.500000)--(1.900000,2.500000);
\draw[-Latex,draw=black!100] (1.500000,2.500000)--(1.500000,2.100000);
\draw[-Latex,draw=black!100] (2.500000,2.500000)--(2.900000,2.500000);
\draw[-Latex,draw=black!100] (2.500000,2.500000)--(2.500000,2.100000);
\draw[-Latex,draw=black!100] (3.500000,2.500000)--(3.900000,2.500000);
\draw[-Latex,draw=black!100] (3.500000,2.500000)--(3.500000,2.100000);
\draw[-Latex,draw=black!100] (4.500000,2.500000)--(4.500000,2.100000);
\draw[-Latex,draw=black!100] (9.500000,2.500000)--(9.500000,2.100000);
\draw[-Latex,draw=black!100] (1.500000,1.500000)--(1.900000,1.500000);
\draw[-Latex,draw=black!100] (2.500000,1.500000)--(2.900000,1.500000);
\draw[-Latex,draw=black!100] (3.500000,1.500000)--(3.900000,1.500000);
\draw[-Latex,draw=black!100] (4.500000,1.500000)--(4.900000,1.500000);
\draw[-Latex,draw=black!100] (5.500000,1.500000)--(5.900000,1.500000);
\draw[-Latex,draw=black!100] (6.500000,1.500000)--(6.900000,1.500000);
\draw[-Latex,draw=black!100] (7.500000,1.500000)--(7.900000,1.500000);
\draw[-Latex,draw=black!100] (8.500000,1.500000)--(8.900000,1.500000);
\draw[-Latex,draw=black!100] (9.500000,1.500000)--(9.100000,1.500000);
\draw[-Latex,draw=black!100] (9.500000,1.500000)--(9.500000,1.900000);
\draw[-Latex,draw=black!100] (9.500000,1.500000)--(9.900000,1.500000);
\draw[-Latex,draw=black!100] (9.500000,1.500000)--(9.500000,1.100000);
\end{tikzpicture}
      }
    \end{minipage}
    \hspace{-.10cm}{\small $M_{3}$}\hspace{-.15cm}
    \begin{minipage}{\mpwidth}
      \centering
      \adjustbox{width=\aboxwidth}{
        \begin{tikzpicture}
\draw (0,0) grid (11,11);
\draw[fill=gray] (0,0) rectangle (1,1);
\draw[fill=gray] (1,0) rectangle (2,1);
\draw[fill=gray] (2,0) rectangle (3,1);
\draw[fill=gray] (3,0) rectangle (4,1);
\draw[fill=gray] (4,0) rectangle (5,1);
\draw[fill=gray] (5,0) rectangle (6,1);
\draw[fill=gray] (6,0) rectangle (7,1);
\draw[fill=gray] (7,0) rectangle (8,1);
\draw[fill=gray] (8,0) rectangle (9,1);
\draw[fill=gray] (9,0) rectangle (10,1);
\draw[fill=gray] (10,0) rectangle (11,1);
\draw[fill=gray] (0,1) rectangle (1,2);
\draw[fill=pink] (9.500000,1.500000) circle (0.500000);
\draw[fill=gray] (10,1) rectangle (11,2);
\draw[fill=gray] (0,2) rectangle (1,3);
\draw[fill=gray] (5,2) rectangle (6,3);
\draw[fill=gray] (6,2) rectangle (7,3);
\draw[fill=gray] (7,2) rectangle (8,3);
\draw[fill=gray] (8,2) rectangle (9,3);
\draw[fill=gray] (10,2) rectangle (11,3);
\draw[fill=gray] (0,3) rectangle (1,4);
\draw[fill=gray] (5,3) rectangle (6,4);
\draw[fill=gray] (10,3) rectangle (11,4);
\draw[fill=gray] (0,4) rectangle (1,5);
\draw[fill=gray] (5,4) rectangle (6,5);
\draw[fill=gray] (10,4) rectangle (11,5);
\draw[fill=gray] (0,5) rectangle (1,6);
\draw[fill=gray] (5,5) rectangle (6,6);
\draw[fill=gray] (10,5) rectangle (11,6);
\draw[fill=gray] (0,6) rectangle (1,7);
\draw[fill=gray] (2,6) rectangle (3,7);
\draw[fill=gray] (3,6) rectangle (4,7);
\draw[fill=gray] (4,6) rectangle (5,7);
\draw[fill=gray] (5,6) rectangle (6,7);
\draw[fill=gray] (10,6) rectangle (11,7);
\draw[fill=gray] (0,7) rectangle (1,8);
\draw[fill=gray] (10,7) rectangle (11,8);
\draw[fill=gray] (0,8) rectangle (1,9);
\draw[fill=gray] (2,8) rectangle (3,9);
\draw[fill=gray] (3,8) rectangle (4,9);
\draw[fill=gray] (4,8) rectangle (5,9);
\draw[fill=gray] (5,8) rectangle (6,9);
\draw[fill=gray] (6,8) rectangle (7,9);
\draw[fill=gray] (7,8) rectangle (8,9);
\draw[fill=gray] (8,8) rectangle (9,9);
\draw[fill=gray] (9,8) rectangle (10,9);
\draw[fill=gray] (10,8) rectangle (11,9);
\draw[fill=gray] (0,9) rectangle (1,10);
\draw[line width=1mm] (1.500000,9.500000) circle (0.300000);
\draw[fill=gray] (2,9) rectangle (3,10);
\draw[fill=gray] (3,9) rectangle (4,10);
\draw[fill=gray] (4,9) rectangle (5,10);
\draw[fill=gray] (5,9) rectangle (6,10);
\draw[fill=gray] (6,9) rectangle (7,10);
\draw[fill=gray] (7,9) rectangle (8,10);
\draw[fill=gray] (8,9) rectangle (9,10);
\draw[fill=gray] (9,9) rectangle (10,10);
\draw[fill=gray] (10,9) rectangle (11,10);
\draw[fill=gray] (0,10) rectangle (1,11);
\draw[fill=gray] (1,10) rectangle (2,11);
\draw[fill=gray] (2,10) rectangle (3,11);
\draw[fill=gray] (3,10) rectangle (4,11);
\draw[fill=gray] (4,10) rectangle (5,11);
\draw[fill=gray] (5,10) rectangle (6,11);
\draw[fill=gray] (6,10) rectangle (7,11);
\draw[fill=gray] (7,10) rectangle (8,11);
\draw[fill=gray] (8,10) rectangle (9,11);
\draw[fill=gray] (9,10) rectangle (10,11);
\draw[fill=gray] (10,10) rectangle (11,11);
\node [align=center] at(-0.500000,0.500000) {0};
\node [align=center] at(-0.500000,1.500000) {1};
\node [align=center] at(-0.500000,2.500000) {2};
\node [align=center] at(-0.500000,3.500000) {3};
\node [align=center] at(-0.500000,4.500000) {4};
\node [align=center] at(-0.500000,5.500000) {5};
\node [align=center] at(-0.500000,6.500000) {6};
\node [align=center] at(-0.500000,7.500000) {7};
\node [align=center] at(-0.500000,8.500000) {8};
\node [align=center] at(-0.500000,9.500000) {9};
\node [align=center] at(-0.500000,10.500000) {10};
\node [align=center] at(0.500000,-0.500000) {A};
\node [align=center] at(1.500000,-0.500000) {B};
\node [align=center] at(2.500000,-0.500000) {C};
\node [align=center] at(3.500000,-0.500000) {D};
\node [align=center] at(4.500000,-0.500000) {E};
\node [align=center] at(5.500000,-0.500000) {F};
\node [align=center] at(6.500000,-0.500000) {G};
\node [align=center] at(7.500000,-0.500000) {H};
\node [align=center] at(8.500000,-0.500000) {I};
\node [align=center] at(9.500000,-0.500000) {J};
\node [align=center] at(10.500000,-0.500000) {K};
\draw[-Latex,draw=black!100] (1.500000,9.500000)--(1.500000,9.100000);
\draw[-Latex,draw=black!100] (1.500000,8.500000)--(1.500000,8.100000);
\draw[-Latex,draw=black!100] (1.500000,7.500000)--(1.900000,7.500000);
\draw[-Latex,draw=black!100] (1.500000,7.500000)--(1.500000,7.100000);
\draw[-Latex,draw=black!100] (2.500000,7.500000)--(2.900000,7.500000);
\draw[-Latex,draw=black!100] (3.500000,7.500000)--(3.900000,7.500000);
\draw[-Latex,draw=black!100] (4.500000,7.500000)--(4.900000,7.500000);
\draw[-Latex,draw=black!100] (5.500000,7.500000)--(5.900000,7.500000);
\draw[-Latex,draw=black!100] (6.500000,7.500000)--(6.900000,7.500000);
\draw[-Latex,draw=black!100] (7.500000,7.500000)--(7.900000,7.500000);
\draw[-Latex,draw=black!100] (8.500000,7.500000)--(8.900000,7.500000);
\draw[-Latex,draw=black!100] (9.500000,7.500000)--(9.500000,7.100000);
\draw[-Latex,draw=black!100] (1.500000,6.500000)--(1.500000,6.100000);
\draw[-Latex,draw=black!100] (6.500000,6.500000)--(6.900000,6.500000);
\draw[-Latex,draw=black!100] (6.500000,6.500000)--(6.500000,6.100000);
\draw[-Latex,draw=black!100] (7.500000,6.500000)--(7.900000,6.500000);
\draw[-Latex,draw=black!100] (8.500000,6.500000)--(8.900000,6.500000);
\draw[-Latex,draw=black!100] (9.500000,6.500000)--(9.500000,6.100000);
\draw[-Latex,draw=black!100] (1.500000,5.500000)--(1.900000,5.500000);
\draw[-Latex,draw=black!100] (2.500000,5.500000)--(2.900000,5.500000);
\draw[-Latex,draw=black!100] (3.500000,5.500000)--(3.900000,5.500000);
\draw[-Latex,draw=black!100] (4.500000,5.500000)--(4.500000,5.100000);
\draw[-Latex,draw=black!100] (6.500000,5.500000)--(6.500000,5.100000);
\draw[-Latex,draw=black!100] (7.500000,5.500000)--(7.900000,5.500000);
\draw[-Latex,draw=black!100] (7.500000,5.500000)--(7.500000,5.100000);
\draw[-Latex,draw=black!100] (8.500000,5.500000)--(8.900000,5.500000);
\draw[-Latex,draw=black!100] (9.500000,5.500000)--(9.500000,5.100000);
\draw[-Latex,draw=black!100] (1.500000,4.500000)--(1.900000,4.500000);
\draw[-Latex,draw=black!100] (1.500000,4.500000)--(1.500000,4.100000);
\draw[-Latex,draw=black!100] (2.500000,4.500000)--(2.900000,4.500000);
\draw[-Latex,draw=black!100] (3.500000,4.500000)--(3.900000,4.500000);
\draw[-Latex,draw=black!100] (4.500000,4.500000)--(4.500000,4.100000);
\draw[-Latex,draw=black!100] (6.500000,4.500000)--(6.500000,4.100000);
\draw[-Latex,draw=black!100] (7.500000,4.500000)--(7.500000,4.100000);
\draw[-Latex,draw=black!100] (8.500000,4.500000)--(8.900000,4.500000);
\draw[-Latex,draw=black!100] (8.500000,4.500000)--(8.500000,4.100000);
\draw[-Latex,draw=black!100] (9.500000,4.500000)--(9.500000,4.100000);
\draw[-Latex,draw=black!100] (1.500000,3.500000)--(1.500000,3.100000);
\draw[-Latex,draw=black!100] (2.500000,3.500000)--(2.900000,3.500000);
\draw[-Latex,draw=black!100] (2.500000,3.500000)--(2.500000,3.100000);
\draw[-Latex,draw=black!100] (3.500000,3.500000)--(3.900000,3.500000);
\draw[-Latex,draw=black!100] (4.500000,3.500000)--(4.500000,3.100000);
\draw[-Latex,draw=black!100] (6.500000,3.500000)--(6.900000,3.500000);
\draw[-Latex,draw=black!100] (7.500000,3.500000)--(7.900000,3.500000);
\draw[-Latex,draw=black!100] (8.500000,3.500000)--(8.900000,3.500000);
\draw[-Latex,draw=black!100] (9.500000,3.500000)--(9.500000,3.100000);
\draw[-Latex,draw=black!100] (1.500000,2.500000)--(1.500000,2.100000);
\draw[-Latex,draw=black!100] (2.500000,2.500000)--(2.500000,2.100000);
\draw[-Latex,draw=black!100] (3.500000,2.500000)--(3.900000,2.500000);
\draw[-Latex,draw=black!100] (3.500000,2.500000)--(3.500000,2.100000);
\draw[-Latex,draw=black!100] (4.500000,2.500000)--(4.500000,2.100000);
\draw[-Latex,draw=black!100] (9.500000,2.500000)--(9.500000,2.100000);
\draw[-Latex,draw=black!100] (1.500000,1.500000)--(1.900000,1.500000);
\draw[-Latex,draw=black!100] (2.500000,1.500000)--(2.900000,1.500000);
\draw[-Latex,draw=black!100] (3.500000,1.500000)--(3.900000,1.500000);
\draw[-Latex,draw=black!100] (4.500000,1.500000)--(4.900000,1.500000);
\draw[-Latex,draw=black!100] (5.500000,1.500000)--(5.900000,1.500000);
\draw[-Latex,draw=black!100] (6.500000,1.500000)--(6.900000,1.500000);
\draw[-Latex,draw=black!100] (7.500000,1.500000)--(7.900000,1.500000);
\draw[-Latex,draw=black!100] (8.500000,1.500000)--(8.900000,1.500000);
\draw[-Latex,draw=black!100] (9.500000,1.500000)--(9.100000,1.500000);
\draw[-Latex,draw=black!100] (9.500000,1.500000)--(9.500000,1.900000);
\draw[-Latex,draw=black!100] (9.500000,1.500000)--(9.900000,1.500000);
\draw[-Latex,draw=black!100] (9.500000,1.500000)--(9.500000,1.100000);
\end{tikzpicture} }
    \end{minipage}
  
  \hfill
  
    \begin{minipage}{\mpwidth}
      \centering
      \adjustbox{width=\aboxwidth}{
        \begin{tikzpicture}
\draw (0,0) grid (11,11);
\draw[fill=gray] (0,0) rectangle (1,1);
\draw[fill=gray] (1,0) rectangle (2,1);
\draw[fill=gray] (2,0) rectangle (3,1);
\draw[fill=gray] (3,0) rectangle (4,1);
\draw[fill=gray] (4,0) rectangle (5,1);
\draw[fill=gray] (5,0) rectangle (6,1);
\draw[fill=gray] (6,0) rectangle (7,1);
\draw[fill=gray] (7,0) rectangle (8,1);
\draw[fill=gray] (8,0) rectangle (9,1);
\draw[fill=gray] (9,0) rectangle (10,1);
\draw[fill=gray] (10,0) rectangle (11,1);
\draw[fill=gray] (0,1) rectangle (1,2);
\draw[fill=pink] (9.500000,1.500000) circle (0.500000);
\draw[fill=gray] (10,1) rectangle (11,2);
\draw[fill=gray] (0,2) rectangle (1,3);
\draw[fill=gray] (5,2) rectangle (6,3);
\draw[fill=gray] (6,2) rectangle (7,3);
\draw[fill=gray] (7,2) rectangle (8,3);
\draw[fill=gray] (8,2) rectangle (9,3);
\draw[fill=gray] (10,2) rectangle (11,3);
\draw[fill=gray] (0,3) rectangle (1,4);
\draw[fill=gray] (5,3) rectangle (6,4);
\draw[fill=gray] (10,3) rectangle (11,4);
\draw[fill=gray] (0,4) rectangle (1,5);
\draw[fill=gray] (10,4) rectangle (11,5);
\draw[fill=gray] (0,5) rectangle (1,6);
\draw[fill=gray] (5,5) rectangle (6,6);
\draw[fill=gray] (10,5) rectangle (11,6);
\draw[fill=gray] (0,6) rectangle (1,7);
\draw[fill=gray] (2,6) rectangle (3,7);
\draw[fill=gray] (3,6) rectangle (4,7);
\draw[fill=gray] (4,6) rectangle (5,7);
\draw[fill=gray] (5,6) rectangle (6,7);
\draw[fill=gray] (10,6) rectangle (11,7);
\draw[fill=gray] (0,7) rectangle (1,8);
\draw[fill=gray] (10,7) rectangle (11,8);
\draw[fill=gray] (0,8) rectangle (1,9);
\draw[fill=gray] (2,8) rectangle (3,9);
\draw[fill=gray] (3,8) rectangle (4,9);
\draw[fill=gray] (4,8) rectangle (5,9);
\draw[fill=gray] (5,8) rectangle (6,9);
\draw[fill=gray] (6,8) rectangle (7,9);
\draw[fill=gray] (7,8) rectangle (8,9);
\draw[fill=gray] (8,8) rectangle (9,9);
\draw[fill=gray] (9,8) rectangle (10,9);
\draw[fill=gray] (10,8) rectangle (11,9);
\draw[fill=gray] (0,9) rectangle (1,10);
\draw[line width=1mm] (1.500000,9.500000) circle (0.300000);
\draw[fill=gray] (2,9) rectangle (3,10);
\draw[fill=gray] (3,9) rectangle (4,10);
\draw[fill=gray] (4,9) rectangle (5,10);
\draw[fill=gray] (5,9) rectangle (6,10);
\draw[fill=gray] (6,9) rectangle (7,10);
\draw[fill=gray] (7,9) rectangle (8,10);
\draw[fill=gray] (8,9) rectangle (9,10);
\draw[fill=gray] (9,9) rectangle (10,10);
\draw[fill=gray] (10,9) rectangle (11,10);
\draw[fill=gray] (0,10) rectangle (1,11);
\draw[fill=gray] (1,10) rectangle (2,11);
\draw[fill=gray] (2,10) rectangle (3,11);
\draw[fill=gray] (3,10) rectangle (4,11);
\draw[fill=gray] (4,10) rectangle (5,11);
\draw[fill=gray] (5,10) rectangle (6,11);
\draw[fill=gray] (6,10) rectangle (7,11);
\draw[fill=gray] (7,10) rectangle (8,11);
\draw[fill=gray] (8,10) rectangle (9,11);
\draw[fill=gray] (9,10) rectangle (10,11);
\draw[fill=gray] (10,10) rectangle (11,11);
\node [align=center] at(-0.500000,0.500000) {0};
\node [align=center] at(-0.500000,1.500000) {1};
\node [align=center] at(-0.500000,2.500000) {2};
\node [align=center] at(-0.500000,3.500000) {3};
\node [align=center] at(-0.500000,4.500000) {4};
\node [align=center] at(-0.500000,5.500000) {5};
\node [align=center] at(-0.500000,6.500000) {6};
\node [align=center] at(-0.500000,7.500000) {7};
\node [align=center] at(-0.500000,8.500000) {8};
\node [align=center] at(-0.500000,9.500000) {9};
\node [align=center] at(-0.500000,10.500000) {10};
\node [align=center] at(0.500000,-0.500000) {A};
\node [align=center] at(1.500000,-0.500000) {B};
\node [align=center] at(2.500000,-0.500000) {C};
\node [align=center] at(3.500000,-0.500000) {D};
\node [align=center] at(4.500000,-0.500000) {E};
\node [align=center] at(5.500000,-0.500000) {F};
\node [align=center] at(6.500000,-0.500000) {G};
\node [align=center] at(7.500000,-0.500000) {H};
\node [align=center] at(8.500000,-0.500000) {I};
\node [align=center] at(9.500000,-0.500000) {J};
\node [align=center] at(10.500000,-0.500000) {K};
\draw[-Latex,draw=black!100] (1.500000,9.500000)--(1.500000,9.100000);
\draw[-Latex,draw=black!100] (1.500000,8.500000)--(1.500000,8.100000);
\draw[-Latex,draw=black!100] (1.500000,7.500000)--(1.900000,7.500000);
\draw[-Latex,draw=black!100] (1.500000,7.500000)--(1.500000,7.100000);
\draw[-Latex,draw=black!100] (2.500000,7.500000)--(2.900000,7.500000);
\draw[-Latex,draw=black!100] (3.500000,7.500000)--(3.900000,7.500000);
\draw[-Latex,draw=black!100] (4.500000,7.500000)--(4.900000,7.500000);
\draw[-Latex,draw=black!100] (5.500000,7.500000)--(5.900000,7.500000);
\draw[-Latex,draw=black!100] (6.500000,7.500000)--(6.900000,7.500000);
\draw[-Latex,draw=black!100] (6.500000,7.500000)--(6.500000,7.100000);
\draw[-Latex,draw=black!100] (7.500000,7.500000)--(7.900000,7.500000);
\draw[-Latex,draw=black!100] (7.500000,7.500000)--(7.500000,7.100000);
\draw[-Latex,draw=black!100] (8.500000,7.500000)--(8.900000,7.500000);
\draw[-Latex,draw=black!100] (8.500000,7.500000)--(8.500000,7.100000);
\draw[-Latex,draw=black!100] (9.500000,7.500000)--(9.500000,7.100000);
\draw[-Latex,draw=black!100] (1.500000,6.500000)--(1.500000,6.100000);
\draw[-Latex,draw=black!100] (6.500000,6.500000)--(6.900000,6.500000);
\draw[-Latex,draw=black!100] (6.500000,6.500000)--(6.500000,6.100000);
\draw[-Latex,draw=black!100] (7.500000,6.500000)--(7.900000,6.500000);
\draw[-Latex,draw=black!100] (7.500000,6.500000)--(7.500000,6.100000);
\draw[-Latex,draw=black!100] (8.500000,6.500000)--(8.900000,6.500000);
\draw[-Latex,draw=black!100] (8.500000,6.500000)--(8.500000,6.100000);
\draw[-Latex,draw=black!100] (9.500000,6.500000)--(9.500000,6.100000);
\draw[-Latex,draw=black!100] (1.500000,5.500000)--(1.900000,5.500000);
\draw[-Latex,draw=black!100] (1.500000,5.500000)--(1.500000,5.100000);
\draw[-Latex,draw=black!100] (2.500000,5.500000)--(2.900000,5.500000);
\draw[-Latex,draw=black!100] (2.500000,5.500000)--(2.500000,5.100000);
\draw[-Latex,draw=black!100] (3.500000,5.500000)--(3.900000,5.500000);
\draw[-Latex,draw=black!100] (3.500000,5.500000)--(3.500000,5.100000);
\draw[-Latex,draw=black!100] (4.500000,5.500000)--(4.500000,5.100000);
\draw[-Latex,draw=black!100] (6.500000,5.500000)--(6.900000,5.500000);
\draw[-Latex,draw=black!100] (6.500000,5.500000)--(6.500000,5.100000);
\draw[-Latex,draw=black!100] (7.500000,5.500000)--(7.900000,5.500000);
\draw[-Latex,draw=black!100] (7.500000,5.500000)--(7.500000,5.100000);
\draw[-Latex,draw=black!100] (8.500000,5.500000)--(8.900000,5.500000);
\draw[-Latex,draw=black!100] (8.500000,5.500000)--(8.500000,5.100000);
\draw[-Latex,draw=black!100] (9.500000,5.500000)--(9.500000,5.100000);
\draw[-Latex,draw=black!100] (1.500000,4.500000)--(1.900000,4.500000);
\draw[-Latex,draw=black!100] (1.500000,4.500000)--(1.500000,4.100000);
\draw[-Latex,draw=black!100] (2.500000,4.500000)--(2.900000,4.500000);
\draw[-Latex,draw=black!100] (2.500000,4.500000)--(2.500000,4.100000);
\draw[-Latex,draw=black!100] (3.500000,4.500000)--(3.900000,4.500000);
\draw[-Latex,draw=black!100] (3.500000,4.500000)--(3.500000,4.100000);
\draw[-Latex,draw=black!100] (4.500000,4.500000)--(4.900000,4.500000);
\draw[-Latex,draw=black!100] (4.500000,4.500000)--(4.500000,4.100000);
\draw[-Latex,draw=black!100] (5.500000,4.500000)--(5.900000,4.500000);
\draw[-Latex,draw=black!100] (6.500000,4.500000)--(6.900000,4.500000);
\draw[-Latex,draw=black!100] (6.500000,4.500000)--(6.500000,4.100000);
\draw[-Latex,draw=black!100] (7.500000,4.500000)--(7.900000,4.500000);
\draw[-Latex,draw=black!100] (7.500000,4.500000)--(7.500000,4.100000);
\draw[-Latex,draw=black!100] (8.500000,4.500000)--(8.900000,4.500000);
\draw[-Latex,draw=black!100] (8.500000,4.500000)--(8.500000,4.100000);
\draw[-Latex,draw=black!100] (9.500000,4.500000)--(9.500000,4.100000);
\draw[-Latex,draw=black!100] (1.500000,3.500000)--(1.900000,3.500000);
\draw[-Latex,draw=black!100] (1.500000,3.500000)--(1.500000,3.100000);
\draw[-Latex,draw=black!100] (2.500000,3.500000)--(2.900000,3.500000);
\draw[-Latex,draw=black!100] (2.500000,3.500000)--(2.500000,3.100000);
\draw[-Latex,draw=black!100] (3.500000,3.500000)--(3.900000,3.500000);
\draw[-Latex,draw=black!100] (3.500000,3.500000)--(3.500000,3.100000);
\draw[-Latex,draw=black!100] (4.500000,3.500000)--(4.500000,3.100000);
\draw[-Latex,draw=black!100] (6.500000,3.500000)--(6.900000,3.500000);
\draw[-Latex,draw=black!100] (7.500000,3.500000)--(7.900000,3.500000);
\draw[-Latex,draw=black!100] (8.500000,3.500000)--(8.900000,3.500000);
\draw[-Latex,draw=black!100] (9.500000,3.500000)--(9.500000,3.100000);
\draw[-Latex,draw=black!100] (1.500000,2.500000)--(1.900000,2.500000);
\draw[-Latex,draw=black!100] (1.500000,2.500000)--(1.500000,2.100000);
\draw[-Latex,draw=black!100] (2.500000,2.500000)--(2.900000,2.500000);
\draw[-Latex,draw=black!100] (2.500000,2.500000)--(2.500000,2.100000);
\draw[-Latex,draw=black!100] (3.500000,2.500000)--(3.900000,2.500000);
\draw[-Latex,draw=black!100] (3.500000,2.500000)--(3.500000,2.100000);
\draw[-Latex,draw=black!100] (4.500000,2.500000)--(4.500000,2.100000);
\draw[-Latex,draw=black!100] (9.500000,2.500000)--(9.500000,2.100000);
\draw[-Latex,draw=black!100] (1.500000,1.500000)--(1.900000,1.500000);
\draw[-Latex,draw=black!100] (2.500000,1.500000)--(2.900000,1.500000);
\draw[-Latex,draw=black!100] (3.500000,1.500000)--(3.900000,1.500000);
\draw[-Latex,draw=black!100] (4.500000,1.500000)--(4.900000,1.500000);
\draw[-Latex,draw=black!100] (5.500000,1.500000)--(5.900000,1.500000);
\draw[-Latex,draw=black!100] (6.500000,1.500000)--(6.900000,1.500000);
\draw[-Latex,draw=black!100] (7.500000,1.500000)--(7.900000,1.500000);
\draw[-Latex,draw=black!100] (8.500000,1.500000)--(8.900000,1.500000);
\draw[-Latex,draw=black!100] (9.500000,1.500000)--(9.100000,1.500000);
\draw[-Latex,draw=black!100] (9.500000,1.500000)--(9.500000,1.900000);
\draw[-Latex,draw=black!100] (9.500000,1.500000)--(9.900000,1.500000);
\draw[-Latex,draw=black!100] (9.500000,1.500000)--(9.500000,1.100000);
\end{tikzpicture}
      }
    \end{minipage}
    \hspace{-.10cm}{\small $M_{4}$}\hspace{-.15cm}
    \begin{minipage}{\mpwidth}
      \centering
      \adjustbox{width=\aboxwidth}{
        \begin{tikzpicture}
\draw (0,0) grid (11,11);
\draw[fill=gray] (0,0) rectangle (1,1);
\draw[fill=gray] (1,0) rectangle (2,1);
\draw[fill=gray] (2,0) rectangle (3,1);
\draw[fill=gray] (3,0) rectangle (4,1);
\draw[fill=gray] (4,0) rectangle (5,1);
\draw[fill=gray] (5,0) rectangle (6,1);
\draw[fill=gray] (6,0) rectangle (7,1);
\draw[fill=gray] (7,0) rectangle (8,1);
\draw[fill=gray] (8,0) rectangle (9,1);
\draw[fill=gray] (9,0) rectangle (10,1);
\draw[fill=gray] (10,0) rectangle (11,1);
\draw[fill=gray] (0,1) rectangle (1,2);
\draw[fill=pink] (9.500000,1.500000) circle (0.500000);
\draw[fill=gray] (10,1) rectangle (11,2);
\draw[fill=gray] (0,2) rectangle (1,3);
\draw[fill=gray] (5,2) rectangle (6,3);
\draw[fill=gray] (6,2) rectangle (7,3);
\draw[fill=gray] (7,2) rectangle (8,3);
\draw[fill=gray] (8,2) rectangle (9,3);
\draw[fill=gray] (10,2) rectangle (11,3);
\draw[fill=gray] (0,3) rectangle (1,4);
\draw[fill=gray] (5,3) rectangle (6,4);
\draw[fill=gray] (10,3) rectangle (11,4);
\draw[fill=gray] (0,4) rectangle (1,5);
\draw[fill=gray] (10,4) rectangle (11,5);
\draw[fill=gray] (0,5) rectangle (1,6);
\draw[fill=gray] (5,5) rectangle (6,6);
\draw[fill=gray] (10,5) rectangle (11,6);
\draw[fill=gray] (0,6) rectangle (1,7);
\draw[fill=gray] (2,6) rectangle (3,7);
\draw[fill=gray] (3,6) rectangle (4,7);
\draw[fill=gray] (4,6) rectangle (5,7);
\draw[fill=gray] (5,6) rectangle (6,7);
\draw[fill=gray] (10,6) rectangle (11,7);
\draw[fill=gray] (0,7) rectangle (1,8);
\draw[fill=gray] (10,7) rectangle (11,8);
\draw[fill=gray] (0,8) rectangle (1,9);
\draw[fill=gray] (2,8) rectangle (3,9);
\draw[fill=gray] (3,8) rectangle (4,9);
\draw[fill=gray] (4,8) rectangle (5,9);
\draw[fill=gray] (5,8) rectangle (6,9);
\draw[fill=gray] (6,8) rectangle (7,9);
\draw[fill=gray] (7,8) rectangle (8,9);
\draw[fill=gray] (8,8) rectangle (9,9);
\draw[fill=gray] (9,8) rectangle (10,9);
\draw[fill=gray] (10,8) rectangle (11,9);
\draw[fill=gray] (0,9) rectangle (1,10);
\draw[line width=1mm] (1.500000,9.500000) circle (0.300000);
\draw[fill=gray] (2,9) rectangle (3,10);
\draw[fill=gray] (3,9) rectangle (4,10);
\draw[fill=gray] (4,9) rectangle (5,10);
\draw[fill=gray] (5,9) rectangle (6,10);
\draw[fill=gray] (6,9) rectangle (7,10);
\draw[fill=gray] (7,9) rectangle (8,10);
\draw[fill=gray] (8,9) rectangle (9,10);
\draw[fill=gray] (9,9) rectangle (10,10);
\draw[fill=gray] (10,9) rectangle (11,10);
\draw[fill=gray] (0,10) rectangle (1,11);
\draw[fill=gray] (1,10) rectangle (2,11);
\draw[fill=gray] (2,10) rectangle (3,11);
\draw[fill=gray] (3,10) rectangle (4,11);
\draw[fill=gray] (4,10) rectangle (5,11);
\draw[fill=gray] (5,10) rectangle (6,11);
\draw[fill=gray] (6,10) rectangle (7,11);
\draw[fill=gray] (7,10) rectangle (8,11);
\draw[fill=gray] (8,10) rectangle (9,11);
\draw[fill=gray] (9,10) rectangle (10,11);
\draw[fill=gray] (10,10) rectangle (11,11);
\node [align=center] at(-0.500000,0.500000) {0};
\node [align=center] at(-0.500000,1.500000) {1};
\node [align=center] at(-0.500000,2.500000) {2};
\node [align=center] at(-0.500000,3.500000) {3};
\node [align=center] at(-0.500000,4.500000) {4};
\node [align=center] at(-0.500000,5.500000) {5};
\node [align=center] at(-0.500000,6.500000) {6};
\node [align=center] at(-0.500000,7.500000) {7};
\node [align=center] at(-0.500000,8.500000) {8};
\node [align=center] at(-0.500000,9.500000) {9};
\node [align=center] at(-0.500000,10.500000) {10};
\node [align=center] at(0.500000,-0.500000) {A};
\node [align=center] at(1.500000,-0.500000) {B};
\node [align=center] at(2.500000,-0.500000) {C};
\node [align=center] at(3.500000,-0.500000) {D};
\node [align=center] at(4.500000,-0.500000) {E};
\node [align=center] at(5.500000,-0.500000) {F};
\node [align=center] at(6.500000,-0.500000) {G};
\node [align=center] at(7.500000,-0.500000) {H};
\node [align=center] at(8.500000,-0.500000) {I};
\node [align=center] at(9.500000,-0.500000) {J};
\node [align=center] at(10.500000,-0.500000) {K};
\draw[-Latex,draw=black!100] (1.500000,9.500000)--(1.500000,9.100000);
\draw[-Latex,draw=black!100] (1.500000,8.500000)--(1.500000,8.100000);
\draw[-Latex,draw=black!100] (1.500000,7.500000)--(1.900000,7.500000);
\draw[-Latex,draw=black!100] (2.500000,7.500000)--(2.900000,7.500000);
\draw[-Latex,draw=black!100] (3.500000,7.500000)--(3.900000,7.500000);
\draw[-Latex,draw=black!100] (4.500000,7.500000)--(4.900000,7.500000);
\draw[-Latex,draw=black!100] (5.500000,7.500000)--(5.900000,7.500000);
\draw[-Latex,draw=black!100] (6.500000,7.500000)--(6.900000,7.500000);
\draw[-Latex,draw=black!100] (7.500000,7.500000)--(7.900000,7.500000);
\draw[-Latex,draw=black!100] (8.500000,7.500000)--(8.900000,7.500000);
\draw[-Latex,draw=black!100] (9.500000,7.500000)--(9.500000,7.100000);
\draw[-Latex,draw=black!100] (1.500000,6.500000)--(1.500000,6.100000);
\draw[-Latex,draw=black!100] (6.500000,6.500000)--(6.900000,6.500000);
\draw[-Latex,draw=black!100] (6.500000,6.500000)--(6.500000,6.100000);
\draw[-Latex,draw=black!100] (7.500000,6.500000)--(7.900000,6.500000);
\draw[-Latex,draw=black!100] (8.500000,6.500000)--(8.900000,6.500000);
\draw[-Latex,draw=black!100] (9.500000,6.500000)--(9.500000,6.100000);
\draw[-Latex,draw=black!100] (1.500000,5.500000)--(1.900000,5.500000);
\draw[-Latex,draw=black!100] (1.500000,5.500000)--(1.500000,5.100000);
\draw[-Latex,draw=black!100] (2.500000,5.500000)--(2.900000,5.500000);
\draw[-Latex,draw=black!100] (3.500000,5.500000)--(3.900000,5.500000);
\draw[-Latex,draw=black!100] (4.500000,5.500000)--(4.500000,5.100000);
\draw[-Latex,draw=black!100] (6.500000,5.500000)--(6.500000,5.100000);
\draw[-Latex,draw=black!100] (7.500000,5.500000)--(7.900000,5.500000);
\draw[-Latex,draw=black!100] (7.500000,5.500000)--(7.500000,5.100000);
\draw[-Latex,draw=black!100] (8.500000,5.500000)--(8.900000,5.500000);
\draw[-Latex,draw=black!100] (9.500000,5.500000)--(9.500000,5.100000);
\draw[-Latex,draw=black!100] (1.500000,4.500000)--(1.500000,4.100000);
\draw[-Latex,draw=black!100] (2.500000,4.500000)--(2.900000,4.500000);
\draw[-Latex,draw=black!100] (2.500000,4.500000)--(2.500000,4.100000);
\draw[-Latex,draw=black!100] (3.500000,4.500000)--(3.900000,4.500000);
\draw[-Latex,draw=black!100] (3.500000,4.500000)--(3.500000,4.100000);
\draw[-Latex,draw=black!100] (4.500000,4.500000)--(4.500000,4.100000);
\draw[-Latex,draw=black!100] (5.500000,4.500000)--(5.900000,4.500000);
\draw[-Latex,draw=black!100] (6.500000,4.500000)--(6.500000,4.100000);
\draw[-Latex,draw=black!100] (7.500000,4.500000)--(7.500000,4.100000);
\draw[-Latex,draw=black!100] (8.500000,4.500000)--(8.900000,4.500000);
\draw[-Latex,draw=black!100] (8.500000,4.500000)--(8.500000,4.100000);
\draw[-Latex,draw=black!100] (9.500000,4.500000)--(9.500000,4.100000);
\draw[-Latex,draw=black!100] (1.500000,3.500000)--(1.500000,3.100000);
\draw[-Latex,draw=black!100] (2.500000,3.500000)--(2.900000,3.500000);
\draw[-Latex,draw=black!100] (2.500000,3.500000)--(2.500000,3.100000);
\draw[-Latex,draw=black!100] (3.500000,3.500000)--(3.900000,3.500000);
\draw[-Latex,draw=black!100] (4.500000,3.500000)--(4.500000,3.100000);
\draw[-Latex,draw=black!100] (6.500000,3.500000)--(6.900000,3.500000);
\draw[-Latex,draw=black!100] (7.500000,3.500000)--(7.900000,3.500000);
\draw[-Latex,draw=black!100] (8.500000,3.500000)--(8.900000,3.500000);
\draw[-Latex,draw=black!100] (9.500000,3.500000)--(9.500000,3.100000);
\draw[-Latex,draw=black!100] (1.500000,2.500000)--(1.500000,2.100000);
\draw[-Latex,draw=black!100] (2.500000,2.500000)--(2.500000,2.100000);
\draw[-Latex,draw=black!100] (3.500000,2.500000)--(3.900000,2.500000);
\draw[-Latex,draw=black!100] (3.500000,2.500000)--(3.500000,2.100000);
\draw[-Latex,draw=black!100] (4.500000,2.500000)--(4.500000,2.100000);
\draw[-Latex,draw=black!100] (9.500000,2.500000)--(9.500000,2.100000);
\draw[-Latex,draw=black!100] (1.500000,1.500000)--(1.900000,1.500000);
\draw[-Latex,draw=black!100] (2.500000,1.500000)--(2.900000,1.500000);
\draw[-Latex,draw=black!100] (3.500000,1.500000)--(3.900000,1.500000);
\draw[-Latex,draw=black!100] (4.500000,1.500000)--(4.900000,1.500000);
\draw[-Latex,draw=black!100] (5.500000,1.500000)--(5.900000,1.500000);
\draw[-Latex,draw=black!100] (6.500000,1.500000)--(6.900000,1.500000);
\draw[-Latex,draw=black!100] (7.500000,1.500000)--(7.900000,1.500000);
\draw[-Latex,draw=black!100] (8.500000,1.500000)--(8.900000,1.500000);
\draw[-Latex,draw=black!100] (9.500000,1.500000)--(9.100000,1.500000);
\draw[-Latex,draw=black!100] (9.500000,1.500000)--(9.500000,1.900000);
\draw[-Latex,draw=black!100] (9.500000,1.500000)--(9.900000,1.500000);
\draw[-Latex,draw=black!100] (9.500000,1.500000)--(9.500000,1.100000);
\end{tikzpicture} }
    \end{minipage}

    \begin{minipage}{\mpwidth}
      \centering
      \adjustbox{width=\aboxwidth}{
        \input{expFinalVers/6_MDP_policyStoch.tikz}
      }
    \end{minipage}
    \hspace{-.10cm}{\small $M_{5}$}\hspace{-.15cm}
    \begin{minipage}{\mpwidth}
      \centering
      \adjustbox{width=\aboxwidth}{
        \begin{tikzpicture}
\draw (0,0) grid (11,12);
\draw[fill=gray] (0,0) rectangle (1,1);
\draw[fill=gray] (1,0) rectangle (2,1);
\draw[fill=gray] (2,0) rectangle (3,1);
\draw[fill=gray] (3,0) rectangle (4,1);
\draw[fill=gray] (4,0) rectangle (5,1);
\draw[fill=gray] (5,0) rectangle (6,1);
\draw[fill=gray] (6,0) rectangle (7,1);
\draw[fill=gray] (7,0) rectangle (8,1);
\draw[fill=gray] (8,0) rectangle (9,1);
\draw[fill=gray] (9,0) rectangle (10,1);
\draw[fill=gray] (10,0) rectangle (11,1);
\draw[fill=gray] (0,1) rectangle (1,2);
\draw[line width=1mm] (1.500000,1.500000) circle (0.300000);
\draw[fill=gray] (2,1) rectangle (3,2);
\draw[fill=gray] (3,1) rectangle (4,2);
\draw[fill=gray] (4,1) rectangle (5,2);
\draw[fill=gray] (5,1) rectangle (6,2);
\draw[fill=gray] (6,1) rectangle (7,2);
\draw[fill=gray] (7,1) rectangle (8,2);
\draw[fill=gray] (8,1) rectangle (9,2);
\draw[fill=gray] (9,1) rectangle (10,2);
\draw[fill=gray] (10,1) rectangle (11,2);
\draw[fill=gray] (0,2) rectangle (1,3);
\draw[fill=gray] (2,2) rectangle (3,3);
\draw[fill=gray] (3,2) rectangle (4,3);
\draw[fill=gray] (4,2) rectangle (5,3);
\draw[fill=gray] (5,2) rectangle (6,3);
\draw[fill=gray] (6,2) rectangle (7,3);
\draw[fill=gray] (7,2) rectangle (8,3);
\draw[fill=gray] (8,2) rectangle (9,3);
\draw[fill=gray] (9,2) rectangle (10,3);
\draw[fill=gray] (10,2) rectangle (11,3);
\draw[fill=gray] (0,3) rectangle (1,4);
\draw[fill=gray] (10,3) rectangle (11,4);
\draw[fill=gray] (0,4) rectangle (1,5);
\draw[fill=gray] (2,4) rectangle (3,5);
\draw[fill=gray] (3,4) rectangle (4,5);
\draw[fill=gray] (4,4) rectangle (5,5);
\draw[fill=gray] (5,4) rectangle (6,5);
\draw[fill=gray] (10,4) rectangle (11,5);
\draw[fill=gray] (0,5) rectangle (1,6);
\draw[fill=gray] (5,5) rectangle (6,6);
\draw[fill=gray] (10,5) rectangle (11,6);
\draw[fill=gray] (0,6) rectangle (1,7);
\draw[fill=gray] (5,6) rectangle (6,7);
\draw[fill=gray] (10,6) rectangle (11,7);
\draw[fill=gray] (0,7) rectangle (1,8);
\draw[fill=gray] (5,7) rectangle (6,8);
\draw[fill=gray] (10,7) rectangle (11,8);
\draw[fill=gray] (0,8) rectangle (1,9);
\draw[fill=gray] (5,8) rectangle (6,9);
\draw[fill=gray] (10,8) rectangle (11,9);
\draw[fill=gray] (0,9) rectangle (1,10);
\draw[fill=gray] (5,9) rectangle (6,10);
\draw[fill=gray] (10,9) rectangle (11,10);
\draw[fill=gray] (0,10) rectangle (1,11);
\draw[fill=pink] (9.500000,10.500000) circle (0.500000);
\draw[fill=gray] (10,10) rectangle (11,11);
\draw[fill=gray] (0,11) rectangle (1,12);
\draw[fill=gray] (1,11) rectangle (2,12);
\draw[fill=gray] (2,11) rectangle (3,12);
\draw[fill=gray] (3,11) rectangle (4,12);
\draw[fill=gray] (4,11) rectangle (5,12);
\draw[fill=gray] (5,11) rectangle (6,12);
\draw[fill=gray] (6,11) rectangle (7,12);
\draw[fill=gray] (7,11) rectangle (8,12);
\draw[fill=gray] (8,11) rectangle (9,12);
\draw[fill=gray] (9,11) rectangle (10,12);
\draw[fill=gray] (10,11) rectangle (11,12);
\node [align=center] at(-0.500000,0.500000) {0};
\node [align=center] at(-0.500000,1.500000) {1};
\node [align=center] at(-0.500000,2.500000) {2};
\node [align=center] at(-0.500000,3.500000) {3};
\node [align=center] at(-0.500000,4.500000) {4};
\node [align=center] at(-0.500000,5.500000) {5};
\node [align=center] at(-0.500000,6.500000) {6};
\node [align=center] at(-0.500000,7.500000) {7};
\node [align=center] at(-0.500000,8.500000) {8};
\node [align=center] at(-0.500000,9.500000) {9};
\node [align=center] at(-0.500000,10.500000) {10};
\node [align=center] at(-0.500000,11.500000) {11};
\node [align=center] at(0.500000,-0.500000) {A};
\node [align=center] at(1.500000,-0.500000) {B};
\node [align=center] at(2.500000,-0.500000) {C};
\node [align=center] at(3.500000,-0.500000) {D};
\node [align=center] at(4.500000,-0.500000) {E};
\node [align=center] at(5.500000,-0.500000) {F};
\node [align=center] at(6.500000,-0.500000) {G};
\node [align=center] at(7.500000,-0.500000) {H};
\node [align=center] at(8.500000,-0.500000) {I};
\node [align=center] at(9.500000,-0.500000) {J};
\node [align=center] at(10.500000,-0.500000) {K};
\draw[-Latex,draw=black!100] (1.500000,10.500000)--(1.900000,10.500000);
\draw[-Latex,draw=black!100] (2.500000,10.500000)--(2.900000,10.500000);
\draw[-Latex,draw=black!100] (3.500000,10.500000)--(3.900000,10.500000);
\draw[-Latex,draw=black!100] (4.500000,10.500000)--(4.900000,10.500000);
\draw[-Latex,draw=black!100] (5.500000,10.500000)--(5.900000,10.500000);
\draw[-Latex,draw=black!100] (6.500000,10.500000)--(6.900000,10.500000);
\draw[-Latex,draw=black!100] (7.500000,10.500000)--(7.900000,10.500000);
\draw[-Latex,draw=black!100] (8.500000,10.500000)--(8.900000,10.500000);
\draw[-Latex,draw=black!100] (9.500000,10.500000)--(9.100000,10.500000);
\draw[-Latex,draw=black!100] (9.500000,10.500000)--(9.500000,10.900000);
\draw[-Latex,draw=black!100] (9.500000,10.500000)--(9.900000,10.500000);
\draw[-Latex,draw=black!100] (9.500000,10.500000)--(9.500000,10.100000);
\draw[-Latex,draw=black!100] (1.500000,9.500000)--(1.500000,9.900000);
\draw[-Latex,draw=black!100] (2.500000,9.500000)--(2.500000,9.900000);
\draw[-Latex,draw=black!100] (3.500000,9.500000)--(3.500000,9.900000);
\draw[-Latex,draw=black!100] (3.500000,9.500000)--(3.900000,9.500000);
\draw[-Latex,draw=black!100] (4.500000,9.500000)--(4.500000,9.900000);
\draw[-Latex,draw=black!100] (6.500000,9.500000)--(6.500000,9.900000);
\draw[-Latex,draw=black!100] (7.500000,9.500000)--(7.500000,9.900000);
\draw[-Latex,draw=black!100] (8.500000,9.500000)--(8.500000,9.900000);
\draw[-Latex,draw=black!100] (8.500000,9.500000)--(8.900000,9.500000);
\draw[-Latex,draw=black!100] (9.500000,9.500000)--(9.500000,9.900000);
\draw[-Latex,draw=black!100] (1.500000,8.500000)--(1.500000,8.900000);
\draw[-Latex,draw=black!100] (2.500000,8.500000)--(2.500000,8.900000);
\draw[-Latex,draw=black!100] (2.500000,8.500000)--(2.900000,8.500000);
\draw[-Latex,draw=black!100] (3.500000,8.500000)--(3.900000,8.500000);
\draw[-Latex,draw=black!100] (4.500000,8.500000)--(4.500000,8.900000);
\draw[-Latex,draw=black!100] (6.500000,8.500000)--(6.500000,8.900000);
\draw[-Latex,draw=black!100] (7.500000,8.500000)--(7.500000,8.900000);
\draw[-Latex,draw=black!100] (7.500000,8.500000)--(7.900000,8.500000);
\draw[-Latex,draw=black!100] (8.500000,8.500000)--(8.900000,8.500000);
\draw[-Latex,draw=black!100] (9.500000,8.500000)--(9.500000,8.900000);
\draw[-Latex,draw=black!100] (1.500000,7.500000)--(1.500000,7.900000);
\draw[-Latex,draw=black!100] (1.500000,7.500000)--(1.900000,7.500000);
\draw[-Latex,draw=black!100] (2.500000,7.500000)--(2.900000,7.500000);
\draw[-Latex,draw=black!100] (3.500000,7.500000)--(3.900000,7.500000);
\draw[-Latex,draw=black!100] (4.500000,7.500000)--(4.500000,7.900000);
\draw[-Latex,draw=black!100] (6.500000,7.500000)--(6.500000,7.900000);
\draw[-Latex,draw=black!100] (6.500000,7.500000)--(6.900000,7.500000);
\draw[-Latex,draw=black!100] (7.500000,7.500000)--(7.900000,7.500000);
\draw[-Latex,draw=black!100] (8.500000,7.500000)--(8.900000,7.500000);
\draw[-Latex,draw=black!100] (9.500000,7.500000)--(9.500000,7.900000);
\draw[-Latex,draw=black!100] (1.500000,6.500000)--(1.900000,6.500000);
\draw[-Latex,draw=black!100] (2.500000,6.500000)--(2.900000,6.500000);
\draw[-Latex,draw=black!100] (3.500000,6.500000)--(3.900000,6.500000);
\draw[-Latex,draw=black!100] (4.500000,6.500000)--(4.500000,6.900000);
\draw[-Latex,draw=black!100] (6.500000,6.500000)--(6.900000,6.500000);
\draw[-Latex,draw=black!100] (7.500000,6.500000)--(7.900000,6.500000);
\draw[-Latex,draw=black!100] (8.500000,6.500000)--(8.900000,6.500000);
\draw[-Latex,draw=black!100] (9.500000,6.500000)--(9.500000,6.900000);
\draw[-Latex,draw=black!100] (1.500000,5.500000)--(1.900000,5.500000);
\draw[-Latex,draw=black!100] (2.500000,5.500000)--(2.900000,5.500000);
\draw[-Latex,draw=black!100] (3.500000,5.500000)--(3.900000,5.500000);
\draw[-Latex,draw=black!100] (4.500000,5.500000)--(4.500000,5.900000);
\draw[-Latex,draw=black!100] (6.500000,5.500000)--(6.900000,5.500000);
\draw[-Latex,draw=black!100] (7.500000,5.500000)--(7.900000,5.500000);
\draw[-Latex,draw=black!100] (8.500000,5.500000)--(8.900000,5.500000);
\draw[-Latex,draw=black!100] (9.500000,5.500000)--(9.500000,5.900000);
\draw[-Latex,draw=black!100] (1.500000,4.500000)--(1.500000,4.900000);
\draw[-Latex,draw=black!100] (6.500000,4.500000)--(6.900000,4.500000);
\draw[-Latex,draw=black!100] (7.500000,4.500000)--(7.900000,4.500000);
\draw[-Latex,draw=black!100] (8.500000,4.500000)--(8.900000,4.500000);
\draw[-Latex,draw=black!100] (9.500000,4.500000)--(9.500000,4.900000);
\draw[-Latex,draw=black!100] (1.500000,3.500000)--(1.500000,3.900000);
\draw[-Latex,draw=black!100] (1.500000,3.500000)--(1.900000,3.500000);
\draw[-Latex,draw=black!100] (2.500000,3.500000)--(2.900000,3.500000);
\draw[-Latex,draw=black!100] (3.500000,3.500000)--(3.900000,3.500000);
\draw[-Latex,draw=black!100] (4.500000,3.500000)--(4.900000,3.500000);
\draw[-Latex,draw=black!100] (5.500000,3.500000)--(5.900000,3.500000);
\draw[-Latex,draw=black!100] (6.500000,3.500000)--(6.900000,3.500000);
\draw[-Latex,draw=black!100] (7.500000,3.500000)--(7.900000,3.500000);
\draw[-Latex,draw=black!100] (8.500000,3.500000)--(8.900000,3.500000);
\draw[-Latex,draw=black!100] (9.500000,3.500000)--(9.500000,3.900000);
\draw[-Latex,draw=black!100] (1.500000,2.500000)--(1.500000,2.900000);
\draw[-Latex,draw=black!100] (1.500000,1.500000)--(1.500000,1.900000);
\end{tikzpicture} }
    \end{minipage}
  
  \hfill
  
    \begin{minipage}{\mpwidth}
      \centering
      \adjustbox{width=\aboxwidth}{
        \input{expFinalVers/4_MDP_policyStoch.tikz}
      }
    \end{minipage}
    \hspace{-.10cm}{\small $M_{6}$}\hspace{-.15cm}
    \begin{minipage}{\mpwidth}
      \centering
      \adjustbox{width=\aboxwidth}{
        \input{expFinalVers/4_OAMDP_Action_Pred.tikz} }
    \end{minipage}

    \begin{minipage}{\mpwidth}
      \centering
      \adjustbox{width=\aboxwidth}{
        \input{expFinalVers/7_MDP_policyStoch.tikz}
      }
    \end{minipage}
    \hspace{-.10cm}{\small $M_{7}$}\hspace{-.15cm}
    \begin{minipage}{\mpwidth}
      \centering
      \adjustbox{width=\aboxwidth}{
        \input{expFinalVers/7_OAMDP_Action_Pred.tikz} }
    \end{minipage}
  
  \hfill
  \begin{minipage}{.5\linewidth}
    \caption{Action predictability results showing, for mazes $M_1$--$M_7$, the stochastic policy $\pi_\mdps$ (left) (which ``covers'' all deterministic policies $\pi_\mdpb$) and the OAMDP policy $\piApred$ (right).
      All policies have been computed using $\gamma=1$.
      \label{fig|labyS}}
  \end{minipage}

\end{figure*}

\def\mpwidth{.6\columnwidth}
\def\aboxwidth{.9\textwidth}

\begin{figure}[ht!]
  \centering

	\begin{minipage}{\mpwidth}
		\centering
		\adjustbox{width=\aboxwidth}{
\begin{tikzpicture}
\draw (0,0) grid (12,5);
\draw[fill=gray] (0,0) rectangle (1,1);
\draw[fill=gray] (1,0) rectangle (2,1);
\draw[fill=gray] (2,0) rectangle (3,1);
\draw[fill=gray] (3,0) rectangle (4,1);
\draw[fill=gray] (4,0) rectangle (5,1);
\draw[fill=gray] (5,0) rectangle (6,1);
\draw[fill=gray] (6,0) rectangle (7,1);
\draw[fill=gray] (7,0) rectangle (8,1);
\draw[fill=gray] (8,0) rectangle (9,1);
\draw[fill=gray] (9,0) rectangle (10,1);
\draw[fill=gray] (10,0) rectangle (11,1);
\draw[fill=gray] (11,0) rectangle (12,1);
\draw[fill=gray] (0,1) rectangle (1,2);
\draw[fill=cyan] (2,1) rectangle (3,2);
\draw[fill=cyan] (4,1) rectangle (5,2);
\draw[fill=cyan] (6,1) rectangle (7,2);
\draw[fill=cyan] (8,1) rectangle (9,2);
\draw[fill=pink] (10.500000,1.500000) circle (0.500000);
\draw[fill=gray] (11,1) rectangle (12,2);
\draw[fill=gray] (0,2) rectangle (1,3);
\draw[fill=gray] (2,2) rectangle (3,3);
\draw[fill=gray] (3,2) rectangle (4,3);
\draw[fill=gray] (4,2) rectangle (5,3);
\draw[fill=gray] (5,2) rectangle (6,3);
\draw[fill=gray] (6,2) rectangle (7,3);
\draw[fill=gray] (7,2) rectangle (8,3);
\draw[fill=gray] (8,2) rectangle (9,3);
\draw[fill=gray] (9,2) rectangle (10,3);
\draw[fill=gray] (10,2) rectangle (11,3);
\draw[fill=gray] (11,2) rectangle (12,3);
\draw[fill=gray] (0,3) rectangle (1,4);
\draw[fill=pink] (8.500000,3.500000) circle (0.500000);
\draw[fill=gray] (9,3) rectangle (10,4);
\draw[fill=gray] (10,3) rectangle (11,4);
\draw[fill=gray] (11,3) rectangle (12,4);
\draw[fill=gray] (0,4) rectangle (1,5);
\draw[fill=gray] (1,4) rectangle (2,5);
\draw[fill=gray] (2,4) rectangle (3,5);
\draw[fill=gray] (3,4) rectangle (4,5);
\draw[fill=gray] (4,4) rectangle (5,5);
\draw[fill=gray] (5,4) rectangle (6,5);
\draw[fill=gray] (6,4) rectangle (7,5);
\draw[fill=gray] (7,4) rectangle (8,5);
\draw[fill=gray] (8,4) rectangle (9,5);
\draw[fill=gray] (9,4) rectangle (10,5);
\draw[fill=gray] (10,4) rectangle (11,5);
\draw[fill=gray] (11,4) rectangle (12,5);
\node [align=center] at(-0.500000,0.500000) {0};
\node [align=center] at(-0.500000,1.500000) {1};
\node [align=center] at(-0.500000,2.500000) {2};
\node [align=center] at(-0.500000,3.500000) {3};
\node [align=center] at(-0.500000,4.500000) {4};
\node [align=center] at(0.500000,-0.500000) {A};
\node [align=center] at(1.500000,-0.500000) {B};
\node [align=center] at(2.500000,-0.500000) {C};
\node [align=center] at(3.500000,-0.500000) {D};
\node [align=center] at(4.500000,-0.500000) {E};
\node [align=center] at(5.500000,-0.500000) {F};
\node [align=center] at(6.500000,-0.500000) {G};
\node [align=center] at(7.500000,-0.500000) {H};
\node [align=center] at(8.500000,-0.500000) {I};
\node [align=center] at(9.500000,-0.500000) {J};
\node [align=center] at(10.500000,-0.500000) {K};
\node [align=center] at(11.500000,-0.500000) {L};
\draw[-Latex,draw=black!100] (1.500000,3.500000)--(1.900000,3.500000);
\draw[-Latex,draw=black!100] (2.500000,3.500000)--(2.900000,3.500000);
\draw[-Latex,draw=black!100] (3.500000,3.500000)--(3.900000,3.500000);
\draw[-Latex,draw=black!100] (4.500000,3.500000)--(4.900000,3.500000);
\draw[-Latex,draw=black!100] (5.500000,3.500000)--(5.900000,3.500000);
\draw[-Latex,draw=black!100] (6.500000,3.500000)--(6.900000,3.500000);
\draw[-Latex,draw=black!100] (7.500000,3.500000)--(7.900000,3.500000);
\draw[-Latex,draw=black!100] (8.500000,3.500000)--(8.100000,3.500000);
\draw[-Latex,draw=black!100] (8.500000,3.500000)--(8.500000,3.900000);
\draw[-Latex,draw=black!100] (8.500000,3.500000)--(8.900000,3.500000);
\draw[-Latex,draw=black!100] (8.500000,3.500000)--(8.500000,3.100000);
\draw[-Latex,draw=black!100] (1.500000,2.500000)--(1.500000,2.900000);
\draw[-Latex,draw=black!100] (1.500000,2.500000)--(1.500000,2.100000);
\draw[-Latex,draw=black!100] (1.500000,1.500000)--(1.900000,1.500000);
\draw[-Latex,draw=black!100] (2.500000,1.500000)--(2.900000,1.500000);
\draw[-Latex,draw=black!100] (3.500000,1.500000)--(3.900000,1.500000);
\draw[-Latex,draw=black!100] (4.500000,1.500000)--(4.900000,1.500000);
\draw[-Latex,draw=black!100] (5.500000,1.500000)--(5.900000,1.500000);
\draw[-Latex,draw=black!100] (6.500000,1.500000)--(6.900000,1.500000);
\draw[-Latex,draw=black!100] (7.500000,1.500000)--(7.900000,1.500000);
\draw[-Latex,draw=black!100] (8.500000,1.500000)--(8.900000,1.500000);
\draw[-Latex,draw=black!100] (9.500000,1.500000)--(9.900000,1.500000);
\draw[-Latex,draw=black!100] (10.500000,1.500000)--(10.100000,1.500000);
\draw[-Latex,draw=black!100] (10.500000,1.500000)--(10.500000,1.900000);
\draw[-Latex,draw=black!100] (10.500000,1.500000)--(10.900000,1.500000);
\draw[-Latex,draw=black!100] (10.500000,1.500000)--(10.500000,1.100000);
\end{tikzpicture}
 		}
	\end{minipage}

	\begin{minipage}{\mpwidth}
          \centering
          (a) Stochastic policy $\pi_\mdps$  ($\gamma=1$)
       \end{minipage}

       \medskip

       \begin{minipage}{\mpwidth}
		\centering
		\adjustbox{width=\aboxwidth}{
\begin{tikzpicture}
\draw (0,0) grid (12,5);
\draw[fill=gray] (0,0) rectangle (1,1);
\draw[fill=gray] (1,0) rectangle (2,1);
\draw[fill=gray] (2,0) rectangle (3,1);
\draw[fill=gray] (3,0) rectangle (4,1);
\draw[fill=gray] (4,0) rectangle (5,1);
\draw[fill=gray] (5,0) rectangle (6,1);
\draw[fill=gray] (6,0) rectangle (7,1);
\draw[fill=gray] (7,0) rectangle (8,1);
\draw[fill=gray] (8,0) rectangle (9,1);
\draw[fill=gray] (9,0) rectangle (10,1);
\draw[fill=gray] (10,0) rectangle (11,1);
\draw[fill=gray] (11,0) rectangle (12,1);
\draw[fill=gray] (0,1) rectangle (1,2);
\draw[fill=cyan] (2,1) rectangle (3,2);
\draw[fill=cyan] (4,1) rectangle (5,2);
\draw[fill=cyan] (6,1) rectangle (7,2);
\draw[fill=cyan] (8,1) rectangle (9,2);
\draw[fill=pink] (10.500000,1.500000) circle (0.500000);
\draw[fill=gray] (11,1) rectangle (12,2);
\draw[fill=gray] (0,2) rectangle (1,3);
\draw[fill=gray] (2,2) rectangle (3,3);
\draw[fill=gray] (3,2) rectangle (4,3);
\draw[fill=gray] (4,2) rectangle (5,3);
\draw[fill=gray] (5,2) rectangle (6,3);
\draw[fill=gray] (6,2) rectangle (7,3);
\draw[fill=gray] (7,2) rectangle (8,3);
\draw[fill=gray] (8,2) rectangle (9,3);
\draw[fill=gray] (9,2) rectangle (10,3);
\draw[fill=gray] (10,2) rectangle (11,3);
\draw[fill=gray] (11,2) rectangle (12,3);
\draw[fill=gray] (0,3) rectangle (1,4);
\draw[fill=pink] (8.500000,3.500000) circle (0.500000);
\draw[fill=gray] (9,3) rectangle (10,4);
\draw[fill=gray] (10,3) rectangle (11,4);
\draw[fill=gray] (11,3) rectangle (12,4);
\draw[fill=gray] (0,4) rectangle (1,5);
\draw[fill=gray] (1,4) rectangle (2,5);
\draw[fill=gray] (2,4) rectangle (3,5);
\draw[fill=gray] (3,4) rectangle (4,5);
\draw[fill=gray] (4,4) rectangle (5,5);
\draw[fill=gray] (5,4) rectangle (6,5);
\draw[fill=gray] (6,4) rectangle (7,5);
\draw[fill=gray] (7,4) rectangle (8,5);
\draw[fill=gray] (8,4) rectangle (9,5);
\draw[fill=gray] (9,4) rectangle (10,5);
\draw[fill=gray] (10,4) rectangle (11,5);
\draw[fill=gray] (11,4) rectangle (12,5);
\node [align=center] at(-0.500000,0.500000) {0};
\node [align=center] at(-0.500000,1.500000) {1};
\node [align=center] at(-0.500000,2.500000) {2};
\node [align=center] at(-0.500000,3.500000) {3};
\node [align=center] at(-0.500000,4.500000) {4};
\node [align=center] at(0.500000,-0.500000) {A};
\node [align=center] at(1.500000,-0.500000) {B};
\node [align=center] at(2.500000,-0.500000) {C};
\node [align=center] at(3.500000,-0.500000) {D};
\node [align=center] at(4.500000,-0.500000) {E};
\node [align=center] at(5.500000,-0.500000) {F};
\node [align=center] at(6.500000,-0.500000) {G};
\node [align=center] at(7.500000,-0.500000) {H};
\node [align=center] at(8.500000,-0.500000) {I};
\node [align=center] at(9.500000,-0.500000) {J};
\node [align=center] at(10.500000,-0.500000) {K};
\node [align=center] at(11.500000,-0.500000) {L};
\draw[-Latex,draw=black!100] (1.500000,3.500000)--(1.900000,3.500000);
\draw[-Latex,draw=black!100] (2.500000,3.500000)--(2.900000,3.500000);
\draw[-Latex,draw=black!100] (3.500000,3.500000)--(3.900000,3.500000);
\draw[-Latex,draw=black!100] (4.500000,3.500000)--(4.900000,3.500000);
\draw[-Latex,draw=black!100] (5.500000,3.500000)--(5.900000,3.500000);
\draw[-Latex,draw=black!100] (6.500000,3.500000)--(6.900000,3.500000);
\draw[-Latex,draw=black!100] (7.500000,3.500000)--(7.900000,3.500000);
\draw[-Latex,draw=black!100] (8.500000,3.500000)--(8.100000,3.500000);
\draw[-Latex,draw=black!100] (8.500000,3.500000)--(8.500000,3.900000);
\draw[-Latex,draw=black!100] (8.500000,3.500000)--(8.900000,3.500000);
\draw[-Latex,draw=black!100] (8.500000,3.500000)--(8.500000,3.100000);
\draw[-Latex,draw=black!100] (1.500000,2.500000)--(1.500000,2.900000);
\draw[-Latex,draw=black!100] (1.500000,2.500000)--(1.500000,2.100000);
\draw[-Latex,draw=black!100] (1.500000,1.500000)--(1.900000,1.500000);
\draw[-Latex,draw=black!100] (2.500000,1.500000)--(2.900000,1.500000);
\draw[-Latex,draw=black!100] (3.500000,1.500000)--(3.900000,1.500000);
\draw[-Latex,draw=black!100] (4.500000,1.500000)--(4.900000,1.500000);
\draw[-Latex,draw=black!100] (5.500000,1.500000)--(5.900000,1.500000);
\draw[-Latex,draw=black!100] (6.500000,1.500000)--(6.900000,1.500000);
\draw[-Latex,draw=black!100] (7.500000,1.500000)--(7.900000,1.500000);
\draw[-Latex,draw=black!100] (8.500000,1.500000)--(8.900000,1.500000);
\draw[-Latex,draw=black!100] (9.500000,1.500000)--(9.900000,1.500000);
\draw[-Latex,draw=black!100] (10.500000,1.500000)--(10.100000,1.500000);
\draw[-Latex,draw=black!100] (10.500000,1.500000)--(10.500000,1.900000);
\draw[-Latex,draw=black!100] (10.500000,1.500000)--(10.900000,1.500000);
\draw[-Latex,draw=black!100] (10.500000,1.500000)--(10.500000,1.100000);
\end{tikzpicture}
 		}
	\end{minipage}

	\begin{minipage}{\mpwidth}
		\centering
		(b) pOAMDP policy $\piApred$ ($\gamma=1$)
	\end{minipage}

	\medskip

	\begin{minipage}{\mpwidth}
		\centering
		\adjustbox{width=\aboxwidth}{
\begin{tikzpicture}
\draw (0,0) grid (12,5);
\draw[fill=gray] (0,0) rectangle (1,1);
\draw[fill=gray] (1,0) rectangle (2,1);
\draw[fill=gray] (2,0) rectangle (3,1);
\draw[fill=gray] (3,0) rectangle (4,1);
\draw[fill=gray] (4,0) rectangle (5,1);
\draw[fill=gray] (5,0) rectangle (6,1);
\draw[fill=gray] (6,0) rectangle (7,1);
\draw[fill=gray] (7,0) rectangle (8,1);
\draw[fill=gray] (8,0) rectangle (9,1);
\draw[fill=gray] (9,0) rectangle (10,1);
\draw[fill=gray] (10,0) rectangle (11,1);
\draw[fill=gray] (11,0) rectangle (12,1);
\draw[fill=gray] (0,1) rectangle (1,2);
\draw[fill=cyan] (2,1) rectangle (3,2);
\draw[fill=cyan] (4,1) rectangle (5,2);
\draw[fill=cyan] (6,1) rectangle (7,2);
\draw[fill=cyan] (8,1) rectangle (9,2);
\draw[fill=pink] (10.500000,1.500000) circle (0.500000);
\draw[fill=gray] (11,1) rectangle (12,2);
\draw[fill=gray] (0,2) rectangle (1,3);
\draw[fill=gray] (2,2) rectangle (3,3);
\draw[fill=gray] (3,2) rectangle (4,3);
\draw[fill=gray] (4,2) rectangle (5,3);
\draw[fill=gray] (5,2) rectangle (6,3);
\draw[fill=gray] (6,2) rectangle (7,3);
\draw[fill=gray] (7,2) rectangle (8,3);
\draw[fill=gray] (8,2) rectangle (9,3);
\draw[fill=gray] (9,2) rectangle (10,3);
\draw[fill=gray] (10,2) rectangle (11,3);
\draw[fill=gray] (11,2) rectangle (12,3);
\draw[fill=gray] (0,3) rectangle (1,4);
\draw[fill=pink] (8.500000,3.500000) circle (0.500000);
\draw[fill=gray] (9,3) rectangle (10,4);
\draw[fill=gray] (10,3) rectangle (11,4);
\draw[fill=gray] (11,3) rectangle (12,4);
\draw[fill=gray] (0,4) rectangle (1,5);
\draw[fill=gray] (1,4) rectangle (2,5);
\draw[fill=gray] (2,4) rectangle (3,5);
\draw[fill=gray] (3,4) rectangle (4,5);
\draw[fill=gray] (4,4) rectangle (5,5);
\draw[fill=gray] (5,4) rectangle (6,5);
\draw[fill=gray] (6,4) rectangle (7,5);
\draw[fill=gray] (7,4) rectangle (8,5);
\draw[fill=gray] (8,4) rectangle (9,5);
\draw[fill=gray] (9,4) rectangle (10,5);
\draw[fill=gray] (10,4) rectangle (11,5);
\draw[fill=gray] (11,4) rectangle (12,5);
\node [align=center] at(-0.500000,0.500000) {0};
\node [align=center] at(-0.500000,1.500000) {1};
\node [align=center] at(-0.500000,2.500000) {2};
\node [align=center] at(-0.500000,3.500000) {3};
\node [align=center] at(-0.500000,4.500000) {4};
\node [align=center] at(0.500000,-0.500000) {A};
\node [align=center] at(1.500000,-0.500000) {B};
\node [align=center] at(2.500000,-0.500000) {C};
\node [align=center] at(3.500000,-0.500000) {D};
\node [align=center] at(4.500000,-0.500000) {E};
\node [align=center] at(5.500000,-0.500000) {F};
\node [align=center] at(6.500000,-0.500000) {G};
\node [align=center] at(7.500000,-0.500000) {H};
\node [align=center] at(8.500000,-0.500000) {I};
\node [align=center] at(9.500000,-0.500000) {J};
\node [align=center] at(10.500000,-0.500000) {K};
\node [align=center] at(11.500000,-0.500000) {L};
\draw[-Latex,draw=black!100] (1.500000,3.500000)--(1.900000,3.500000);
\draw[-Latex,draw=black!100] (2.500000,3.500000)--(2.900000,3.500000);
\draw[-Latex,draw=black!100] (3.500000,3.500000)--(3.900000,3.500000);
\draw[-Latex,draw=black!100] (4.500000,3.500000)--(4.900000,3.500000);
\draw[-Latex,draw=black!100] (5.500000,3.500000)--(5.900000,3.500000);
\draw[-Latex,draw=black!100] (6.500000,3.500000)--(6.900000,3.500000);
\draw[-Latex,draw=black!100] (7.500000,3.500000)--(7.900000,3.500000);
\draw[-Latex,draw=black!100] (8.500000,3.500000)--(8.100000,3.500000);
\draw[-Latex,draw=black!100] (8.500000,3.500000)--(8.500000,3.900000);
\draw[-Latex,draw=black!100] (8.500000,3.500000)--(8.900000,3.500000);
\draw[-Latex,draw=black!100] (8.500000,3.500000)--(8.500000,3.100000);
\draw[-Latex,draw=black!100] (1.500000,2.500000)--(1.500000,2.900000);
\draw[-Latex,draw=black!100] (1.500000,1.500000)--(1.500000,1.900000);
\draw[-Latex,draw=black!100] (2.500000,1.500000)--(2.900000,1.500000);
\draw[-Latex,draw=black!100] (3.500000,1.500000)--(3.900000,1.500000);
\draw[-Latex,draw=black!100] (4.500000,1.500000)--(4.900000,1.500000);
\draw[-Latex,draw=black!100] (5.500000,1.500000)--(5.900000,1.500000);
\draw[-Latex,draw=black!100] (6.500000,1.500000)--(6.900000,1.500000);
\draw[-Latex,draw=black!100] (7.500000,1.500000)--(7.900000,1.500000);
\draw[-Latex,draw=black!100] (8.500000,1.500000)--(8.900000,1.500000);
\draw[-Latex,draw=black!100] (9.500000,1.500000)--(9.900000,1.500000);
\draw[-Latex,draw=black!100] (10.500000,1.500000)--(10.100000,1.500000);
\draw[-Latex,draw=black!100] (10.500000,1.500000)--(10.500000,1.900000);
\draw[-Latex,draw=black!100] (10.500000,1.500000)--(10.900000,1.500000);
\draw[-Latex,draw=black!100] (10.500000,1.500000)--(10.500000,1.100000);
\end{tikzpicture}
 		}
	\end{minipage}

	\begin{minipage}{\mpwidth}
		\centering
		(c) pOAMDP policy $\piSpred$  ($\gamma=1$)
	\end{minipage}

	\caption{Results for maze $M_8$
		\label{fig|laby4}}
\end{figure}

\begin{table}[ht!]
  \caption{Results for Maze problems $M_1$--$M_7$ with actual human observers against 3 agents: $\pi_\mdps$, $\pi_\mdpb$, $\pi_\pred$, indicating: [\#Err.p] the predicted average number of errors when evaluating the policy using $\RApred$;
    [\#Err.h] the actual average number of errors per trajectory with human observers;
    [\#steps] the number of time steps to reach the goal;
[time/step] the average response time of the human observer per time step.
  }
    \label{tab|HumanXPs}

  \sisetup{
    round-mode = places,
    round-precision = 1,
    table-format=2.1,
}

  \centering

  \adjustbox{max width=1.\linewidth}
  {
\begin{tabular}{lrS[table-format=2.0]
  S[table-format=2.1]
  S[table-format=2.1]@{$\pm$}>{\scriptsize}S[table-format=1.1]
  S@{$\pm$}>{\scriptsize}S[table-format=3.2]
  }
\toprule
& & {\#steps} &  {\#Err.p} & \multicolumn{2}{c}{\#Err.h}   & \multicolumn{2}{c}{time/step (ms)}  \\ 
\midrule
$\pi_1$
& $M_1$ & 15&2.90625&3.3684210526315788&1.738790330334712&561.3263157894737&188.21447431874074\\
& $M_2$ & 13&3.26953125&3.9473684210526314&1.5446568914424659&600.65991902834&279.20705241252386\\
& $M_3$ & 15&2.90625&2.8947368421052633&1.5597270716416047&490.7894736842105&167.49651665905617\\
& $M_4$ & 15&3.052734375&3.3157894736842106&1.293257367998745&574.5157894736842&188.16859244115778\\
& $M_5$ & 16&3.25&3.0526315789473686&1.5802139169719438&506.4375&124.18371142797986\\
& $M_6$ & 84&10.5&12.789473684210526&2.070398441750365&481.6632775119617&101.57244978178646\\
& $M_7$ & 29&2.6284722222222223&2.6842105263157894&1.529438225803745&437.5473684210526&100.8069009222794\\
\cmidrule(lr){2-8}
 & $\bigoplus_i M_i$ &189&28.51323784722222&30.45&{-}&499.0825109649123&{-}\\ 
\midrule
$\pi_2$
& $M_1$ & 15&2.0&1.105263157894737&0.936585811581694&379.3438596491228&110.55520550063598\\
& $M_2$ & 13&2.125&1.0&0.8164965809277261&350.69635627530363&94.00119767056411\\
& $M_3$ & 15&2.125&1.0&0.9428090415820635&370.9017543859649&120.87928958705923\\
& $M_4$ & 15&2.125&0.8947368421052632&0.8093026382225119&334.0736842105263&86.32492707461702\\
& $M_5$ & 16&2.5&0.631578947368421&0.6839855680567695&361.3717105263158&95.95485156533482\\
& $M_6$ & 84&10.25&10.105263157894736&2.5142866636285897&436.76794258373207&81.54144146606014\\
& $M_7$ & 29&2.8333333333333335&2.473684210526316&0.7723284457212329&392.4614035087719&109.65077294664763\\
  \cmidrule(lr){2-8} & $\bigoplus_i M_i$ &189&23.958333333333332&16.35&{-}&400.07922149122805&{-}\\ 
 \midrule$\pi_3$& $M_1$ & 17&1.5&1.0526315789473684&0.9112679939102141&310.9783281733746&74.83572220128599\\
& $M_2$ & 13&2.0&1.4736842105263157&1.1239029738980326&364.39676113360326&125.85690575127221\\
& $M_3$ & 15&2.0&1.736842105263158&1.147078669352809&353.2701754385965&94.73168730470888\\
& $M_4$ & 15&2.0&0.47368421052631576&0.6966922684794661&346.2526315789474&82.76805186396939\\
& $M_5$ & 16&2.0&1.368421052631579&1.0651304727481081&371.8486842105263&142.95609471279502\\
& $M_6$ & 86&2.666666666666667&2.263157894736842&1.5578512717186201&310.3953216374269&62.60358026678134\\
& $M_7$ & 29&1.6666666666666667&1.8421052631578947&1.3442535266309141&376.2280701754386&94.16076104926891\\
 \cmidrule(lr){2-8} & $\bigoplus_i M_i$ &192&13.833333333333334&9.7&{-}&335.14607948442534&{-}\\ 
 \bottomrule\end{tabular}
   }
  \end{table}

\subsubsection{Maze problem}

\paragraph{Grids used}
The mazes mainly consist of corridors and (empty) rooms.
For action predictability, we expect the pOAMDP policies to prefer corridors over rooms (which allow for more possible optimal actions).
\Cref{fig|labyS} shows mazes $M_1$--$M_7$, which have been used for action predicatility (including experiments with humans discussed in \Cref{sec|XPinVivo}).
They all consist in a number of corridors and rooms, have a starting state $s_0$ (marked by a circle), and overall increase in complexity from $M_1$ to $M_7$.
The maze $M_8$ in \Cref{fig|laby4} consists of 2 corridors that lead to a terminal state.
  One of these corridors contains slippery cells, but the average traversal time is the same for both.
  This maze's goal is to observe differences between $\RApred$ and $\RSpred$.

Each SSP is solved with $\gamma=1$ and $\epsilon=0.001$.
As expected, when crossing a room of size $n\times m$ from one corner to the opposite corner, $\pi_\mdps$ randomly picks one of the ${n+m \choose n}$ optimal paths, while the only two possible $\pi_\mdpb$ policies follow the walls (clockwise or counterclockwise).

\medskip

Note:
In the following, we mainly focus on action predictability because, here, solution policies turn out to be identical for state predictability.
This is favored in deterministic environments, where predicting the next state is often equivalent to predicting the next action.

\paragraph{Analysis of $\piApred$ and $\piSpred$}

We observe several interesting behaviors with $\RApred(s,a,s')$:
\begin{enumerate}
\item The
pOAMDP agent
will plan a longer path through a narrow corridor, where its next action will be easy to predict, rather than a shorter path going through one or multiple rooms as illustrated on $M_1$ and $M_6$.
\item
In rooms,  $\pi_\mdps$ has two optimal actions except along the two walls near the exit, with a single optimal action.
The 
   pOAMDP agent
 behaves thus more predictably by going towards the closest of these two exit walls and following it, as visible in $M_1$--$M_7$.
\item In $M_3$, the 
pOAMDP agent
 can choose between
  \begin{enumerate*}[label=(\roman*)]
  \item a corridor leading to a room, and
  \item a room leading to a corridor.
  \end{enumerate*}
  When $\gamma=1$, the 
  pOAMDP agent
 has no preference.
  When $\gamma<1$ (policy not shown here), the 
  pOAMDP agent
 prefers to go through a corridor first because the discount puts more importance on early rewards (see cell $(B,7)$).
\item In $M_4$, adding a door compared to $M_3$ makes for more uncertainty in the left room, so that the agent prefers going towards the right room.
\item In \Cref{fig|laby4}, cell $(B,2)$, $\piApred$ has no preference between going up and down as, in both cases, there is no ambiguity about optimal actions afterwards.
\end{enumerate}

Quantitative results in the first column of \Cref{tab|HumanXPs} are obtained by computing the value of each policy wrt $\RApred$ and displaying $-V^\pi_{\RApred}(s_0)$.
They show that $\pi_\mdps$'s expected number of errors per trajectory is worse than for the two other agent policies, in particular when large rooms exist.
Also, $\piApred$ has significantly better results than the two other policies on problems $M_6$ \& $M_7$, which have multiple rooms and are more complex.

In most of these problems, $\piApred$ and $\piSpred$ exhibit identical behaviors.
This is not the case in maze $M_8$ (\Cref{fig|laby4}), as $\piSpred$ prefers going up in cell $(B,1)$, which goes against the observer's predictions, to follow the path with no slippery cells (as slippery cells induce state uncertainties).

\subsubsection{Firefighter problem}

\def\mpwidth{\columnwidth}
\def\aboxwidth{0.8\textwidth}

\begin{figure}[ht!]
	\begin{minipage}{\mpwidth}
		\centering
		\adjustbox{width=\aboxwidth}{

 		}
	\end{minipage}

	\begin{minipage}{\mpwidth}
		\centering
		(b) pOAMDP policy $\piApred$  ($\gamma=\myDiscount$)
	\end{minipage}

	\caption{Results for firefighter problem F1
          \label{fig|pompier1}}

\end{figure}

\begin{figure}[ht!]
	\begin{minipage}{\mpwidth}
		\centering
		\adjustbox{width=\aboxwidth}{
                  %
 		}
	\end{minipage}

	\begin{minipage}{\mpwidth}
		\centering
		(b) pOAMDP policy $\piApred$ ($\gamma=\myDiscount$) \end{minipage}

	\caption{Results for firefighter problem F2
		\label{fig|pompier2}}

\end{figure}

\begin{figure}[ht!]
	\begin{minipage}{\mpwidth}
		\centering
		\adjustbox{width=\aboxwidth}{
                  %
 		}
	\end{minipage}

	\begin{minipage}{\mpwidth}
		\centering
		(b) pOAMDP policy $\piApred$ ($\gamma=\myDiscount$) \end{minipage}

	\caption{Results for firefighter problem F3
		\label{fig|pompier3}}

\end{figure}

\paragraph{Grids used}
The following grids where used to test the reward functions:
\begin{enumerate}
	\item the grid in \Cref{fig|pompier1}
	contains 1 fire and 1 water source linked by a room and by a corridor;
	\item the grid in \Cref{fig|pompier2}
	is a room with 2 fires and 2 water sources;
	\item the grid in \Cref{fig|pompier3}
          contains 2 fires and 2 water sources;
          a part of the map is a room and the other part is a corridor.
\end{enumerate}

The underlying MDPs are not SSPs anymore, so that we use $\gamma=\myDiscount$-discounted pOAMDPs.

\paragraph{Analysis of $\piApred$} A behavior similar to the maze problem can be observed.
In \Cref{fig|pompier1}, $\piApred$ prefers the corridor over the open room.
In such rooms, $\piApred$, as $\pi_\mdpb$ (see \Cref{fig|heatmapError}, page  \pageref{fig|heatmapError}), tries to reach a wall and walk along it (\Cref{fig|pompier1,fig|pompier3}).
In \Cref{fig|pompier2}, the pOAMDP agent tries to be more predictable by walking along the wall or by reaching Row~5 or Column~F to reduce the number of optimal paths to reach the fire in the middle.
In \Cref{fig|pompier3}, the pOAMDP agent prefers the fire located in $(B,1)$ and the water source located in $(E,8)$ even if another water source ou fire spot is closer.
This is particularly visible on the ``without water'' side of the figure, where $\piApred$ goes from $(G,5)$ to $(E,8)$ to refill.

\section{Experimenting with Humans} \label{sec|XPinVivo}

The objective of pOAMDP solution policies is to make it easier for an observer to predict actions or states.
Of particular interest is the case of human observers.
An experiment was thus conducted to confront actual human participants with stochastic and biased MDP policies ($\pi_\mdps$ and $\pi_\mdpb$), and with pOAMDP policies ($\piApred$).
We were interested in particular in:
\begin{itemize}
\item assessing how predictable each type of policy was for humans, by measuring the number of prediction errors;
\item assessing whether predictions were easy to make, by measuring their response times; and
\item knowing how the various agent behaviors were perceived by humans. \end{itemize}

\subsection{Protocol}
\label{sec|inVivoProtocol}

\paragraph{Participants}
Experiments have been conducted with 20 human participants (4 women; aged $28.9 \pm 7.7$ years) to assess the actual predictability of the 3 policies at hand on mazes $M_1$--$M_7$ (\Cref{fig|labyS}).
All participants provided written consent prior to their participation.

\paragraph{Task}
Participants were seated in front of a computer displaying a maze containing a robot and the robot's goal.
For each position of the robot, at each time step, the participant had to indicate the next action by pressing one of the four arrow keys. The robot then moved to the next position according to the policy, independently of the participant's response, and the participant had to indicate the next action, and so on along the trajectory to the goal.

\paragraph{Experimental process}
Participants began with a learning phase lasting about one minute, consisting of a maze with a random policy.
Then came the test phase, consisting of 3 sequences of 7 mazes each, each sequence associated with a policy.
Participants were told that the robot behavior was going to change at each sequence.
Each robot was identified by a color.
The ordering of policies was randomized, as well as the ordering of mazes within a sequence, with the exception that $M_6$, the largest maze, was always presented in 4th position.
For $\pi_\mdpb$, 4 different orderings over actions were used as biases (out of $4!=24$ possibilities), and randomly sampled before each trajectory.
All previously mentionned randomizations were controlled (hand-written) to prevent unwanted regularities.

At the end of each sequence, the participant completed a 3-item questionnaire. For each item, the participant answered on a 7-point Lickert scale from ``strongly disagree'' to ``strongly agree''. The 3 items related to the policy they had just seen, and were as follows:
\begin{enumerate*}
	\item this robot was easy to anticipate (\emph{Anticipation});
	\item its decisions seemed generally logical (\emph{Logic});
	\item some of its decisions surprised me (\emph{Surprise}).
\end{enumerate*}
Each participant completed this questionnaire 3 times, once per policy. Once the test phase was over, they completed a questionnaire including: socio-demographic questions; a request to rank the three policies from easiest to most difficult, and another from most logical to most unexpected.

On average, the experiment lasted 30 minutes.

\paragraph{Data analysis}

The data recorded during each maze were: the number of errors, \emph{i.e.} the number of times the next move predicted by the participant did not correspond to the move subsequently chosen by the robot ; the response time (in ms), \emph{i.e.} from the instant when the robot finished a move to the instant when the participant indicated the position he thought would be the next. Each maze began with a two-square corridor to control the start of each trajectory, and the first square (the first response given by the participants) of each maze was removed from the analyses. One participant was removed from the analyses, as his response times were more than 3 standard deviations above the overall mean. Data processing on errors and response times was therefore carried out on 19 participants.

For the questionnaire, each of the 3 items was analyzed (Anticipation, Logic and Surprise).

To determine whether there were any significant differences between the three policies, standard errors were calculated for the quantitative variables, as well as for the questionnaire.

\subsection{Results}

\subsubsection{Numbers of Errors and Response Times}

\begin{figure}
  \centering
	\includegraphics[width=.8\textwidth]{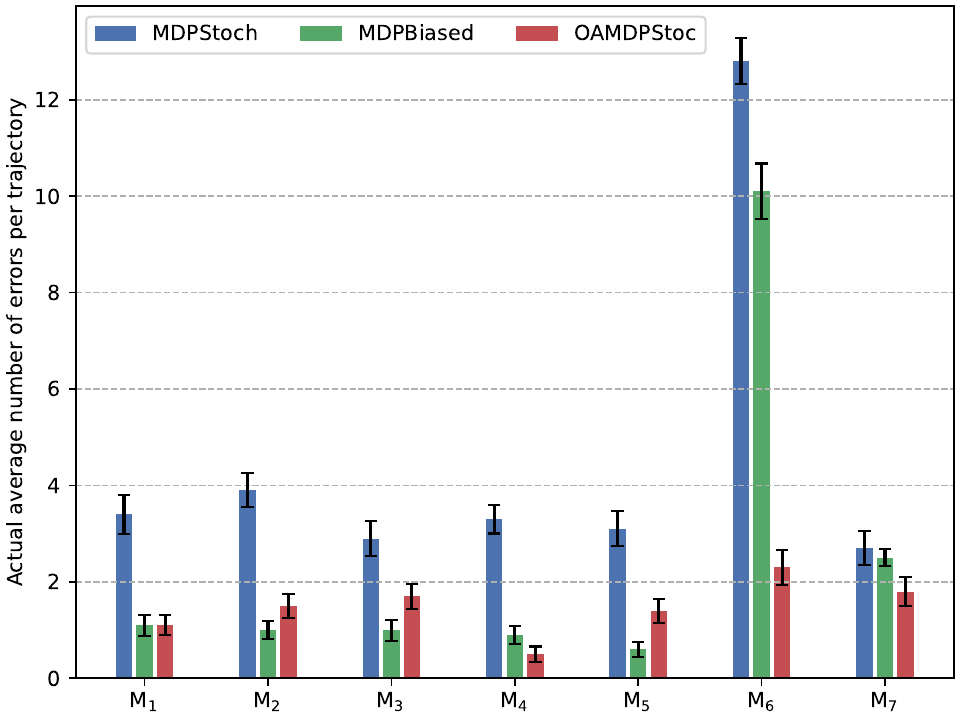}
	\caption{Graph representing the average number of errors made by the participants for each maze
\label{fig|graphError}}
\end{figure}

\begin{figure}
  \centering
	\includegraphics[width=.8\textwidth]{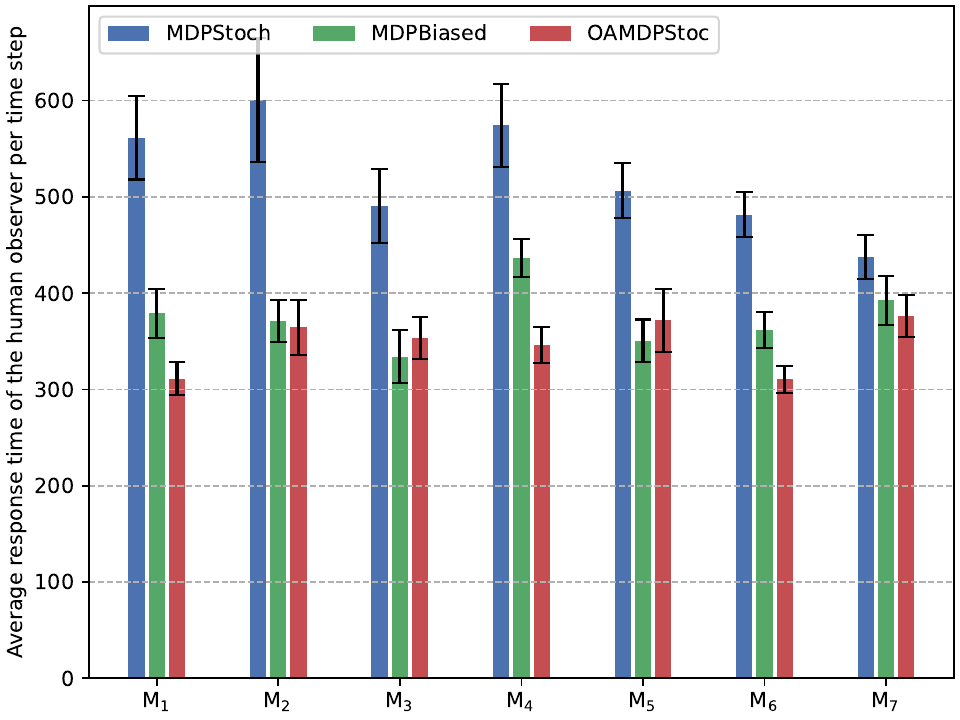}
	\caption{Graph representing the average response time of the participants for each maze
		\label{fig|graphTime}}
\end{figure}

The main quantitative results are presented in \Cref{tab|HumanXPs}, page \pageref{tab|HumanXPs}, as well as in \Cref{fig|graphError} for errors and in \Cref{fig|graphTime} for response times, for each policy-maze combination, plus a fake maze $\bigoplus_i M_i$ whose results are obtained by assuming that the other mazes have been concatenated.
The 1st column shows the expected number of errors per trajectory according to our model ($-V^\pi_{\RApred}(s_0)$), which can be compared with the measured values with human observers in the 2nd column.
Values are rather similar for $\pi_\mdps$, with typically a few more errors made by humans.
Human scores are notably better than anticipated for $\pi_\mdpb$ (and also better than human scores with $\pi_\mdps$), because humans very quickly learn the agent's bias, which facilitates predictions in large rooms.
The benefit of learning is very limited in complex mazes with many small rooms as $M_6$.
Human scores with $\piApred$ are worse than with $\pi_\mdpb$ on simple mazes (where learning biases helps), but notably better on complex mazes $M_6$+$M_7$.

As complementary information, the 3rd column provides the (constant) lengths of trajectories in each case as an indicator of the problem size.
As anticipated, $\pi_\mdps$ and $\pi_\mdpb$ generate minimal-length trajectories, while $\piApred$ generates slightly longer ones in some cases ($M_1$+$M_6$) to follow more predictable paths.

The 4th column indicates the average response time (in ms) per cell, which appears to be inversely related to the difficulty to make predictions.
These average response times are lower for $\pi_\mdpb$ and $\piApred$ than for $\pi_\mdps$.
An important difference between response times of $\pi_\mdpb$ and $\piApred$ can be observed for $M_6$.
In this maze, it is harder for the human to learn the agent's bias of $\pi_\mdpb$, while $\piApred$ plans its actions to go through states with reduced action ambiguity.

\subsubsection{Heatmaps}

Heatmaps allow us to visualize where in the maze the participants made mistakes or were faster/ slower to take a decision.

\paragraph{Error Rate Heatmaps}
Error rate heatmaps for each maze in each policy are shown \Cref{fig|heatmapError}.
They are defined as follows:
\begin{itemize}
\item the blue color represents the average number of visits computed as $\frac{\#visits}{\#participants}$, and
\item the red color represents the average error rate computed as  $\frac{\#errors}{\#visits}$.
\end{itemize}
Note that, in cells with a low number of visits (light blue background color), the error rate estimate is poor compared to often-visited cells (dark blue background color).
In unvisited cells (white background color), there is no error rate to estimate, hence the lack of inner square.

\newcommand{\threeMazes}[2]{
  \begin{minipage}{\colwidth}
    \begin{minipage}{\mpwidth}
      \centering
      \adjustbox{width=\aboxwidth}{
        \input{\dir #1_MDP_policyStochHeatmap.tikz}
      }
    \end{minipage}
    \hfill
    \begin{minipage}{\mpwidth}
      \centering
      \adjustbox{width=\aboxwidth}{
        \input{\dir #1_MDP_policyBiasedHeatMap.tikz}
      }
    \end{minipage}
    \hfill
    \begin{minipage}{\mpwidth}
      \centering
      \adjustbox{width=\aboxwidth}{
        \input{\dir #1_OAMDP_policyStochHeatMap.tikz}
      }
    \end{minipage}

    \centerline{\small $M_{#2}$}
  \end{minipage}
}

\def\colwidth{.48\textwidth}
\def\mpwidth{.32\textwidth}
\def\aboxwidth{\textwidth}

\def\dir{expFinalVers/heatmap/error/}

\begin{figure}[ht!]

  \begin{minipage}{\colwidth}
    \begin{minipage}{\mpwidth}
      \centering
      \adjustbox{width=\aboxwidth}{
        \input{\dir 1_MDP_policyStochHeatmap.tikz}
      }
    \end{minipage}
    \hfill
    \begin{minipage}{\mpwidth}
      \centering
      \adjustbox{width=\aboxwidth}{
        \begin{tikzpicture}
\draw (0,0) grid (11,12);
\draw[fill=gray] (0,0) rectangle (1,1);
\draw[fill=gray] (1,0) rectangle (2,1);
\draw[fill=gray] (2,0) rectangle (3,1);
\draw[fill=gray] (3,0) rectangle (4,1);
\draw[fill=gray] (4,0) rectangle (5,1);
\draw[fill=gray] (5,0) rectangle (6,1);
\draw[fill=gray] (6,0) rectangle (7,1);
\draw[fill=gray] (7,0) rectangle (8,1);
\draw[fill=gray] (8,0) rectangle (9,1);
\draw[fill=gray] (9,0) rectangle (10,1);
\draw[fill=gray] (10,0) rectangle (11,1);
\draw[fill=gray] (0,1) rectangle (1,2);
\draw[fill=gray] (1,1) rectangle (2,2);
\draw[fill={rgb,255:red,0.0;green,0.0;blue,255.0}]  (2,1) rectangle (3,2);
\draw[fill=gray] (3,1) rectangle (4,2);
\draw[fill=gray] (4,1) rectangle (5,2);
\draw[fill=gray] (5,1) rectangle (6,2);
\draw[fill=gray] (6,1) rectangle (7,2);
\draw[fill=gray] (7,1) rectangle (8,2);
\draw[fill=gray] (8,1) rectangle (9,2);
\draw[fill=gray] (9,1) rectangle (10,2);
\draw[fill=gray] (10,1) rectangle (11,2);
\draw[fill=gray] (0,2) rectangle (1,3);
\draw[fill=gray] (1,2) rectangle (2,3);
\draw[fill={rgb,255:red,0.0;green,0.0;blue,255.0}]  (2,2) rectangle (3,3);
\draw[fill={rgb,255:red,255.0;green,241.0;blue,228.0}]  (2.250000,2.250000) rectangle (2.750000,2.750000);
\draw[fill=gray] (3,2) rectangle (4,3);
\draw[fill=gray] (4,2) rectangle (5,3);
\draw[fill=gray] (5,2) rectangle (6,3);
\draw[fill=gray] (6,2) rectangle (7,3);
\draw[fill=gray] (7,2) rectangle (8,3);
\draw[fill=gray] (8,2) rectangle (9,3);
\draw[fill=gray] (9,2) rectangle (10,3);
\draw[fill=gray] (10,2) rectangle (11,3);
\draw[fill=gray] (0,3) rectangle (1,4);
\draw[fill={rgb,255:red,0.0;green,0.0;blue,255.0}]  (2,3) rectangle (3,4);
\draw[fill={rgb,255:red,255.0;green,214.0;blue,174.0}]  (2.250000,3.250000) rectangle (2.750000,3.750000);
\draw[fill={rgb,255:red,0.0;green,0.0;blue,255.0}]  (3,3) rectangle (4,4);
\draw[fill={rgb,255:red,255.0;green,228.0;blue,201.0}]  (3.250000,3.250000) rectangle (3.750000,3.750000);
\draw[fill={rgb,255:red,0.0;green,0.0;blue,255.0}]  (4,3) rectangle (5,4);
\draw[fill={rgb,255:red,255.0;green,241.0;blue,228.0}]  (4.250000,3.250000) rectangle (4.750000,3.750000);
\draw[fill={rgb,255:red,0.0;green,0.0;blue,255.0}]  (5,3) rectangle (6,4);
\draw[fill={rgb,255:red,255.0;green,241.0;blue,228.0}]  (5.250000,3.250000) rectangle (5.750000,3.750000);
\draw[fill=gray] (6,3) rectangle (7,4);
\draw[fill=gray] (7,3) rectangle (8,4);
\draw[fill=gray] (8,3) rectangle (9,4);
\draw[fill=gray] (9,3) rectangle (10,4);
\draw[fill=gray] (10,3) rectangle (11,4);
\draw[fill=gray] (0,4) rectangle (1,5);
\draw[fill=gray] (2,4) rectangle (3,5);
\draw[fill=gray] (3,4) rectangle (4,5);
\draw[fill=gray] (4,4) rectangle (5,5);
\draw[fill={rgb,255:red,0.0;green,0.0;blue,255.0}]  (5,4) rectangle (6,5);
\draw[fill={rgb,255:red,255.0;green,228.0;blue,201.0}]  (5.250000,4.250000) rectangle (5.750000,4.750000);
\draw[fill=gray] (10,4) rectangle (11,5);
\draw[fill=gray] (0,5) rectangle (1,6);
\draw[fill=gray] (2,5) rectangle (3,6);
\draw[fill=gray] (3,5) rectangle (4,6);
\draw[fill=gray] (4,5) rectangle (5,6);
\draw[fill={rgb,255:red,0.0;green,0.0;blue,255.0}]  (5,5) rectangle (6,6);
\draw[fill=gray] (10,5) rectangle (11,6);
\draw[fill=gray] (0,6) rectangle (1,7);
\draw[fill=gray] (2,6) rectangle (3,7);
\draw[fill=gray] (3,6) rectangle (4,7);
\draw[fill=gray] (4,6) rectangle (5,7);
\draw[fill={rgb,255:red,0.0;green,0.0;blue,255.0}]  (5,6) rectangle (6,7);
\draw[fill=gray] (10,6) rectangle (11,7);
\draw[fill=gray] (0,7) rectangle (1,8);
\draw[fill=gray] (2,7) rectangle (3,8);
\draw[fill=gray] (3,7) rectangle (4,8);
\draw[fill=gray] (4,7) rectangle (5,8);
\draw[fill={rgb,255:red,0.0;green,0.0;blue,255.0}]  (5,7) rectangle (6,8);
\draw[fill={rgb,255:red,255.0;green,241.0;blue,228.0}]  (5.250000,7.250000) rectangle (5.750000,7.750000);
\draw[fill=gray] (10,7) rectangle (11,8);
\draw[fill=gray] (0,8) rectangle (1,9);
\draw[fill=gray] (2,8) rectangle (3,9);
\draw[fill=gray] (3,8) rectangle (4,9);
\draw[fill=gray] (4,8) rectangle (5,9);
\draw[fill={rgb,255:red,0.0;green,0.0;blue,255.0}]  (5,8) rectangle (6,9);
\draw[fill={rgb,255:red,255.0;green,241.0;blue,228.0}]  (5.250000,8.250000) rectangle (5.750000,8.750000);
\draw[fill={rgb,255:red,0.0;green,0.0;blue,255.0}]  (6,8) rectangle (7,9);
\draw[fill={rgb,255:red,255.0;green,214.0;blue,174.0}]  (6.250000,8.250000) rectangle (6.750000,8.750000);
\draw[fill={rgb,255:red,0.0;green,0.0;blue,255.0}]  (7,8) rectangle (8,9);
\draw[fill={rgb,255:red,255.0;green,214.0;blue,174.0}]  (7.250000,8.250000) rectangle (7.750000,8.750000);
\draw[fill={rgb,255:red,0.0;green,0.0;blue,255.0}]  (8,8) rectangle (9,9);
\draw[fill={rgb,255:red,255.0;green,228.0;blue,201.0}]  (8.250000,8.250000) rectangle (8.750000,8.750000);
\draw[fill={rgb,255:red,0.0;green,0.0;blue,255.0}]  (9,8) rectangle (10,9);
\draw[fill={rgb,255:red,255.0;green,214.0;blue,174.0}]  (9.250000,8.250000) rectangle (9.750000,8.750000);
\draw[fill=gray] (10,8) rectangle (11,9);
\draw[fill=gray] (0,9) rectangle (1,10);
\draw[fill=gray] (2,9) rectangle (3,10);
\draw[fill=gray] (3,9) rectangle (4,10);
\draw[fill=gray] (4,9) rectangle (5,10);
\draw[fill=gray] (5,9) rectangle (6,10);
\draw[fill=gray] (6,9) rectangle (7,10);
\draw[fill=gray] (7,9) rectangle (8,10);
\draw[fill=gray] (8,9) rectangle (9,10);
\draw[fill={rgb,255:red,0.0;green,0.0;blue,255.0}]  (9,9) rectangle (10,10);
\draw[fill={rgb,255:red,255.0;green,228.0;blue,201.0}]  (9.250000,9.250000) rectangle (9.750000,9.750000);
\draw[fill=gray] (10,9) rectangle (11,10);
\draw[fill=gray] (0,10) rectangle (1,11);
\draw[fill=pink] (9.500000,10.500000) circle (0.500000);
\draw[fill=gray] (10,10) rectangle (11,11);
\draw[fill=gray] (0,11) rectangle (1,12);
\draw[fill=gray] (1,11) rectangle (2,12);
\draw[fill=gray] (2,11) rectangle (3,12);
\draw[fill=gray] (3,11) rectangle (4,12);
\draw[fill=gray] (4,11) rectangle (5,12);
\draw[fill=gray] (5,11) rectangle (6,12);
\draw[fill=gray] (6,11) rectangle (7,12);
\draw[fill=gray] (7,11) rectangle (8,12);
\draw[fill=gray] (8,11) rectangle (9,12);
\draw[fill=gray] (9,11) rectangle (10,12);
\draw[fill=gray] (10,11) rectangle (11,12);
\node [align=center] at(-0.500000,0.500000) {0};
\node [align=center] at(-0.500000,1.500000) {1};
\node [align=center] at(-0.500000,2.500000) {2};
\node [align=center] at(-0.500000,3.500000) {3};
\node [align=center] at(-0.500000,4.500000) {4};
\node [align=center] at(-0.500000,5.500000) {5};
\node [align=center] at(-0.500000,6.500000) {6};
\node [align=center] at(-0.500000,7.500000) {7};
\node [align=center] at(-0.500000,8.500000) {8};
\node [align=center] at(-0.500000,9.500000) {9};
\node [align=center] at(-0.500000,10.500000) {10};
\node [align=center] at(-0.500000,11.500000) {11};
\node [align=center] at(0.500000,-0.500000) {A};
\node [align=center] at(1.500000,-0.500000) {B};
\node [align=center] at(2.500000,-0.500000) {C};
\node [align=center] at(3.500000,-0.500000) {D};
\node [align=center] at(4.500000,-0.500000) {E};
\node [align=center] at(5.500000,-0.500000) {F};
\node [align=center] at(6.500000,-0.500000) {G};
\node [align=center] at(7.500000,-0.500000) {H};
\node [align=center] at(8.500000,-0.500000) {I};
\node [align=center] at(9.500000,-0.500000) {J};
\node [align=center] at(10.500000,-0.500000) {K};
\end{tikzpicture}
      }
    \end{minipage}
    \hfill
    \begin{minipage}{\mpwidth}
      \centering
      \adjustbox{width=\aboxwidth}{
        \begin{tikzpicture}
\draw (0,0) grid (11,12);
\draw[fill=gray] (0,0) rectangle (1,1);
\draw[fill=gray] (1,0) rectangle (2,1);
\draw[fill=gray] (2,0) rectangle (3,1);
\draw[fill=gray] (3,0) rectangle (4,1);
\draw[fill=gray] (4,0) rectangle (5,1);
\draw[fill=gray] (5,0) rectangle (6,1);
\draw[fill=gray] (6,0) rectangle (7,1);
\draw[fill=gray] (7,0) rectangle (8,1);
\draw[fill=gray] (8,0) rectangle (9,1);
\draw[fill=gray] (9,0) rectangle (10,1);
\draw[fill=gray] (10,0) rectangle (11,1);
\draw[fill=gray] (0,1) rectangle (1,2);
\draw[fill=gray] (1,1) rectangle (2,2);
\draw[fill={rgb,255:red,0.0;green,0.0;blue,255.0}]  (2,1) rectangle (3,2);
\draw[fill=gray] (3,1) rectangle (4,2);
\draw[fill=gray] (4,1) rectangle (5,2);
\draw[fill=gray] (5,1) rectangle (6,2);
\draw[fill=gray] (6,1) rectangle (7,2);
\draw[fill=gray] (7,1) rectangle (8,2);
\draw[fill=gray] (8,1) rectangle (9,2);
\draw[fill=gray] (9,1) rectangle (10,2);
\draw[fill=gray] (10,1) rectangle (11,2);
\draw[fill=gray] (0,2) rectangle (1,3);
\draw[fill=gray] (1,2) rectangle (2,3);
\draw[fill={rgb,255:red,0.0;green,0.0;blue,255.0}]  (2,2) rectangle (3,3);
\draw[fill={rgb,255:red,255.0;green,241.0;blue,228.0}]  (2.250000,2.250000) rectangle (2.750000,2.750000);
\draw[fill=gray] (3,2) rectangle (4,3);
\draw[fill=gray] (4,2) rectangle (5,3);
\draw[fill=gray] (5,2) rectangle (6,3);
\draw[fill=gray] (6,2) rectangle (7,3);
\draw[fill=gray] (7,2) rectangle (8,3);
\draw[fill=gray] (8,2) rectangle (9,3);
\draw[fill=gray] (9,2) rectangle (10,3);
\draw[fill=gray] (10,2) rectangle (11,3);
\draw[fill=gray] (0,3) rectangle (1,4);
\draw[fill={rgb,255:red,0.0;green,0.0;blue,255.0}]  (1,3) rectangle (2,4);
\draw[fill={rgb,255:red,0.0;green,0.0;blue,255.0}]  (2,3) rectangle (3,4);
\draw[fill={rgb,255:red,255.0;green,241.0;blue,228.0}]  (2.250000,3.250000) rectangle (2.750000,3.750000);
\draw[fill=gray] (6,3) rectangle (7,4);
\draw[fill=gray] (7,3) rectangle (8,4);
\draw[fill=gray] (8,3) rectangle (9,4);
\draw[fill=gray] (9,3) rectangle (10,4);
\draw[fill=gray] (10,3) rectangle (11,4);
\draw[fill=gray] (0,4) rectangle (1,5);
\draw[fill={rgb,255:red,0.0;green,0.0;blue,255.0}]  (1,4) rectangle (2,5);
\draw[fill={rgb,255:red,255.0;green,201.0;blue,147.0}]  (1.250000,4.250000) rectangle (1.750000,4.750000);
\draw[fill=gray] (2,4) rectangle (3,5);
\draw[fill=gray] (3,4) rectangle (4,5);
\draw[fill=gray] (4,4) rectangle (5,5);
\draw[fill=gray] (10,4) rectangle (11,5);
\draw[fill=gray] (0,5) rectangle (1,6);
\draw[fill={rgb,255:red,0.0;green,0.0;blue,255.0}]  (1,5) rectangle (2,6);
\draw[fill={rgb,255:red,255.0;green,187.0;blue,120.0}]  (1.250000,5.250000) rectangle (1.750000,5.750000);
\draw[fill=gray] (2,5) rectangle (3,6);
\draw[fill=gray] (3,5) rectangle (4,6);
\draw[fill=gray] (4,5) rectangle (5,6);
\draw[fill=gray] (10,5) rectangle (11,6);
\draw[fill=gray] (0,6) rectangle (1,7);
\draw[fill={rgb,255:red,0.0;green,0.0;blue,255.0}]  (1,6) rectangle (2,7);
\draw[fill={rgb,255:red,255.0;green,201.0;blue,147.0}]  (1.250000,6.250000) rectangle (1.750000,6.750000);
\draw[fill=gray] (2,6) rectangle (3,7);
\draw[fill=gray] (3,6) rectangle (4,7);
\draw[fill=gray] (4,6) rectangle (5,7);
\draw[fill=gray] (10,6) rectangle (11,7);
\draw[fill=gray] (0,7) rectangle (1,8);
\draw[fill={rgb,255:red,0.0;green,0.0;blue,255.0}]  (1,7) rectangle (2,8);
\draw[fill={rgb,255:red,255.0;green,161.0;blue,67.0}]  (1.250000,7.250000) rectangle (1.750000,7.750000);
\draw[fill=gray] (2,7) rectangle (3,8);
\draw[fill=gray] (3,7) rectangle (4,8);
\draw[fill=gray] (4,7) rectangle (5,8);
\draw[fill=gray] (10,7) rectangle (11,8);
\draw[fill=gray] (0,8) rectangle (1,9);
\draw[fill={rgb,255:red,0.0;green,0.0;blue,255.0}]  (1,8) rectangle (2,9);
\draw[fill={rgb,255:red,255.0;green,161.0;blue,67.0}]  (1.250000,8.250000) rectangle (1.750000,8.750000);
\draw[fill=gray] (2,8) rectangle (3,9);
\draw[fill=gray] (3,8) rectangle (4,9);
\draw[fill=gray] (4,8) rectangle (5,9);
\draw[fill=gray] (10,8) rectangle (11,9);
\draw[fill=gray] (0,9) rectangle (1,10);
\draw[fill={rgb,255:red,0.0;green,0.0;blue,255.0}]  (1,9) rectangle (2,10);
\draw[fill={rgb,255:red,255.0;green,228.0;blue,201.0}]  (1.250000,9.250000) rectangle (1.750000,9.750000);
\draw[fill=gray] (2,9) rectangle (3,10);
\draw[fill=gray] (3,9) rectangle (4,10);
\draw[fill=gray] (4,9) rectangle (5,10);
\draw[fill=gray] (5,9) rectangle (6,10);
\draw[fill=gray] (6,9) rectangle (7,10);
\draw[fill=gray] (7,9) rectangle (8,10);
\draw[fill=gray] (8,9) rectangle (9,10);
\draw[fill=gray] (10,9) rectangle (11,10);
\draw[fill=gray] (0,10) rectangle (1,11);
\draw[fill={rgb,255:red,0.0;green,0.0;blue,255.0}]  (1,10) rectangle (2,11);
\draw[fill={rgb,255:red,255.0;green,201.0;blue,147.0}]  (1.250000,10.250000) rectangle (1.750000,10.750000);
\draw[fill={rgb,255:red,0.0;green,0.0;blue,255.0}]  (2,10) rectangle (3,11);
\draw[fill={rgb,255:red,255.0;green,174.0;blue,93.0}]  (2.250000,10.250000) rectangle (2.750000,10.750000);
\draw[fill={rgb,255:red,0.0;green,0.0;blue,255.0}]  (3,10) rectangle (4,11);
\draw[fill={rgb,255:red,255.0;green,161.0;blue,67.0}]  (3.250000,10.250000) rectangle (3.750000,10.750000);
\draw[fill={rgb,255:red,0.0;green,0.0;blue,255.0}]  (4,10) rectangle (5,11);
\draw[fill={rgb,255:red,255.0;green,147.0;blue,40.0}]  (4.250000,10.250000) rectangle (4.750000,10.750000);
\draw[fill={rgb,255:red,0.0;green,0.0;blue,255.0}]  (5,10) rectangle (6,11);
\draw[fill={rgb,255:red,255.0;green,161.0;blue,67.0}]  (5.250000,10.250000) rectangle (5.750000,10.750000);
\draw[fill={rgb,255:red,0.0;green,0.0;blue,255.0}]  (6,10) rectangle (7,11);
\draw[fill={rgb,255:red,255.0;green,187.0;blue,120.0}]  (6.250000,10.250000) rectangle (6.750000,10.750000);
\draw[fill={rgb,255:red,0.0;green,0.0;blue,255.0}]  (7,10) rectangle (8,11);
\draw[fill={rgb,255:red,255.0;green,161.0;blue,67.0}]  (7.250000,10.250000) rectangle (7.750000,10.750000);
\draw[fill={rgb,255:red,0.0;green,0.0;blue,255.0}]  (8,10) rectangle (9,11);
\draw[fill={rgb,255:red,255.0;green,201.0;blue,147.0}]  (8.250000,10.250000) rectangle (8.750000,10.750000);
\draw[fill=pink] (9.500000,10.500000) circle (0.500000);
\draw[fill=gray] (10,10) rectangle (11,11);
\draw[fill=gray] (0,11) rectangle (1,12);
\draw[fill=gray] (1,11) rectangle (2,12);
\draw[fill=gray] (2,11) rectangle (3,12);
\draw[fill=gray] (3,11) rectangle (4,12);
\draw[fill=gray] (4,11) rectangle (5,12);
\draw[fill=gray] (5,11) rectangle (6,12);
\draw[fill=gray] (6,11) rectangle (7,12);
\draw[fill=gray] (7,11) rectangle (8,12);
\draw[fill=gray] (8,11) rectangle (9,12);
\draw[fill=gray] (9,11) rectangle (10,12);
\draw[fill=gray] (10,11) rectangle (11,12);
\node [align=center] at(-0.500000,0.500000) {0};
\node [align=center] at(-0.500000,1.500000) {1};
\node [align=center] at(-0.500000,2.500000) {2};
\node [align=center] at(-0.500000,3.500000) {3};
\node [align=center] at(-0.500000,4.500000) {4};
\node [align=center] at(-0.500000,5.500000) {5};
\node [align=center] at(-0.500000,6.500000) {6};
\node [align=center] at(-0.500000,7.500000) {7};
\node [align=center] at(-0.500000,8.500000) {8};
\node [align=center] at(-0.500000,9.500000) {9};
\node [align=center] at(-0.500000,10.500000) {10};
\node [align=center] at(-0.500000,11.500000) {11};
\node [align=center] at(0.500000,-0.500000) {A};
\node [align=center] at(1.500000,-0.500000) {B};
\node [align=center] at(2.500000,-0.500000) {C};
\node [align=center] at(3.500000,-0.500000) {D};
\node [align=center] at(4.500000,-0.500000) {E};
\node [align=center] at(5.500000,-0.500000) {F};
\node [align=center] at(6.500000,-0.500000) {G};
\node [align=center] at(7.500000,-0.500000) {H};
\node [align=center] at(8.500000,-0.500000) {I};
\node [align=center] at(9.500000,-0.500000) {J};
\node [align=center] at(10.500000,-0.500000) {K};
\end{tikzpicture}
      }
    \end{minipage}

    \centerline{\small $M_{1}$}
  \end{minipage}

  \hfill
  
  \begin{minipage}{\colwidth}
    \begin{minipage}{\mpwidth}
      \centering
      \adjustbox{width=\aboxwidth}{
        \input{\dir 5_MDP_policyStochHeatmap.tikz}
      }
    \end{minipage}
    \hfill
    \begin{minipage}{\mpwidth}
      \centering
      \adjustbox{width=\aboxwidth}{
        \begin{tikzpicture}
\draw (0,0) grid (11,13);
\draw[fill=gray] (0,0) rectangle (1,1);
\draw[fill=gray] (1,0) rectangle (2,1);
\draw[fill=gray] (2,0) rectangle (3,1);
\draw[fill=gray] (3,0) rectangle (4,1);
\draw[fill=gray] (4,0) rectangle (5,1);
\draw[fill=gray] (5,0) rectangle (6,1);
\draw[fill=gray] (6,0) rectangle (7,1);
\draw[fill=gray] (7,0) rectangle (8,1);
\draw[fill=gray] (8,0) rectangle (9,1);
\draw[fill=gray] (9,0) rectangle (10,1);
\draw[fill=gray] (10,0) rectangle (11,1);
\draw[fill=gray] (0,1) rectangle (1,2);
\draw[fill=pink] (9.500000,1.500000) circle (0.500000);
\draw[fill=gray] (10,1) rectangle (11,2);
\draw[fill=gray] (0,2) rectangle (1,3);
\draw[fill=gray] (2,2) rectangle (3,3);
\draw[fill=gray] (3,2) rectangle (4,3);
\draw[fill=gray] (4,2) rectangle (5,3);
\draw[fill=gray] (5,2) rectangle (6,3);
\draw[fill=gray] (6,2) rectangle (7,3);
\draw[fill=gray] (7,2) rectangle (8,3);
\draw[fill=gray] (8,2) rectangle (9,3);
\draw[fill={rgb,255:red,0.0;green,0.0;blue,255.0}]  (9,2) rectangle (10,3);
\draw[fill={rgb,255:red,255.0;green,214.0;blue,174.0}]  (9.250000,2.250000) rectangle (9.750000,2.750000);
\draw[fill=gray] (10,2) rectangle (11,3);
\draw[fill=gray] (0,3) rectangle (1,4);
\draw[fill=gray] (2,3) rectangle (3,4);
\draw[fill=gray] (3,3) rectangle (4,4);
\draw[fill=gray] (4,3) rectangle (5,4);
\draw[fill={rgb,255:red,0.0;green,0.0;blue,255.0}]  (9,3) rectangle (10,4);
\draw[fill={rgb,255:red,255.0;green,187.0;blue,120.0}]  (9.250000,3.250000) rectangle (9.750000,3.750000);
\draw[fill=gray] (10,3) rectangle (11,4);
\draw[fill=gray] (0,4) rectangle (1,5);
\draw[fill=gray] (2,4) rectangle (3,5);
\draw[fill=gray] (3,4) rectangle (4,5);
\draw[fill=gray] (4,4) rectangle (5,5);
\draw[fill={rgb,255:red,0.0;green,0.0;blue,255.0}]  (9,4) rectangle (10,5);
\draw[fill={rgb,255:red,255.0;green,174.0;blue,93.0}]  (9.250000,4.250000) rectangle (9.750000,4.750000);
\draw[fill=gray] (10,4) rectangle (11,5);
\draw[fill=gray] (0,5) rectangle (1,6);
\draw[fill=gray] (2,5) rectangle (3,6);
\draw[fill=gray] (3,5) rectangle (4,6);
\draw[fill=gray] (4,5) rectangle (5,6);
\draw[fill={rgb,255:red,0.0;green,0.0;blue,255.0}]  (9,5) rectangle (10,6);
\draw[fill={rgb,255:red,255.0;green,134.0;blue,13.0}]  (9.250000,5.250000) rectangle (9.750000,5.750000);
\draw[fill=gray] (10,5) rectangle (11,6);
\draw[fill=gray] (0,6) rectangle (1,7);
\draw[fill=gray] (2,6) rectangle (3,7);
\draw[fill=gray] (3,6) rectangle (4,7);
\draw[fill=gray] (4,6) rectangle (5,7);
\draw[fill={rgb,255:red,0.0;green,0.0;blue,255.0}]  (9,6) rectangle (10,7);
\draw[fill={rgb,255:red,255.0;green,161.0;blue,67.0}]  (9.250000,6.250000) rectangle (9.750000,6.750000);
\draw[fill=gray] (10,6) rectangle (11,7);
\draw[fill=gray] (0,7) rectangle (1,8);
\draw[fill=gray] (2,7) rectangle (3,8);
\draw[fill=gray] (3,7) rectangle (4,8);
\draw[fill=gray] (4,7) rectangle (5,8);
\draw[fill={rgb,255:red,0.0;green,0.0;blue,255.0}]  (9,7) rectangle (10,8);
\draw[fill={rgb,255:red,255.0;green,228.0;blue,201.0}]  (9.250000,7.250000) rectangle (9.750000,7.750000);
\draw[fill=gray] (10,7) rectangle (11,8);
\draw[fill=gray] (0,8) rectangle (1,9);
\draw[fill=gray] (2,8) rectangle (3,9);
\draw[fill=gray] (3,8) rectangle (4,9);
\draw[fill=gray] (4,8) rectangle (5,9);
\draw[fill={rgb,255:red,0.0;green,0.0;blue,255.0}]  (5,8) rectangle (6,9);
\draw[fill={rgb,255:red,255.0;green,241.0;blue,228.0}]  (5.250000,8.250000) rectangle (5.750000,8.750000);
\draw[fill={rgb,255:red,0.0;green,0.0;blue,255.0}]  (6,8) rectangle (7,9);
\draw[fill={rgb,255:red,0.0;green,0.0;blue,255.0}]  (7,8) rectangle (8,9);
\draw[fill={rgb,255:red,0.0;green,0.0;blue,255.0}]  (8,8) rectangle (9,9);
\draw[fill={rgb,255:red,255.0;green,228.0;blue,201.0}]  (8.250000,8.250000) rectangle (8.750000,8.750000);
\draw[fill={rgb,255:red,0.0;green,0.0;blue,255.0}]  (9,8) rectangle (10,9);
\draw[fill={rgb,255:red,255.0;green,228.0;blue,201.0}]  (9.250000,8.250000) rectangle (9.750000,8.750000);
\draw[fill=gray] (10,8) rectangle (11,9);
\draw[fill=gray] (0,9) rectangle (1,10);
\draw[fill={rgb,255:red,0.0;green,0.0;blue,255.0}]  (5,9) rectangle (6,10);
\draw[fill={rgb,255:red,255.0;green,241.0;blue,228.0}]  (5.250000,9.250000) rectangle (5.750000,9.750000);
\draw[fill=gray] (6,9) rectangle (7,10);
\draw[fill=gray] (7,9) rectangle (8,10);
\draw[fill=gray] (8,9) rectangle (9,10);
\draw[fill=gray] (9,9) rectangle (10,10);
\draw[fill=gray] (10,9) rectangle (11,10);
\draw[fill=gray] (0,10) rectangle (1,11);
\draw[fill=gray] (1,10) rectangle (2,11);
\draw[fill=gray] (2,10) rectangle (3,11);
\draw[fill=gray] (3,10) rectangle (4,11);
\draw[fill=gray] (4,10) rectangle (5,11);
\draw[fill={rgb,255:red,0.0;green,0.0;blue,255.0}]  (5,10) rectangle (6,11);
\draw[fill={rgb,255:red,255.0;green,241.0;blue,228.0}]  (5.250000,10.250000) rectangle (5.750000,10.750000);
\draw[fill=gray] (6,10) rectangle (7,11);
\draw[fill=gray] (7,10) rectangle (8,11);
\draw[fill=gray] (8,10) rectangle (9,11);
\draw[fill=gray] (9,10) rectangle (10,11);
\draw[fill=gray] (10,10) rectangle (11,11);
\draw[fill=gray] (0,11) rectangle (1,12);
\draw[fill=gray] (1,11) rectangle (2,12);
\draw[fill=gray] (2,11) rectangle (3,12);
\draw[fill=gray] (3,11) rectangle (4,12);
\draw[fill=gray] (4,11) rectangle (5,12);
\draw[fill={rgb,255:red,0.0;green,0.0;blue,255.0}]  (5,11) rectangle (6,12);
\draw[fill=gray] (6,11) rectangle (7,12);
\draw[fill=gray] (7,11) rectangle (8,12);
\draw[fill=gray] (8,11) rectangle (9,12);
\draw[fill=gray] (9,11) rectangle (10,12);
\draw[fill=gray] (10,11) rectangle (11,12);
\draw[fill=gray] (0,12) rectangle (1,13);
\draw[fill=gray] (1,12) rectangle (2,13);
\draw[fill=gray] (2,12) rectangle (3,13);
\draw[fill=gray] (3,12) rectangle (4,13);
\draw[fill=gray] (4,12) rectangle (5,13);
\draw[fill=gray] (5,12) rectangle (6,13);
\draw[fill=gray] (6,12) rectangle (7,13);
\draw[fill=gray] (7,12) rectangle (8,13);
\draw[fill=gray] (8,12) rectangle (9,13);
\draw[fill=gray] (9,12) rectangle (10,13);
\draw[fill=gray] (10,12) rectangle (11,13);
\node [align=center] at(-0.500000,0.500000) {0};
\node [align=center] at(-0.500000,1.500000) {1};
\node [align=center] at(-0.500000,2.500000) {2};
\node [align=center] at(-0.500000,3.500000) {3};
\node [align=center] at(-0.500000,4.500000) {4};
\node [align=center] at(-0.500000,5.500000) {5};
\node [align=center] at(-0.500000,6.500000) {6};
\node [align=center] at(-0.500000,7.500000) {7};
\node [align=center] at(-0.500000,8.500000) {8};
\node [align=center] at(-0.500000,9.500000) {9};
\node [align=center] at(-0.500000,10.500000) {10};
\node [align=center] at(-0.500000,11.500000) {11};
\node [align=center] at(-0.500000,12.500000) {12};
\node [align=center] at(0.500000,-0.500000) {A};
\node [align=center] at(1.500000,-0.500000) {B};
\node [align=center] at(2.500000,-0.500000) {C};
\node [align=center] at(3.500000,-0.500000) {D};
\node [align=center] at(4.500000,-0.500000) {E};
\node [align=center] at(5.500000,-0.500000) {F};
\node [align=center] at(6.500000,-0.500000) {G};
\node [align=center] at(7.500000,-0.500000) {H};
\node [align=center] at(8.500000,-0.500000) {I};
\node [align=center] at(9.500000,-0.500000) {J};
\node [align=center] at(10.500000,-0.500000) {K};
\end{tikzpicture}
      }
    \end{minipage}
    \hfill
    \begin{minipage}{\mpwidth}
      \centering
      \adjustbox{width=\aboxwidth}{
        \begin{tikzpicture}
\draw (0,0) grid (11,13);
\draw[fill=gray] (0,0) rectangle (1,1);
\draw[fill=gray] (1,0) rectangle (2,1);
\draw[fill=gray] (2,0) rectangle (3,1);
\draw[fill=gray] (3,0) rectangle (4,1);
\draw[fill=gray] (4,0) rectangle (5,1);
\draw[fill=gray] (5,0) rectangle (6,1);
\draw[fill=gray] (6,0) rectangle (7,1);
\draw[fill=gray] (7,0) rectangle (8,1);
\draw[fill=gray] (8,0) rectangle (9,1);
\draw[fill=gray] (9,0) rectangle (10,1);
\draw[fill=gray] (10,0) rectangle (11,1);
\draw[fill=gray] (0,1) rectangle (1,2);
\draw[fill=pink] (9.500000,1.500000) circle (0.500000);
\draw[fill=gray] (10,1) rectangle (11,2);
\draw[fill=gray] (0,2) rectangle (1,3);
\draw[fill=gray] (2,2) rectangle (3,3);
\draw[fill=gray] (3,2) rectangle (4,3);
\draw[fill=gray] (4,2) rectangle (5,3);
\draw[fill=gray] (5,2) rectangle (6,3);
\draw[fill=gray] (6,2) rectangle (7,3);
\draw[fill=gray] (7,2) rectangle (8,3);
\draw[fill=gray] (8,2) rectangle (9,3);
\draw[fill={rgb,255:red,0.0;green,0.0;blue,255.0}]  (9,2) rectangle (10,3);
\draw[fill={rgb,255:red,255.0;green,214.0;blue,174.0}]  (9.250000,2.250000) rectangle (9.750000,2.750000);
\draw[fill=gray] (10,2) rectangle (11,3);
\draw[fill=gray] (0,3) rectangle (1,4);
\draw[fill=gray] (2,3) rectangle (3,4);
\draw[fill=gray] (3,3) rectangle (4,4);
\draw[fill=gray] (4,3) rectangle (5,4);
\draw[fill={rgb,255:red,0.0;green,0.0;blue,255.0}]  (9,3) rectangle (10,4);
\draw[fill={rgb,255:red,255.0;green,161.0;blue,67.0}]  (9.250000,3.250000) rectangle (9.750000,3.750000);
\draw[fill=gray] (10,3) rectangle (11,4);
\draw[fill=gray] (0,4) rectangle (1,5);
\draw[fill=gray] (2,4) rectangle (3,5);
\draw[fill=gray] (3,4) rectangle (4,5);
\draw[fill=gray] (4,4) rectangle (5,5);
\draw[fill={rgb,255:red,0.0;green,0.0;blue,255.0}]  (9,4) rectangle (10,5);
\draw[fill={rgb,255:red,255.0;green,174.0;blue,93.0}]  (9.250000,4.250000) rectangle (9.750000,4.750000);
\draw[fill=gray] (10,4) rectangle (11,5);
\draw[fill=gray] (0,5) rectangle (1,6);
\draw[fill=gray] (2,5) rectangle (3,6);
\draw[fill=gray] (3,5) rectangle (4,6);
\draw[fill=gray] (4,5) rectangle (5,6);
\draw[fill={rgb,255:red,0.0;green,0.0;blue,255.0}]  (9,5) rectangle (10,6);
\draw[fill={rgb,255:red,255.0;green,187.0;blue,120.0}]  (9.250000,5.250000) rectangle (9.750000,5.750000);
\draw[fill=gray] (10,5) rectangle (11,6);
\draw[fill=gray] (0,6) rectangle (1,7);
\draw[fill=gray] (2,6) rectangle (3,7);
\draw[fill=gray] (3,6) rectangle (4,7);
\draw[fill=gray] (4,6) rectangle (5,7);
\draw[fill={rgb,255:red,0.0;green,0.0;blue,255.0}]  (9,6) rectangle (10,7);
\draw[fill={rgb,255:red,255.0;green,161.0;blue,67.0}]  (9.250000,6.250000) rectangle (9.750000,6.750000);
\draw[fill=gray] (10,6) rectangle (11,7);
\draw[fill=gray] (0,7) rectangle (1,8);
\draw[fill=gray] (2,7) rectangle (3,8);
\draw[fill=gray] (3,7) rectangle (4,8);
\draw[fill=gray] (4,7) rectangle (5,8);
\draw[fill={rgb,255:red,0.0;green,0.0;blue,255.0}]  (9,7) rectangle (10,8);
\draw[fill={rgb,255:red,255.0;green,161.0;blue,67.0}]  (9.250000,7.250000) rectangle (9.750000,7.750000);
\draw[fill=gray] (10,7) rectangle (11,8);
\draw[fill=gray] (0,8) rectangle (1,9);
\draw[fill=gray] (2,8) rectangle (3,9);
\draw[fill=gray] (3,8) rectangle (4,9);
\draw[fill=gray] (4,8) rectangle (5,9);
\draw[fill={rgb,255:red,0.0;green,0.0;blue,255.0}]  (5,8) rectangle (6,9);
\draw[fill={rgb,255:red,0.0;green,0.0;blue,255.0}]  (6,8) rectangle (7,9);
\draw[fill={rgb,255:red,0.0;green,0.0;blue,255.0}]  (7,8) rectangle (8,9);
\draw[fill={rgb,255:red,0.0;green,0.0;blue,255.0}]  (8,8) rectangle (9,9);
\draw[fill={rgb,255:red,255.0;green,241.0;blue,228.0}]  (8.250000,8.250000) rectangle (8.750000,8.750000);
\draw[fill={rgb,255:red,0.0;green,0.0;blue,255.0}]  (9,8) rectangle (10,9);
\draw[fill={rgb,255:red,255.0;green,228.0;blue,201.0}]  (9.250000,8.250000) rectangle (9.750000,8.750000);
\draw[fill=gray] (10,8) rectangle (11,9);
\draw[fill=gray] (0,9) rectangle (1,10);
\draw[fill={rgb,255:red,0.0;green,0.0;blue,255.0}]  (5,9) rectangle (6,10);
\draw[fill=gray] (6,9) rectangle (7,10);
\draw[fill=gray] (7,9) rectangle (8,10);
\draw[fill=gray] (8,9) rectangle (9,10);
\draw[fill=gray] (9,9) rectangle (10,10);
\draw[fill=gray] (10,9) rectangle (11,10);
\draw[fill=gray] (0,10) rectangle (1,11);
\draw[fill=gray] (1,10) rectangle (2,11);
\draw[fill=gray] (2,10) rectangle (3,11);
\draw[fill=gray] (3,10) rectangle (4,11);
\draw[fill=gray] (4,10) rectangle (5,11);
\draw[fill={rgb,255:red,0.0;green,0.0;blue,255.0}]  (5,10) rectangle (6,11);
\draw[fill=gray] (6,10) rectangle (7,11);
\draw[fill=gray] (7,10) rectangle (8,11);
\draw[fill=gray] (8,10) rectangle (9,11);
\draw[fill=gray] (9,10) rectangle (10,11);
\draw[fill=gray] (10,10) rectangle (11,11);
\draw[fill=gray] (0,11) rectangle (1,12);
\draw[fill=gray] (1,11) rectangle (2,12);
\draw[fill=gray] (2,11) rectangle (3,12);
\draw[fill=gray] (3,11) rectangle (4,12);
\draw[fill=gray] (4,11) rectangle (5,12);
\draw[fill={rgb,255:red,0.0;green,0.0;blue,255.0}]  (5,11) rectangle (6,12);
\draw[fill=gray] (6,11) rectangle (7,12);
\draw[fill=gray] (7,11) rectangle (8,12);
\draw[fill=gray] (8,11) rectangle (9,12);
\draw[fill=gray] (9,11) rectangle (10,12);
\draw[fill=gray] (10,11) rectangle (11,12);
\draw[fill=gray] (0,12) rectangle (1,13);
\draw[fill=gray] (1,12) rectangle (2,13);
\draw[fill=gray] (2,12) rectangle (3,13);
\draw[fill=gray] (3,12) rectangle (4,13);
\draw[fill=gray] (4,12) rectangle (5,13);
\draw[fill=gray] (5,12) rectangle (6,13);
\draw[fill=gray] (6,12) rectangle (7,13);
\draw[fill=gray] (7,12) rectangle (8,13);
\draw[fill=gray] (8,12) rectangle (9,13);
\draw[fill=gray] (9,12) rectangle (10,13);
\draw[fill=gray] (10,12) rectangle (11,13);
\node [align=center] at(-0.500000,0.500000) {0};
\node [align=center] at(-0.500000,1.500000) {1};
\node [align=center] at(-0.500000,2.500000) {2};
\node [align=center] at(-0.500000,3.500000) {3};
\node [align=center] at(-0.500000,4.500000) {4};
\node [align=center] at(-0.500000,5.500000) {5};
\node [align=center] at(-0.500000,6.500000) {6};
\node [align=center] at(-0.500000,7.500000) {7};
\node [align=center] at(-0.500000,8.500000) {8};
\node [align=center] at(-0.500000,9.500000) {9};
\node [align=center] at(-0.500000,10.500000) {10};
\node [align=center] at(-0.500000,11.500000) {11};
\node [align=center] at(-0.500000,12.500000) {12};
\node [align=center] at(0.500000,-0.500000) {A};
\node [align=center] at(1.500000,-0.500000) {B};
\node [align=center] at(2.500000,-0.500000) {C};
\node [align=center] at(3.500000,-0.500000) {D};
\node [align=center] at(4.500000,-0.500000) {E};
\node [align=center] at(5.500000,-0.500000) {F};
\node [align=center] at(6.500000,-0.500000) {G};
\node [align=center] at(7.500000,-0.500000) {H};
\node [align=center] at(8.500000,-0.500000) {I};
\node [align=center] at(9.500000,-0.500000) {J};
\node [align=center] at(10.500000,-0.500000) {K};
\end{tikzpicture}
      }
    \end{minipage}

    \centerline{\small $M_{2}$}
  \end{minipage}

  \medskip

  \begin{minipage}{\colwidth}
    \begin{minipage}{\mpwidth}
      \centering
      \adjustbox{width=\aboxwidth}{
        \input{\dir 2_MDP_policyStochHeatmap.tikz}
      }
    \end{minipage}
    \hfill
    \begin{minipage}{\mpwidth}
      \centering
      \adjustbox{width=\aboxwidth}{
        \begin{tikzpicture}
\draw (0,0) grid (11,11);
\draw[fill=gray] (0,0) rectangle (1,1);
\draw[fill=gray] (1,0) rectangle (2,1);
\draw[fill=gray] (2,0) rectangle (3,1);
\draw[fill=gray] (3,0) rectangle (4,1);
\draw[fill=gray] (4,0) rectangle (5,1);
\draw[fill=gray] (5,0) rectangle (6,1);
\draw[fill=gray] (6,0) rectangle (7,1);
\draw[fill=gray] (7,0) rectangle (8,1);
\draw[fill=gray] (8,0) rectangle (9,1);
\draw[fill=gray] (9,0) rectangle (10,1);
\draw[fill=gray] (10,0) rectangle (11,1);
\draw[fill=gray] (0,1) rectangle (1,2);
\draw[fill=pink] (9.500000,1.500000) circle (0.500000);
\draw[fill=gray] (10,1) rectangle (11,2);
\draw[fill=gray] (0,2) rectangle (1,3);
\draw[fill=gray] (5,2) rectangle (6,3);
\draw[fill=gray] (6,2) rectangle (7,3);
\draw[fill=gray] (7,2) rectangle (8,3);
\draw[fill=gray] (8,2) rectangle (9,3);
\draw[fill={rgb,255:red,0.0;green,0.0;blue,255.0}]  (9,2) rectangle (10,3);
\draw[fill={rgb,255:red,255.0;green,241.0;blue,228.0}]  (9.250000,2.250000) rectangle (9.750000,2.750000);
\draw[fill=gray] (10,2) rectangle (11,3);
\draw[fill=gray] (0,3) rectangle (1,4);
\draw[fill=gray] (5,3) rectangle (6,4);
\draw[fill={rgb,255:red,0.0;green,0.0;blue,255.0}]  (9,3) rectangle (10,4);
\draw[fill={rgb,255:red,255.0;green,201.0;blue,147.0}]  (9.250000,3.250000) rectangle (9.750000,3.750000);
\draw[fill=gray] (10,3) rectangle (11,4);
\draw[fill=gray] (0,4) rectangle (1,5);
\draw[fill=gray] (5,4) rectangle (6,5);
\draw[fill={rgb,255:red,0.0;green,0.0;blue,255.0}]  (9,4) rectangle (10,5);
\draw[fill={rgb,255:red,255.0;green,174.0;blue,93.0}]  (9.250000,4.250000) rectangle (9.750000,4.750000);
\draw[fill=gray] (10,4) rectangle (11,5);
\draw[fill=gray] (0,5) rectangle (1,6);
\draw[fill=gray] (5,5) rectangle (6,6);
\draw[fill={rgb,255:red,0.0;green,0.0;blue,255.0}]  (9,5) rectangle (10,6);
\draw[fill={rgb,255:red,255.0;green,187.0;blue,120.0}]  (9.250000,5.250000) rectangle (9.750000,5.750000);
\draw[fill=gray] (10,5) rectangle (11,6);
\draw[fill=gray] (0,6) rectangle (1,7);
\draw[fill=gray] (2,6) rectangle (3,7);
\draw[fill=gray] (3,6) rectangle (4,7);
\draw[fill=gray] (4,6) rectangle (5,7);
\draw[fill=gray] (5,6) rectangle (6,7);
\draw[fill={rgb,255:red,0.0;green,0.0;blue,255.0}]  (9,6) rectangle (10,7);
\draw[fill={rgb,255:red,255.0;green,201.0;blue,147.0}]  (9.250000,6.250000) rectangle (9.750000,6.750000);
\draw[fill=gray] (10,6) rectangle (11,7);
\draw[fill=gray] (0,7) rectangle (1,8);
\draw[fill={rgb,255:red,0.0;green,0.0;blue,255.0}]  (1,7) rectangle (2,8);
\draw[fill={rgb,255:red,255.0;green,214.0;blue,174.0}]  (1.250000,7.250000) rectangle (1.750000,7.750000);
\draw[fill={rgb,255:red,0.0;green,0.0;blue,255.0}]  (2,7) rectangle (3,8);
\draw[fill={rgb,255:red,0.0;green,0.0;blue,255.0}]  (3,7) rectangle (4,8);
\draw[fill={rgb,255:red,255.0;green,228.0;blue,201.0}]  (3.250000,7.250000) rectangle (3.750000,7.750000);
\draw[fill={rgb,255:red,0.0;green,0.0;blue,255.0}]  (4,7) rectangle (5,8);
\draw[fill={rgb,255:red,255.0;green,241.0;blue,228.0}]  (4.250000,7.250000) rectangle (4.750000,7.750000);
\draw[fill={rgb,255:red,0.0;green,0.0;blue,255.0}]  (5,7) rectangle (6,8);
\draw[fill={rgb,255:red,255.0;green,201.0;blue,147.0}]  (5.250000,7.250000) rectangle (5.750000,7.750000);
\draw[fill={rgb,255:red,0.0;green,0.0;blue,255.0}]  (6,7) rectangle (7,8);
\draw[fill={rgb,255:red,255.0;green,214.0;blue,174.0}]  (6.250000,7.250000) rectangle (6.750000,7.750000);
\draw[fill={rgb,255:red,0.0;green,0.0;blue,255.0}]  (7,7) rectangle (8,8);
\draw[fill={rgb,255:red,255.0;green,228.0;blue,201.0}]  (7.250000,7.250000) rectangle (7.750000,7.750000);
\draw[fill={rgb,255:red,0.0;green,0.0;blue,255.0}]  (8,7) rectangle (9,8);
\draw[fill={rgb,255:red,255.0;green,241.0;blue,228.0}]  (8.250000,7.250000) rectangle (8.750000,7.750000);
\draw[fill={rgb,255:red,0.0;green,0.0;blue,255.0}]  (9,7) rectangle (10,8);
\draw[fill=gray] (10,7) rectangle (11,8);
\draw[fill=gray] (0,8) rectangle (1,9);
\draw[fill={rgb,255:red,0.0;green,0.0;blue,255.0}]  (1,8) rectangle (2,9);
\draw[fill=gray] (2,8) rectangle (3,9);
\draw[fill=gray] (3,8) rectangle (4,9);
\draw[fill=gray] (4,8) rectangle (5,9);
\draw[fill=gray] (5,8) rectangle (6,9);
\draw[fill=gray] (6,8) rectangle (7,9);
\draw[fill=gray] (7,8) rectangle (8,9);
\draw[fill=gray] (8,8) rectangle (9,9);
\draw[fill=gray] (9,8) rectangle (10,9);
\draw[fill=gray] (10,8) rectangle (11,9);
\draw[fill=gray] (0,9) rectangle (1,10);
\draw[fill={rgb,255:red,0.0;green,0.0;blue,255.0}]  (1,9) rectangle (2,10);
\draw[fill=gray] (2,9) rectangle (3,10);
\draw[fill=gray] (3,9) rectangle (4,10);
\draw[fill=gray] (4,9) rectangle (5,10);
\draw[fill=gray] (5,9) rectangle (6,10);
\draw[fill=gray] (6,9) rectangle (7,10);
\draw[fill=gray] (7,9) rectangle (8,10);
\draw[fill=gray] (8,9) rectangle (9,10);
\draw[fill=gray] (9,9) rectangle (10,10);
\draw[fill=gray] (10,9) rectangle (11,10);
\draw[fill=gray] (0,10) rectangle (1,11);
\draw[fill=gray] (1,10) rectangle (2,11);
\draw[fill=gray] (2,10) rectangle (3,11);
\draw[fill=gray] (3,10) rectangle (4,11);
\draw[fill=gray] (4,10) rectangle (5,11);
\draw[fill=gray] (5,10) rectangle (6,11);
\draw[fill=gray] (6,10) rectangle (7,11);
\draw[fill=gray] (7,10) rectangle (8,11);
\draw[fill=gray] (8,10) rectangle (9,11);
\draw[fill=gray] (9,10) rectangle (10,11);
\draw[fill=gray] (10,10) rectangle (11,11);
\node [align=center] at(-0.500000,0.500000) {0};
\node [align=center] at(-0.500000,1.500000) {1};
\node [align=center] at(-0.500000,2.500000) {2};
\node [align=center] at(-0.500000,3.500000) {3};
\node [align=center] at(-0.500000,4.500000) {4};
\node [align=center] at(-0.500000,5.500000) {5};
\node [align=center] at(-0.500000,6.500000) {6};
\node [align=center] at(-0.500000,7.500000) {7};
\node [align=center] at(-0.500000,8.500000) {8};
\node [align=center] at(-0.500000,9.500000) {9};
\node [align=center] at(-0.500000,10.500000) {10};
\node [align=center] at(0.500000,-0.500000) {A};
\node [align=center] at(1.500000,-0.500000) {B};
\node [align=center] at(2.500000,-0.500000) {C};
\node [align=center] at(3.500000,-0.500000) {D};
\node [align=center] at(4.500000,-0.500000) {E};
\node [align=center] at(5.500000,-0.500000) {F};
\node [align=center] at(6.500000,-0.500000) {G};
\node [align=center] at(7.500000,-0.500000) {H};
\node [align=center] at(8.500000,-0.500000) {I};
\node [align=center] at(9.500000,-0.500000) {J};
\node [align=center] at(10.500000,-0.500000) {K};
\end{tikzpicture}
      }
    \end{minipage}
    \hfill
    \begin{minipage}{\mpwidth}
      \centering
      \adjustbox{width=\aboxwidth}{
        \begin{tikzpicture}
\draw (0,0) grid (11,11);
\draw[fill=gray] (0,0) rectangle (1,1);
\draw[fill=gray] (1,0) rectangle (2,1);
\draw[fill=gray] (2,0) rectangle (3,1);
\draw[fill=gray] (3,0) rectangle (4,1);
\draw[fill=gray] (4,0) rectangle (5,1);
\draw[fill=gray] (5,0) rectangle (6,1);
\draw[fill=gray] (6,0) rectangle (7,1);
\draw[fill=gray] (7,0) rectangle (8,1);
\draw[fill=gray] (8,0) rectangle (9,1);
\draw[fill=gray] (9,0) rectangle (10,1);
\draw[fill=gray] (10,0) rectangle (11,1);
\draw[fill=gray] (0,1) rectangle (1,2);
\draw[fill={rgb,255:red,93.0;green,93.0;blue,255.0}]  (4,1) rectangle (5,2);
\draw[fill={rgb,255:red,255.0;green,148.0;blue,42.0}]  (4.250000,1.250000) rectangle (4.750000,1.750000);
\draw[fill={rgb,255:red,93.0;green,93.0;blue,255.0}]  (5,1) rectangle (6,2);
\draw[fill={rgb,255:red,255.0;green,212.0;blue,170.0}]  (5.250000,1.250000) rectangle (5.750000,1.750000);
\draw[fill={rgb,255:red,93.0;green,93.0;blue,255.0}]  (6,1) rectangle (7,2);
\draw[fill={rgb,255:red,255.0;green,212.0;blue,170.0}]  (6.250000,1.250000) rectangle (6.750000,1.750000);
\draw[fill={rgb,255:red,93.0;green,93.0;blue,255.0}]  (7,1) rectangle (8,2);
\draw[fill={rgb,255:red,255.0;green,212.0;blue,170.0}]  (7.250000,1.250000) rectangle (7.750000,1.750000);
\draw[fill={rgb,255:red,93.0;green,93.0;blue,255.0}]  (8,1) rectangle (9,2);
\draw[fill={rgb,255:red,255.0;green,212.0;blue,170.0}]  (8.250000,1.250000) rectangle (8.750000,1.750000);
\draw[fill=pink] (9.500000,1.500000) circle (0.500000);
\draw[fill=gray] (10,1) rectangle (11,2);
\draw[fill=gray] (0,2) rectangle (1,3);
\draw[fill={rgb,255:red,93.0;green,93.0;blue,255.0}]  (4,2) rectangle (5,3);
\draw[fill=gray] (5,2) rectangle (6,3);
\draw[fill=gray] (6,2) rectangle (7,3);
\draw[fill=gray] (7,2) rectangle (8,3);
\draw[fill=gray] (8,2) rectangle (9,3);
\draw[fill={rgb,255:red,161.0;green,161.0;blue,255.0}]  (9,2) rectangle (10,3);
\draw[fill=gray] (10,2) rectangle (11,3);
\draw[fill=gray] (0,3) rectangle (1,4);
\draw[fill={rgb,255:red,93.0;green,93.0;blue,255.0}]  (4,3) rectangle (5,4);
\draw[fill={rgb,255:red,255.0;green,170.0;blue,85.0}]  (4.250000,3.250000) rectangle (4.750000,3.750000);
\draw[fill=gray] (5,3) rectangle (6,4);
\draw[fill={rgb,255:red,161.0;green,161.0;blue,255.0}]  (9,3) rectangle (10,4);
\draw[fill={rgb,255:red,255.0;green,218.0;blue,182.0}]  (9.250000,3.250000) rectangle (9.750000,3.750000);
\draw[fill=gray] (10,3) rectangle (11,4);
\draw[fill=gray] (0,4) rectangle (1,5);
\draw[fill={rgb,255:red,93.0;green,93.0;blue,255.0}]  (4,4) rectangle (5,5);
\draw[fill={rgb,255:red,255.0;green,191.0;blue,127.0}]  (4.250000,4.250000) rectangle (4.750000,4.750000);
\draw[fill=gray] (5,4) rectangle (6,5);
\draw[fill={rgb,255:red,161.0;green,161.0;blue,255.0}]  (9,4) rectangle (10,5);
\draw[fill=gray] (10,4) rectangle (11,5);
\draw[fill=gray] (0,5) rectangle (1,6);
\draw[fill={rgb,255:red,93.0;green,93.0;blue,255.0}]  (1,5) rectangle (2,6);
\draw[fill={rgb,255:red,255.0;green,212.0;blue,170.0}]  (1.250000,5.250000) rectangle (1.750000,5.750000);
\draw[fill={rgb,255:red,93.0;green,93.0;blue,255.0}]  (2,5) rectangle (3,6);
\draw[fill={rgb,255:red,255.0;green,170.0;blue,85.0}]  (2.250000,5.250000) rectangle (2.750000,5.750000);
\draw[fill={rgb,255:red,93.0;green,93.0;blue,255.0}]  (3,5) rectangle (4,6);
\draw[fill={rgb,255:red,93.0;green,93.0;blue,255.0}]  (4,5) rectangle (5,6);
\draw[fill={rgb,255:red,255.0;green,212.0;blue,170.0}]  (4.250000,5.250000) rectangle (4.750000,5.750000);
\draw[fill=gray] (5,5) rectangle (6,6);
\draw[fill={rgb,255:red,161.0;green,161.0;blue,255.0}]  (9,5) rectangle (10,6);
\draw[fill=gray] (10,5) rectangle (11,6);
\draw[fill=gray] (0,6) rectangle (1,7);
\draw[fill={rgb,255:red,93.0;green,93.0;blue,255.0}]  (1,6) rectangle (2,7);
\draw[fill=gray] (2,6) rectangle (3,7);
\draw[fill=gray] (3,6) rectangle (4,7);
\draw[fill=gray] (4,6) rectangle (5,7);
\draw[fill=gray] (5,6) rectangle (6,7);
\draw[fill={rgb,255:red,161.0;green,161.0;blue,255.0}]  (9,6) rectangle (10,7);
\draw[fill={rgb,255:red,255.0;green,182.0;blue,109.0}]  (9.250000,6.250000) rectangle (9.750000,6.750000);
\draw[fill=gray] (10,6) rectangle (11,7);
\draw[fill=gray] (0,7) rectangle (1,8);
\draw[fill={rgb,255:red,0.0;green,0.0;blue,255.0}]  (1,7) rectangle (2,8);
\draw[fill={rgb,255:red,255.0;green,228.0;blue,201.0}]  (1.250000,7.250000) rectangle (1.750000,7.750000);
\draw[fill={rgb,255:red,161.0;green,161.0;blue,255.0}]  (2,7) rectangle (3,8);
\draw[fill={rgb,255:red,161.0;green,161.0;blue,255.0}]  (3,7) rectangle (4,8);
\draw[fill={rgb,255:red,255.0;green,218.0;blue,182.0}]  (3.250000,7.250000) rectangle (3.750000,7.750000);
\draw[fill={rgb,255:red,161.0;green,161.0;blue,255.0}]  (4,7) rectangle (5,8);
\draw[fill={rgb,255:red,255.0;green,218.0;blue,182.0}]  (4.250000,7.250000) rectangle (4.750000,7.750000);
\draw[fill={rgb,255:red,161.0;green,161.0;blue,255.0}]  (5,7) rectangle (6,8);
\draw[fill={rgb,255:red,255.0;green,218.0;blue,182.0}]  (5.250000,7.250000) rectangle (5.750000,7.750000);
\draw[fill={rgb,255:red,161.0;green,161.0;blue,255.0}]  (6,7) rectangle (7,8);
\draw[fill={rgb,255:red,161.0;green,161.0;blue,255.0}]  (7,7) rectangle (8,8);
\draw[fill={rgb,255:red,161.0;green,161.0;blue,255.0}]  (8,7) rectangle (9,8);
\draw[fill={rgb,255:red,161.0;green,161.0;blue,255.0}]  (9,7) rectangle (10,8);
\draw[fill={rgb,255:red,255.0;green,218.0;blue,182.0}]  (9.250000,7.250000) rectangle (9.750000,7.750000);
\draw[fill=gray] (10,7) rectangle (11,8);
\draw[fill=gray] (0,8) rectangle (1,9);
\draw[fill={rgb,255:red,0.0;green,0.0;blue,255.0}]  (1,8) rectangle (2,9);
\draw[fill={rgb,255:red,255.0;green,241.0;blue,228.0}]  (1.250000,8.250000) rectangle (1.750000,8.750000);
\draw[fill=gray] (2,8) rectangle (3,9);
\draw[fill=gray] (3,8) rectangle (4,9);
\draw[fill=gray] (4,8) rectangle (5,9);
\draw[fill=gray] (5,8) rectangle (6,9);
\draw[fill=gray] (6,8) rectangle (7,9);
\draw[fill=gray] (7,8) rectangle (8,9);
\draw[fill=gray] (8,8) rectangle (9,9);
\draw[fill=gray] (9,8) rectangle (10,9);
\draw[fill=gray] (10,8) rectangle (11,9);
\draw[fill=gray] (0,9) rectangle (1,10);
\draw[fill={rgb,255:red,0.0;green,0.0;blue,255.0}]  (1,9) rectangle (2,10);
\draw[fill=gray] (2,9) rectangle (3,10);
\draw[fill=gray] (3,9) rectangle (4,10);
\draw[fill=gray] (4,9) rectangle (5,10);
\draw[fill=gray] (5,9) rectangle (6,10);
\draw[fill=gray] (6,9) rectangle (7,10);
\draw[fill=gray] (7,9) rectangle (8,10);
\draw[fill=gray] (8,9) rectangle (9,10);
\draw[fill=gray] (9,9) rectangle (10,10);
\draw[fill=gray] (10,9) rectangle (11,10);
\draw[fill=gray] (0,10) rectangle (1,11);
\draw[fill=gray] (1,10) rectangle (2,11);
\draw[fill=gray] (2,10) rectangle (3,11);
\draw[fill=gray] (3,10) rectangle (4,11);
\draw[fill=gray] (4,10) rectangle (5,11);
\draw[fill=gray] (5,10) rectangle (6,11);
\draw[fill=gray] (6,10) rectangle (7,11);
\draw[fill=gray] (7,10) rectangle (8,11);
\draw[fill=gray] (8,10) rectangle (9,11);
\draw[fill=gray] (9,10) rectangle (10,11);
\draw[fill=gray] (10,10) rectangle (11,11);
\node [align=center] at(-0.500000,0.500000) {0};
\node [align=center] at(-0.500000,1.500000) {1};
\node [align=center] at(-0.500000,2.500000) {2};
\node [align=center] at(-0.500000,3.500000) {3};
\node [align=center] at(-0.500000,4.500000) {4};
\node [align=center] at(-0.500000,5.500000) {5};
\node [align=center] at(-0.500000,6.500000) {6};
\node [align=center] at(-0.500000,7.500000) {7};
\node [align=center] at(-0.500000,8.500000) {8};
\node [align=center] at(-0.500000,9.500000) {9};
\node [align=center] at(-0.500000,10.500000) {10};
\node [align=center] at(0.500000,-0.500000) {A};
\node [align=center] at(1.500000,-0.500000) {B};
\node [align=center] at(2.500000,-0.500000) {C};
\node [align=center] at(3.500000,-0.500000) {D};
\node [align=center] at(4.500000,-0.500000) {E};
\node [align=center] at(5.500000,-0.500000) {F};
\node [align=center] at(6.500000,-0.500000) {G};
\node [align=center] at(7.500000,-0.500000) {H};
\node [align=center] at(8.500000,-0.500000) {I};
\node [align=center] at(9.500000,-0.500000) {J};
\node [align=center] at(10.500000,-0.500000) {K};
\end{tikzpicture}
      }
    \end{minipage}

    \centerline{\small $M_{3}$}
  \end{minipage}

  \hfill
  
  \begin{minipage}{\colwidth}
    \begin{minipage}{\mpwidth}
      \centering
      \adjustbox{width=\aboxwidth}{
        \input{\dir 3_MDP_policyStochHeatmap.tikz}
      }
    \end{minipage}
    \hfill
    \begin{minipage}{\mpwidth}
      \centering
      \adjustbox{width=\aboxwidth}{
        \begin{tikzpicture}
\draw (0,0) grid (11,11);
\draw[fill=gray] (0,0) rectangle (1,1);
\draw[fill=gray] (1,0) rectangle (2,1);
\draw[fill=gray] (2,0) rectangle (3,1);
\draw[fill=gray] (3,0) rectangle (4,1);
\draw[fill=gray] (4,0) rectangle (5,1);
\draw[fill=gray] (5,0) rectangle (6,1);
\draw[fill=gray] (6,0) rectangle (7,1);
\draw[fill=gray] (7,0) rectangle (8,1);
\draw[fill=gray] (8,0) rectangle (9,1);
\draw[fill=gray] (9,0) rectangle (10,1);
\draw[fill=gray] (10,0) rectangle (11,1);
\draw[fill=gray] (0,1) rectangle (1,2);
\draw[fill=pink] (9.500000,1.500000) circle (0.500000);
\draw[fill=gray] (10,1) rectangle (11,2);
\draw[fill=gray] (0,2) rectangle (1,3);
\draw[fill=gray] (5,2) rectangle (6,3);
\draw[fill=gray] (6,2) rectangle (7,3);
\draw[fill=gray] (7,2) rectangle (8,3);
\draw[fill=gray] (8,2) rectangle (9,3);
\draw[fill={rgb,255:red,0.0;green,0.0;blue,255.0}]  (9,2) rectangle (10,3);
\draw[fill={rgb,255:red,255.0;green,228.0;blue,201.0}]  (9.250000,2.250000) rectangle (9.750000,2.750000);
\draw[fill=gray] (10,2) rectangle (11,3);
\draw[fill=gray] (0,3) rectangle (1,4);
\draw[fill=gray] (5,3) rectangle (6,4);
\draw[fill={rgb,255:red,0.0;green,0.0;blue,255.0}]  (9,3) rectangle (10,4);
\draw[fill={rgb,255:red,255.0;green,187.0;blue,120.0}]  (9.250000,3.250000) rectangle (9.750000,3.750000);
\draw[fill=gray] (10,3) rectangle (11,4);
\draw[fill=gray] (0,4) rectangle (1,5);
\draw[fill={rgb,255:red,0.0;green,0.0;blue,255.0}]  (9,4) rectangle (10,5);
\draw[fill={rgb,255:red,255.0;green,187.0;blue,120.0}]  (9.250000,4.250000) rectangle (9.750000,4.750000);
\draw[fill=gray] (10,4) rectangle (11,5);
\draw[fill=gray] (0,5) rectangle (1,6);
\draw[fill=gray] (5,5) rectangle (6,6);
\draw[fill={rgb,255:red,0.0;green,0.0;blue,255.0}]  (9,5) rectangle (10,6);
\draw[fill={rgb,255:red,255.0;green,147.0;blue,40.0}]  (9.250000,5.250000) rectangle (9.750000,5.750000);
\draw[fill=gray] (10,5) rectangle (11,6);
\draw[fill=gray] (0,6) rectangle (1,7);
\draw[fill=gray] (2,6) rectangle (3,7);
\draw[fill=gray] (3,6) rectangle (4,7);
\draw[fill=gray] (4,6) rectangle (5,7);
\draw[fill=gray] (5,6) rectangle (6,7);
\draw[fill={rgb,255:red,0.0;green,0.0;blue,255.0}]  (9,6) rectangle (10,7);
\draw[fill={rgb,255:red,255.0;green,228.0;blue,201.0}]  (9.250000,6.250000) rectangle (9.750000,6.750000);
\draw[fill=gray] (10,6) rectangle (11,7);
\draw[fill=gray] (0,7) rectangle (1,8);
\draw[fill={rgb,255:red,0.0;green,0.0;blue,255.0}]  (1,7) rectangle (2,8);
\draw[fill={rgb,255:red,255.0;green,228.0;blue,201.0}]  (1.250000,7.250000) rectangle (1.750000,7.750000);
\draw[fill={rgb,255:red,0.0;green,0.0;blue,255.0}]  (2,7) rectangle (3,8);
\draw[fill={rgb,255:red,0.0;green,0.0;blue,255.0}]  (3,7) rectangle (4,8);
\draw[fill={rgb,255:red,255.0;green,201.0;blue,147.0}]  (3.250000,7.250000) rectangle (3.750000,7.750000);
\draw[fill={rgb,255:red,0.0;green,0.0;blue,255.0}]  (4,7) rectangle (5,8);
\draw[fill={rgb,255:red,255.0;green,187.0;blue,120.0}]  (4.250000,7.250000) rectangle (4.750000,7.750000);
\draw[fill={rgb,255:red,0.0;green,0.0;blue,255.0}]  (5,7) rectangle (6,8);
\draw[fill={rgb,255:red,255.0;green,174.0;blue,93.0}]  (5.250000,7.250000) rectangle (5.750000,7.750000);
\draw[fill={rgb,255:red,0.0;green,0.0;blue,255.0}]  (6,7) rectangle (7,8);
\draw[fill={rgb,255:red,255.0;green,228.0;blue,201.0}]  (6.250000,7.250000) rectangle (6.750000,7.750000);
\draw[fill={rgb,255:red,0.0;green,0.0;blue,255.0}]  (7,7) rectangle (8,8);
\draw[fill={rgb,255:red,255.0;green,214.0;blue,174.0}]  (7.250000,7.250000) rectangle (7.750000,7.750000);
\draw[fill={rgb,255:red,0.0;green,0.0;blue,255.0}]  (8,7) rectangle (9,8);
\draw[fill={rgb,255:red,255.0;green,241.0;blue,228.0}]  (8.250000,7.250000) rectangle (8.750000,7.750000);
\draw[fill={rgb,255:red,0.0;green,0.0;blue,255.0}]  (9,7) rectangle (10,8);
\draw[fill={rgb,255:red,255.0;green,228.0;blue,201.0}]  (9.250000,7.250000) rectangle (9.750000,7.750000);
\draw[fill=gray] (10,7) rectangle (11,8);
\draw[fill=gray] (0,8) rectangle (1,9);
\draw[fill={rgb,255:red,0.0;green,0.0;blue,255.0}]  (1,8) rectangle (2,9);
\draw[fill={rgb,255:red,255.0;green,241.0;blue,228.0}]  (1.250000,8.250000) rectangle (1.750000,8.750000);
\draw[fill=gray] (2,8) rectangle (3,9);
\draw[fill=gray] (3,8) rectangle (4,9);
\draw[fill=gray] (4,8) rectangle (5,9);
\draw[fill=gray] (5,8) rectangle (6,9);
\draw[fill=gray] (6,8) rectangle (7,9);
\draw[fill=gray] (7,8) rectangle (8,9);
\draw[fill=gray] (8,8) rectangle (9,9);
\draw[fill=gray] (9,8) rectangle (10,9);
\draw[fill=gray] (10,8) rectangle (11,9);
\draw[fill=gray] (0,9) rectangle (1,10);
\draw[fill={rgb,255:red,0.0;green,0.0;blue,255.0}]  (1,9) rectangle (2,10);
\draw[fill=gray] (2,9) rectangle (3,10);
\draw[fill=gray] (3,9) rectangle (4,10);
\draw[fill=gray] (4,9) rectangle (5,10);
\draw[fill=gray] (5,9) rectangle (6,10);
\draw[fill=gray] (6,9) rectangle (7,10);
\draw[fill=gray] (7,9) rectangle (8,10);
\draw[fill=gray] (8,9) rectangle (9,10);
\draw[fill=gray] (9,9) rectangle (10,10);
\draw[fill=gray] (10,9) rectangle (11,10);
\draw[fill=gray] (0,10) rectangle (1,11);
\draw[fill=gray] (1,10) rectangle (2,11);
\draw[fill=gray] (2,10) rectangle (3,11);
\draw[fill=gray] (3,10) rectangle (4,11);
\draw[fill=gray] (4,10) rectangle (5,11);
\draw[fill=gray] (5,10) rectangle (6,11);
\draw[fill=gray] (6,10) rectangle (7,11);
\draw[fill=gray] (7,10) rectangle (8,11);
\draw[fill=gray] (8,10) rectangle (9,11);
\draw[fill=gray] (9,10) rectangle (10,11);
\draw[fill=gray] (10,10) rectangle (11,11);
\node [align=center] at(-0.500000,0.500000) {0};
\node [align=center] at(-0.500000,1.500000) {1};
\node [align=center] at(-0.500000,2.500000) {2};
\node [align=center] at(-0.500000,3.500000) {3};
\node [align=center] at(-0.500000,4.500000) {4};
\node [align=center] at(-0.500000,5.500000) {5};
\node [align=center] at(-0.500000,6.500000) {6};
\node [align=center] at(-0.500000,7.500000) {7};
\node [align=center] at(-0.500000,8.500000) {8};
\node [align=center] at(-0.500000,9.500000) {9};
\node [align=center] at(-0.500000,10.500000) {10};
\node [align=center] at(0.500000,-0.500000) {A};
\node [align=center] at(1.500000,-0.500000) {B};
\node [align=center] at(2.500000,-0.500000) {C};
\node [align=center] at(3.500000,-0.500000) {D};
\node [align=center] at(4.500000,-0.500000) {E};
\node [align=center] at(5.500000,-0.500000) {F};
\node [align=center] at(6.500000,-0.500000) {G};
\node [align=center] at(7.500000,-0.500000) {H};
\node [align=center] at(8.500000,-0.500000) {I};
\node [align=center] at(9.500000,-0.500000) {J};
\node [align=center] at(10.500000,-0.500000) {K};
\end{tikzpicture}
      }
    \end{minipage}
    \hfill
    \begin{minipage}{\mpwidth}
      \centering
      \adjustbox{width=\aboxwidth}{
        \begin{tikzpicture}
\draw (0,0) grid (11,11);
\draw[fill=gray] (0,0) rectangle (1,1);
\draw[fill=gray] (1,0) rectangle (2,1);
\draw[fill=gray] (2,0) rectangle (3,1);
\draw[fill=gray] (3,0) rectangle (4,1);
\draw[fill=gray] (4,0) rectangle (5,1);
\draw[fill=gray] (5,0) rectangle (6,1);
\draw[fill=gray] (6,0) rectangle (7,1);
\draw[fill=gray] (7,0) rectangle (8,1);
\draw[fill=gray] (8,0) rectangle (9,1);
\draw[fill=gray] (9,0) rectangle (10,1);
\draw[fill=gray] (10,0) rectangle (11,1);
\draw[fill=gray] (0,1) rectangle (1,2);
\draw[fill=pink] (9.500000,1.500000) circle (0.500000);
\draw[fill=gray] (10,1) rectangle (11,2);
\draw[fill=gray] (0,2) rectangle (1,3);
\draw[fill=gray] (5,2) rectangle (6,3);
\draw[fill=gray] (6,2) rectangle (7,3);
\draw[fill=gray] (7,2) rectangle (8,3);
\draw[fill=gray] (8,2) rectangle (9,3);
\draw[fill={rgb,255:red,0.0;green,0.0;blue,255.0}]  (9,2) rectangle (10,3);
\draw[fill={rgb,255:red,255.0;green,214.0;blue,174.0}]  (9.250000,2.250000) rectangle (9.750000,2.750000);
\draw[fill=gray] (10,2) rectangle (11,3);
\draw[fill=gray] (0,3) rectangle (1,4);
\draw[fill=gray] (5,3) rectangle (6,4);
\draw[fill={rgb,255:red,0.0;green,0.0;blue,255.0}]  (9,3) rectangle (10,4);
\draw[fill={rgb,255:red,255.0;green,187.0;blue,120.0}]  (9.250000,3.250000) rectangle (9.750000,3.750000);
\draw[fill=gray] (10,3) rectangle (11,4);
\draw[fill=gray] (0,4) rectangle (1,5);
\draw[fill={rgb,255:red,0.0;green,0.0;blue,255.0}]  (9,4) rectangle (10,5);
\draw[fill={rgb,255:red,255.0;green,161.0;blue,67.0}]  (9.250000,4.250000) rectangle (9.750000,4.750000);
\draw[fill=gray] (10,4) rectangle (11,5);
\draw[fill=gray] (0,5) rectangle (1,6);
\draw[fill=gray] (5,5) rectangle (6,6);
\draw[fill={rgb,255:red,0.0;green,0.0;blue,255.0}]  (9,5) rectangle (10,6);
\draw[fill={rgb,255:red,255.0;green,147.0;blue,40.0}]  (9.250000,5.250000) rectangle (9.750000,5.750000);
\draw[fill=gray] (10,5) rectangle (11,6);
\draw[fill=gray] (0,6) rectangle (1,7);
\draw[fill=gray] (2,6) rectangle (3,7);
\draw[fill=gray] (3,6) rectangle (4,7);
\draw[fill=gray] (4,6) rectangle (5,7);
\draw[fill=gray] (5,6) rectangle (6,7);
\draw[fill={rgb,255:red,0.0;green,0.0;blue,255.0}]  (9,6) rectangle (10,7);
\draw[fill={rgb,255:red,255.0;green,174.0;blue,93.0}]  (9.250000,6.250000) rectangle (9.750000,6.750000);
\draw[fill=gray] (10,6) rectangle (11,7);
\draw[fill=gray] (0,7) rectangle (1,8);
\draw[fill={rgb,255:red,0.0;green,0.0;blue,255.0}]  (1,7) rectangle (2,8);
\draw[fill={rgb,255:red,255.0;green,228.0;blue,201.0}]  (1.250000,7.250000) rectangle (1.750000,7.750000);
\draw[fill={rgb,255:red,0.0;green,0.0;blue,255.0}]  (2,7) rectangle (3,8);
\draw[fill={rgb,255:red,0.0;green,0.0;blue,255.0}]  (3,7) rectangle (4,8);
\draw[fill={rgb,255:red,255.0;green,214.0;blue,174.0}]  (3.250000,7.250000) rectangle (3.750000,7.750000);
\draw[fill={rgb,255:red,0.0;green,0.0;blue,255.0}]  (4,7) rectangle (5,8);
\draw[fill={rgb,255:red,255.0;green,214.0;blue,174.0}]  (4.250000,7.250000) rectangle (4.750000,7.750000);
\draw[fill={rgb,255:red,0.0;green,0.0;blue,255.0}]  (5,7) rectangle (6,8);
\draw[fill={rgb,255:red,255.0;green,201.0;blue,147.0}]  (5.250000,7.250000) rectangle (5.750000,7.750000);
\draw[fill={rgb,255:red,0.0;green,0.0;blue,255.0}]  (6,7) rectangle (7,8);
\draw[fill={rgb,255:red,255.0;green,214.0;blue,174.0}]  (6.250000,7.250000) rectangle (6.750000,7.750000);
\draw[fill={rgb,255:red,0.0;green,0.0;blue,255.0}]  (7,7) rectangle (8,8);
\draw[fill={rgb,255:red,255.0;green,241.0;blue,228.0}]  (7.250000,7.250000) rectangle (7.750000,7.750000);
\draw[fill={rgb,255:red,0.0;green,0.0;blue,255.0}]  (8,7) rectangle (9,8);
\draw[fill={rgb,255:red,255.0;green,241.0;blue,228.0}]  (8.250000,7.250000) rectangle (8.750000,7.750000);
\draw[fill={rgb,255:red,0.0;green,0.0;blue,255.0}]  (9,7) rectangle (10,8);
\draw[fill={rgb,255:red,255.0;green,228.0;blue,201.0}]  (9.250000,7.250000) rectangle (9.750000,7.750000);
\draw[fill=gray] (10,7) rectangle (11,8);
\draw[fill=gray] (0,8) rectangle (1,9);
\draw[fill={rgb,255:red,0.0;green,0.0;blue,255.0}]  (1,8) rectangle (2,9);
\draw[fill={rgb,255:red,255.0;green,241.0;blue,228.0}]  (1.250000,8.250000) rectangle (1.750000,8.750000);
\draw[fill=gray] (2,8) rectangle (3,9);
\draw[fill=gray] (3,8) rectangle (4,9);
\draw[fill=gray] (4,8) rectangle (5,9);
\draw[fill=gray] (5,8) rectangle (6,9);
\draw[fill=gray] (6,8) rectangle (7,9);
\draw[fill=gray] (7,8) rectangle (8,9);
\draw[fill=gray] (8,8) rectangle (9,9);
\draw[fill=gray] (9,8) rectangle (10,9);
\draw[fill=gray] (10,8) rectangle (11,9);
\draw[fill=gray] (0,9) rectangle (1,10);
\draw[fill={rgb,255:red,0.0;green,0.0;blue,255.0}]  (1,9) rectangle (2,10);
\draw[fill=gray] (2,9) rectangle (3,10);
\draw[fill=gray] (3,9) rectangle (4,10);
\draw[fill=gray] (4,9) rectangle (5,10);
\draw[fill=gray] (5,9) rectangle (6,10);
\draw[fill=gray] (6,9) rectangle (7,10);
\draw[fill=gray] (7,9) rectangle (8,10);
\draw[fill=gray] (8,9) rectangle (9,10);
\draw[fill=gray] (9,9) rectangle (10,10);
\draw[fill=gray] (10,9) rectangle (11,10);
\draw[fill=gray] (0,10) rectangle (1,11);
\draw[fill=gray] (1,10) rectangle (2,11);
\draw[fill=gray] (2,10) rectangle (3,11);
\draw[fill=gray] (3,10) rectangle (4,11);
\draw[fill=gray] (4,10) rectangle (5,11);
\draw[fill=gray] (5,10) rectangle (6,11);
\draw[fill=gray] (6,10) rectangle (7,11);
\draw[fill=gray] (7,10) rectangle (8,11);
\draw[fill=gray] (8,10) rectangle (9,11);
\draw[fill=gray] (9,10) rectangle (10,11);
\draw[fill=gray] (10,10) rectangle (11,11);
\node [align=center] at(-0.500000,0.500000) {0};
\node [align=center] at(-0.500000,1.500000) {1};
\node [align=center] at(-0.500000,2.500000) {2};
\node [align=center] at(-0.500000,3.500000) {3};
\node [align=center] at(-0.500000,4.500000) {4};
\node [align=center] at(-0.500000,5.500000) {5};
\node [align=center] at(-0.500000,6.500000) {6};
\node [align=center] at(-0.500000,7.500000) {7};
\node [align=center] at(-0.500000,8.500000) {8};
\node [align=center] at(-0.500000,9.500000) {9};
\node [align=center] at(-0.500000,10.500000) {10};
\node [align=center] at(0.500000,-0.500000) {A};
\node [align=center] at(1.500000,-0.500000) {B};
\node [align=center] at(2.500000,-0.500000) {C};
\node [align=center] at(3.500000,-0.500000) {D};
\node [align=center] at(4.500000,-0.500000) {E};
\node [align=center] at(5.500000,-0.500000) {F};
\node [align=center] at(6.500000,-0.500000) {G};
\node [align=center] at(7.500000,-0.500000) {H};
\node [align=center] at(8.500000,-0.500000) {I};
\node [align=center] at(9.500000,-0.500000) {J};
\node [align=center] at(10.500000,-0.500000) {K};
\end{tikzpicture}
      }
    \end{minipage}

    \centerline{\small $M_{4}$}
  \end{minipage}

  \medskip

  \begin{minipage}{\colwidth}
    \begin{minipage}{\mpwidth}
      \centering
      \adjustbox{width=\aboxwidth}{
        \input{\dir 6_MDP_policyStochHeatmap.tikz}
      }
    \end{minipage}
    \hfill
    \begin{minipage}{\mpwidth}
      \centering
      \adjustbox{width=\aboxwidth}{
        \begin{tikzpicture}
\draw (0,0) grid (11,12);
\draw[fill=gray] (0,0) rectangle (1,1);
\draw[fill=gray] (1,0) rectangle (2,1);
\draw[fill=gray] (2,0) rectangle (3,1);
\draw[fill=gray] (3,0) rectangle (4,1);
\draw[fill=gray] (4,0) rectangle (5,1);
\draw[fill=gray] (5,0) rectangle (6,1);
\draw[fill=gray] (6,0) rectangle (7,1);
\draw[fill=gray] (7,0) rectangle (8,1);
\draw[fill=gray] (8,0) rectangle (9,1);
\draw[fill=gray] (9,0) rectangle (10,1);
\draw[fill=gray] (10,0) rectangle (11,1);
\draw[fill=gray] (0,1) rectangle (1,2);
\draw[fill={rgb,255:red,0.0;green,0.0;blue,255.0}]  (1,1) rectangle (2,2);
\draw[fill=gray] (2,1) rectangle (3,2);
\draw[fill=gray] (3,1) rectangle (4,2);
\draw[fill=gray] (4,1) rectangle (5,2);
\draw[fill=gray] (5,1) rectangle (6,2);
\draw[fill=gray] (6,1) rectangle (7,2);
\draw[fill=gray] (7,1) rectangle (8,2);
\draw[fill=gray] (8,1) rectangle (9,2);
\draw[fill=gray] (9,1) rectangle (10,2);
\draw[fill=gray] (10,1) rectangle (11,2);
\draw[fill=gray] (0,2) rectangle (1,3);
\draw[fill={rgb,255:red,0.0;green,0.0;blue,255.0}]  (1,2) rectangle (2,3);
\draw[fill={rgb,255:red,255.0;green,228.0;blue,201.0}]  (1.250000,2.250000) rectangle (1.750000,2.750000);
\draw[fill=gray] (2,2) rectangle (3,3);
\draw[fill=gray] (3,2) rectangle (4,3);
\draw[fill=gray] (4,2) rectangle (5,3);
\draw[fill=gray] (5,2) rectangle (6,3);
\draw[fill=gray] (6,2) rectangle (7,3);
\draw[fill=gray] (7,2) rectangle (8,3);
\draw[fill=gray] (8,2) rectangle (9,3);
\draw[fill=gray] (9,2) rectangle (10,3);
\draw[fill=gray] (10,2) rectangle (11,3);
\draw[fill=gray] (0,3) rectangle (1,4);
\draw[fill={rgb,255:red,0.0;green,0.0;blue,255.0}]  (1,3) rectangle (2,4);
\draw[fill={rgb,255:red,0.0;green,0.0;blue,255.0}]  (2,3) rectangle (3,4);
\draw[fill={rgb,255:red,255.0;green,241.0;blue,228.0}]  (2.250000,3.250000) rectangle (2.750000,3.750000);
\draw[fill={rgb,255:red,0.0;green,0.0;blue,255.0}]  (3,3) rectangle (4,4);
\draw[fill={rgb,255:red,255.0;green,187.0;blue,120.0}]  (3.250000,3.250000) rectangle (3.750000,3.750000);
\draw[fill={rgb,255:red,0.0;green,0.0;blue,255.0}]  (4,3) rectangle (5,4);
\draw[fill={rgb,255:red,255.0;green,201.0;blue,147.0}]  (4.250000,3.250000) rectangle (4.750000,3.750000);
\draw[fill={rgb,255:red,0.0;green,0.0;blue,255.0}]  (5,3) rectangle (6,4);
\draw[fill={rgb,255:red,255.0;green,187.0;blue,120.0}]  (5.250000,3.250000) rectangle (5.750000,3.750000);
\draw[fill={rgb,255:red,0.0;green,0.0;blue,255.0}]  (6,3) rectangle (7,4);
\draw[fill={rgb,255:red,255.0;green,228.0;blue,201.0}]  (6.250000,3.250000) rectangle (6.750000,3.750000);
\draw[fill={rgb,255:red,0.0;green,0.0;blue,255.0}]  (7,3) rectangle (8,4);
\draw[fill={rgb,255:red,0.0;green,0.0;blue,255.0}]  (8,3) rectangle (9,4);
\draw[fill={rgb,255:red,255.0;green,241.0;blue,228.0}]  (8.250000,3.250000) rectangle (8.750000,3.750000);
\draw[fill={rgb,255:red,0.0;green,0.0;blue,255.0}]  (9,3) rectangle (10,4);
\draw[fill={rgb,255:red,255.0;green,241.0;blue,228.0}]  (9.250000,3.250000) rectangle (9.750000,3.750000);
\draw[fill=gray] (10,3) rectangle (11,4);
\draw[fill=gray] (0,4) rectangle (1,5);
\draw[fill=gray] (2,4) rectangle (3,5);
\draw[fill=gray] (3,4) rectangle (4,5);
\draw[fill=gray] (4,4) rectangle (5,5);
\draw[fill=gray] (5,4) rectangle (6,5);
\draw[fill={rgb,255:red,0.0;green,0.0;blue,255.0}]  (9,4) rectangle (10,5);
\draw[fill={rgb,255:red,255.0;green,201.0;blue,147.0}]  (9.250000,4.250000) rectangle (9.750000,4.750000);
\draw[fill=gray] (10,4) rectangle (11,5);
\draw[fill=gray] (0,5) rectangle (1,6);
\draw[fill=gray] (5,5) rectangle (6,6);
\draw[fill={rgb,255:red,0.0;green,0.0;blue,255.0}]  (9,5) rectangle (10,6);
\draw[fill={rgb,255:red,255.0;green,201.0;blue,147.0}]  (9.250000,5.250000) rectangle (9.750000,5.750000);
\draw[fill=gray] (10,5) rectangle (11,6);
\draw[fill=gray] (0,6) rectangle (1,7);
\draw[fill=gray] (5,6) rectangle (6,7);
\draw[fill={rgb,255:red,0.0;green,0.0;blue,255.0}]  (9,6) rectangle (10,7);
\draw[fill={rgb,255:red,255.0;green,174.0;blue,93.0}]  (9.250000,6.250000) rectangle (9.750000,6.750000);
\draw[fill=gray] (10,6) rectangle (11,7);
\draw[fill=gray] (0,7) rectangle (1,8);
\draw[fill=gray] (5,7) rectangle (6,8);
\draw[fill={rgb,255:red,0.0;green,0.0;blue,255.0}]  (9,7) rectangle (10,8);
\draw[fill={rgb,255:red,255.0;green,187.0;blue,120.0}]  (9.250000,7.250000) rectangle (9.750000,7.750000);
\draw[fill=gray] (10,7) rectangle (11,8);
\draw[fill=gray] (0,8) rectangle (1,9);
\draw[fill=gray] (5,8) rectangle (6,9);
\draw[fill={rgb,255:red,0.0;green,0.0;blue,255.0}]  (9,8) rectangle (10,9);
\draw[fill={rgb,255:red,255.0;green,201.0;blue,147.0}]  (9.250000,8.250000) rectangle (9.750000,8.750000);
\draw[fill=gray] (10,8) rectangle (11,9);
\draw[fill=gray] (0,9) rectangle (1,10);
\draw[fill=gray] (5,9) rectangle (6,10);
\draw[fill={rgb,255:red,0.0;green,0.0;blue,255.0}]  (9,9) rectangle (10,10);
\draw[fill={rgb,255:red,255.0;green,214.0;blue,174.0}]  (9.250000,9.250000) rectangle (9.750000,9.750000);
\draw[fill=gray] (10,9) rectangle (11,10);
\draw[fill=gray] (0,10) rectangle (1,11);
\draw[fill=pink] (9.500000,10.500000) circle (0.500000);
\draw[fill=gray] (10,10) rectangle (11,11);
\draw[fill=gray] (0,11) rectangle (1,12);
\draw[fill=gray] (1,11) rectangle (2,12);
\draw[fill=gray] (2,11) rectangle (3,12);
\draw[fill=gray] (3,11) rectangle (4,12);
\draw[fill=gray] (4,11) rectangle (5,12);
\draw[fill=gray] (5,11) rectangle (6,12);
\draw[fill=gray] (6,11) rectangle (7,12);
\draw[fill=gray] (7,11) rectangle (8,12);
\draw[fill=gray] (8,11) rectangle (9,12);
\draw[fill=gray] (9,11) rectangle (10,12);
\draw[fill=gray] (10,11) rectangle (11,12);
\node [align=center] at(-0.500000,0.500000) {0};
\node [align=center] at(-0.500000,1.500000) {1};
\node [align=center] at(-0.500000,2.500000) {2};
\node [align=center] at(-0.500000,3.500000) {3};
\node [align=center] at(-0.500000,4.500000) {4};
\node [align=center] at(-0.500000,5.500000) {5};
\node [align=center] at(-0.500000,6.500000) {6};
\node [align=center] at(-0.500000,7.500000) {7};
\node [align=center] at(-0.500000,8.500000) {8};
\node [align=center] at(-0.500000,9.500000) {9};
\node [align=center] at(-0.500000,10.500000) {10};
\node [align=center] at(-0.500000,11.500000) {11};
\node [align=center] at(0.500000,-0.500000) {A};
\node [align=center] at(1.500000,-0.500000) {B};
\node [align=center] at(2.500000,-0.500000) {C};
\node [align=center] at(3.500000,-0.500000) {D};
\node [align=center] at(4.500000,-0.500000) {E};
\node [align=center] at(5.500000,-0.500000) {F};
\node [align=center] at(6.500000,-0.500000) {G};
\node [align=center] at(7.500000,-0.500000) {H};
\node [align=center] at(8.500000,-0.500000) {I};
\node [align=center] at(9.500000,-0.500000) {J};
\node [align=center] at(10.500000,-0.500000) {K};
\end{tikzpicture}
      }
    \end{minipage}
    \hfill
    \begin{minipage}{\mpwidth}
      \centering
      \adjustbox{width=\aboxwidth}{
        \begin{tikzpicture}
\draw (0,0) grid (11,12);
\draw[fill=gray] (0,0) rectangle (1,1);
\draw[fill=gray] (1,0) rectangle (2,1);
\draw[fill=gray] (2,0) rectangle (3,1);
\draw[fill=gray] (3,0) rectangle (4,1);
\draw[fill=gray] (4,0) rectangle (5,1);
\draw[fill=gray] (5,0) rectangle (6,1);
\draw[fill=gray] (6,0) rectangle (7,1);
\draw[fill=gray] (7,0) rectangle (8,1);
\draw[fill=gray] (8,0) rectangle (9,1);
\draw[fill=gray] (9,0) rectangle (10,1);
\draw[fill=gray] (10,0) rectangle (11,1);
\draw[fill=gray] (0,1) rectangle (1,2);
\draw[fill={rgb,255:red,0.0;green,0.0;blue,255.0}]  (1,1) rectangle (2,2);
\draw[fill=gray] (2,1) rectangle (3,2);
\draw[fill=gray] (3,1) rectangle (4,2);
\draw[fill=gray] (4,1) rectangle (5,2);
\draw[fill=gray] (5,1) rectangle (6,2);
\draw[fill=gray] (6,1) rectangle (7,2);
\draw[fill=gray] (7,1) rectangle (8,2);
\draw[fill=gray] (8,1) rectangle (9,2);
\draw[fill=gray] (9,1) rectangle (10,2);
\draw[fill=gray] (10,1) rectangle (11,2);
\draw[fill=gray] (0,2) rectangle (1,3);
\draw[fill={rgb,255:red,0.0;green,0.0;blue,255.0}]  (1,2) rectangle (2,3);
\draw[fill={rgb,255:red,255.0;green,228.0;blue,201.0}]  (1.250000,2.250000) rectangle (1.750000,2.750000);
\draw[fill=gray] (2,2) rectangle (3,3);
\draw[fill=gray] (3,2) rectangle (4,3);
\draw[fill=gray] (4,2) rectangle (5,3);
\draw[fill=gray] (5,2) rectangle (6,3);
\draw[fill=gray] (6,2) rectangle (7,3);
\draw[fill=gray] (7,2) rectangle (8,3);
\draw[fill=gray] (8,2) rectangle (9,3);
\draw[fill=gray] (9,2) rectangle (10,3);
\draw[fill=gray] (10,2) rectangle (11,3);
\draw[fill=gray] (0,3) rectangle (1,4);
\draw[fill={rgb,255:red,0.0;green,0.0;blue,255.0}]  (1,3) rectangle (2,4);
\draw[fill={rgb,255:red,134.0;green,134.0;blue,255.0}]  (2,3) rectangle (3,4);
\draw[fill={rgb,255:red,134.0;green,134.0;blue,255.0}]  (3,3) rectangle (4,4);
\draw[fill={rgb,255:red,255.0;green,226.0;blue,198.0}]  (3.250000,3.250000) rectangle (3.750000,3.750000);
\draw[fill={rgb,255:red,134.0;green,134.0;blue,255.0}]  (4,3) rectangle (5,4);
\draw[fill={rgb,255:red,255.0;green,198.0;blue,141.0}]  (4.250000,3.250000) rectangle (4.750000,3.750000);
\draw[fill={rgb,255:red,134.0;green,134.0;blue,255.0}]  (5,3) rectangle (6,4);
\draw[fill={rgb,255:red,255.0;green,170.0;blue,85.0}]  (5.250000,3.250000) rectangle (5.750000,3.750000);
\draw[fill={rgb,255:red,134.0;green,134.0;blue,255.0}]  (6,3) rectangle (7,4);
\draw[fill={rgb,255:red,134.0;green,134.0;blue,255.0}]  (7,3) rectangle (8,4);
\draw[fill={rgb,255:red,134.0;green,134.0;blue,255.0}]  (8,3) rectangle (9,4);
\draw[fill={rgb,255:red,134.0;green,134.0;blue,255.0}]  (9,3) rectangle (10,4);
\draw[fill=gray] (10,3) rectangle (11,4);
\draw[fill=gray] (0,4) rectangle (1,5);
\draw[fill={rgb,255:red,120.0;green,120.0;blue,255.0}]  (1,4) rectangle (2,5);
\draw[fill={rgb,255:red,255.0;green,229.0;blue,204.0}]  (1.250000,4.250000) rectangle (1.750000,4.750000);
\draw[fill=gray] (2,4) rectangle (3,5);
\draw[fill=gray] (3,4) rectangle (4,5);
\draw[fill=gray] (4,4) rectangle (5,5);
\draw[fill=gray] (5,4) rectangle (6,5);
\draw[fill={rgb,255:red,134.0;green,134.0;blue,255.0}]  (9,4) rectangle (10,5);
\draw[fill={rgb,255:red,255.0;green,170.0;blue,85.0}]  (9.250000,4.250000) rectangle (9.750000,4.750000);
\draw[fill=gray] (10,4) rectangle (11,5);
\draw[fill=gray] (0,5) rectangle (1,6);
\draw[fill={rgb,255:red,120.0;green,120.0;blue,255.0}]  (1,5) rectangle (2,6);
\draw[fill={rgb,255:red,255.0;green,229.0;blue,204.0}]  (1.250000,5.250000) rectangle (1.750000,5.750000);
\draw[fill={rgb,255:red,120.0;green,120.0;blue,255.0}]  (2,5) rectangle (3,6);
\draw[fill={rgb,255:red,255.0;green,178.0;blue,102.0}]  (2.250000,5.250000) rectangle (2.750000,5.750000);
\draw[fill={rgb,255:red,120.0;green,120.0;blue,255.0}]  (3,5) rectangle (4,6);
\draw[fill={rgb,255:red,120.0;green,120.0;blue,255.0}]  (4,5) rectangle (5,6);
\draw[fill={rgb,255:red,255.0;green,204.0;blue,153.0}]  (4.250000,5.250000) rectangle (4.750000,5.750000);
\draw[fill=gray] (5,5) rectangle (6,6);
\draw[fill={rgb,255:red,134.0;green,134.0;blue,255.0}]  (9,5) rectangle (10,6);
\draw[fill={rgb,255:red,255.0;green,170.0;blue,85.0}]  (9.250000,5.250000) rectangle (9.750000,5.750000);
\draw[fill=gray] (10,5) rectangle (11,6);
\draw[fill=gray] (0,6) rectangle (1,7);
\draw[fill={rgb,255:red,120.0;green,120.0;blue,255.0}]  (4,6) rectangle (5,7);
\draw[fill={rgb,255:red,255.0;green,178.0;blue,102.0}]  (4.250000,6.250000) rectangle (4.750000,6.750000);
\draw[fill=gray] (5,6) rectangle (6,7);
\draw[fill={rgb,255:red,134.0;green,134.0;blue,255.0}]  (9,6) rectangle (10,7);
\draw[fill={rgb,255:red,255.0;green,226.0;blue,198.0}]  (9.250000,6.250000) rectangle (9.750000,6.750000);
\draw[fill=gray] (10,6) rectangle (11,7);
\draw[fill=gray] (0,7) rectangle (1,8);
\draw[fill={rgb,255:red,120.0;green,120.0;blue,255.0}]  (4,7) rectangle (5,8);
\draw[fill=gray] (5,7) rectangle (6,8);
\draw[fill={rgb,255:red,134.0;green,134.0;blue,255.0}]  (9,7) rectangle (10,8);
\draw[fill={rgb,255:red,255.0;green,226.0;blue,198.0}]  (9.250000,7.250000) rectangle (9.750000,7.750000);
\draw[fill=gray] (10,7) rectangle (11,8);
\draw[fill=gray] (0,8) rectangle (1,9);
\draw[fill={rgb,255:red,120.0;green,120.0;blue,255.0}]  (4,8) rectangle (5,9);
\draw[fill=gray] (5,8) rectangle (6,9);
\draw[fill={rgb,255:red,134.0;green,134.0;blue,255.0}]  (9,8) rectangle (10,9);
\draw[fill={rgb,255:red,255.0;green,226.0;blue,198.0}]  (9.250000,8.250000) rectangle (9.750000,8.750000);
\draw[fill=gray] (10,8) rectangle (11,9);
\draw[fill=gray] (0,9) rectangle (1,10);
\draw[fill={rgb,255:red,120.0;green,120.0;blue,255.0}]  (4,9) rectangle (5,10);
\draw[fill={rgb,255:red,255.0;green,229.0;blue,204.0}]  (4.250000,9.250000) rectangle (4.750000,9.750000);
\draw[fill=gray] (5,9) rectangle (6,10);
\draw[fill={rgb,255:red,134.0;green,134.0;blue,255.0}]  (9,9) rectangle (10,10);
\draw[fill=gray] (10,9) rectangle (11,10);
\draw[fill=gray] (0,10) rectangle (1,11);
\draw[fill={rgb,255:red,120.0;green,120.0;blue,255.0}]  (4,10) rectangle (5,11);
\draw[fill={rgb,255:red,120.0;green,120.0;blue,255.0}]  (5,10) rectangle (6,11);
\draw[fill={rgb,255:red,255.0;green,229.0;blue,204.0}]  (5.250000,10.250000) rectangle (5.750000,10.750000);
\draw[fill={rgb,255:red,120.0;green,120.0;blue,255.0}]  (6,10) rectangle (7,11);
\draw[fill={rgb,255:red,255.0;green,153.0;blue,50.0}]  (6.250000,10.250000) rectangle (6.750000,10.750000);
\draw[fill={rgb,255:red,120.0;green,120.0;blue,255.0}]  (7,10) rectangle (8,11);
\draw[fill={rgb,255:red,255.0;green,204.0;blue,153.0}]  (7.250000,10.250000) rectangle (7.750000,10.750000);
\draw[fill={rgb,255:red,120.0;green,120.0;blue,255.0}]  (8,10) rectangle (9,11);
\draw[fill=pink] (9.500000,10.500000) circle (0.500000);
\draw[fill=gray] (10,10) rectangle (11,11);
\draw[fill=gray] (0,11) rectangle (1,12);
\draw[fill=gray] (1,11) rectangle (2,12);
\draw[fill=gray] (2,11) rectangle (3,12);
\draw[fill=gray] (3,11) rectangle (4,12);
\draw[fill=gray] (4,11) rectangle (5,12);
\draw[fill=gray] (5,11) rectangle (6,12);
\draw[fill=gray] (6,11) rectangle (7,12);
\draw[fill=gray] (7,11) rectangle (8,12);
\draw[fill=gray] (8,11) rectangle (9,12);
\draw[fill=gray] (9,11) rectangle (10,12);
\draw[fill=gray] (10,11) rectangle (11,12);
\node [align=center] at(-0.500000,0.500000) {0};
\node [align=center] at(-0.500000,1.500000) {1};
\node [align=center] at(-0.500000,2.500000) {2};
\node [align=center] at(-0.500000,3.500000) {3};
\node [align=center] at(-0.500000,4.500000) {4};
\node [align=center] at(-0.500000,5.500000) {5};
\node [align=center] at(-0.500000,6.500000) {6};
\node [align=center] at(-0.500000,7.500000) {7};
\node [align=center] at(-0.500000,8.500000) {8};
\node [align=center] at(-0.500000,9.500000) {9};
\node [align=center] at(-0.500000,10.500000) {10};
\node [align=center] at(-0.500000,11.500000) {11};
\node [align=center] at(0.500000,-0.500000) {A};
\node [align=center] at(1.500000,-0.500000) {B};
\node [align=center] at(2.500000,-0.500000) {C};
\node [align=center] at(3.500000,-0.500000) {D};
\node [align=center] at(4.500000,-0.500000) {E};
\node [align=center] at(5.500000,-0.500000) {F};
\node [align=center] at(6.500000,-0.500000) {G};
\node [align=center] at(7.500000,-0.500000) {H};
\node [align=center] at(8.500000,-0.500000) {I};
\node [align=center] at(9.500000,-0.500000) {J};
\node [align=center] at(10.500000,-0.500000) {K};
\end{tikzpicture}
      }
    \end{minipage}

    \centerline{\small $M_{5}$}
  \end{minipage}

  \hfill
  
  \begin{minipage}{\colwidth}
    \begin{minipage}{\mpwidth}
      \centering
      \adjustbox{width=\aboxwidth}{
        \input{\dir 7_MDP_policyStochHeatmap.tikz}
      }
    \end{minipage}
    \hfill
    \begin{minipage}{\mpwidth}
      \centering
      \adjustbox{width=\aboxwidth}{
        \input{\dir 7_MDP_policyBiasedHeatMap.tikz}
      }
    \end{minipage}
    \hfill
    \begin{minipage}{\mpwidth}
      \centering
      \adjustbox{width=\aboxwidth}{
        \input{\dir 7_OAMDP_policyStochHeatMap.tikz}
      }
    \end{minipage}

    \centerline{\small $M_{7}$}
  \end{minipage}

  \medskip

  \def\colwidth{.8\textwidth}

  \begin{minipage}{\colwidth}
    \begin{minipage}{\mpwidth}
      \centering
      \adjustbox{width=\aboxwidth}{
        \input{\dir 4_MDP_policyStochHeatmap.tikz}
      }
    \end{minipage}
    \hfill
    \begin{minipage}{\mpwidth}
      \centering
      \adjustbox{width=\aboxwidth}{
        \input{\dir 4_MDP_policyBiasedHeatMap.tikz}
      }
    \end{minipage}
    \hfill
    \begin{minipage}{\mpwidth}
      \centering
      \adjustbox{width=\aboxwidth}{
        \input{\dir 4_OAMDP_policyStochHeatMap.tikz}
      }
    \end{minipage}

    \centerline{\small $M_{6}$}
  \end{minipage}

  \caption{[Error Rate] Heatmaps showing, for each cell,
    (1) [in background] the probability of visit during a trajectory from dark blue ($P=1$) to white ($P=0$), and
    (2) [in the middle] the action-prediction error rate from dark red ($P=1$) to white ($P=0$).
    These heatmaps are provided for each maze and each policy, from left to right: stochastic MDP policy, biased MDP policy, and pOAMDP policy.
    \label{fig|heatmapError}}
\end{figure}

The results given by the error rate heatmaps show that:
\begin{itemize}
\item as expected, participants made many mistakes in open areas with the stochastic MDP policy;
\item in the OAMDP policy, mistakes were mostly made in the beginning when participants needed to make a choice for example in maze $M_1$, cell $(C,3)$;
\item in some cases, as in maze $M_1$, cell $(B,3)$, participants make mistake even if there is no ambiguity over the robot next action.
  one hypothesis to explain this observation is that the participants try to anticipate the robot's next action so that, when they make a mistake, they are likely to also make another one for the next action.
This is coherent with some remarks made by the participants afterward: they anticipate the robot behavior over several time steps and, when the robot behavior did not match their expectation, they were unable to correct their predictions;
\item for the OAMDP policy, participants tend to make mistakes whenever the robot changes direction.
It seems that, for a human observer, changing direction is more costly than going forward;
\item except for maze $M_6$ (which was designed specifically to minimize the usefulness of the bias), participants perform well with the biased policy.
The chosen bias for the maze problem probably facilitates a lot the human prediction.
\end{itemize}

\paragraph{Response-Time Heatmaps}

Response-Time heatmaps for each maze in each policy are shown on \Cref{fig|heatmapTime}.
The response-time heatmap is defined as follows:
\begin{itemize}
\item the blue color represents the average number of visits computed as $\frac{\#visits}{\#participants}$, and
\item the green color represents the average response time computed as $\frac{\#time}{\#visits}$.
\end{itemize}
To make the heatmaps more readable, we cap the values to 1000\,ms, any value above 1000\,ms being indicated by a black cell.
The goal of these heatmaps was to see where the participants decision making is slow, if there was any kind of anticipation, and the overall response time depending on the policies.

\def\colwidth{.48\textwidth}
\def\mpwidth{.32\textwidth}
\def\aboxwidth{\textwidth}

\def\dir{expFinalVers/heatmap/time/}

\begin{figure}[ht!]

  \begin{minipage}{\colwidth}
    \begin{minipage}{\mpwidth}
      \centering
      \adjustbox{width=\aboxwidth}{
        \input{\dir 1_MDP_policyStochHeatmap.tikz}
      }
    \end{minipage}
    \hfill
    \begin{minipage}{\mpwidth}
      \centering
      \adjustbox{width=\aboxwidth}{
        \begin{tikzpicture}
\draw (0,0) grid (11,12);
\draw[fill=gray] (0,0) rectangle (1,1);
\draw[fill=gray] (1,0) rectangle (2,1);
\draw[fill=gray] (2,0) rectangle (3,1);
\draw[fill=gray] (3,0) rectangle (4,1);
\draw[fill=gray] (4,0) rectangle (5,1);
\draw[fill=gray] (5,0) rectangle (6,1);
\draw[fill=gray] (6,0) rectangle (7,1);
\draw[fill=gray] (7,0) rectangle (8,1);
\draw[fill=gray] (8,0) rectangle (9,1);
\draw[fill=gray] (9,0) rectangle (10,1);
\draw[fill=gray] (10,0) rectangle (11,1);
\draw[fill=gray] (0,1) rectangle (1,2);
\draw[fill=gray] (1,1) rectangle (2,2);
\draw[fill={rgb,255:red,0.0;green,0.0;blue,255.0}]  (2,1) rectangle (3,2);
\draw[fill=gray] (3,1) rectangle (4,2);
\draw[fill=gray] (4,1) rectangle (5,2);
\draw[fill=gray] (5,1) rectangle (6,2);
\draw[fill=gray] (6,1) rectangle (7,2);
\draw[fill=gray] (7,1) rectangle (8,2);
\draw[fill=gray] (8,1) rectangle (9,2);
\draw[fill=gray] (9,1) rectangle (10,2);
\draw[fill=gray] (10,1) rectangle (11,2);
\draw[fill=gray] (0,2) rectangle (1,3);
\draw[fill=gray] (1,2) rectangle (2,3);
\draw[fill={rgb,255:red,0.0;green,0.0;blue,255.0}]  (2,2) rectangle (3,3);
\draw[fill={rgb,255:red,255.0;green,241.0;blue,228.0}]  (2.250000,2.250000) rectangle (2.750000,2.750000);
\draw[fill=gray] (3,2) rectangle (4,3);
\draw[fill=gray] (4,2) rectangle (5,3);
\draw[fill=gray] (5,2) rectangle (6,3);
\draw[fill=gray] (6,2) rectangle (7,3);
\draw[fill=gray] (7,2) rectangle (8,3);
\draw[fill=gray] (8,2) rectangle (9,3);
\draw[fill=gray] (9,2) rectangle (10,3);
\draw[fill=gray] (10,2) rectangle (11,3);
\draw[fill=gray] (0,3) rectangle (1,4);
\draw[fill={rgb,255:red,0.0;green,0.0;blue,255.0}]  (2,3) rectangle (3,4);
\draw[fill={rgb,255:red,255.0;green,214.0;blue,174.0}]  (2.250000,3.250000) rectangle (2.750000,3.750000);
\draw[fill={rgb,255:red,0.0;green,0.0;blue,255.0}]  (3,3) rectangle (4,4);
\draw[fill={rgb,255:red,255.0;green,228.0;blue,201.0}]  (3.250000,3.250000) rectangle (3.750000,3.750000);
\draw[fill={rgb,255:red,0.0;green,0.0;blue,255.0}]  (4,3) rectangle (5,4);
\draw[fill={rgb,255:red,255.0;green,241.0;blue,228.0}]  (4.250000,3.250000) rectangle (4.750000,3.750000);
\draw[fill={rgb,255:red,0.0;green,0.0;blue,255.0}]  (5,3) rectangle (6,4);
\draw[fill={rgb,255:red,255.0;green,241.0;blue,228.0}]  (5.250000,3.250000) rectangle (5.750000,3.750000);
\draw[fill=gray] (6,3) rectangle (7,4);
\draw[fill=gray] (7,3) rectangle (8,4);
\draw[fill=gray] (8,3) rectangle (9,4);
\draw[fill=gray] (9,3) rectangle (10,4);
\draw[fill=gray] (10,3) rectangle (11,4);
\draw[fill=gray] (0,4) rectangle (1,5);
\draw[fill=gray] (2,4) rectangle (3,5);
\draw[fill=gray] (3,4) rectangle (4,5);
\draw[fill=gray] (4,4) rectangle (5,5);
\draw[fill={rgb,255:red,0.0;green,0.0;blue,255.0}]  (5,4) rectangle (6,5);
\draw[fill={rgb,255:red,255.0;green,228.0;blue,201.0}]  (5.250000,4.250000) rectangle (5.750000,4.750000);
\draw[fill=gray] (10,4) rectangle (11,5);
\draw[fill=gray] (0,5) rectangle (1,6);
\draw[fill=gray] (2,5) rectangle (3,6);
\draw[fill=gray] (3,5) rectangle (4,6);
\draw[fill=gray] (4,5) rectangle (5,6);
\draw[fill={rgb,255:red,0.0;green,0.0;blue,255.0}]  (5,5) rectangle (6,6);
\draw[fill=gray] (10,5) rectangle (11,6);
\draw[fill=gray] (0,6) rectangle (1,7);
\draw[fill=gray] (2,6) rectangle (3,7);
\draw[fill=gray] (3,6) rectangle (4,7);
\draw[fill=gray] (4,6) rectangle (5,7);
\draw[fill={rgb,255:red,0.0;green,0.0;blue,255.0}]  (5,6) rectangle (6,7);
\draw[fill=gray] (10,6) rectangle (11,7);
\draw[fill=gray] (0,7) rectangle (1,8);
\draw[fill=gray] (2,7) rectangle (3,8);
\draw[fill=gray] (3,7) rectangle (4,8);
\draw[fill=gray] (4,7) rectangle (5,8);
\draw[fill={rgb,255:red,0.0;green,0.0;blue,255.0}]  (5,7) rectangle (6,8);
\draw[fill={rgb,255:red,255.0;green,241.0;blue,228.0}]  (5.250000,7.250000) rectangle (5.750000,7.750000);
\draw[fill=gray] (10,7) rectangle (11,8);
\draw[fill=gray] (0,8) rectangle (1,9);
\draw[fill=gray] (2,8) rectangle (3,9);
\draw[fill=gray] (3,8) rectangle (4,9);
\draw[fill=gray] (4,8) rectangle (5,9);
\draw[fill={rgb,255:red,0.0;green,0.0;blue,255.0}]  (5,8) rectangle (6,9);
\draw[fill={rgb,255:red,255.0;green,241.0;blue,228.0}]  (5.250000,8.250000) rectangle (5.750000,8.750000);
\draw[fill={rgb,255:red,0.0;green,0.0;blue,255.0}]  (6,8) rectangle (7,9);
\draw[fill={rgb,255:red,255.0;green,214.0;blue,174.0}]  (6.250000,8.250000) rectangle (6.750000,8.750000);
\draw[fill={rgb,255:red,0.0;green,0.0;blue,255.0}]  (7,8) rectangle (8,9);
\draw[fill={rgb,255:red,255.0;green,214.0;blue,174.0}]  (7.250000,8.250000) rectangle (7.750000,8.750000);
\draw[fill={rgb,255:red,0.0;green,0.0;blue,255.0}]  (8,8) rectangle (9,9);
\draw[fill={rgb,255:red,255.0;green,228.0;blue,201.0}]  (8.250000,8.250000) rectangle (8.750000,8.750000);
\draw[fill={rgb,255:red,0.0;green,0.0;blue,255.0}]  (9,8) rectangle (10,9);
\draw[fill={rgb,255:red,255.0;green,214.0;blue,174.0}]  (9.250000,8.250000) rectangle (9.750000,8.750000);
\draw[fill=gray] (10,8) rectangle (11,9);
\draw[fill=gray] (0,9) rectangle (1,10);
\draw[fill=gray] (2,9) rectangle (3,10);
\draw[fill=gray] (3,9) rectangle (4,10);
\draw[fill=gray] (4,9) rectangle (5,10);
\draw[fill=gray] (5,9) rectangle (6,10);
\draw[fill=gray] (6,9) rectangle (7,10);
\draw[fill=gray] (7,9) rectangle (8,10);
\draw[fill=gray] (8,9) rectangle (9,10);
\draw[fill={rgb,255:red,0.0;green,0.0;blue,255.0}]  (9,9) rectangle (10,10);
\draw[fill={rgb,255:red,255.0;green,228.0;blue,201.0}]  (9.250000,9.250000) rectangle (9.750000,9.750000);
\draw[fill=gray] (10,9) rectangle (11,10);
\draw[fill=gray] (0,10) rectangle (1,11);
\draw[fill=pink] (9.500000,10.500000) circle (0.500000);
\draw[fill=gray] (10,10) rectangle (11,11);
\draw[fill=gray] (0,11) rectangle (1,12);
\draw[fill=gray] (1,11) rectangle (2,12);
\draw[fill=gray] (2,11) rectangle (3,12);
\draw[fill=gray] (3,11) rectangle (4,12);
\draw[fill=gray] (4,11) rectangle (5,12);
\draw[fill=gray] (5,11) rectangle (6,12);
\draw[fill=gray] (6,11) rectangle (7,12);
\draw[fill=gray] (7,11) rectangle (8,12);
\draw[fill=gray] (8,11) rectangle (9,12);
\draw[fill=gray] (9,11) rectangle (10,12);
\draw[fill=gray] (10,11) rectangle (11,12);
\node [align=center] at(-0.500000,0.500000) {0};
\node [align=center] at(-0.500000,1.500000) {1};
\node [align=center] at(-0.500000,2.500000) {2};
\node [align=center] at(-0.500000,3.500000) {3};
\node [align=center] at(-0.500000,4.500000) {4};
\node [align=center] at(-0.500000,5.500000) {5};
\node [align=center] at(-0.500000,6.500000) {6};
\node [align=center] at(-0.500000,7.500000) {7};
\node [align=center] at(-0.500000,8.500000) {8};
\node [align=center] at(-0.500000,9.500000) {9};
\node [align=center] at(-0.500000,10.500000) {10};
\node [align=center] at(-0.500000,11.500000) {11};
\node [align=center] at(0.500000,-0.500000) {A};
\node [align=center] at(1.500000,-0.500000) {B};
\node [align=center] at(2.500000,-0.500000) {C};
\node [align=center] at(3.500000,-0.500000) {D};
\node [align=center] at(4.500000,-0.500000) {E};
\node [align=center] at(5.500000,-0.500000) {F};
\node [align=center] at(6.500000,-0.500000) {G};
\node [align=center] at(7.500000,-0.500000) {H};
\node [align=center] at(8.500000,-0.500000) {I};
\node [align=center] at(9.500000,-0.500000) {J};
\node [align=center] at(10.500000,-0.500000) {K};
\end{tikzpicture}
      }
    \end{minipage}
    \hfill
    \begin{minipage}{\mpwidth}
      \centering
      \adjustbox{width=\aboxwidth}{
        \begin{tikzpicture}
\draw (0,0) grid (11,12);
\draw[fill=gray] (0,0) rectangle (1,1);
\draw[fill=gray] (1,0) rectangle (2,1);
\draw[fill=gray] (2,0) rectangle (3,1);
\draw[fill=gray] (3,0) rectangle (4,1);
\draw[fill=gray] (4,0) rectangle (5,1);
\draw[fill=gray] (5,0) rectangle (6,1);
\draw[fill=gray] (6,0) rectangle (7,1);
\draw[fill=gray] (7,0) rectangle (8,1);
\draw[fill=gray] (8,0) rectangle (9,1);
\draw[fill=gray] (9,0) rectangle (10,1);
\draw[fill=gray] (10,0) rectangle (11,1);
\draw[fill=gray] (0,1) rectangle (1,2);
\draw[fill=gray] (1,1) rectangle (2,2);
\draw[fill={rgb,255:red,0.0;green,0.0;blue,255.0}]  (2,1) rectangle (3,2);
\draw[fill=gray] (3,1) rectangle (4,2);
\draw[fill=gray] (4,1) rectangle (5,2);
\draw[fill=gray] (5,1) rectangle (6,2);
\draw[fill=gray] (6,1) rectangle (7,2);
\draw[fill=gray] (7,1) rectangle (8,2);
\draw[fill=gray] (8,1) rectangle (9,2);
\draw[fill=gray] (9,1) rectangle (10,2);
\draw[fill=gray] (10,1) rectangle (11,2);
\draw[fill=gray] (0,2) rectangle (1,3);
\draw[fill=gray] (1,2) rectangle (2,3);
\draw[fill={rgb,255:red,0.0;green,0.0;blue,255.0}]  (2,2) rectangle (3,3);
\draw[fill={rgb,255:red,255.0;green,241.0;blue,228.0}]  (2.250000,2.250000) rectangle (2.750000,2.750000);
\draw[fill=gray] (3,2) rectangle (4,3);
\draw[fill=gray] (4,2) rectangle (5,3);
\draw[fill=gray] (5,2) rectangle (6,3);
\draw[fill=gray] (6,2) rectangle (7,3);
\draw[fill=gray] (7,2) rectangle (8,3);
\draw[fill=gray] (8,2) rectangle (9,3);
\draw[fill=gray] (9,2) rectangle (10,3);
\draw[fill=gray] (10,2) rectangle (11,3);
\draw[fill=gray] (0,3) rectangle (1,4);
\draw[fill={rgb,255:red,0.0;green,0.0;blue,255.0}]  (1,3) rectangle (2,4);
\draw[fill={rgb,255:red,0.0;green,0.0;blue,255.0}]  (2,3) rectangle (3,4);
\draw[fill={rgb,255:red,255.0;green,241.0;blue,228.0}]  (2.250000,3.250000) rectangle (2.750000,3.750000);
\draw[fill=gray] (6,3) rectangle (7,4);
\draw[fill=gray] (7,3) rectangle (8,4);
\draw[fill=gray] (8,3) rectangle (9,4);
\draw[fill=gray] (9,3) rectangle (10,4);
\draw[fill=gray] (10,3) rectangle (11,4);
\draw[fill=gray] (0,4) rectangle (1,5);
\draw[fill={rgb,255:red,0.0;green,0.0;blue,255.0}]  (1,4) rectangle (2,5);
\draw[fill={rgb,255:red,255.0;green,201.0;blue,147.0}]  (1.250000,4.250000) rectangle (1.750000,4.750000);
\draw[fill=gray] (2,4) rectangle (3,5);
\draw[fill=gray] (3,4) rectangle (4,5);
\draw[fill=gray] (4,4) rectangle (5,5);
\draw[fill=gray] (10,4) rectangle (11,5);
\draw[fill=gray] (0,5) rectangle (1,6);
\draw[fill={rgb,255:red,0.0;green,0.0;blue,255.0}]  (1,5) rectangle (2,6);
\draw[fill={rgb,255:red,255.0;green,187.0;blue,120.0}]  (1.250000,5.250000) rectangle (1.750000,5.750000);
\draw[fill=gray] (2,5) rectangle (3,6);
\draw[fill=gray] (3,5) rectangle (4,6);
\draw[fill=gray] (4,5) rectangle (5,6);
\draw[fill=gray] (10,5) rectangle (11,6);
\draw[fill=gray] (0,6) rectangle (1,7);
\draw[fill={rgb,255:red,0.0;green,0.0;blue,255.0}]  (1,6) rectangle (2,7);
\draw[fill={rgb,255:red,255.0;green,201.0;blue,147.0}]  (1.250000,6.250000) rectangle (1.750000,6.750000);
\draw[fill=gray] (2,6) rectangle (3,7);
\draw[fill=gray] (3,6) rectangle (4,7);
\draw[fill=gray] (4,6) rectangle (5,7);
\draw[fill=gray] (10,6) rectangle (11,7);
\draw[fill=gray] (0,7) rectangle (1,8);
\draw[fill={rgb,255:red,0.0;green,0.0;blue,255.0}]  (1,7) rectangle (2,8);
\draw[fill={rgb,255:red,255.0;green,161.0;blue,67.0}]  (1.250000,7.250000) rectangle (1.750000,7.750000);
\draw[fill=gray] (2,7) rectangle (3,8);
\draw[fill=gray] (3,7) rectangle (4,8);
\draw[fill=gray] (4,7) rectangle (5,8);
\draw[fill=gray] (10,7) rectangle (11,8);
\draw[fill=gray] (0,8) rectangle (1,9);
\draw[fill={rgb,255:red,0.0;green,0.0;blue,255.0}]  (1,8) rectangle (2,9);
\draw[fill={rgb,255:red,255.0;green,161.0;blue,67.0}]  (1.250000,8.250000) rectangle (1.750000,8.750000);
\draw[fill=gray] (2,8) rectangle (3,9);
\draw[fill=gray] (3,8) rectangle (4,9);
\draw[fill=gray] (4,8) rectangle (5,9);
\draw[fill=gray] (10,8) rectangle (11,9);
\draw[fill=gray] (0,9) rectangle (1,10);
\draw[fill={rgb,255:red,0.0;green,0.0;blue,255.0}]  (1,9) rectangle (2,10);
\draw[fill={rgb,255:red,255.0;green,228.0;blue,201.0}]  (1.250000,9.250000) rectangle (1.750000,9.750000);
\draw[fill=gray] (2,9) rectangle (3,10);
\draw[fill=gray] (3,9) rectangle (4,10);
\draw[fill=gray] (4,9) rectangle (5,10);
\draw[fill=gray] (5,9) rectangle (6,10);
\draw[fill=gray] (6,9) rectangle (7,10);
\draw[fill=gray] (7,9) rectangle (8,10);
\draw[fill=gray] (8,9) rectangle (9,10);
\draw[fill=gray] (10,9) rectangle (11,10);
\draw[fill=gray] (0,10) rectangle (1,11);
\draw[fill={rgb,255:red,0.0;green,0.0;blue,255.0}]  (1,10) rectangle (2,11);
\draw[fill={rgb,255:red,255.0;green,201.0;blue,147.0}]  (1.250000,10.250000) rectangle (1.750000,10.750000);
\draw[fill={rgb,255:red,0.0;green,0.0;blue,255.0}]  (2,10) rectangle (3,11);
\draw[fill={rgb,255:red,255.0;green,174.0;blue,93.0}]  (2.250000,10.250000) rectangle (2.750000,10.750000);
\draw[fill={rgb,255:red,0.0;green,0.0;blue,255.0}]  (3,10) rectangle (4,11);
\draw[fill={rgb,255:red,255.0;green,161.0;blue,67.0}]  (3.250000,10.250000) rectangle (3.750000,10.750000);
\draw[fill={rgb,255:red,0.0;green,0.0;blue,255.0}]  (4,10) rectangle (5,11);
\draw[fill={rgb,255:red,255.0;green,147.0;blue,40.0}]  (4.250000,10.250000) rectangle (4.750000,10.750000);
\draw[fill={rgb,255:red,0.0;green,0.0;blue,255.0}]  (5,10) rectangle (6,11);
\draw[fill={rgb,255:red,255.0;green,161.0;blue,67.0}]  (5.250000,10.250000) rectangle (5.750000,10.750000);
\draw[fill={rgb,255:red,0.0;green,0.0;blue,255.0}]  (6,10) rectangle (7,11);
\draw[fill={rgb,255:red,255.0;green,187.0;blue,120.0}]  (6.250000,10.250000) rectangle (6.750000,10.750000);
\draw[fill={rgb,255:red,0.0;green,0.0;blue,255.0}]  (7,10) rectangle (8,11);
\draw[fill={rgb,255:red,255.0;green,161.0;blue,67.0}]  (7.250000,10.250000) rectangle (7.750000,10.750000);
\draw[fill={rgb,255:red,0.0;green,0.0;blue,255.0}]  (8,10) rectangle (9,11);
\draw[fill={rgb,255:red,255.0;green,201.0;blue,147.0}]  (8.250000,10.250000) rectangle (8.750000,10.750000);
\draw[fill=pink] (9.500000,10.500000) circle (0.500000);
\draw[fill=gray] (10,10) rectangle (11,11);
\draw[fill=gray] (0,11) rectangle (1,12);
\draw[fill=gray] (1,11) rectangle (2,12);
\draw[fill=gray] (2,11) rectangle (3,12);
\draw[fill=gray] (3,11) rectangle (4,12);
\draw[fill=gray] (4,11) rectangle (5,12);
\draw[fill=gray] (5,11) rectangle (6,12);
\draw[fill=gray] (6,11) rectangle (7,12);
\draw[fill=gray] (7,11) rectangle (8,12);
\draw[fill=gray] (8,11) rectangle (9,12);
\draw[fill=gray] (9,11) rectangle (10,12);
\draw[fill=gray] (10,11) rectangle (11,12);
\node [align=center] at(-0.500000,0.500000) {0};
\node [align=center] at(-0.500000,1.500000) {1};
\node [align=center] at(-0.500000,2.500000) {2};
\node [align=center] at(-0.500000,3.500000) {3};
\node [align=center] at(-0.500000,4.500000) {4};
\node [align=center] at(-0.500000,5.500000) {5};
\node [align=center] at(-0.500000,6.500000) {6};
\node [align=center] at(-0.500000,7.500000) {7};
\node [align=center] at(-0.500000,8.500000) {8};
\node [align=center] at(-0.500000,9.500000) {9};
\node [align=center] at(-0.500000,10.500000) {10};
\node [align=center] at(-0.500000,11.500000) {11};
\node [align=center] at(0.500000,-0.500000) {A};
\node [align=center] at(1.500000,-0.500000) {B};
\node [align=center] at(2.500000,-0.500000) {C};
\node [align=center] at(3.500000,-0.500000) {D};
\node [align=center] at(4.500000,-0.500000) {E};
\node [align=center] at(5.500000,-0.500000) {F};
\node [align=center] at(6.500000,-0.500000) {G};
\node [align=center] at(7.500000,-0.500000) {H};
\node [align=center] at(8.500000,-0.500000) {I};
\node [align=center] at(9.500000,-0.500000) {J};
\node [align=center] at(10.500000,-0.500000) {K};
\end{tikzpicture}
      }
    \end{minipage}

    \centerline{\small $M_{1}$}
  \end{minipage}

  \hfill
  
  \begin{minipage}{\colwidth}
    \begin{minipage}{\mpwidth}
      \centering
      \adjustbox{width=\aboxwidth}{
        \input{\dir 5_MDP_policyStochHeatmap.tikz}
      }
    \end{minipage}
    \hfill
    \begin{minipage}{\mpwidth}
      \centering
      \adjustbox{width=\aboxwidth}{
        \begin{tikzpicture}
\draw (0,0) grid (11,13);
\draw[fill=gray] (0,0) rectangle (1,1);
\draw[fill=gray] (1,0) rectangle (2,1);
\draw[fill=gray] (2,0) rectangle (3,1);
\draw[fill=gray] (3,0) rectangle (4,1);
\draw[fill=gray] (4,0) rectangle (5,1);
\draw[fill=gray] (5,0) rectangle (6,1);
\draw[fill=gray] (6,0) rectangle (7,1);
\draw[fill=gray] (7,0) rectangle (8,1);
\draw[fill=gray] (8,0) rectangle (9,1);
\draw[fill=gray] (9,0) rectangle (10,1);
\draw[fill=gray] (10,0) rectangle (11,1);
\draw[fill=gray] (0,1) rectangle (1,2);
\draw[fill=pink] (9.500000,1.500000) circle (0.500000);
\draw[fill=gray] (10,1) rectangle (11,2);
\draw[fill=gray] (0,2) rectangle (1,3);
\draw[fill=gray] (2,2) rectangle (3,3);
\draw[fill=gray] (3,2) rectangle (4,3);
\draw[fill=gray] (4,2) rectangle (5,3);
\draw[fill=gray] (5,2) rectangle (6,3);
\draw[fill=gray] (6,2) rectangle (7,3);
\draw[fill=gray] (7,2) rectangle (8,3);
\draw[fill=gray] (8,2) rectangle (9,3);
\draw[fill={rgb,255:red,0.0;green,0.0;blue,255.0}]  (9,2) rectangle (10,3);
\draw[fill={rgb,255:red,255.0;green,214.0;blue,174.0}]  (9.250000,2.250000) rectangle (9.750000,2.750000);
\draw[fill=gray] (10,2) rectangle (11,3);
\draw[fill=gray] (0,3) rectangle (1,4);
\draw[fill=gray] (2,3) rectangle (3,4);
\draw[fill=gray] (3,3) rectangle (4,4);
\draw[fill=gray] (4,3) rectangle (5,4);
\draw[fill={rgb,255:red,0.0;green,0.0;blue,255.0}]  (9,3) rectangle (10,4);
\draw[fill={rgb,255:red,255.0;green,187.0;blue,120.0}]  (9.250000,3.250000) rectangle (9.750000,3.750000);
\draw[fill=gray] (10,3) rectangle (11,4);
\draw[fill=gray] (0,4) rectangle (1,5);
\draw[fill=gray] (2,4) rectangle (3,5);
\draw[fill=gray] (3,4) rectangle (4,5);
\draw[fill=gray] (4,4) rectangle (5,5);
\draw[fill={rgb,255:red,0.0;green,0.0;blue,255.0}]  (9,4) rectangle (10,5);
\draw[fill={rgb,255:red,255.0;green,174.0;blue,93.0}]  (9.250000,4.250000) rectangle (9.750000,4.750000);
\draw[fill=gray] (10,4) rectangle (11,5);
\draw[fill=gray] (0,5) rectangle (1,6);
\draw[fill=gray] (2,5) rectangle (3,6);
\draw[fill=gray] (3,5) rectangle (4,6);
\draw[fill=gray] (4,5) rectangle (5,6);
\draw[fill={rgb,255:red,0.0;green,0.0;blue,255.0}]  (9,5) rectangle (10,6);
\draw[fill={rgb,255:red,255.0;green,134.0;blue,13.0}]  (9.250000,5.250000) rectangle (9.750000,5.750000);
\draw[fill=gray] (10,5) rectangle (11,6);
\draw[fill=gray] (0,6) rectangle (1,7);
\draw[fill=gray] (2,6) rectangle (3,7);
\draw[fill=gray] (3,6) rectangle (4,7);
\draw[fill=gray] (4,6) rectangle (5,7);
\draw[fill={rgb,255:red,0.0;green,0.0;blue,255.0}]  (9,6) rectangle (10,7);
\draw[fill={rgb,255:red,255.0;green,161.0;blue,67.0}]  (9.250000,6.250000) rectangle (9.750000,6.750000);
\draw[fill=gray] (10,6) rectangle (11,7);
\draw[fill=gray] (0,7) rectangle (1,8);
\draw[fill=gray] (2,7) rectangle (3,8);
\draw[fill=gray] (3,7) rectangle (4,8);
\draw[fill=gray] (4,7) rectangle (5,8);
\draw[fill={rgb,255:red,0.0;green,0.0;blue,255.0}]  (9,7) rectangle (10,8);
\draw[fill={rgb,255:red,255.0;green,228.0;blue,201.0}]  (9.250000,7.250000) rectangle (9.750000,7.750000);
\draw[fill=gray] (10,7) rectangle (11,8);
\draw[fill=gray] (0,8) rectangle (1,9);
\draw[fill=gray] (2,8) rectangle (3,9);
\draw[fill=gray] (3,8) rectangle (4,9);
\draw[fill=gray] (4,8) rectangle (5,9);
\draw[fill={rgb,255:red,0.0;green,0.0;blue,255.0}]  (5,8) rectangle (6,9);
\draw[fill={rgb,255:red,255.0;green,241.0;blue,228.0}]  (5.250000,8.250000) rectangle (5.750000,8.750000);
\draw[fill={rgb,255:red,0.0;green,0.0;blue,255.0}]  (6,8) rectangle (7,9);
\draw[fill={rgb,255:red,0.0;green,0.0;blue,255.0}]  (7,8) rectangle (8,9);
\draw[fill={rgb,255:red,0.0;green,0.0;blue,255.0}]  (8,8) rectangle (9,9);
\draw[fill={rgb,255:red,255.0;green,228.0;blue,201.0}]  (8.250000,8.250000) rectangle (8.750000,8.750000);
\draw[fill={rgb,255:red,0.0;green,0.0;blue,255.0}]  (9,8) rectangle (10,9);
\draw[fill={rgb,255:red,255.0;green,228.0;blue,201.0}]  (9.250000,8.250000) rectangle (9.750000,8.750000);
\draw[fill=gray] (10,8) rectangle (11,9);
\draw[fill=gray] (0,9) rectangle (1,10);
\draw[fill={rgb,255:red,0.0;green,0.0;blue,255.0}]  (5,9) rectangle (6,10);
\draw[fill={rgb,255:red,255.0;green,241.0;blue,228.0}]  (5.250000,9.250000) rectangle (5.750000,9.750000);
\draw[fill=gray] (6,9) rectangle (7,10);
\draw[fill=gray] (7,9) rectangle (8,10);
\draw[fill=gray] (8,9) rectangle (9,10);
\draw[fill=gray] (9,9) rectangle (10,10);
\draw[fill=gray] (10,9) rectangle (11,10);
\draw[fill=gray] (0,10) rectangle (1,11);
\draw[fill=gray] (1,10) rectangle (2,11);
\draw[fill=gray] (2,10) rectangle (3,11);
\draw[fill=gray] (3,10) rectangle (4,11);
\draw[fill=gray] (4,10) rectangle (5,11);
\draw[fill={rgb,255:red,0.0;green,0.0;blue,255.0}]  (5,10) rectangle (6,11);
\draw[fill={rgb,255:red,255.0;green,241.0;blue,228.0}]  (5.250000,10.250000) rectangle (5.750000,10.750000);
\draw[fill=gray] (6,10) rectangle (7,11);
\draw[fill=gray] (7,10) rectangle (8,11);
\draw[fill=gray] (8,10) rectangle (9,11);
\draw[fill=gray] (9,10) rectangle (10,11);
\draw[fill=gray] (10,10) rectangle (11,11);
\draw[fill=gray] (0,11) rectangle (1,12);
\draw[fill=gray] (1,11) rectangle (2,12);
\draw[fill=gray] (2,11) rectangle (3,12);
\draw[fill=gray] (3,11) rectangle (4,12);
\draw[fill=gray] (4,11) rectangle (5,12);
\draw[fill={rgb,255:red,0.0;green,0.0;blue,255.0}]  (5,11) rectangle (6,12);
\draw[fill=gray] (6,11) rectangle (7,12);
\draw[fill=gray] (7,11) rectangle (8,12);
\draw[fill=gray] (8,11) rectangle (9,12);
\draw[fill=gray] (9,11) rectangle (10,12);
\draw[fill=gray] (10,11) rectangle (11,12);
\draw[fill=gray] (0,12) rectangle (1,13);
\draw[fill=gray] (1,12) rectangle (2,13);
\draw[fill=gray] (2,12) rectangle (3,13);
\draw[fill=gray] (3,12) rectangle (4,13);
\draw[fill=gray] (4,12) rectangle (5,13);
\draw[fill=gray] (5,12) rectangle (6,13);
\draw[fill=gray] (6,12) rectangle (7,13);
\draw[fill=gray] (7,12) rectangle (8,13);
\draw[fill=gray] (8,12) rectangle (9,13);
\draw[fill=gray] (9,12) rectangle (10,13);
\draw[fill=gray] (10,12) rectangle (11,13);
\node [align=center] at(-0.500000,0.500000) {0};
\node [align=center] at(-0.500000,1.500000) {1};
\node [align=center] at(-0.500000,2.500000) {2};
\node [align=center] at(-0.500000,3.500000) {3};
\node [align=center] at(-0.500000,4.500000) {4};
\node [align=center] at(-0.500000,5.500000) {5};
\node [align=center] at(-0.500000,6.500000) {6};
\node [align=center] at(-0.500000,7.500000) {7};
\node [align=center] at(-0.500000,8.500000) {8};
\node [align=center] at(-0.500000,9.500000) {9};
\node [align=center] at(-0.500000,10.500000) {10};
\node [align=center] at(-0.500000,11.500000) {11};
\node [align=center] at(-0.500000,12.500000) {12};
\node [align=center] at(0.500000,-0.500000) {A};
\node [align=center] at(1.500000,-0.500000) {B};
\node [align=center] at(2.500000,-0.500000) {C};
\node [align=center] at(3.500000,-0.500000) {D};
\node [align=center] at(4.500000,-0.500000) {E};
\node [align=center] at(5.500000,-0.500000) {F};
\node [align=center] at(6.500000,-0.500000) {G};
\node [align=center] at(7.500000,-0.500000) {H};
\node [align=center] at(8.500000,-0.500000) {I};
\node [align=center] at(9.500000,-0.500000) {J};
\node [align=center] at(10.500000,-0.500000) {K};
\end{tikzpicture}
      }
    \end{minipage}
    \hfill
    \begin{minipage}{\mpwidth}
      \centering
      \adjustbox{width=\aboxwidth}{
        \begin{tikzpicture}
\draw (0,0) grid (11,13);
\draw[fill=gray] (0,0) rectangle (1,1);
\draw[fill=gray] (1,0) rectangle (2,1);
\draw[fill=gray] (2,0) rectangle (3,1);
\draw[fill=gray] (3,0) rectangle (4,1);
\draw[fill=gray] (4,0) rectangle (5,1);
\draw[fill=gray] (5,0) rectangle (6,1);
\draw[fill=gray] (6,0) rectangle (7,1);
\draw[fill=gray] (7,0) rectangle (8,1);
\draw[fill=gray] (8,0) rectangle (9,1);
\draw[fill=gray] (9,0) rectangle (10,1);
\draw[fill=gray] (10,0) rectangle (11,1);
\draw[fill=gray] (0,1) rectangle (1,2);
\draw[fill=pink] (9.500000,1.500000) circle (0.500000);
\draw[fill=gray] (10,1) rectangle (11,2);
\draw[fill=gray] (0,2) rectangle (1,3);
\draw[fill=gray] (2,2) rectangle (3,3);
\draw[fill=gray] (3,2) rectangle (4,3);
\draw[fill=gray] (4,2) rectangle (5,3);
\draw[fill=gray] (5,2) rectangle (6,3);
\draw[fill=gray] (6,2) rectangle (7,3);
\draw[fill=gray] (7,2) rectangle (8,3);
\draw[fill=gray] (8,2) rectangle (9,3);
\draw[fill={rgb,255:red,0.0;green,0.0;blue,255.0}]  (9,2) rectangle (10,3);
\draw[fill={rgb,255:red,255.0;green,214.0;blue,174.0}]  (9.250000,2.250000) rectangle (9.750000,2.750000);
\draw[fill=gray] (10,2) rectangle (11,3);
\draw[fill=gray] (0,3) rectangle (1,4);
\draw[fill=gray] (2,3) rectangle (3,4);
\draw[fill=gray] (3,3) rectangle (4,4);
\draw[fill=gray] (4,3) rectangle (5,4);
\draw[fill={rgb,255:red,0.0;green,0.0;blue,255.0}]  (9,3) rectangle (10,4);
\draw[fill={rgb,255:red,255.0;green,161.0;blue,67.0}]  (9.250000,3.250000) rectangle (9.750000,3.750000);
\draw[fill=gray] (10,3) rectangle (11,4);
\draw[fill=gray] (0,4) rectangle (1,5);
\draw[fill=gray] (2,4) rectangle (3,5);
\draw[fill=gray] (3,4) rectangle (4,5);
\draw[fill=gray] (4,4) rectangle (5,5);
\draw[fill={rgb,255:red,0.0;green,0.0;blue,255.0}]  (9,4) rectangle (10,5);
\draw[fill={rgb,255:red,255.0;green,174.0;blue,93.0}]  (9.250000,4.250000) rectangle (9.750000,4.750000);
\draw[fill=gray] (10,4) rectangle (11,5);
\draw[fill=gray] (0,5) rectangle (1,6);
\draw[fill=gray] (2,5) rectangle (3,6);
\draw[fill=gray] (3,5) rectangle (4,6);
\draw[fill=gray] (4,5) rectangle (5,6);
\draw[fill={rgb,255:red,0.0;green,0.0;blue,255.0}]  (9,5) rectangle (10,6);
\draw[fill={rgb,255:red,255.0;green,187.0;blue,120.0}]  (9.250000,5.250000) rectangle (9.750000,5.750000);
\draw[fill=gray] (10,5) rectangle (11,6);
\draw[fill=gray] (0,6) rectangle (1,7);
\draw[fill=gray] (2,6) rectangle (3,7);
\draw[fill=gray] (3,6) rectangle (4,7);
\draw[fill=gray] (4,6) rectangle (5,7);
\draw[fill={rgb,255:red,0.0;green,0.0;blue,255.0}]  (9,6) rectangle (10,7);
\draw[fill={rgb,255:red,255.0;green,161.0;blue,67.0}]  (9.250000,6.250000) rectangle (9.750000,6.750000);
\draw[fill=gray] (10,6) rectangle (11,7);
\draw[fill=gray] (0,7) rectangle (1,8);
\draw[fill=gray] (2,7) rectangle (3,8);
\draw[fill=gray] (3,7) rectangle (4,8);
\draw[fill=gray] (4,7) rectangle (5,8);
\draw[fill={rgb,255:red,0.0;green,0.0;blue,255.0}]  (9,7) rectangle (10,8);
\draw[fill={rgb,255:red,255.0;green,161.0;blue,67.0}]  (9.250000,7.250000) rectangle (9.750000,7.750000);
\draw[fill=gray] (10,7) rectangle (11,8);
\draw[fill=gray] (0,8) rectangle (1,9);
\draw[fill=gray] (2,8) rectangle (3,9);
\draw[fill=gray] (3,8) rectangle (4,9);
\draw[fill=gray] (4,8) rectangle (5,9);
\draw[fill={rgb,255:red,0.0;green,0.0;blue,255.0}]  (5,8) rectangle (6,9);
\draw[fill={rgb,255:red,0.0;green,0.0;blue,255.0}]  (6,8) rectangle (7,9);
\draw[fill={rgb,255:red,0.0;green,0.0;blue,255.0}]  (7,8) rectangle (8,9);
\draw[fill={rgb,255:red,0.0;green,0.0;blue,255.0}]  (8,8) rectangle (9,9);
\draw[fill={rgb,255:red,255.0;green,241.0;blue,228.0}]  (8.250000,8.250000) rectangle (8.750000,8.750000);
\draw[fill={rgb,255:red,0.0;green,0.0;blue,255.0}]  (9,8) rectangle (10,9);
\draw[fill={rgb,255:red,255.0;green,228.0;blue,201.0}]  (9.250000,8.250000) rectangle (9.750000,8.750000);
\draw[fill=gray] (10,8) rectangle (11,9);
\draw[fill=gray] (0,9) rectangle (1,10);
\draw[fill={rgb,255:red,0.0;green,0.0;blue,255.0}]  (5,9) rectangle (6,10);
\draw[fill=gray] (6,9) rectangle (7,10);
\draw[fill=gray] (7,9) rectangle (8,10);
\draw[fill=gray] (8,9) rectangle (9,10);
\draw[fill=gray] (9,9) rectangle (10,10);
\draw[fill=gray] (10,9) rectangle (11,10);
\draw[fill=gray] (0,10) rectangle (1,11);
\draw[fill=gray] (1,10) rectangle (2,11);
\draw[fill=gray] (2,10) rectangle (3,11);
\draw[fill=gray] (3,10) rectangle (4,11);
\draw[fill=gray] (4,10) rectangle (5,11);
\draw[fill={rgb,255:red,0.0;green,0.0;blue,255.0}]  (5,10) rectangle (6,11);
\draw[fill=gray] (6,10) rectangle (7,11);
\draw[fill=gray] (7,10) rectangle (8,11);
\draw[fill=gray] (8,10) rectangle (9,11);
\draw[fill=gray] (9,10) rectangle (10,11);
\draw[fill=gray] (10,10) rectangle (11,11);
\draw[fill=gray] (0,11) rectangle (1,12);
\draw[fill=gray] (1,11) rectangle (2,12);
\draw[fill=gray] (2,11) rectangle (3,12);
\draw[fill=gray] (3,11) rectangle (4,12);
\draw[fill=gray] (4,11) rectangle (5,12);
\draw[fill={rgb,255:red,0.0;green,0.0;blue,255.0}]  (5,11) rectangle (6,12);
\draw[fill=gray] (6,11) rectangle (7,12);
\draw[fill=gray] (7,11) rectangle (8,12);
\draw[fill=gray] (8,11) rectangle (9,12);
\draw[fill=gray] (9,11) rectangle (10,12);
\draw[fill=gray] (10,11) rectangle (11,12);
\draw[fill=gray] (0,12) rectangle (1,13);
\draw[fill=gray] (1,12) rectangle (2,13);
\draw[fill=gray] (2,12) rectangle (3,13);
\draw[fill=gray] (3,12) rectangle (4,13);
\draw[fill=gray] (4,12) rectangle (5,13);
\draw[fill=gray] (5,12) rectangle (6,13);
\draw[fill=gray] (6,12) rectangle (7,13);
\draw[fill=gray] (7,12) rectangle (8,13);
\draw[fill=gray] (8,12) rectangle (9,13);
\draw[fill=gray] (9,12) rectangle (10,13);
\draw[fill=gray] (10,12) rectangle (11,13);
\node [align=center] at(-0.500000,0.500000) {0};
\node [align=center] at(-0.500000,1.500000) {1};
\node [align=center] at(-0.500000,2.500000) {2};
\node [align=center] at(-0.500000,3.500000) {3};
\node [align=center] at(-0.500000,4.500000) {4};
\node [align=center] at(-0.500000,5.500000) {5};
\node [align=center] at(-0.500000,6.500000) {6};
\node [align=center] at(-0.500000,7.500000) {7};
\node [align=center] at(-0.500000,8.500000) {8};
\node [align=center] at(-0.500000,9.500000) {9};
\node [align=center] at(-0.500000,10.500000) {10};
\node [align=center] at(-0.500000,11.500000) {11};
\node [align=center] at(-0.500000,12.500000) {12};
\node [align=center] at(0.500000,-0.500000) {A};
\node [align=center] at(1.500000,-0.500000) {B};
\node [align=center] at(2.500000,-0.500000) {C};
\node [align=center] at(3.500000,-0.500000) {D};
\node [align=center] at(4.500000,-0.500000) {E};
\node [align=center] at(5.500000,-0.500000) {F};
\node [align=center] at(6.500000,-0.500000) {G};
\node [align=center] at(7.500000,-0.500000) {H};
\node [align=center] at(8.500000,-0.500000) {I};
\node [align=center] at(9.500000,-0.500000) {J};
\node [align=center] at(10.500000,-0.500000) {K};
\end{tikzpicture}
      }
    \end{minipage}

    \centerline{\small $M_{2}$}
  \end{minipage}

  \medskip

  \begin{minipage}{\colwidth}
    \begin{minipage}{\mpwidth}
      \centering
      \adjustbox{width=\aboxwidth}{
        \input{\dir 2_MDP_policyStochHeatmap.tikz}
      }
    \end{minipage}
    \hfill
    \begin{minipage}{\mpwidth}
      \centering
      \adjustbox{width=\aboxwidth}{
        \begin{tikzpicture}
\draw (0,0) grid (11,11);
\draw[fill=gray] (0,0) rectangle (1,1);
\draw[fill=gray] (1,0) rectangle (2,1);
\draw[fill=gray] (2,0) rectangle (3,1);
\draw[fill=gray] (3,0) rectangle (4,1);
\draw[fill=gray] (4,0) rectangle (5,1);
\draw[fill=gray] (5,0) rectangle (6,1);
\draw[fill=gray] (6,0) rectangle (7,1);
\draw[fill=gray] (7,0) rectangle (8,1);
\draw[fill=gray] (8,0) rectangle (9,1);
\draw[fill=gray] (9,0) rectangle (10,1);
\draw[fill=gray] (10,0) rectangle (11,1);
\draw[fill=gray] (0,1) rectangle (1,2);
\draw[fill=pink] (9.500000,1.500000) circle (0.500000);
\draw[fill=gray] (10,1) rectangle (11,2);
\draw[fill=gray] (0,2) rectangle (1,3);
\draw[fill=gray] (5,2) rectangle (6,3);
\draw[fill=gray] (6,2) rectangle (7,3);
\draw[fill=gray] (7,2) rectangle (8,3);
\draw[fill=gray] (8,2) rectangle (9,3);
\draw[fill={rgb,255:red,0.0;green,0.0;blue,255.0}]  (9,2) rectangle (10,3);
\draw[fill={rgb,255:red,255.0;green,241.0;blue,228.0}]  (9.250000,2.250000) rectangle (9.750000,2.750000);
\draw[fill=gray] (10,2) rectangle (11,3);
\draw[fill=gray] (0,3) rectangle (1,4);
\draw[fill=gray] (5,3) rectangle (6,4);
\draw[fill={rgb,255:red,0.0;green,0.0;blue,255.0}]  (9,3) rectangle (10,4);
\draw[fill={rgb,255:red,255.0;green,201.0;blue,147.0}]  (9.250000,3.250000) rectangle (9.750000,3.750000);
\draw[fill=gray] (10,3) rectangle (11,4);
\draw[fill=gray] (0,4) rectangle (1,5);
\draw[fill=gray] (5,4) rectangle (6,5);
\draw[fill={rgb,255:red,0.0;green,0.0;blue,255.0}]  (9,4) rectangle (10,5);
\draw[fill={rgb,255:red,255.0;green,174.0;blue,93.0}]  (9.250000,4.250000) rectangle (9.750000,4.750000);
\draw[fill=gray] (10,4) rectangle (11,5);
\draw[fill=gray] (0,5) rectangle (1,6);
\draw[fill=gray] (5,5) rectangle (6,6);
\draw[fill={rgb,255:red,0.0;green,0.0;blue,255.0}]  (9,5) rectangle (10,6);
\draw[fill={rgb,255:red,255.0;green,187.0;blue,120.0}]  (9.250000,5.250000) rectangle (9.750000,5.750000);
\draw[fill=gray] (10,5) rectangle (11,6);
\draw[fill=gray] (0,6) rectangle (1,7);
\draw[fill=gray] (2,6) rectangle (3,7);
\draw[fill=gray] (3,6) rectangle (4,7);
\draw[fill=gray] (4,6) rectangle (5,7);
\draw[fill=gray] (5,6) rectangle (6,7);
\draw[fill={rgb,255:red,0.0;green,0.0;blue,255.0}]  (9,6) rectangle (10,7);
\draw[fill={rgb,255:red,255.0;green,201.0;blue,147.0}]  (9.250000,6.250000) rectangle (9.750000,6.750000);
\draw[fill=gray] (10,6) rectangle (11,7);
\draw[fill=gray] (0,7) rectangle (1,8);
\draw[fill={rgb,255:red,0.0;green,0.0;blue,255.0}]  (1,7) rectangle (2,8);
\draw[fill={rgb,255:red,255.0;green,214.0;blue,174.0}]  (1.250000,7.250000) rectangle (1.750000,7.750000);
\draw[fill={rgb,255:red,0.0;green,0.0;blue,255.0}]  (2,7) rectangle (3,8);
\draw[fill={rgb,255:red,0.0;green,0.0;blue,255.0}]  (3,7) rectangle (4,8);
\draw[fill={rgb,255:red,255.0;green,228.0;blue,201.0}]  (3.250000,7.250000) rectangle (3.750000,7.750000);
\draw[fill={rgb,255:red,0.0;green,0.0;blue,255.0}]  (4,7) rectangle (5,8);
\draw[fill={rgb,255:red,255.0;green,241.0;blue,228.0}]  (4.250000,7.250000) rectangle (4.750000,7.750000);
\draw[fill={rgb,255:red,0.0;green,0.0;blue,255.0}]  (5,7) rectangle (6,8);
\draw[fill={rgb,255:red,255.0;green,201.0;blue,147.0}]  (5.250000,7.250000) rectangle (5.750000,7.750000);
\draw[fill={rgb,255:red,0.0;green,0.0;blue,255.0}]  (6,7) rectangle (7,8);
\draw[fill={rgb,255:red,255.0;green,214.0;blue,174.0}]  (6.250000,7.250000) rectangle (6.750000,7.750000);
\draw[fill={rgb,255:red,0.0;green,0.0;blue,255.0}]  (7,7) rectangle (8,8);
\draw[fill={rgb,255:red,255.0;green,228.0;blue,201.0}]  (7.250000,7.250000) rectangle (7.750000,7.750000);
\draw[fill={rgb,255:red,0.0;green,0.0;blue,255.0}]  (8,7) rectangle (9,8);
\draw[fill={rgb,255:red,255.0;green,241.0;blue,228.0}]  (8.250000,7.250000) rectangle (8.750000,7.750000);
\draw[fill={rgb,255:red,0.0;green,0.0;blue,255.0}]  (9,7) rectangle (10,8);
\draw[fill=gray] (10,7) rectangle (11,8);
\draw[fill=gray] (0,8) rectangle (1,9);
\draw[fill={rgb,255:red,0.0;green,0.0;blue,255.0}]  (1,8) rectangle (2,9);
\draw[fill=gray] (2,8) rectangle (3,9);
\draw[fill=gray] (3,8) rectangle (4,9);
\draw[fill=gray] (4,8) rectangle (5,9);
\draw[fill=gray] (5,8) rectangle (6,9);
\draw[fill=gray] (6,8) rectangle (7,9);
\draw[fill=gray] (7,8) rectangle (8,9);
\draw[fill=gray] (8,8) rectangle (9,9);
\draw[fill=gray] (9,8) rectangle (10,9);
\draw[fill=gray] (10,8) rectangle (11,9);
\draw[fill=gray] (0,9) rectangle (1,10);
\draw[fill={rgb,255:red,0.0;green,0.0;blue,255.0}]  (1,9) rectangle (2,10);
\draw[fill=gray] (2,9) rectangle (3,10);
\draw[fill=gray] (3,9) rectangle (4,10);
\draw[fill=gray] (4,9) rectangle (5,10);
\draw[fill=gray] (5,9) rectangle (6,10);
\draw[fill=gray] (6,9) rectangle (7,10);
\draw[fill=gray] (7,9) rectangle (8,10);
\draw[fill=gray] (8,9) rectangle (9,10);
\draw[fill=gray] (9,9) rectangle (10,10);
\draw[fill=gray] (10,9) rectangle (11,10);
\draw[fill=gray] (0,10) rectangle (1,11);
\draw[fill=gray] (1,10) rectangle (2,11);
\draw[fill=gray] (2,10) rectangle (3,11);
\draw[fill=gray] (3,10) rectangle (4,11);
\draw[fill=gray] (4,10) rectangle (5,11);
\draw[fill=gray] (5,10) rectangle (6,11);
\draw[fill=gray] (6,10) rectangle (7,11);
\draw[fill=gray] (7,10) rectangle (8,11);
\draw[fill=gray] (8,10) rectangle (9,11);
\draw[fill=gray] (9,10) rectangle (10,11);
\draw[fill=gray] (10,10) rectangle (11,11);
\node [align=center] at(-0.500000,0.500000) {0};
\node [align=center] at(-0.500000,1.500000) {1};
\node [align=center] at(-0.500000,2.500000) {2};
\node [align=center] at(-0.500000,3.500000) {3};
\node [align=center] at(-0.500000,4.500000) {4};
\node [align=center] at(-0.500000,5.500000) {5};
\node [align=center] at(-0.500000,6.500000) {6};
\node [align=center] at(-0.500000,7.500000) {7};
\node [align=center] at(-0.500000,8.500000) {8};
\node [align=center] at(-0.500000,9.500000) {9};
\node [align=center] at(-0.500000,10.500000) {10};
\node [align=center] at(0.500000,-0.500000) {A};
\node [align=center] at(1.500000,-0.500000) {B};
\node [align=center] at(2.500000,-0.500000) {C};
\node [align=center] at(3.500000,-0.500000) {D};
\node [align=center] at(4.500000,-0.500000) {E};
\node [align=center] at(5.500000,-0.500000) {F};
\node [align=center] at(6.500000,-0.500000) {G};
\node [align=center] at(7.500000,-0.500000) {H};
\node [align=center] at(8.500000,-0.500000) {I};
\node [align=center] at(9.500000,-0.500000) {J};
\node [align=center] at(10.500000,-0.500000) {K};
\end{tikzpicture}
      }
    \end{minipage}
    \hfill
    \begin{minipage}{\mpwidth}
      \centering
      \adjustbox{width=\aboxwidth}{
        \begin{tikzpicture}
\draw (0,0) grid (11,11);
\draw[fill=gray] (0,0) rectangle (1,1);
\draw[fill=gray] (1,0) rectangle (2,1);
\draw[fill=gray] (2,0) rectangle (3,1);
\draw[fill=gray] (3,0) rectangle (4,1);
\draw[fill=gray] (4,0) rectangle (5,1);
\draw[fill=gray] (5,0) rectangle (6,1);
\draw[fill=gray] (6,0) rectangle (7,1);
\draw[fill=gray] (7,0) rectangle (8,1);
\draw[fill=gray] (8,0) rectangle (9,1);
\draw[fill=gray] (9,0) rectangle (10,1);
\draw[fill=gray] (10,0) rectangle (11,1);
\draw[fill=gray] (0,1) rectangle (1,2);
\draw[fill={rgb,255:red,93.0;green,93.0;blue,255.0}]  (4,1) rectangle (5,2);
\draw[fill={rgb,255:red,255.0;green,148.0;blue,42.0}]  (4.250000,1.250000) rectangle (4.750000,1.750000);
\draw[fill={rgb,255:red,93.0;green,93.0;blue,255.0}]  (5,1) rectangle (6,2);
\draw[fill={rgb,255:red,255.0;green,212.0;blue,170.0}]  (5.250000,1.250000) rectangle (5.750000,1.750000);
\draw[fill={rgb,255:red,93.0;green,93.0;blue,255.0}]  (6,1) rectangle (7,2);
\draw[fill={rgb,255:red,255.0;green,212.0;blue,170.0}]  (6.250000,1.250000) rectangle (6.750000,1.750000);
\draw[fill={rgb,255:red,93.0;green,93.0;blue,255.0}]  (7,1) rectangle (8,2);
\draw[fill={rgb,255:red,255.0;green,212.0;blue,170.0}]  (7.250000,1.250000) rectangle (7.750000,1.750000);
\draw[fill={rgb,255:red,93.0;green,93.0;blue,255.0}]  (8,1) rectangle (9,2);
\draw[fill={rgb,255:red,255.0;green,212.0;blue,170.0}]  (8.250000,1.250000) rectangle (8.750000,1.750000);
\draw[fill=pink] (9.500000,1.500000) circle (0.500000);
\draw[fill=gray] (10,1) rectangle (11,2);
\draw[fill=gray] (0,2) rectangle (1,3);
\draw[fill={rgb,255:red,93.0;green,93.0;blue,255.0}]  (4,2) rectangle (5,3);
\draw[fill=gray] (5,2) rectangle (6,3);
\draw[fill=gray] (6,2) rectangle (7,3);
\draw[fill=gray] (7,2) rectangle (8,3);
\draw[fill=gray] (8,2) rectangle (9,3);
\draw[fill={rgb,255:red,161.0;green,161.0;blue,255.0}]  (9,2) rectangle (10,3);
\draw[fill=gray] (10,2) rectangle (11,3);
\draw[fill=gray] (0,3) rectangle (1,4);
\draw[fill={rgb,255:red,93.0;green,93.0;blue,255.0}]  (4,3) rectangle (5,4);
\draw[fill={rgb,255:red,255.0;green,170.0;blue,85.0}]  (4.250000,3.250000) rectangle (4.750000,3.750000);
\draw[fill=gray] (5,3) rectangle (6,4);
\draw[fill={rgb,255:red,161.0;green,161.0;blue,255.0}]  (9,3) rectangle (10,4);
\draw[fill={rgb,255:red,255.0;green,218.0;blue,182.0}]  (9.250000,3.250000) rectangle (9.750000,3.750000);
\draw[fill=gray] (10,3) rectangle (11,4);
\draw[fill=gray] (0,4) rectangle (1,5);
\draw[fill={rgb,255:red,93.0;green,93.0;blue,255.0}]  (4,4) rectangle (5,5);
\draw[fill={rgb,255:red,255.0;green,191.0;blue,127.0}]  (4.250000,4.250000) rectangle (4.750000,4.750000);
\draw[fill=gray] (5,4) rectangle (6,5);
\draw[fill={rgb,255:red,161.0;green,161.0;blue,255.0}]  (9,4) rectangle (10,5);
\draw[fill=gray] (10,4) rectangle (11,5);
\draw[fill=gray] (0,5) rectangle (1,6);
\draw[fill={rgb,255:red,93.0;green,93.0;blue,255.0}]  (1,5) rectangle (2,6);
\draw[fill={rgb,255:red,255.0;green,212.0;blue,170.0}]  (1.250000,5.250000) rectangle (1.750000,5.750000);
\draw[fill={rgb,255:red,93.0;green,93.0;blue,255.0}]  (2,5) rectangle (3,6);
\draw[fill={rgb,255:red,255.0;green,170.0;blue,85.0}]  (2.250000,5.250000) rectangle (2.750000,5.750000);
\draw[fill={rgb,255:red,93.0;green,93.0;blue,255.0}]  (3,5) rectangle (4,6);
\draw[fill={rgb,255:red,93.0;green,93.0;blue,255.0}]  (4,5) rectangle (5,6);
\draw[fill={rgb,255:red,255.0;green,212.0;blue,170.0}]  (4.250000,5.250000) rectangle (4.750000,5.750000);
\draw[fill=gray] (5,5) rectangle (6,6);
\draw[fill={rgb,255:red,161.0;green,161.0;blue,255.0}]  (9,5) rectangle (10,6);
\draw[fill=gray] (10,5) rectangle (11,6);
\draw[fill=gray] (0,6) rectangle (1,7);
\draw[fill={rgb,255:red,93.0;green,93.0;blue,255.0}]  (1,6) rectangle (2,7);
\draw[fill=gray] (2,6) rectangle (3,7);
\draw[fill=gray] (3,6) rectangle (4,7);
\draw[fill=gray] (4,6) rectangle (5,7);
\draw[fill=gray] (5,6) rectangle (6,7);
\draw[fill={rgb,255:red,161.0;green,161.0;blue,255.0}]  (9,6) rectangle (10,7);
\draw[fill={rgb,255:red,255.0;green,182.0;blue,109.0}]  (9.250000,6.250000) rectangle (9.750000,6.750000);
\draw[fill=gray] (10,6) rectangle (11,7);
\draw[fill=gray] (0,7) rectangle (1,8);
\draw[fill={rgb,255:red,0.0;green,0.0;blue,255.0}]  (1,7) rectangle (2,8);
\draw[fill={rgb,255:red,255.0;green,228.0;blue,201.0}]  (1.250000,7.250000) rectangle (1.750000,7.750000);
\draw[fill={rgb,255:red,161.0;green,161.0;blue,255.0}]  (2,7) rectangle (3,8);
\draw[fill={rgb,255:red,161.0;green,161.0;blue,255.0}]  (3,7) rectangle (4,8);
\draw[fill={rgb,255:red,255.0;green,218.0;blue,182.0}]  (3.250000,7.250000) rectangle (3.750000,7.750000);
\draw[fill={rgb,255:red,161.0;green,161.0;blue,255.0}]  (4,7) rectangle (5,8);
\draw[fill={rgb,255:red,255.0;green,218.0;blue,182.0}]  (4.250000,7.250000) rectangle (4.750000,7.750000);
\draw[fill={rgb,255:red,161.0;green,161.0;blue,255.0}]  (5,7) rectangle (6,8);
\draw[fill={rgb,255:red,255.0;green,218.0;blue,182.0}]  (5.250000,7.250000) rectangle (5.750000,7.750000);
\draw[fill={rgb,255:red,161.0;green,161.0;blue,255.0}]  (6,7) rectangle (7,8);
\draw[fill={rgb,255:red,161.0;green,161.0;blue,255.0}]  (7,7) rectangle (8,8);
\draw[fill={rgb,255:red,161.0;green,161.0;blue,255.0}]  (8,7) rectangle (9,8);
\draw[fill={rgb,255:red,161.0;green,161.0;blue,255.0}]  (9,7) rectangle (10,8);
\draw[fill={rgb,255:red,255.0;green,218.0;blue,182.0}]  (9.250000,7.250000) rectangle (9.750000,7.750000);
\draw[fill=gray] (10,7) rectangle (11,8);
\draw[fill=gray] (0,8) rectangle (1,9);
\draw[fill={rgb,255:red,0.0;green,0.0;blue,255.0}]  (1,8) rectangle (2,9);
\draw[fill={rgb,255:red,255.0;green,241.0;blue,228.0}]  (1.250000,8.250000) rectangle (1.750000,8.750000);
\draw[fill=gray] (2,8) rectangle (3,9);
\draw[fill=gray] (3,8) rectangle (4,9);
\draw[fill=gray] (4,8) rectangle (5,9);
\draw[fill=gray] (5,8) rectangle (6,9);
\draw[fill=gray] (6,8) rectangle (7,9);
\draw[fill=gray] (7,8) rectangle (8,9);
\draw[fill=gray] (8,8) rectangle (9,9);
\draw[fill=gray] (9,8) rectangle (10,9);
\draw[fill=gray] (10,8) rectangle (11,9);
\draw[fill=gray] (0,9) rectangle (1,10);
\draw[fill={rgb,255:red,0.0;green,0.0;blue,255.0}]  (1,9) rectangle (2,10);
\draw[fill=gray] (2,9) rectangle (3,10);
\draw[fill=gray] (3,9) rectangle (4,10);
\draw[fill=gray] (4,9) rectangle (5,10);
\draw[fill=gray] (5,9) rectangle (6,10);
\draw[fill=gray] (6,9) rectangle (7,10);
\draw[fill=gray] (7,9) rectangle (8,10);
\draw[fill=gray] (8,9) rectangle (9,10);
\draw[fill=gray] (9,9) rectangle (10,10);
\draw[fill=gray] (10,9) rectangle (11,10);
\draw[fill=gray] (0,10) rectangle (1,11);
\draw[fill=gray] (1,10) rectangle (2,11);
\draw[fill=gray] (2,10) rectangle (3,11);
\draw[fill=gray] (3,10) rectangle (4,11);
\draw[fill=gray] (4,10) rectangle (5,11);
\draw[fill=gray] (5,10) rectangle (6,11);
\draw[fill=gray] (6,10) rectangle (7,11);
\draw[fill=gray] (7,10) rectangle (8,11);
\draw[fill=gray] (8,10) rectangle (9,11);
\draw[fill=gray] (9,10) rectangle (10,11);
\draw[fill=gray] (10,10) rectangle (11,11);
\node [align=center] at(-0.500000,0.500000) {0};
\node [align=center] at(-0.500000,1.500000) {1};
\node [align=center] at(-0.500000,2.500000) {2};
\node [align=center] at(-0.500000,3.500000) {3};
\node [align=center] at(-0.500000,4.500000) {4};
\node [align=center] at(-0.500000,5.500000) {5};
\node [align=center] at(-0.500000,6.500000) {6};
\node [align=center] at(-0.500000,7.500000) {7};
\node [align=center] at(-0.500000,8.500000) {8};
\node [align=center] at(-0.500000,9.500000) {9};
\node [align=center] at(-0.500000,10.500000) {10};
\node [align=center] at(0.500000,-0.500000) {A};
\node [align=center] at(1.500000,-0.500000) {B};
\node [align=center] at(2.500000,-0.500000) {C};
\node [align=center] at(3.500000,-0.500000) {D};
\node [align=center] at(4.500000,-0.500000) {E};
\node [align=center] at(5.500000,-0.500000) {F};
\node [align=center] at(6.500000,-0.500000) {G};
\node [align=center] at(7.500000,-0.500000) {H};
\node [align=center] at(8.500000,-0.500000) {I};
\node [align=center] at(9.500000,-0.500000) {J};
\node [align=center] at(10.500000,-0.500000) {K};
\end{tikzpicture}
      }
    \end{minipage}

    \centerline{\small $M_{3}$}
  \end{minipage}

  \hfill
  
  \begin{minipage}{\colwidth}
    \begin{minipage}{\mpwidth}
      \centering
      \adjustbox{width=\aboxwidth}{
        \input{\dir 3_MDP_policyStochHeatmap.tikz}
      }
    \end{minipage}
    \hfill
    \begin{minipage}{\mpwidth}
      \centering
      \adjustbox{width=\aboxwidth}{
        \begin{tikzpicture}
\draw (0,0) grid (11,11);
\draw[fill=gray] (0,0) rectangle (1,1);
\draw[fill=gray] (1,0) rectangle (2,1);
\draw[fill=gray] (2,0) rectangle (3,1);
\draw[fill=gray] (3,0) rectangle (4,1);
\draw[fill=gray] (4,0) rectangle (5,1);
\draw[fill=gray] (5,0) rectangle (6,1);
\draw[fill=gray] (6,0) rectangle (7,1);
\draw[fill=gray] (7,0) rectangle (8,1);
\draw[fill=gray] (8,0) rectangle (9,1);
\draw[fill=gray] (9,0) rectangle (10,1);
\draw[fill=gray] (10,0) rectangle (11,1);
\draw[fill=gray] (0,1) rectangle (1,2);
\draw[fill=pink] (9.500000,1.500000) circle (0.500000);
\draw[fill=gray] (10,1) rectangle (11,2);
\draw[fill=gray] (0,2) rectangle (1,3);
\draw[fill=gray] (5,2) rectangle (6,3);
\draw[fill=gray] (6,2) rectangle (7,3);
\draw[fill=gray] (7,2) rectangle (8,3);
\draw[fill=gray] (8,2) rectangle (9,3);
\draw[fill={rgb,255:red,0.0;green,0.0;blue,255.0}]  (9,2) rectangle (10,3);
\draw[fill={rgb,255:red,255.0;green,228.0;blue,201.0}]  (9.250000,2.250000) rectangle (9.750000,2.750000);
\draw[fill=gray] (10,2) rectangle (11,3);
\draw[fill=gray] (0,3) rectangle (1,4);
\draw[fill=gray] (5,3) rectangle (6,4);
\draw[fill={rgb,255:red,0.0;green,0.0;blue,255.0}]  (9,3) rectangle (10,4);
\draw[fill={rgb,255:red,255.0;green,187.0;blue,120.0}]  (9.250000,3.250000) rectangle (9.750000,3.750000);
\draw[fill=gray] (10,3) rectangle (11,4);
\draw[fill=gray] (0,4) rectangle (1,5);
\draw[fill={rgb,255:red,0.0;green,0.0;blue,255.0}]  (9,4) rectangle (10,5);
\draw[fill={rgb,255:red,255.0;green,187.0;blue,120.0}]  (9.250000,4.250000) rectangle (9.750000,4.750000);
\draw[fill=gray] (10,4) rectangle (11,5);
\draw[fill=gray] (0,5) rectangle (1,6);
\draw[fill=gray] (5,5) rectangle (6,6);
\draw[fill={rgb,255:red,0.0;green,0.0;blue,255.0}]  (9,5) rectangle (10,6);
\draw[fill={rgb,255:red,255.0;green,147.0;blue,40.0}]  (9.250000,5.250000) rectangle (9.750000,5.750000);
\draw[fill=gray] (10,5) rectangle (11,6);
\draw[fill=gray] (0,6) rectangle (1,7);
\draw[fill=gray] (2,6) rectangle (3,7);
\draw[fill=gray] (3,6) rectangle (4,7);
\draw[fill=gray] (4,6) rectangle (5,7);
\draw[fill=gray] (5,6) rectangle (6,7);
\draw[fill={rgb,255:red,0.0;green,0.0;blue,255.0}]  (9,6) rectangle (10,7);
\draw[fill={rgb,255:red,255.0;green,228.0;blue,201.0}]  (9.250000,6.250000) rectangle (9.750000,6.750000);
\draw[fill=gray] (10,6) rectangle (11,7);
\draw[fill=gray] (0,7) rectangle (1,8);
\draw[fill={rgb,255:red,0.0;green,0.0;blue,255.0}]  (1,7) rectangle (2,8);
\draw[fill={rgb,255:red,255.0;green,228.0;blue,201.0}]  (1.250000,7.250000) rectangle (1.750000,7.750000);
\draw[fill={rgb,255:red,0.0;green,0.0;blue,255.0}]  (2,7) rectangle (3,8);
\draw[fill={rgb,255:red,0.0;green,0.0;blue,255.0}]  (3,7) rectangle (4,8);
\draw[fill={rgb,255:red,255.0;green,201.0;blue,147.0}]  (3.250000,7.250000) rectangle (3.750000,7.750000);
\draw[fill={rgb,255:red,0.0;green,0.0;blue,255.0}]  (4,7) rectangle (5,8);
\draw[fill={rgb,255:red,255.0;green,187.0;blue,120.0}]  (4.250000,7.250000) rectangle (4.750000,7.750000);
\draw[fill={rgb,255:red,0.0;green,0.0;blue,255.0}]  (5,7) rectangle (6,8);
\draw[fill={rgb,255:red,255.0;green,174.0;blue,93.0}]  (5.250000,7.250000) rectangle (5.750000,7.750000);
\draw[fill={rgb,255:red,0.0;green,0.0;blue,255.0}]  (6,7) rectangle (7,8);
\draw[fill={rgb,255:red,255.0;green,228.0;blue,201.0}]  (6.250000,7.250000) rectangle (6.750000,7.750000);
\draw[fill={rgb,255:red,0.0;green,0.0;blue,255.0}]  (7,7) rectangle (8,8);
\draw[fill={rgb,255:red,255.0;green,214.0;blue,174.0}]  (7.250000,7.250000) rectangle (7.750000,7.750000);
\draw[fill={rgb,255:red,0.0;green,0.0;blue,255.0}]  (8,7) rectangle (9,8);
\draw[fill={rgb,255:red,255.0;green,241.0;blue,228.0}]  (8.250000,7.250000) rectangle (8.750000,7.750000);
\draw[fill={rgb,255:red,0.0;green,0.0;blue,255.0}]  (9,7) rectangle (10,8);
\draw[fill={rgb,255:red,255.0;green,228.0;blue,201.0}]  (9.250000,7.250000) rectangle (9.750000,7.750000);
\draw[fill=gray] (10,7) rectangle (11,8);
\draw[fill=gray] (0,8) rectangle (1,9);
\draw[fill={rgb,255:red,0.0;green,0.0;blue,255.0}]  (1,8) rectangle (2,9);
\draw[fill={rgb,255:red,255.0;green,241.0;blue,228.0}]  (1.250000,8.250000) rectangle (1.750000,8.750000);
\draw[fill=gray] (2,8) rectangle (3,9);
\draw[fill=gray] (3,8) rectangle (4,9);
\draw[fill=gray] (4,8) rectangle (5,9);
\draw[fill=gray] (5,8) rectangle (6,9);
\draw[fill=gray] (6,8) rectangle (7,9);
\draw[fill=gray] (7,8) rectangle (8,9);
\draw[fill=gray] (8,8) rectangle (9,9);
\draw[fill=gray] (9,8) rectangle (10,9);
\draw[fill=gray] (10,8) rectangle (11,9);
\draw[fill=gray] (0,9) rectangle (1,10);
\draw[fill={rgb,255:red,0.0;green,0.0;blue,255.0}]  (1,9) rectangle (2,10);
\draw[fill=gray] (2,9) rectangle (3,10);
\draw[fill=gray] (3,9) rectangle (4,10);
\draw[fill=gray] (4,9) rectangle (5,10);
\draw[fill=gray] (5,9) rectangle (6,10);
\draw[fill=gray] (6,9) rectangle (7,10);
\draw[fill=gray] (7,9) rectangle (8,10);
\draw[fill=gray] (8,9) rectangle (9,10);
\draw[fill=gray] (9,9) rectangle (10,10);
\draw[fill=gray] (10,9) rectangle (11,10);
\draw[fill=gray] (0,10) rectangle (1,11);
\draw[fill=gray] (1,10) rectangle (2,11);
\draw[fill=gray] (2,10) rectangle (3,11);
\draw[fill=gray] (3,10) rectangle (4,11);
\draw[fill=gray] (4,10) rectangle (5,11);
\draw[fill=gray] (5,10) rectangle (6,11);
\draw[fill=gray] (6,10) rectangle (7,11);
\draw[fill=gray] (7,10) rectangle (8,11);
\draw[fill=gray] (8,10) rectangle (9,11);
\draw[fill=gray] (9,10) rectangle (10,11);
\draw[fill=gray] (10,10) rectangle (11,11);
\node [align=center] at(-0.500000,0.500000) {0};
\node [align=center] at(-0.500000,1.500000) {1};
\node [align=center] at(-0.500000,2.500000) {2};
\node [align=center] at(-0.500000,3.500000) {3};
\node [align=center] at(-0.500000,4.500000) {4};
\node [align=center] at(-0.500000,5.500000) {5};
\node [align=center] at(-0.500000,6.500000) {6};
\node [align=center] at(-0.500000,7.500000) {7};
\node [align=center] at(-0.500000,8.500000) {8};
\node [align=center] at(-0.500000,9.500000) {9};
\node [align=center] at(-0.500000,10.500000) {10};
\node [align=center] at(0.500000,-0.500000) {A};
\node [align=center] at(1.500000,-0.500000) {B};
\node [align=center] at(2.500000,-0.500000) {C};
\node [align=center] at(3.500000,-0.500000) {D};
\node [align=center] at(4.500000,-0.500000) {E};
\node [align=center] at(5.500000,-0.500000) {F};
\node [align=center] at(6.500000,-0.500000) {G};
\node [align=center] at(7.500000,-0.500000) {H};
\node [align=center] at(8.500000,-0.500000) {I};
\node [align=center] at(9.500000,-0.500000) {J};
\node [align=center] at(10.500000,-0.500000) {K};
\end{tikzpicture}
      }
    \end{minipage}
    \hfill
    \begin{minipage}{\mpwidth}
      \centering
      \adjustbox{width=\aboxwidth}{
        \begin{tikzpicture}
\draw (0,0) grid (11,11);
\draw[fill=gray] (0,0) rectangle (1,1);
\draw[fill=gray] (1,0) rectangle (2,1);
\draw[fill=gray] (2,0) rectangle (3,1);
\draw[fill=gray] (3,0) rectangle (4,1);
\draw[fill=gray] (4,0) rectangle (5,1);
\draw[fill=gray] (5,0) rectangle (6,1);
\draw[fill=gray] (6,0) rectangle (7,1);
\draw[fill=gray] (7,0) rectangle (8,1);
\draw[fill=gray] (8,0) rectangle (9,1);
\draw[fill=gray] (9,0) rectangle (10,1);
\draw[fill=gray] (10,0) rectangle (11,1);
\draw[fill=gray] (0,1) rectangle (1,2);
\draw[fill=pink] (9.500000,1.500000) circle (0.500000);
\draw[fill=gray] (10,1) rectangle (11,2);
\draw[fill=gray] (0,2) rectangle (1,3);
\draw[fill=gray] (5,2) rectangle (6,3);
\draw[fill=gray] (6,2) rectangle (7,3);
\draw[fill=gray] (7,2) rectangle (8,3);
\draw[fill=gray] (8,2) rectangle (9,3);
\draw[fill={rgb,255:red,0.0;green,0.0;blue,255.0}]  (9,2) rectangle (10,3);
\draw[fill={rgb,255:red,255.0;green,214.0;blue,174.0}]  (9.250000,2.250000) rectangle (9.750000,2.750000);
\draw[fill=gray] (10,2) rectangle (11,3);
\draw[fill=gray] (0,3) rectangle (1,4);
\draw[fill=gray] (5,3) rectangle (6,4);
\draw[fill={rgb,255:red,0.0;green,0.0;blue,255.0}]  (9,3) rectangle (10,4);
\draw[fill={rgb,255:red,255.0;green,187.0;blue,120.0}]  (9.250000,3.250000) rectangle (9.750000,3.750000);
\draw[fill=gray] (10,3) rectangle (11,4);
\draw[fill=gray] (0,4) rectangle (1,5);
\draw[fill={rgb,255:red,0.0;green,0.0;blue,255.0}]  (9,4) rectangle (10,5);
\draw[fill={rgb,255:red,255.0;green,161.0;blue,67.0}]  (9.250000,4.250000) rectangle (9.750000,4.750000);
\draw[fill=gray] (10,4) rectangle (11,5);
\draw[fill=gray] (0,5) rectangle (1,6);
\draw[fill=gray] (5,5) rectangle (6,6);
\draw[fill={rgb,255:red,0.0;green,0.0;blue,255.0}]  (9,5) rectangle (10,6);
\draw[fill={rgb,255:red,255.0;green,147.0;blue,40.0}]  (9.250000,5.250000) rectangle (9.750000,5.750000);
\draw[fill=gray] (10,5) rectangle (11,6);
\draw[fill=gray] (0,6) rectangle (1,7);
\draw[fill=gray] (2,6) rectangle (3,7);
\draw[fill=gray] (3,6) rectangle (4,7);
\draw[fill=gray] (4,6) rectangle (5,7);
\draw[fill=gray] (5,6) rectangle (6,7);
\draw[fill={rgb,255:red,0.0;green,0.0;blue,255.0}]  (9,6) rectangle (10,7);
\draw[fill={rgb,255:red,255.0;green,174.0;blue,93.0}]  (9.250000,6.250000) rectangle (9.750000,6.750000);
\draw[fill=gray] (10,6) rectangle (11,7);
\draw[fill=gray] (0,7) rectangle (1,8);
\draw[fill={rgb,255:red,0.0;green,0.0;blue,255.0}]  (1,7) rectangle (2,8);
\draw[fill={rgb,255:red,255.0;green,228.0;blue,201.0}]  (1.250000,7.250000) rectangle (1.750000,7.750000);
\draw[fill={rgb,255:red,0.0;green,0.0;blue,255.0}]  (2,7) rectangle (3,8);
\draw[fill={rgb,255:red,0.0;green,0.0;blue,255.0}]  (3,7) rectangle (4,8);
\draw[fill={rgb,255:red,255.0;green,214.0;blue,174.0}]  (3.250000,7.250000) rectangle (3.750000,7.750000);
\draw[fill={rgb,255:red,0.0;green,0.0;blue,255.0}]  (4,7) rectangle (5,8);
\draw[fill={rgb,255:red,255.0;green,214.0;blue,174.0}]  (4.250000,7.250000) rectangle (4.750000,7.750000);
\draw[fill={rgb,255:red,0.0;green,0.0;blue,255.0}]  (5,7) rectangle (6,8);
\draw[fill={rgb,255:red,255.0;green,201.0;blue,147.0}]  (5.250000,7.250000) rectangle (5.750000,7.750000);
\draw[fill={rgb,255:red,0.0;green,0.0;blue,255.0}]  (6,7) rectangle (7,8);
\draw[fill={rgb,255:red,255.0;green,214.0;blue,174.0}]  (6.250000,7.250000) rectangle (6.750000,7.750000);
\draw[fill={rgb,255:red,0.0;green,0.0;blue,255.0}]  (7,7) rectangle (8,8);
\draw[fill={rgb,255:red,255.0;green,241.0;blue,228.0}]  (7.250000,7.250000) rectangle (7.750000,7.750000);
\draw[fill={rgb,255:red,0.0;green,0.0;blue,255.0}]  (8,7) rectangle (9,8);
\draw[fill={rgb,255:red,255.0;green,241.0;blue,228.0}]  (8.250000,7.250000) rectangle (8.750000,7.750000);
\draw[fill={rgb,255:red,0.0;green,0.0;blue,255.0}]  (9,7) rectangle (10,8);
\draw[fill={rgb,255:red,255.0;green,228.0;blue,201.0}]  (9.250000,7.250000) rectangle (9.750000,7.750000);
\draw[fill=gray] (10,7) rectangle (11,8);
\draw[fill=gray] (0,8) rectangle (1,9);
\draw[fill={rgb,255:red,0.0;green,0.0;blue,255.0}]  (1,8) rectangle (2,9);
\draw[fill={rgb,255:red,255.0;green,241.0;blue,228.0}]  (1.250000,8.250000) rectangle (1.750000,8.750000);
\draw[fill=gray] (2,8) rectangle (3,9);
\draw[fill=gray] (3,8) rectangle (4,9);
\draw[fill=gray] (4,8) rectangle (5,9);
\draw[fill=gray] (5,8) rectangle (6,9);
\draw[fill=gray] (6,8) rectangle (7,9);
\draw[fill=gray] (7,8) rectangle (8,9);
\draw[fill=gray] (8,8) rectangle (9,9);
\draw[fill=gray] (9,8) rectangle (10,9);
\draw[fill=gray] (10,8) rectangle (11,9);
\draw[fill=gray] (0,9) rectangle (1,10);
\draw[fill={rgb,255:red,0.0;green,0.0;blue,255.0}]  (1,9) rectangle (2,10);
\draw[fill=gray] (2,9) rectangle (3,10);
\draw[fill=gray] (3,9) rectangle (4,10);
\draw[fill=gray] (4,9) rectangle (5,10);
\draw[fill=gray] (5,9) rectangle (6,10);
\draw[fill=gray] (6,9) rectangle (7,10);
\draw[fill=gray] (7,9) rectangle (8,10);
\draw[fill=gray] (8,9) rectangle (9,10);
\draw[fill=gray] (9,9) rectangle (10,10);
\draw[fill=gray] (10,9) rectangle (11,10);
\draw[fill=gray] (0,10) rectangle (1,11);
\draw[fill=gray] (1,10) rectangle (2,11);
\draw[fill=gray] (2,10) rectangle (3,11);
\draw[fill=gray] (3,10) rectangle (4,11);
\draw[fill=gray] (4,10) rectangle (5,11);
\draw[fill=gray] (5,10) rectangle (6,11);
\draw[fill=gray] (6,10) rectangle (7,11);
\draw[fill=gray] (7,10) rectangle (8,11);
\draw[fill=gray] (8,10) rectangle (9,11);
\draw[fill=gray] (9,10) rectangle (10,11);
\draw[fill=gray] (10,10) rectangle (11,11);
\node [align=center] at(-0.500000,0.500000) {0};
\node [align=center] at(-0.500000,1.500000) {1};
\node [align=center] at(-0.500000,2.500000) {2};
\node [align=center] at(-0.500000,3.500000) {3};
\node [align=center] at(-0.500000,4.500000) {4};
\node [align=center] at(-0.500000,5.500000) {5};
\node [align=center] at(-0.500000,6.500000) {6};
\node [align=center] at(-0.500000,7.500000) {7};
\node [align=center] at(-0.500000,8.500000) {8};
\node [align=center] at(-0.500000,9.500000) {9};
\node [align=center] at(-0.500000,10.500000) {10};
\node [align=center] at(0.500000,-0.500000) {A};
\node [align=center] at(1.500000,-0.500000) {B};
\node [align=center] at(2.500000,-0.500000) {C};
\node [align=center] at(3.500000,-0.500000) {D};
\node [align=center] at(4.500000,-0.500000) {E};
\node [align=center] at(5.500000,-0.500000) {F};
\node [align=center] at(6.500000,-0.500000) {G};
\node [align=center] at(7.500000,-0.500000) {H};
\node [align=center] at(8.500000,-0.500000) {I};
\node [align=center] at(9.500000,-0.500000) {J};
\node [align=center] at(10.500000,-0.500000) {K};
\end{tikzpicture}
      }
    \end{minipage}

    \centerline{\small $M_{4}$}
  \end{minipage}

  \medskip

  \begin{minipage}{\colwidth}
    \begin{minipage}{\mpwidth}
      \centering
      \adjustbox{width=\aboxwidth}{
        \input{\dir 6_MDP_policyStochHeatmap.tikz}
      }
    \end{minipage}
    \hfill
    \begin{minipage}{\mpwidth}
      \centering
      \adjustbox{width=\aboxwidth}{
        \begin{tikzpicture}
\draw (0,0) grid (11,12);
\draw[fill=gray] (0,0) rectangle (1,1);
\draw[fill=gray] (1,0) rectangle (2,1);
\draw[fill=gray] (2,0) rectangle (3,1);
\draw[fill=gray] (3,0) rectangle (4,1);
\draw[fill=gray] (4,0) rectangle (5,1);
\draw[fill=gray] (5,0) rectangle (6,1);
\draw[fill=gray] (6,0) rectangle (7,1);
\draw[fill=gray] (7,0) rectangle (8,1);
\draw[fill=gray] (8,0) rectangle (9,1);
\draw[fill=gray] (9,0) rectangle (10,1);
\draw[fill=gray] (10,0) rectangle (11,1);
\draw[fill=gray] (0,1) rectangle (1,2);
\draw[fill={rgb,255:red,0.0;green,0.0;blue,255.0}]  (1,1) rectangle (2,2);
\draw[fill=gray] (2,1) rectangle (3,2);
\draw[fill=gray] (3,1) rectangle (4,2);
\draw[fill=gray] (4,1) rectangle (5,2);
\draw[fill=gray] (5,1) rectangle (6,2);
\draw[fill=gray] (6,1) rectangle (7,2);
\draw[fill=gray] (7,1) rectangle (8,2);
\draw[fill=gray] (8,1) rectangle (9,2);
\draw[fill=gray] (9,1) rectangle (10,2);
\draw[fill=gray] (10,1) rectangle (11,2);
\draw[fill=gray] (0,2) rectangle (1,3);
\draw[fill={rgb,255:red,0.0;green,0.0;blue,255.0}]  (1,2) rectangle (2,3);
\draw[fill={rgb,255:red,255.0;green,228.0;blue,201.0}]  (1.250000,2.250000) rectangle (1.750000,2.750000);
\draw[fill=gray] (2,2) rectangle (3,3);
\draw[fill=gray] (3,2) rectangle (4,3);
\draw[fill=gray] (4,2) rectangle (5,3);
\draw[fill=gray] (5,2) rectangle (6,3);
\draw[fill=gray] (6,2) rectangle (7,3);
\draw[fill=gray] (7,2) rectangle (8,3);
\draw[fill=gray] (8,2) rectangle (9,3);
\draw[fill=gray] (9,2) rectangle (10,3);
\draw[fill=gray] (10,2) rectangle (11,3);
\draw[fill=gray] (0,3) rectangle (1,4);
\draw[fill={rgb,255:red,0.0;green,0.0;blue,255.0}]  (1,3) rectangle (2,4);
\draw[fill={rgb,255:red,0.0;green,0.0;blue,255.0}]  (2,3) rectangle (3,4);
\draw[fill={rgb,255:red,255.0;green,241.0;blue,228.0}]  (2.250000,3.250000) rectangle (2.750000,3.750000);
\draw[fill={rgb,255:red,0.0;green,0.0;blue,255.0}]  (3,3) rectangle (4,4);
\draw[fill={rgb,255:red,255.0;green,187.0;blue,120.0}]  (3.250000,3.250000) rectangle (3.750000,3.750000);
\draw[fill={rgb,255:red,0.0;green,0.0;blue,255.0}]  (4,3) rectangle (5,4);
\draw[fill={rgb,255:red,255.0;green,201.0;blue,147.0}]  (4.250000,3.250000) rectangle (4.750000,3.750000);
\draw[fill={rgb,255:red,0.0;green,0.0;blue,255.0}]  (5,3) rectangle (6,4);
\draw[fill={rgb,255:red,255.0;green,187.0;blue,120.0}]  (5.250000,3.250000) rectangle (5.750000,3.750000);
\draw[fill={rgb,255:red,0.0;green,0.0;blue,255.0}]  (6,3) rectangle (7,4);
\draw[fill={rgb,255:red,255.0;green,228.0;blue,201.0}]  (6.250000,3.250000) rectangle (6.750000,3.750000);
\draw[fill={rgb,255:red,0.0;green,0.0;blue,255.0}]  (7,3) rectangle (8,4);
\draw[fill={rgb,255:red,0.0;green,0.0;blue,255.0}]  (8,3) rectangle (9,4);
\draw[fill={rgb,255:red,255.0;green,241.0;blue,228.0}]  (8.250000,3.250000) rectangle (8.750000,3.750000);
\draw[fill={rgb,255:red,0.0;green,0.0;blue,255.0}]  (9,3) rectangle (10,4);
\draw[fill={rgb,255:red,255.0;green,241.0;blue,228.0}]  (9.250000,3.250000) rectangle (9.750000,3.750000);
\draw[fill=gray] (10,3) rectangle (11,4);
\draw[fill=gray] (0,4) rectangle (1,5);
\draw[fill=gray] (2,4) rectangle (3,5);
\draw[fill=gray] (3,4) rectangle (4,5);
\draw[fill=gray] (4,4) rectangle (5,5);
\draw[fill=gray] (5,4) rectangle (6,5);
\draw[fill={rgb,255:red,0.0;green,0.0;blue,255.0}]  (9,4) rectangle (10,5);
\draw[fill={rgb,255:red,255.0;green,201.0;blue,147.0}]  (9.250000,4.250000) rectangle (9.750000,4.750000);
\draw[fill=gray] (10,4) rectangle (11,5);
\draw[fill=gray] (0,5) rectangle (1,6);
\draw[fill=gray] (5,5) rectangle (6,6);
\draw[fill={rgb,255:red,0.0;green,0.0;blue,255.0}]  (9,5) rectangle (10,6);
\draw[fill={rgb,255:red,255.0;green,201.0;blue,147.0}]  (9.250000,5.250000) rectangle (9.750000,5.750000);
\draw[fill=gray] (10,5) rectangle (11,6);
\draw[fill=gray] (0,6) rectangle (1,7);
\draw[fill=gray] (5,6) rectangle (6,7);
\draw[fill={rgb,255:red,0.0;green,0.0;blue,255.0}]  (9,6) rectangle (10,7);
\draw[fill={rgb,255:red,255.0;green,174.0;blue,93.0}]  (9.250000,6.250000) rectangle (9.750000,6.750000);
\draw[fill=gray] (10,6) rectangle (11,7);
\draw[fill=gray] (0,7) rectangle (1,8);
\draw[fill=gray] (5,7) rectangle (6,8);
\draw[fill={rgb,255:red,0.0;green,0.0;blue,255.0}]  (9,7) rectangle (10,8);
\draw[fill={rgb,255:red,255.0;green,187.0;blue,120.0}]  (9.250000,7.250000) rectangle (9.750000,7.750000);
\draw[fill=gray] (10,7) rectangle (11,8);
\draw[fill=gray] (0,8) rectangle (1,9);
\draw[fill=gray] (5,8) rectangle (6,9);
\draw[fill={rgb,255:red,0.0;green,0.0;blue,255.0}]  (9,8) rectangle (10,9);
\draw[fill={rgb,255:red,255.0;green,201.0;blue,147.0}]  (9.250000,8.250000) rectangle (9.750000,8.750000);
\draw[fill=gray] (10,8) rectangle (11,9);
\draw[fill=gray] (0,9) rectangle (1,10);
\draw[fill=gray] (5,9) rectangle (6,10);
\draw[fill={rgb,255:red,0.0;green,0.0;blue,255.0}]  (9,9) rectangle (10,10);
\draw[fill={rgb,255:red,255.0;green,214.0;blue,174.0}]  (9.250000,9.250000) rectangle (9.750000,9.750000);
\draw[fill=gray] (10,9) rectangle (11,10);
\draw[fill=gray] (0,10) rectangle (1,11);
\draw[fill=pink] (9.500000,10.500000) circle (0.500000);
\draw[fill=gray] (10,10) rectangle (11,11);
\draw[fill=gray] (0,11) rectangle (1,12);
\draw[fill=gray] (1,11) rectangle (2,12);
\draw[fill=gray] (2,11) rectangle (3,12);
\draw[fill=gray] (3,11) rectangle (4,12);
\draw[fill=gray] (4,11) rectangle (5,12);
\draw[fill=gray] (5,11) rectangle (6,12);
\draw[fill=gray] (6,11) rectangle (7,12);
\draw[fill=gray] (7,11) rectangle (8,12);
\draw[fill=gray] (8,11) rectangle (9,12);
\draw[fill=gray] (9,11) rectangle (10,12);
\draw[fill=gray] (10,11) rectangle (11,12);
\node [align=center] at(-0.500000,0.500000) {0};
\node [align=center] at(-0.500000,1.500000) {1};
\node [align=center] at(-0.500000,2.500000) {2};
\node [align=center] at(-0.500000,3.500000) {3};
\node [align=center] at(-0.500000,4.500000) {4};
\node [align=center] at(-0.500000,5.500000) {5};
\node [align=center] at(-0.500000,6.500000) {6};
\node [align=center] at(-0.500000,7.500000) {7};
\node [align=center] at(-0.500000,8.500000) {8};
\node [align=center] at(-0.500000,9.500000) {9};
\node [align=center] at(-0.500000,10.500000) {10};
\node [align=center] at(-0.500000,11.500000) {11};
\node [align=center] at(0.500000,-0.500000) {A};
\node [align=center] at(1.500000,-0.500000) {B};
\node [align=center] at(2.500000,-0.500000) {C};
\node [align=center] at(3.500000,-0.500000) {D};
\node [align=center] at(4.500000,-0.500000) {E};
\node [align=center] at(5.500000,-0.500000) {F};
\node [align=center] at(6.500000,-0.500000) {G};
\node [align=center] at(7.500000,-0.500000) {H};
\node [align=center] at(8.500000,-0.500000) {I};
\node [align=center] at(9.500000,-0.500000) {J};
\node [align=center] at(10.500000,-0.500000) {K};
\end{tikzpicture}
      }
    \end{minipage}
    \hfill
    \begin{minipage}{\mpwidth}
      \centering
      \adjustbox{width=\aboxwidth}{
        \begin{tikzpicture}
\draw (0,0) grid (11,12);
\draw[fill=gray] (0,0) rectangle (1,1);
\draw[fill=gray] (1,0) rectangle (2,1);
\draw[fill=gray] (2,0) rectangle (3,1);
\draw[fill=gray] (3,0) rectangle (4,1);
\draw[fill=gray] (4,0) rectangle (5,1);
\draw[fill=gray] (5,0) rectangle (6,1);
\draw[fill=gray] (6,0) rectangle (7,1);
\draw[fill=gray] (7,0) rectangle (8,1);
\draw[fill=gray] (8,0) rectangle (9,1);
\draw[fill=gray] (9,0) rectangle (10,1);
\draw[fill=gray] (10,0) rectangle (11,1);
\draw[fill=gray] (0,1) rectangle (1,2);
\draw[fill={rgb,255:red,0.0;green,0.0;blue,255.0}]  (1,1) rectangle (2,2);
\draw[fill=gray] (2,1) rectangle (3,2);
\draw[fill=gray] (3,1) rectangle (4,2);
\draw[fill=gray] (4,1) rectangle (5,2);
\draw[fill=gray] (5,1) rectangle (6,2);
\draw[fill=gray] (6,1) rectangle (7,2);
\draw[fill=gray] (7,1) rectangle (8,2);
\draw[fill=gray] (8,1) rectangle (9,2);
\draw[fill=gray] (9,1) rectangle (10,2);
\draw[fill=gray] (10,1) rectangle (11,2);
\draw[fill=gray] (0,2) rectangle (1,3);
\draw[fill={rgb,255:red,0.0;green,0.0;blue,255.0}]  (1,2) rectangle (2,3);
\draw[fill={rgb,255:red,255.0;green,228.0;blue,201.0}]  (1.250000,2.250000) rectangle (1.750000,2.750000);
\draw[fill=gray] (2,2) rectangle (3,3);
\draw[fill=gray] (3,2) rectangle (4,3);
\draw[fill=gray] (4,2) rectangle (5,3);
\draw[fill=gray] (5,2) rectangle (6,3);
\draw[fill=gray] (6,2) rectangle (7,3);
\draw[fill=gray] (7,2) rectangle (8,3);
\draw[fill=gray] (8,2) rectangle (9,3);
\draw[fill=gray] (9,2) rectangle (10,3);
\draw[fill=gray] (10,2) rectangle (11,3);
\draw[fill=gray] (0,3) rectangle (1,4);
\draw[fill={rgb,255:red,0.0;green,0.0;blue,255.0}]  (1,3) rectangle (2,4);
\draw[fill={rgb,255:red,134.0;green,134.0;blue,255.0}]  (2,3) rectangle (3,4);
\draw[fill={rgb,255:red,134.0;green,134.0;blue,255.0}]  (3,3) rectangle (4,4);
\draw[fill={rgb,255:red,255.0;green,226.0;blue,198.0}]  (3.250000,3.250000) rectangle (3.750000,3.750000);
\draw[fill={rgb,255:red,134.0;green,134.0;blue,255.0}]  (4,3) rectangle (5,4);
\draw[fill={rgb,255:red,255.0;green,198.0;blue,141.0}]  (4.250000,3.250000) rectangle (4.750000,3.750000);
\draw[fill={rgb,255:red,134.0;green,134.0;blue,255.0}]  (5,3) rectangle (6,4);
\draw[fill={rgb,255:red,255.0;green,170.0;blue,85.0}]  (5.250000,3.250000) rectangle (5.750000,3.750000);
\draw[fill={rgb,255:red,134.0;green,134.0;blue,255.0}]  (6,3) rectangle (7,4);
\draw[fill={rgb,255:red,134.0;green,134.0;blue,255.0}]  (7,3) rectangle (8,4);
\draw[fill={rgb,255:red,134.0;green,134.0;blue,255.0}]  (8,3) rectangle (9,4);
\draw[fill={rgb,255:red,134.0;green,134.0;blue,255.0}]  (9,3) rectangle (10,4);
\draw[fill=gray] (10,3) rectangle (11,4);
\draw[fill=gray] (0,4) rectangle (1,5);
\draw[fill={rgb,255:red,120.0;green,120.0;blue,255.0}]  (1,4) rectangle (2,5);
\draw[fill={rgb,255:red,255.0;green,229.0;blue,204.0}]  (1.250000,4.250000) rectangle (1.750000,4.750000);
\draw[fill=gray] (2,4) rectangle (3,5);
\draw[fill=gray] (3,4) rectangle (4,5);
\draw[fill=gray] (4,4) rectangle (5,5);
\draw[fill=gray] (5,4) rectangle (6,5);
\draw[fill={rgb,255:red,134.0;green,134.0;blue,255.0}]  (9,4) rectangle (10,5);
\draw[fill={rgb,255:red,255.0;green,170.0;blue,85.0}]  (9.250000,4.250000) rectangle (9.750000,4.750000);
\draw[fill=gray] (10,4) rectangle (11,5);
\draw[fill=gray] (0,5) rectangle (1,6);
\draw[fill={rgb,255:red,120.0;green,120.0;blue,255.0}]  (1,5) rectangle (2,6);
\draw[fill={rgb,255:red,255.0;green,229.0;blue,204.0}]  (1.250000,5.250000) rectangle (1.750000,5.750000);
\draw[fill={rgb,255:red,120.0;green,120.0;blue,255.0}]  (2,5) rectangle (3,6);
\draw[fill={rgb,255:red,255.0;green,178.0;blue,102.0}]  (2.250000,5.250000) rectangle (2.750000,5.750000);
\draw[fill={rgb,255:red,120.0;green,120.0;blue,255.0}]  (3,5) rectangle (4,6);
\draw[fill={rgb,255:red,120.0;green,120.0;blue,255.0}]  (4,5) rectangle (5,6);
\draw[fill={rgb,255:red,255.0;green,204.0;blue,153.0}]  (4.250000,5.250000) rectangle (4.750000,5.750000);
\draw[fill=gray] (5,5) rectangle (6,6);
\draw[fill={rgb,255:red,134.0;green,134.0;blue,255.0}]  (9,5) rectangle (10,6);
\draw[fill={rgb,255:red,255.0;green,170.0;blue,85.0}]  (9.250000,5.250000) rectangle (9.750000,5.750000);
\draw[fill=gray] (10,5) rectangle (11,6);
\draw[fill=gray] (0,6) rectangle (1,7);
\draw[fill={rgb,255:red,120.0;green,120.0;blue,255.0}]  (4,6) rectangle (5,7);
\draw[fill={rgb,255:red,255.0;green,178.0;blue,102.0}]  (4.250000,6.250000) rectangle (4.750000,6.750000);
\draw[fill=gray] (5,6) rectangle (6,7);
\draw[fill={rgb,255:red,134.0;green,134.0;blue,255.0}]  (9,6) rectangle (10,7);
\draw[fill={rgb,255:red,255.0;green,226.0;blue,198.0}]  (9.250000,6.250000) rectangle (9.750000,6.750000);
\draw[fill=gray] (10,6) rectangle (11,7);
\draw[fill=gray] (0,7) rectangle (1,8);
\draw[fill={rgb,255:red,120.0;green,120.0;blue,255.0}]  (4,7) rectangle (5,8);
\draw[fill=gray] (5,7) rectangle (6,8);
\draw[fill={rgb,255:red,134.0;green,134.0;blue,255.0}]  (9,7) rectangle (10,8);
\draw[fill={rgb,255:red,255.0;green,226.0;blue,198.0}]  (9.250000,7.250000) rectangle (9.750000,7.750000);
\draw[fill=gray] (10,7) rectangle (11,8);
\draw[fill=gray] (0,8) rectangle (1,9);
\draw[fill={rgb,255:red,120.0;green,120.0;blue,255.0}]  (4,8) rectangle (5,9);
\draw[fill=gray] (5,8) rectangle (6,9);
\draw[fill={rgb,255:red,134.0;green,134.0;blue,255.0}]  (9,8) rectangle (10,9);
\draw[fill={rgb,255:red,255.0;green,226.0;blue,198.0}]  (9.250000,8.250000) rectangle (9.750000,8.750000);
\draw[fill=gray] (10,8) rectangle (11,9);
\draw[fill=gray] (0,9) rectangle (1,10);
\draw[fill={rgb,255:red,120.0;green,120.0;blue,255.0}]  (4,9) rectangle (5,10);
\draw[fill={rgb,255:red,255.0;green,229.0;blue,204.0}]  (4.250000,9.250000) rectangle (4.750000,9.750000);
\draw[fill=gray] (5,9) rectangle (6,10);
\draw[fill={rgb,255:red,134.0;green,134.0;blue,255.0}]  (9,9) rectangle (10,10);
\draw[fill=gray] (10,9) rectangle (11,10);
\draw[fill=gray] (0,10) rectangle (1,11);
\draw[fill={rgb,255:red,120.0;green,120.0;blue,255.0}]  (4,10) rectangle (5,11);
\draw[fill={rgb,255:red,120.0;green,120.0;blue,255.0}]  (5,10) rectangle (6,11);
\draw[fill={rgb,255:red,255.0;green,229.0;blue,204.0}]  (5.250000,10.250000) rectangle (5.750000,10.750000);
\draw[fill={rgb,255:red,120.0;green,120.0;blue,255.0}]  (6,10) rectangle (7,11);
\draw[fill={rgb,255:red,255.0;green,153.0;blue,50.0}]  (6.250000,10.250000) rectangle (6.750000,10.750000);
\draw[fill={rgb,255:red,120.0;green,120.0;blue,255.0}]  (7,10) rectangle (8,11);
\draw[fill={rgb,255:red,255.0;green,204.0;blue,153.0}]  (7.250000,10.250000) rectangle (7.750000,10.750000);
\draw[fill={rgb,255:red,120.0;green,120.0;blue,255.0}]  (8,10) rectangle (9,11);
\draw[fill=pink] (9.500000,10.500000) circle (0.500000);
\draw[fill=gray] (10,10) rectangle (11,11);
\draw[fill=gray] (0,11) rectangle (1,12);
\draw[fill=gray] (1,11) rectangle (2,12);
\draw[fill=gray] (2,11) rectangle (3,12);
\draw[fill=gray] (3,11) rectangle (4,12);
\draw[fill=gray] (4,11) rectangle (5,12);
\draw[fill=gray] (5,11) rectangle (6,12);
\draw[fill=gray] (6,11) rectangle (7,12);
\draw[fill=gray] (7,11) rectangle (8,12);
\draw[fill=gray] (8,11) rectangle (9,12);
\draw[fill=gray] (9,11) rectangle (10,12);
\draw[fill=gray] (10,11) rectangle (11,12);
\node [align=center] at(-0.500000,0.500000) {0};
\node [align=center] at(-0.500000,1.500000) {1};
\node [align=center] at(-0.500000,2.500000) {2};
\node [align=center] at(-0.500000,3.500000) {3};
\node [align=center] at(-0.500000,4.500000) {4};
\node [align=center] at(-0.500000,5.500000) {5};
\node [align=center] at(-0.500000,6.500000) {6};
\node [align=center] at(-0.500000,7.500000) {7};
\node [align=center] at(-0.500000,8.500000) {8};
\node [align=center] at(-0.500000,9.500000) {9};
\node [align=center] at(-0.500000,10.500000) {10};
\node [align=center] at(-0.500000,11.500000) {11};
\node [align=center] at(0.500000,-0.500000) {A};
\node [align=center] at(1.500000,-0.500000) {B};
\node [align=center] at(2.500000,-0.500000) {C};
\node [align=center] at(3.500000,-0.500000) {D};
\node [align=center] at(4.500000,-0.500000) {E};
\node [align=center] at(5.500000,-0.500000) {F};
\node [align=center] at(6.500000,-0.500000) {G};
\node [align=center] at(7.500000,-0.500000) {H};
\node [align=center] at(8.500000,-0.500000) {I};
\node [align=center] at(9.500000,-0.500000) {J};
\node [align=center] at(10.500000,-0.500000) {K};
\end{tikzpicture}
      }
    \end{minipage}

    \centerline{\small $M_{5}$}
  \end{minipage}

  \hfill
  
  \begin{minipage}{\colwidth}
    \begin{minipage}{\mpwidth}
      \centering
      \adjustbox{width=\aboxwidth}{
        \input{\dir 7_MDP_policyStochHeatmap.tikz}
      }
    \end{minipage}
    \hfill
    \begin{minipage}{\mpwidth}
      \centering
      \adjustbox{width=\aboxwidth}{
        \input{\dir 7_MDP_policyBiasedHeatMap.tikz}
      }
    \end{minipage}
    \hfill
    \begin{minipage}{\mpwidth}
      \centering
      \adjustbox{width=\aboxwidth}{
        \input{\dir 7_OAMDP_policyStochHeatMap.tikz}
      }
    \end{minipage}

    \centerline{\small $M_{7}$}
  \end{minipage}

  \medskip

  \def\colwidth{.8\textwidth}

  \begin{minipage}{\colwidth}
    \begin{minipage}{\mpwidth}
      \centering
      \adjustbox{width=\aboxwidth}{
        \input{\dir 4_MDP_policyStochHeatmap.tikz}
      }
    \end{minipage}
    \hfill
    \begin{minipage}{\mpwidth}
      \centering
      \adjustbox{width=\aboxwidth}{
        \input{\dir 4_MDP_policyBiasedHeatMap.tikz}
      }
    \end{minipage}
    \hfill
    \begin{minipage}{\mpwidth}
      \centering
      \adjustbox{width=\aboxwidth}{
        \input{\dir 4_OAMDP_policyStochHeatMap.tikz}
      }
    \end{minipage}

    \centerline{\small $M_{6}$}
  \end{minipage}

\caption{[Response-Time] Heatmaps showing, for each cell,
      (1) [in background] the probability of visit during a trajectory from dark blue ($P=1$) to white ($P=0$), and
      (2) [in the middle] the response time from green (1000\,ms) to white (0\,ms).
      These heatmaps are provided for each maze and each policy, from left to right: stochastic MDP policy, biased MDP policy, and pOAMDP policy.
      \label{fig|heatmapTime}}

\end{figure}

The response-time heatmaps show that:
\begin{itemize}
\item in open areas, the participants take more time to make a decision;
\item consistent with the error heat map, for the OAMDP policy, participants take more time to make decision in the beginning, where they need to make a choice;
\item in the OAMDP policy, we can notice that, after unexpected actions, the participants take more time for the next prediction.
  For example in $M_1$, cell $(C,3)$, the robot action surprised the participants and their response time is higher in the next few cells.
  These cells also match the cells that were wrongly predicted in the error rate heatmap;
\item the response time in corridors is less important, which is due to the participants anticipation.
\end{itemize}

\paragraph{Heatmaps of ``Reflex'' Responses}
Heatmaps of ``reflex'' responses for each maze in each policy are shown on \Cref{fig|heatmap|timeMaptime150}.
In case of a response time below 150\,ms, the decision is more of a reflex, and the participant will not be able to correct his choice if needed \cite{henry1960increased,marin2019neurosciences}.
These heatmaps show which portion of participants answer in less than 150\,ms in a state.
They are more precisely defined as follows:
\begin{itemize}
	\item the blue color represents the average number of visits computed as $\frac{\#visits}{\#participants}$, and
	\item the orange color represents the rate of response times below 150\,ms, computed as $\frac{N}{\#visits}$ with $N$ the number of times a participant answers in less than 150\,ms.
\end{itemize}
To make the heatmaps more readable, we cap the values to a 50\% rate of response times below 150\,ms.
The goal of this heatmap was to see where humans tend to predict the robot's action using anticipation or reflex responses. 
For example, when asked to predict the robot's move in a corridor at time $t$, some human participants might anticipate the prediction they will have to make at $t+1$ (and afterwards), thus pressing keys as fast as possible along the corridor.
This should result in  a series of response times below 150\,ms.
Note that not all participants will anticipate the robot's next move at $t+1$, and some of them might wait for the screen to redraw the robot's new position at $t+1$ before answering.

This is also the reason why we use the 50\% cap. 

These response time  heatmaps show that:
\begin{itemize}
\item in rooms, most response times are above 150\,ms, except along walls, either because following the wall is optimal, or because the agent is obviously going in a straight line.
Note that the stochastic MDP policy has more chances to go through the center of rooms than to follow walls;
  
\item in the corridors, many participants answer in less than 150\,ms, as can be seen in $M_1$ and in $M_2$ for example.
As explained earlier, in straight line, participants can anticipate the next action and be much faster.
\end{itemize}

\def\colwidth{.48\textwidth}
\def\mpwidth{.32\textwidth}
\def\aboxwidth{\textwidth}

\def\dir{expFinalVers/heatmap/time150/}

\begin{figure}[ht!]

  \begin{minipage}{\colwidth}
    \begin{minipage}{\mpwidth}
      \centering
      \adjustbox{width=\aboxwidth}{
        \input{\dir 1_MDP_policyStochHeatmap.tikz}
      }
    \end{minipage}
    \hfill
    \begin{minipage}{\mpwidth}
      \centering
      \adjustbox{width=\aboxwidth}{
        \begin{tikzpicture}
\draw (0,0) grid (11,12);
\draw[fill=gray] (0,0) rectangle (1,1);
\draw[fill=gray] (1,0) rectangle (2,1);
\draw[fill=gray] (2,0) rectangle (3,1);
\draw[fill=gray] (3,0) rectangle (4,1);
\draw[fill=gray] (4,0) rectangle (5,1);
\draw[fill=gray] (5,0) rectangle (6,1);
\draw[fill=gray] (6,0) rectangle (7,1);
\draw[fill=gray] (7,0) rectangle (8,1);
\draw[fill=gray] (8,0) rectangle (9,1);
\draw[fill=gray] (9,0) rectangle (10,1);
\draw[fill=gray] (10,0) rectangle (11,1);
\draw[fill=gray] (0,1) rectangle (1,2);
\draw[fill=gray] (1,1) rectangle (2,2);
\draw[fill={rgb,255:red,0.0;green,0.0;blue,255.0}]  (2,1) rectangle (3,2);
\draw[fill=gray] (3,1) rectangle (4,2);
\draw[fill=gray] (4,1) rectangle (5,2);
\draw[fill=gray] (5,1) rectangle (6,2);
\draw[fill=gray] (6,1) rectangle (7,2);
\draw[fill=gray] (7,1) rectangle (8,2);
\draw[fill=gray] (8,1) rectangle (9,2);
\draw[fill=gray] (9,1) rectangle (10,2);
\draw[fill=gray] (10,1) rectangle (11,2);
\draw[fill=gray] (0,2) rectangle (1,3);
\draw[fill=gray] (1,2) rectangle (2,3);
\draw[fill={rgb,255:red,0.0;green,0.0;blue,255.0}]  (2,2) rectangle (3,3);
\draw[fill={rgb,255:red,255.0;green,241.0;blue,228.0}]  (2.250000,2.250000) rectangle (2.750000,2.750000);
\draw[fill=gray] (3,2) rectangle (4,3);
\draw[fill=gray] (4,2) rectangle (5,3);
\draw[fill=gray] (5,2) rectangle (6,3);
\draw[fill=gray] (6,2) rectangle (7,3);
\draw[fill=gray] (7,2) rectangle (8,3);
\draw[fill=gray] (8,2) rectangle (9,3);
\draw[fill=gray] (9,2) rectangle (10,3);
\draw[fill=gray] (10,2) rectangle (11,3);
\draw[fill=gray] (0,3) rectangle (1,4);
\draw[fill={rgb,255:red,0.0;green,0.0;blue,255.0}]  (2,3) rectangle (3,4);
\draw[fill={rgb,255:red,255.0;green,214.0;blue,174.0}]  (2.250000,3.250000) rectangle (2.750000,3.750000);
\draw[fill={rgb,255:red,0.0;green,0.0;blue,255.0}]  (3,3) rectangle (4,4);
\draw[fill={rgb,255:red,255.0;green,228.0;blue,201.0}]  (3.250000,3.250000) rectangle (3.750000,3.750000);
\draw[fill={rgb,255:red,0.0;green,0.0;blue,255.0}]  (4,3) rectangle (5,4);
\draw[fill={rgb,255:red,255.0;green,241.0;blue,228.0}]  (4.250000,3.250000) rectangle (4.750000,3.750000);
\draw[fill={rgb,255:red,0.0;green,0.0;blue,255.0}]  (5,3) rectangle (6,4);
\draw[fill={rgb,255:red,255.0;green,241.0;blue,228.0}]  (5.250000,3.250000) rectangle (5.750000,3.750000);
\draw[fill=gray] (6,3) rectangle (7,4);
\draw[fill=gray] (7,3) rectangle (8,4);
\draw[fill=gray] (8,3) rectangle (9,4);
\draw[fill=gray] (9,3) rectangle (10,4);
\draw[fill=gray] (10,3) rectangle (11,4);
\draw[fill=gray] (0,4) rectangle (1,5);
\draw[fill=gray] (2,4) rectangle (3,5);
\draw[fill=gray] (3,4) rectangle (4,5);
\draw[fill=gray] (4,4) rectangle (5,5);
\draw[fill={rgb,255:red,0.0;green,0.0;blue,255.0}]  (5,4) rectangle (6,5);
\draw[fill={rgb,255:red,255.0;green,228.0;blue,201.0}]  (5.250000,4.250000) rectangle (5.750000,4.750000);
\draw[fill=gray] (10,4) rectangle (11,5);
\draw[fill=gray] (0,5) rectangle (1,6);
\draw[fill=gray] (2,5) rectangle (3,6);
\draw[fill=gray] (3,5) rectangle (4,6);
\draw[fill=gray] (4,5) rectangle (5,6);
\draw[fill={rgb,255:red,0.0;green,0.0;blue,255.0}]  (5,5) rectangle (6,6);
\draw[fill=gray] (10,5) rectangle (11,6);
\draw[fill=gray] (0,6) rectangle (1,7);
\draw[fill=gray] (2,6) rectangle (3,7);
\draw[fill=gray] (3,6) rectangle (4,7);
\draw[fill=gray] (4,6) rectangle (5,7);
\draw[fill={rgb,255:red,0.0;green,0.0;blue,255.0}]  (5,6) rectangle (6,7);
\draw[fill=gray] (10,6) rectangle (11,7);
\draw[fill=gray] (0,7) rectangle (1,8);
\draw[fill=gray] (2,7) rectangle (3,8);
\draw[fill=gray] (3,7) rectangle (4,8);
\draw[fill=gray] (4,7) rectangle (5,8);
\draw[fill={rgb,255:red,0.0;green,0.0;blue,255.0}]  (5,7) rectangle (6,8);
\draw[fill={rgb,255:red,255.0;green,241.0;blue,228.0}]  (5.250000,7.250000) rectangle (5.750000,7.750000);
\draw[fill=gray] (10,7) rectangle (11,8);
\draw[fill=gray] (0,8) rectangle (1,9);
\draw[fill=gray] (2,8) rectangle (3,9);
\draw[fill=gray] (3,8) rectangle (4,9);
\draw[fill=gray] (4,8) rectangle (5,9);
\draw[fill={rgb,255:red,0.0;green,0.0;blue,255.0}]  (5,8) rectangle (6,9);
\draw[fill={rgb,255:red,255.0;green,241.0;blue,228.0}]  (5.250000,8.250000) rectangle (5.750000,8.750000);
\draw[fill={rgb,255:red,0.0;green,0.0;blue,255.0}]  (6,8) rectangle (7,9);
\draw[fill={rgb,255:red,255.0;green,214.0;blue,174.0}]  (6.250000,8.250000) rectangle (6.750000,8.750000);
\draw[fill={rgb,255:red,0.0;green,0.0;blue,255.0}]  (7,8) rectangle (8,9);
\draw[fill={rgb,255:red,255.0;green,214.0;blue,174.0}]  (7.250000,8.250000) rectangle (7.750000,8.750000);
\draw[fill={rgb,255:red,0.0;green,0.0;blue,255.0}]  (8,8) rectangle (9,9);
\draw[fill={rgb,255:red,255.0;green,228.0;blue,201.0}]  (8.250000,8.250000) rectangle (8.750000,8.750000);
\draw[fill={rgb,255:red,0.0;green,0.0;blue,255.0}]  (9,8) rectangle (10,9);
\draw[fill={rgb,255:red,255.0;green,214.0;blue,174.0}]  (9.250000,8.250000) rectangle (9.750000,8.750000);
\draw[fill=gray] (10,8) rectangle (11,9);
\draw[fill=gray] (0,9) rectangle (1,10);
\draw[fill=gray] (2,9) rectangle (3,10);
\draw[fill=gray] (3,9) rectangle (4,10);
\draw[fill=gray] (4,9) rectangle (5,10);
\draw[fill=gray] (5,9) rectangle (6,10);
\draw[fill=gray] (6,9) rectangle (7,10);
\draw[fill=gray] (7,9) rectangle (8,10);
\draw[fill=gray] (8,9) rectangle (9,10);
\draw[fill={rgb,255:red,0.0;green,0.0;blue,255.0}]  (9,9) rectangle (10,10);
\draw[fill={rgb,255:red,255.0;green,228.0;blue,201.0}]  (9.250000,9.250000) rectangle (9.750000,9.750000);
\draw[fill=gray] (10,9) rectangle (11,10);
\draw[fill=gray] (0,10) rectangle (1,11);
\draw[fill=pink] (9.500000,10.500000) circle (0.500000);
\draw[fill=gray] (10,10) rectangle (11,11);
\draw[fill=gray] (0,11) rectangle (1,12);
\draw[fill=gray] (1,11) rectangle (2,12);
\draw[fill=gray] (2,11) rectangle (3,12);
\draw[fill=gray] (3,11) rectangle (4,12);
\draw[fill=gray] (4,11) rectangle (5,12);
\draw[fill=gray] (5,11) rectangle (6,12);
\draw[fill=gray] (6,11) rectangle (7,12);
\draw[fill=gray] (7,11) rectangle (8,12);
\draw[fill=gray] (8,11) rectangle (9,12);
\draw[fill=gray] (9,11) rectangle (10,12);
\draw[fill=gray] (10,11) rectangle (11,12);
\node [align=center] at(-0.500000,0.500000) {0};
\node [align=center] at(-0.500000,1.500000) {1};
\node [align=center] at(-0.500000,2.500000) {2};
\node [align=center] at(-0.500000,3.500000) {3};
\node [align=center] at(-0.500000,4.500000) {4};
\node [align=center] at(-0.500000,5.500000) {5};
\node [align=center] at(-0.500000,6.500000) {6};
\node [align=center] at(-0.500000,7.500000) {7};
\node [align=center] at(-0.500000,8.500000) {8};
\node [align=center] at(-0.500000,9.500000) {9};
\node [align=center] at(-0.500000,10.500000) {10};
\node [align=center] at(-0.500000,11.500000) {11};
\node [align=center] at(0.500000,-0.500000) {A};
\node [align=center] at(1.500000,-0.500000) {B};
\node [align=center] at(2.500000,-0.500000) {C};
\node [align=center] at(3.500000,-0.500000) {D};
\node [align=center] at(4.500000,-0.500000) {E};
\node [align=center] at(5.500000,-0.500000) {F};
\node [align=center] at(6.500000,-0.500000) {G};
\node [align=center] at(7.500000,-0.500000) {H};
\node [align=center] at(8.500000,-0.500000) {I};
\node [align=center] at(9.500000,-0.500000) {J};
\node [align=center] at(10.500000,-0.500000) {K};
\end{tikzpicture}
      }
    \end{minipage}
    \hfill
    \begin{minipage}{\mpwidth}
      \centering
      \adjustbox{width=\aboxwidth}{
        \begin{tikzpicture}
\draw (0,0) grid (11,12);
\draw[fill=gray] (0,0) rectangle (1,1);
\draw[fill=gray] (1,0) rectangle (2,1);
\draw[fill=gray] (2,0) rectangle (3,1);
\draw[fill=gray] (3,0) rectangle (4,1);
\draw[fill=gray] (4,0) rectangle (5,1);
\draw[fill=gray] (5,0) rectangle (6,1);
\draw[fill=gray] (6,0) rectangle (7,1);
\draw[fill=gray] (7,0) rectangle (8,1);
\draw[fill=gray] (8,0) rectangle (9,1);
\draw[fill=gray] (9,0) rectangle (10,1);
\draw[fill=gray] (10,0) rectangle (11,1);
\draw[fill=gray] (0,1) rectangle (1,2);
\draw[fill=gray] (1,1) rectangle (2,2);
\draw[fill={rgb,255:red,0.0;green,0.0;blue,255.0}]  (2,1) rectangle (3,2);
\draw[fill=gray] (3,1) rectangle (4,2);
\draw[fill=gray] (4,1) rectangle (5,2);
\draw[fill=gray] (5,1) rectangle (6,2);
\draw[fill=gray] (6,1) rectangle (7,2);
\draw[fill=gray] (7,1) rectangle (8,2);
\draw[fill=gray] (8,1) rectangle (9,2);
\draw[fill=gray] (9,1) rectangle (10,2);
\draw[fill=gray] (10,1) rectangle (11,2);
\draw[fill=gray] (0,2) rectangle (1,3);
\draw[fill=gray] (1,2) rectangle (2,3);
\draw[fill={rgb,255:red,0.0;green,0.0;blue,255.0}]  (2,2) rectangle (3,3);
\draw[fill={rgb,255:red,255.0;green,241.0;blue,228.0}]  (2.250000,2.250000) rectangle (2.750000,2.750000);
\draw[fill=gray] (3,2) rectangle (4,3);
\draw[fill=gray] (4,2) rectangle (5,3);
\draw[fill=gray] (5,2) rectangle (6,3);
\draw[fill=gray] (6,2) rectangle (7,3);
\draw[fill=gray] (7,2) rectangle (8,3);
\draw[fill=gray] (8,2) rectangle (9,3);
\draw[fill=gray] (9,2) rectangle (10,3);
\draw[fill=gray] (10,2) rectangle (11,3);
\draw[fill=gray] (0,3) rectangle (1,4);
\draw[fill={rgb,255:red,0.0;green,0.0;blue,255.0}]  (1,3) rectangle (2,4);
\draw[fill={rgb,255:red,0.0;green,0.0;blue,255.0}]  (2,3) rectangle (3,4);
\draw[fill={rgb,255:red,255.0;green,241.0;blue,228.0}]  (2.250000,3.250000) rectangle (2.750000,3.750000);
\draw[fill=gray] (6,3) rectangle (7,4);
\draw[fill=gray] (7,3) rectangle (8,4);
\draw[fill=gray] (8,3) rectangle (9,4);
\draw[fill=gray] (9,3) rectangle (10,4);
\draw[fill=gray] (10,3) rectangle (11,4);
\draw[fill=gray] (0,4) rectangle (1,5);
\draw[fill={rgb,255:red,0.0;green,0.0;blue,255.0}]  (1,4) rectangle (2,5);
\draw[fill={rgb,255:red,255.0;green,201.0;blue,147.0}]  (1.250000,4.250000) rectangle (1.750000,4.750000);
\draw[fill=gray] (2,4) rectangle (3,5);
\draw[fill=gray] (3,4) rectangle (4,5);
\draw[fill=gray] (4,4) rectangle (5,5);
\draw[fill=gray] (10,4) rectangle (11,5);
\draw[fill=gray] (0,5) rectangle (1,6);
\draw[fill={rgb,255:red,0.0;green,0.0;blue,255.0}]  (1,5) rectangle (2,6);
\draw[fill={rgb,255:red,255.0;green,187.0;blue,120.0}]  (1.250000,5.250000) rectangle (1.750000,5.750000);
\draw[fill=gray] (2,5) rectangle (3,6);
\draw[fill=gray] (3,5) rectangle (4,6);
\draw[fill=gray] (4,5) rectangle (5,6);
\draw[fill=gray] (10,5) rectangle (11,6);
\draw[fill=gray] (0,6) rectangle (1,7);
\draw[fill={rgb,255:red,0.0;green,0.0;blue,255.0}]  (1,6) rectangle (2,7);
\draw[fill={rgb,255:red,255.0;green,201.0;blue,147.0}]  (1.250000,6.250000) rectangle (1.750000,6.750000);
\draw[fill=gray] (2,6) rectangle (3,7);
\draw[fill=gray] (3,6) rectangle (4,7);
\draw[fill=gray] (4,6) rectangle (5,7);
\draw[fill=gray] (10,6) rectangle (11,7);
\draw[fill=gray] (0,7) rectangle (1,8);
\draw[fill={rgb,255:red,0.0;green,0.0;blue,255.0}]  (1,7) rectangle (2,8);
\draw[fill={rgb,255:red,255.0;green,161.0;blue,67.0}]  (1.250000,7.250000) rectangle (1.750000,7.750000);
\draw[fill=gray] (2,7) rectangle (3,8);
\draw[fill=gray] (3,7) rectangle (4,8);
\draw[fill=gray] (4,7) rectangle (5,8);
\draw[fill=gray] (10,7) rectangle (11,8);
\draw[fill=gray] (0,8) rectangle (1,9);
\draw[fill={rgb,255:red,0.0;green,0.0;blue,255.0}]  (1,8) rectangle (2,9);
\draw[fill={rgb,255:red,255.0;green,161.0;blue,67.0}]  (1.250000,8.250000) rectangle (1.750000,8.750000);
\draw[fill=gray] (2,8) rectangle (3,9);
\draw[fill=gray] (3,8) rectangle (4,9);
\draw[fill=gray] (4,8) rectangle (5,9);
\draw[fill=gray] (10,8) rectangle (11,9);
\draw[fill=gray] (0,9) rectangle (1,10);
\draw[fill={rgb,255:red,0.0;green,0.0;blue,255.0}]  (1,9) rectangle (2,10);
\draw[fill={rgb,255:red,255.0;green,228.0;blue,201.0}]  (1.250000,9.250000) rectangle (1.750000,9.750000);
\draw[fill=gray] (2,9) rectangle (3,10);
\draw[fill=gray] (3,9) rectangle (4,10);
\draw[fill=gray] (4,9) rectangle (5,10);
\draw[fill=gray] (5,9) rectangle (6,10);
\draw[fill=gray] (6,9) rectangle (7,10);
\draw[fill=gray] (7,9) rectangle (8,10);
\draw[fill=gray] (8,9) rectangle (9,10);
\draw[fill=gray] (10,9) rectangle (11,10);
\draw[fill=gray] (0,10) rectangle (1,11);
\draw[fill={rgb,255:red,0.0;green,0.0;blue,255.0}]  (1,10) rectangle (2,11);
\draw[fill={rgb,255:red,255.0;green,201.0;blue,147.0}]  (1.250000,10.250000) rectangle (1.750000,10.750000);
\draw[fill={rgb,255:red,0.0;green,0.0;blue,255.0}]  (2,10) rectangle (3,11);
\draw[fill={rgb,255:red,255.0;green,174.0;blue,93.0}]  (2.250000,10.250000) rectangle (2.750000,10.750000);
\draw[fill={rgb,255:red,0.0;green,0.0;blue,255.0}]  (3,10) rectangle (4,11);
\draw[fill={rgb,255:red,255.0;green,161.0;blue,67.0}]  (3.250000,10.250000) rectangle (3.750000,10.750000);
\draw[fill={rgb,255:red,0.0;green,0.0;blue,255.0}]  (4,10) rectangle (5,11);
\draw[fill={rgb,255:red,255.0;green,147.0;blue,40.0}]  (4.250000,10.250000) rectangle (4.750000,10.750000);
\draw[fill={rgb,255:red,0.0;green,0.0;blue,255.0}]  (5,10) rectangle (6,11);
\draw[fill={rgb,255:red,255.0;green,161.0;blue,67.0}]  (5.250000,10.250000) rectangle (5.750000,10.750000);
\draw[fill={rgb,255:red,0.0;green,0.0;blue,255.0}]  (6,10) rectangle (7,11);
\draw[fill={rgb,255:red,255.0;green,187.0;blue,120.0}]  (6.250000,10.250000) rectangle (6.750000,10.750000);
\draw[fill={rgb,255:red,0.0;green,0.0;blue,255.0}]  (7,10) rectangle (8,11);
\draw[fill={rgb,255:red,255.0;green,161.0;blue,67.0}]  (7.250000,10.250000) rectangle (7.750000,10.750000);
\draw[fill={rgb,255:red,0.0;green,0.0;blue,255.0}]  (8,10) rectangle (9,11);
\draw[fill={rgb,255:red,255.0;green,201.0;blue,147.0}]  (8.250000,10.250000) rectangle (8.750000,10.750000);
\draw[fill=pink] (9.500000,10.500000) circle (0.500000);
\draw[fill=gray] (10,10) rectangle (11,11);
\draw[fill=gray] (0,11) rectangle (1,12);
\draw[fill=gray] (1,11) rectangle (2,12);
\draw[fill=gray] (2,11) rectangle (3,12);
\draw[fill=gray] (3,11) rectangle (4,12);
\draw[fill=gray] (4,11) rectangle (5,12);
\draw[fill=gray] (5,11) rectangle (6,12);
\draw[fill=gray] (6,11) rectangle (7,12);
\draw[fill=gray] (7,11) rectangle (8,12);
\draw[fill=gray] (8,11) rectangle (9,12);
\draw[fill=gray] (9,11) rectangle (10,12);
\draw[fill=gray] (10,11) rectangle (11,12);
\node [align=center] at(-0.500000,0.500000) {0};
\node [align=center] at(-0.500000,1.500000) {1};
\node [align=center] at(-0.500000,2.500000) {2};
\node [align=center] at(-0.500000,3.500000) {3};
\node [align=center] at(-0.500000,4.500000) {4};
\node [align=center] at(-0.500000,5.500000) {5};
\node [align=center] at(-0.500000,6.500000) {6};
\node [align=center] at(-0.500000,7.500000) {7};
\node [align=center] at(-0.500000,8.500000) {8};
\node [align=center] at(-0.500000,9.500000) {9};
\node [align=center] at(-0.500000,10.500000) {10};
\node [align=center] at(-0.500000,11.500000) {11};
\node [align=center] at(0.500000,-0.500000) {A};
\node [align=center] at(1.500000,-0.500000) {B};
\node [align=center] at(2.500000,-0.500000) {C};
\node [align=center] at(3.500000,-0.500000) {D};
\node [align=center] at(4.500000,-0.500000) {E};
\node [align=center] at(5.500000,-0.500000) {F};
\node [align=center] at(6.500000,-0.500000) {G};
\node [align=center] at(7.500000,-0.500000) {H};
\node [align=center] at(8.500000,-0.500000) {I};
\node [align=center] at(9.500000,-0.500000) {J};
\node [align=center] at(10.500000,-0.500000) {K};
\end{tikzpicture}
      }
    \end{minipage}

    \centerline{\small $M_{1}$}
  \end{minipage}

  \hfill
  
  \begin{minipage}{\colwidth}
    \begin{minipage}{\mpwidth}
      \centering
      \adjustbox{width=\aboxwidth}{
        \input{\dir 5_MDP_policyStochHeatmap.tikz}
      }
    \end{minipage}
    \hfill
    \begin{minipage}{\mpwidth}
      \centering
      \adjustbox{width=\aboxwidth}{
        \begin{tikzpicture}
\draw (0,0) grid (11,13);
\draw[fill=gray] (0,0) rectangle (1,1);
\draw[fill=gray] (1,0) rectangle (2,1);
\draw[fill=gray] (2,0) rectangle (3,1);
\draw[fill=gray] (3,0) rectangle (4,1);
\draw[fill=gray] (4,0) rectangle (5,1);
\draw[fill=gray] (5,0) rectangle (6,1);
\draw[fill=gray] (6,0) rectangle (7,1);
\draw[fill=gray] (7,0) rectangle (8,1);
\draw[fill=gray] (8,0) rectangle (9,1);
\draw[fill=gray] (9,0) rectangle (10,1);
\draw[fill=gray] (10,0) rectangle (11,1);
\draw[fill=gray] (0,1) rectangle (1,2);
\draw[fill=pink] (9.500000,1.500000) circle (0.500000);
\draw[fill=gray] (10,1) rectangle (11,2);
\draw[fill=gray] (0,2) rectangle (1,3);
\draw[fill=gray] (2,2) rectangle (3,3);
\draw[fill=gray] (3,2) rectangle (4,3);
\draw[fill=gray] (4,2) rectangle (5,3);
\draw[fill=gray] (5,2) rectangle (6,3);
\draw[fill=gray] (6,2) rectangle (7,3);
\draw[fill=gray] (7,2) rectangle (8,3);
\draw[fill=gray] (8,2) rectangle (9,3);
\draw[fill={rgb,255:red,0.0;green,0.0;blue,255.0}]  (9,2) rectangle (10,3);
\draw[fill={rgb,255:red,255.0;green,214.0;blue,174.0}]  (9.250000,2.250000) rectangle (9.750000,2.750000);
\draw[fill=gray] (10,2) rectangle (11,3);
\draw[fill=gray] (0,3) rectangle (1,4);
\draw[fill=gray] (2,3) rectangle (3,4);
\draw[fill=gray] (3,3) rectangle (4,4);
\draw[fill=gray] (4,3) rectangle (5,4);
\draw[fill={rgb,255:red,0.0;green,0.0;blue,255.0}]  (9,3) rectangle (10,4);
\draw[fill={rgb,255:red,255.0;green,187.0;blue,120.0}]  (9.250000,3.250000) rectangle (9.750000,3.750000);
\draw[fill=gray] (10,3) rectangle (11,4);
\draw[fill=gray] (0,4) rectangle (1,5);
\draw[fill=gray] (2,4) rectangle (3,5);
\draw[fill=gray] (3,4) rectangle (4,5);
\draw[fill=gray] (4,4) rectangle (5,5);
\draw[fill={rgb,255:red,0.0;green,0.0;blue,255.0}]  (9,4) rectangle (10,5);
\draw[fill={rgb,255:red,255.0;green,174.0;blue,93.0}]  (9.250000,4.250000) rectangle (9.750000,4.750000);
\draw[fill=gray] (10,4) rectangle (11,5);
\draw[fill=gray] (0,5) rectangle (1,6);
\draw[fill=gray] (2,5) rectangle (3,6);
\draw[fill=gray] (3,5) rectangle (4,6);
\draw[fill=gray] (4,5) rectangle (5,6);
\draw[fill={rgb,255:red,0.0;green,0.0;blue,255.0}]  (9,5) rectangle (10,6);
\draw[fill={rgb,255:red,255.0;green,134.0;blue,13.0}]  (9.250000,5.250000) rectangle (9.750000,5.750000);
\draw[fill=gray] (10,5) rectangle (11,6);
\draw[fill=gray] (0,6) rectangle (1,7);
\draw[fill=gray] (2,6) rectangle (3,7);
\draw[fill=gray] (3,6) rectangle (4,7);
\draw[fill=gray] (4,6) rectangle (5,7);
\draw[fill={rgb,255:red,0.0;green,0.0;blue,255.0}]  (9,6) rectangle (10,7);
\draw[fill={rgb,255:red,255.0;green,161.0;blue,67.0}]  (9.250000,6.250000) rectangle (9.750000,6.750000);
\draw[fill=gray] (10,6) rectangle (11,7);
\draw[fill=gray] (0,7) rectangle (1,8);
\draw[fill=gray] (2,7) rectangle (3,8);
\draw[fill=gray] (3,7) rectangle (4,8);
\draw[fill=gray] (4,7) rectangle (5,8);
\draw[fill={rgb,255:red,0.0;green,0.0;blue,255.0}]  (9,7) rectangle (10,8);
\draw[fill={rgb,255:red,255.0;green,228.0;blue,201.0}]  (9.250000,7.250000) rectangle (9.750000,7.750000);
\draw[fill=gray] (10,7) rectangle (11,8);
\draw[fill=gray] (0,8) rectangle (1,9);
\draw[fill=gray] (2,8) rectangle (3,9);
\draw[fill=gray] (3,8) rectangle (4,9);
\draw[fill=gray] (4,8) rectangle (5,9);
\draw[fill={rgb,255:red,0.0;green,0.0;blue,255.0}]  (5,8) rectangle (6,9);
\draw[fill={rgb,255:red,255.0;green,241.0;blue,228.0}]  (5.250000,8.250000) rectangle (5.750000,8.750000);
\draw[fill={rgb,255:red,0.0;green,0.0;blue,255.0}]  (6,8) rectangle (7,9);
\draw[fill={rgb,255:red,0.0;green,0.0;blue,255.0}]  (7,8) rectangle (8,9);
\draw[fill={rgb,255:red,0.0;green,0.0;blue,255.0}]  (8,8) rectangle (9,9);
\draw[fill={rgb,255:red,255.0;green,228.0;blue,201.0}]  (8.250000,8.250000) rectangle (8.750000,8.750000);
\draw[fill={rgb,255:red,0.0;green,0.0;blue,255.0}]  (9,8) rectangle (10,9);
\draw[fill={rgb,255:red,255.0;green,228.0;blue,201.0}]  (9.250000,8.250000) rectangle (9.750000,8.750000);
\draw[fill=gray] (10,8) rectangle (11,9);
\draw[fill=gray] (0,9) rectangle (1,10);
\draw[fill={rgb,255:red,0.0;green,0.0;blue,255.0}]  (5,9) rectangle (6,10);
\draw[fill={rgb,255:red,255.0;green,241.0;blue,228.0}]  (5.250000,9.250000) rectangle (5.750000,9.750000);
\draw[fill=gray] (6,9) rectangle (7,10);
\draw[fill=gray] (7,9) rectangle (8,10);
\draw[fill=gray] (8,9) rectangle (9,10);
\draw[fill=gray] (9,9) rectangle (10,10);
\draw[fill=gray] (10,9) rectangle (11,10);
\draw[fill=gray] (0,10) rectangle (1,11);
\draw[fill=gray] (1,10) rectangle (2,11);
\draw[fill=gray] (2,10) rectangle (3,11);
\draw[fill=gray] (3,10) rectangle (4,11);
\draw[fill=gray] (4,10) rectangle (5,11);
\draw[fill={rgb,255:red,0.0;green,0.0;blue,255.0}]  (5,10) rectangle (6,11);
\draw[fill={rgb,255:red,255.0;green,241.0;blue,228.0}]  (5.250000,10.250000) rectangle (5.750000,10.750000);
\draw[fill=gray] (6,10) rectangle (7,11);
\draw[fill=gray] (7,10) rectangle (8,11);
\draw[fill=gray] (8,10) rectangle (9,11);
\draw[fill=gray] (9,10) rectangle (10,11);
\draw[fill=gray] (10,10) rectangle (11,11);
\draw[fill=gray] (0,11) rectangle (1,12);
\draw[fill=gray] (1,11) rectangle (2,12);
\draw[fill=gray] (2,11) rectangle (3,12);
\draw[fill=gray] (3,11) rectangle (4,12);
\draw[fill=gray] (4,11) rectangle (5,12);
\draw[fill={rgb,255:red,0.0;green,0.0;blue,255.0}]  (5,11) rectangle (6,12);
\draw[fill=gray] (6,11) rectangle (7,12);
\draw[fill=gray] (7,11) rectangle (8,12);
\draw[fill=gray] (8,11) rectangle (9,12);
\draw[fill=gray] (9,11) rectangle (10,12);
\draw[fill=gray] (10,11) rectangle (11,12);
\draw[fill=gray] (0,12) rectangle (1,13);
\draw[fill=gray] (1,12) rectangle (2,13);
\draw[fill=gray] (2,12) rectangle (3,13);
\draw[fill=gray] (3,12) rectangle (4,13);
\draw[fill=gray] (4,12) rectangle (5,13);
\draw[fill=gray] (5,12) rectangle (6,13);
\draw[fill=gray] (6,12) rectangle (7,13);
\draw[fill=gray] (7,12) rectangle (8,13);
\draw[fill=gray] (8,12) rectangle (9,13);
\draw[fill=gray] (9,12) rectangle (10,13);
\draw[fill=gray] (10,12) rectangle (11,13);
\node [align=center] at(-0.500000,0.500000) {0};
\node [align=center] at(-0.500000,1.500000) {1};
\node [align=center] at(-0.500000,2.500000) {2};
\node [align=center] at(-0.500000,3.500000) {3};
\node [align=center] at(-0.500000,4.500000) {4};
\node [align=center] at(-0.500000,5.500000) {5};
\node [align=center] at(-0.500000,6.500000) {6};
\node [align=center] at(-0.500000,7.500000) {7};
\node [align=center] at(-0.500000,8.500000) {8};
\node [align=center] at(-0.500000,9.500000) {9};
\node [align=center] at(-0.500000,10.500000) {10};
\node [align=center] at(-0.500000,11.500000) {11};
\node [align=center] at(-0.500000,12.500000) {12};
\node [align=center] at(0.500000,-0.500000) {A};
\node [align=center] at(1.500000,-0.500000) {B};
\node [align=center] at(2.500000,-0.500000) {C};
\node [align=center] at(3.500000,-0.500000) {D};
\node [align=center] at(4.500000,-0.500000) {E};
\node [align=center] at(5.500000,-0.500000) {F};
\node [align=center] at(6.500000,-0.500000) {G};
\node [align=center] at(7.500000,-0.500000) {H};
\node [align=center] at(8.500000,-0.500000) {I};
\node [align=center] at(9.500000,-0.500000) {J};
\node [align=center] at(10.500000,-0.500000) {K};
\end{tikzpicture}
      }
    \end{minipage}
    \hfill
    \begin{minipage}{\mpwidth}
      \centering
      \adjustbox{width=\aboxwidth}{
        \begin{tikzpicture}
\draw (0,0) grid (11,13);
\draw[fill=gray] (0,0) rectangle (1,1);
\draw[fill=gray] (1,0) rectangle (2,1);
\draw[fill=gray] (2,0) rectangle (3,1);
\draw[fill=gray] (3,0) rectangle (4,1);
\draw[fill=gray] (4,0) rectangle (5,1);
\draw[fill=gray] (5,0) rectangle (6,1);
\draw[fill=gray] (6,0) rectangle (7,1);
\draw[fill=gray] (7,0) rectangle (8,1);
\draw[fill=gray] (8,0) rectangle (9,1);
\draw[fill=gray] (9,0) rectangle (10,1);
\draw[fill=gray] (10,0) rectangle (11,1);
\draw[fill=gray] (0,1) rectangle (1,2);
\draw[fill=pink] (9.500000,1.500000) circle (0.500000);
\draw[fill=gray] (10,1) rectangle (11,2);
\draw[fill=gray] (0,2) rectangle (1,3);
\draw[fill=gray] (2,2) rectangle (3,3);
\draw[fill=gray] (3,2) rectangle (4,3);
\draw[fill=gray] (4,2) rectangle (5,3);
\draw[fill=gray] (5,2) rectangle (6,3);
\draw[fill=gray] (6,2) rectangle (7,3);
\draw[fill=gray] (7,2) rectangle (8,3);
\draw[fill=gray] (8,2) rectangle (9,3);
\draw[fill={rgb,255:red,0.0;green,0.0;blue,255.0}]  (9,2) rectangle (10,3);
\draw[fill={rgb,255:red,255.0;green,214.0;blue,174.0}]  (9.250000,2.250000) rectangle (9.750000,2.750000);
\draw[fill=gray] (10,2) rectangle (11,3);
\draw[fill=gray] (0,3) rectangle (1,4);
\draw[fill=gray] (2,3) rectangle (3,4);
\draw[fill=gray] (3,3) rectangle (4,4);
\draw[fill=gray] (4,3) rectangle (5,4);
\draw[fill={rgb,255:red,0.0;green,0.0;blue,255.0}]  (9,3) rectangle (10,4);
\draw[fill={rgb,255:red,255.0;green,161.0;blue,67.0}]  (9.250000,3.250000) rectangle (9.750000,3.750000);
\draw[fill=gray] (10,3) rectangle (11,4);
\draw[fill=gray] (0,4) rectangle (1,5);
\draw[fill=gray] (2,4) rectangle (3,5);
\draw[fill=gray] (3,4) rectangle (4,5);
\draw[fill=gray] (4,4) rectangle (5,5);
\draw[fill={rgb,255:red,0.0;green,0.0;blue,255.0}]  (9,4) rectangle (10,5);
\draw[fill={rgb,255:red,255.0;green,174.0;blue,93.0}]  (9.250000,4.250000) rectangle (9.750000,4.750000);
\draw[fill=gray] (10,4) rectangle (11,5);
\draw[fill=gray] (0,5) rectangle (1,6);
\draw[fill=gray] (2,5) rectangle (3,6);
\draw[fill=gray] (3,5) rectangle (4,6);
\draw[fill=gray] (4,5) rectangle (5,6);
\draw[fill={rgb,255:red,0.0;green,0.0;blue,255.0}]  (9,5) rectangle (10,6);
\draw[fill={rgb,255:red,255.0;green,187.0;blue,120.0}]  (9.250000,5.250000) rectangle (9.750000,5.750000);
\draw[fill=gray] (10,5) rectangle (11,6);
\draw[fill=gray] (0,6) rectangle (1,7);
\draw[fill=gray] (2,6) rectangle (3,7);
\draw[fill=gray] (3,6) rectangle (4,7);
\draw[fill=gray] (4,6) rectangle (5,7);
\draw[fill={rgb,255:red,0.0;green,0.0;blue,255.0}]  (9,6) rectangle (10,7);
\draw[fill={rgb,255:red,255.0;green,161.0;blue,67.0}]  (9.250000,6.250000) rectangle (9.750000,6.750000);
\draw[fill=gray] (10,6) rectangle (11,7);
\draw[fill=gray] (0,7) rectangle (1,8);
\draw[fill=gray] (2,7) rectangle (3,8);
\draw[fill=gray] (3,7) rectangle (4,8);
\draw[fill=gray] (4,7) rectangle (5,8);
\draw[fill={rgb,255:red,0.0;green,0.0;blue,255.0}]  (9,7) rectangle (10,8);
\draw[fill={rgb,255:red,255.0;green,161.0;blue,67.0}]  (9.250000,7.250000) rectangle (9.750000,7.750000);
\draw[fill=gray] (10,7) rectangle (11,8);
\draw[fill=gray] (0,8) rectangle (1,9);
\draw[fill=gray] (2,8) rectangle (3,9);
\draw[fill=gray] (3,8) rectangle (4,9);
\draw[fill=gray] (4,8) rectangle (5,9);
\draw[fill={rgb,255:red,0.0;green,0.0;blue,255.0}]  (5,8) rectangle (6,9);
\draw[fill={rgb,255:red,0.0;green,0.0;blue,255.0}]  (6,8) rectangle (7,9);
\draw[fill={rgb,255:red,0.0;green,0.0;blue,255.0}]  (7,8) rectangle (8,9);
\draw[fill={rgb,255:red,0.0;green,0.0;blue,255.0}]  (8,8) rectangle (9,9);
\draw[fill={rgb,255:red,255.0;green,241.0;blue,228.0}]  (8.250000,8.250000) rectangle (8.750000,8.750000);
\draw[fill={rgb,255:red,0.0;green,0.0;blue,255.0}]  (9,8) rectangle (10,9);
\draw[fill={rgb,255:red,255.0;green,228.0;blue,201.0}]  (9.250000,8.250000) rectangle (9.750000,8.750000);
\draw[fill=gray] (10,8) rectangle (11,9);
\draw[fill=gray] (0,9) rectangle (1,10);
\draw[fill={rgb,255:red,0.0;green,0.0;blue,255.0}]  (5,9) rectangle (6,10);
\draw[fill=gray] (6,9) rectangle (7,10);
\draw[fill=gray] (7,9) rectangle (8,10);
\draw[fill=gray] (8,9) rectangle (9,10);
\draw[fill=gray] (9,9) rectangle (10,10);
\draw[fill=gray] (10,9) rectangle (11,10);
\draw[fill=gray] (0,10) rectangle (1,11);
\draw[fill=gray] (1,10) rectangle (2,11);
\draw[fill=gray] (2,10) rectangle (3,11);
\draw[fill=gray] (3,10) rectangle (4,11);
\draw[fill=gray] (4,10) rectangle (5,11);
\draw[fill={rgb,255:red,0.0;green,0.0;blue,255.0}]  (5,10) rectangle (6,11);
\draw[fill=gray] (6,10) rectangle (7,11);
\draw[fill=gray] (7,10) rectangle (8,11);
\draw[fill=gray] (8,10) rectangle (9,11);
\draw[fill=gray] (9,10) rectangle (10,11);
\draw[fill=gray] (10,10) rectangle (11,11);
\draw[fill=gray] (0,11) rectangle (1,12);
\draw[fill=gray] (1,11) rectangle (2,12);
\draw[fill=gray] (2,11) rectangle (3,12);
\draw[fill=gray] (3,11) rectangle (4,12);
\draw[fill=gray] (4,11) rectangle (5,12);
\draw[fill={rgb,255:red,0.0;green,0.0;blue,255.0}]  (5,11) rectangle (6,12);
\draw[fill=gray] (6,11) rectangle (7,12);
\draw[fill=gray] (7,11) rectangle (8,12);
\draw[fill=gray] (8,11) rectangle (9,12);
\draw[fill=gray] (9,11) rectangle (10,12);
\draw[fill=gray] (10,11) rectangle (11,12);
\draw[fill=gray] (0,12) rectangle (1,13);
\draw[fill=gray] (1,12) rectangle (2,13);
\draw[fill=gray] (2,12) rectangle (3,13);
\draw[fill=gray] (3,12) rectangle (4,13);
\draw[fill=gray] (4,12) rectangle (5,13);
\draw[fill=gray] (5,12) rectangle (6,13);
\draw[fill=gray] (6,12) rectangle (7,13);
\draw[fill=gray] (7,12) rectangle (8,13);
\draw[fill=gray] (8,12) rectangle (9,13);
\draw[fill=gray] (9,12) rectangle (10,13);
\draw[fill=gray] (10,12) rectangle (11,13);
\node [align=center] at(-0.500000,0.500000) {0};
\node [align=center] at(-0.500000,1.500000) {1};
\node [align=center] at(-0.500000,2.500000) {2};
\node [align=center] at(-0.500000,3.500000) {3};
\node [align=center] at(-0.500000,4.500000) {4};
\node [align=center] at(-0.500000,5.500000) {5};
\node [align=center] at(-0.500000,6.500000) {6};
\node [align=center] at(-0.500000,7.500000) {7};
\node [align=center] at(-0.500000,8.500000) {8};
\node [align=center] at(-0.500000,9.500000) {9};
\node [align=center] at(-0.500000,10.500000) {10};
\node [align=center] at(-0.500000,11.500000) {11};
\node [align=center] at(-0.500000,12.500000) {12};
\node [align=center] at(0.500000,-0.500000) {A};
\node [align=center] at(1.500000,-0.500000) {B};
\node [align=center] at(2.500000,-0.500000) {C};
\node [align=center] at(3.500000,-0.500000) {D};
\node [align=center] at(4.500000,-0.500000) {E};
\node [align=center] at(5.500000,-0.500000) {F};
\node [align=center] at(6.500000,-0.500000) {G};
\node [align=center] at(7.500000,-0.500000) {H};
\node [align=center] at(8.500000,-0.500000) {I};
\node [align=center] at(9.500000,-0.500000) {J};
\node [align=center] at(10.500000,-0.500000) {K};
\end{tikzpicture}
      }
    \end{minipage}

    \centerline{\small $M_{2}$}
  \end{minipage}

  \medskip

  \begin{minipage}{\colwidth}
    \begin{minipage}{\mpwidth}
      \centering
      \adjustbox{width=\aboxwidth}{
        \input{\dir 2_MDP_policyStochHeatmap.tikz}
      }
    \end{minipage}
    \hfill
    \begin{minipage}{\mpwidth}
      \centering
      \adjustbox{width=\aboxwidth}{
        \begin{tikzpicture}
\draw (0,0) grid (11,11);
\draw[fill=gray] (0,0) rectangle (1,1);
\draw[fill=gray] (1,0) rectangle (2,1);
\draw[fill=gray] (2,0) rectangle (3,1);
\draw[fill=gray] (3,0) rectangle (4,1);
\draw[fill=gray] (4,0) rectangle (5,1);
\draw[fill=gray] (5,0) rectangle (6,1);
\draw[fill=gray] (6,0) rectangle (7,1);
\draw[fill=gray] (7,0) rectangle (8,1);
\draw[fill=gray] (8,0) rectangle (9,1);
\draw[fill=gray] (9,0) rectangle (10,1);
\draw[fill=gray] (10,0) rectangle (11,1);
\draw[fill=gray] (0,1) rectangle (1,2);
\draw[fill=pink] (9.500000,1.500000) circle (0.500000);
\draw[fill=gray] (10,1) rectangle (11,2);
\draw[fill=gray] (0,2) rectangle (1,3);
\draw[fill=gray] (5,2) rectangle (6,3);
\draw[fill=gray] (6,2) rectangle (7,3);
\draw[fill=gray] (7,2) rectangle (8,3);
\draw[fill=gray] (8,2) rectangle (9,3);
\draw[fill={rgb,255:red,0.0;green,0.0;blue,255.0}]  (9,2) rectangle (10,3);
\draw[fill={rgb,255:red,255.0;green,241.0;blue,228.0}]  (9.250000,2.250000) rectangle (9.750000,2.750000);
\draw[fill=gray] (10,2) rectangle (11,3);
\draw[fill=gray] (0,3) rectangle (1,4);
\draw[fill=gray] (5,3) rectangle (6,4);
\draw[fill={rgb,255:red,0.0;green,0.0;blue,255.0}]  (9,3) rectangle (10,4);
\draw[fill={rgb,255:red,255.0;green,201.0;blue,147.0}]  (9.250000,3.250000) rectangle (9.750000,3.750000);
\draw[fill=gray] (10,3) rectangle (11,4);
\draw[fill=gray] (0,4) rectangle (1,5);
\draw[fill=gray] (5,4) rectangle (6,5);
\draw[fill={rgb,255:red,0.0;green,0.0;blue,255.0}]  (9,4) rectangle (10,5);
\draw[fill={rgb,255:red,255.0;green,174.0;blue,93.0}]  (9.250000,4.250000) rectangle (9.750000,4.750000);
\draw[fill=gray] (10,4) rectangle (11,5);
\draw[fill=gray] (0,5) rectangle (1,6);
\draw[fill=gray] (5,5) rectangle (6,6);
\draw[fill={rgb,255:red,0.0;green,0.0;blue,255.0}]  (9,5) rectangle (10,6);
\draw[fill={rgb,255:red,255.0;green,187.0;blue,120.0}]  (9.250000,5.250000) rectangle (9.750000,5.750000);
\draw[fill=gray] (10,5) rectangle (11,6);
\draw[fill=gray] (0,6) rectangle (1,7);
\draw[fill=gray] (2,6) rectangle (3,7);
\draw[fill=gray] (3,6) rectangle (4,7);
\draw[fill=gray] (4,6) rectangle (5,7);
\draw[fill=gray] (5,6) rectangle (6,7);
\draw[fill={rgb,255:red,0.0;green,0.0;blue,255.0}]  (9,6) rectangle (10,7);
\draw[fill={rgb,255:red,255.0;green,201.0;blue,147.0}]  (9.250000,6.250000) rectangle (9.750000,6.750000);
\draw[fill=gray] (10,6) rectangle (11,7);
\draw[fill=gray] (0,7) rectangle (1,8);
\draw[fill={rgb,255:red,0.0;green,0.0;blue,255.0}]  (1,7) rectangle (2,8);
\draw[fill={rgb,255:red,255.0;green,214.0;blue,174.0}]  (1.250000,7.250000) rectangle (1.750000,7.750000);
\draw[fill={rgb,255:red,0.0;green,0.0;blue,255.0}]  (2,7) rectangle (3,8);
\draw[fill={rgb,255:red,0.0;green,0.0;blue,255.0}]  (3,7) rectangle (4,8);
\draw[fill={rgb,255:red,255.0;green,228.0;blue,201.0}]  (3.250000,7.250000) rectangle (3.750000,7.750000);
\draw[fill={rgb,255:red,0.0;green,0.0;blue,255.0}]  (4,7) rectangle (5,8);
\draw[fill={rgb,255:red,255.0;green,241.0;blue,228.0}]  (4.250000,7.250000) rectangle (4.750000,7.750000);
\draw[fill={rgb,255:red,0.0;green,0.0;blue,255.0}]  (5,7) rectangle (6,8);
\draw[fill={rgb,255:red,255.0;green,201.0;blue,147.0}]  (5.250000,7.250000) rectangle (5.750000,7.750000);
\draw[fill={rgb,255:red,0.0;green,0.0;blue,255.0}]  (6,7) rectangle (7,8);
\draw[fill={rgb,255:red,255.0;green,214.0;blue,174.0}]  (6.250000,7.250000) rectangle (6.750000,7.750000);
\draw[fill={rgb,255:red,0.0;green,0.0;blue,255.0}]  (7,7) rectangle (8,8);
\draw[fill={rgb,255:red,255.0;green,228.0;blue,201.0}]  (7.250000,7.250000) rectangle (7.750000,7.750000);
\draw[fill={rgb,255:red,0.0;green,0.0;blue,255.0}]  (8,7) rectangle (9,8);
\draw[fill={rgb,255:red,255.0;green,241.0;blue,228.0}]  (8.250000,7.250000) rectangle (8.750000,7.750000);
\draw[fill={rgb,255:red,0.0;green,0.0;blue,255.0}]  (9,7) rectangle (10,8);
\draw[fill=gray] (10,7) rectangle (11,8);
\draw[fill=gray] (0,8) rectangle (1,9);
\draw[fill={rgb,255:red,0.0;green,0.0;blue,255.0}]  (1,8) rectangle (2,9);
\draw[fill=gray] (2,8) rectangle (3,9);
\draw[fill=gray] (3,8) rectangle (4,9);
\draw[fill=gray] (4,8) rectangle (5,9);
\draw[fill=gray] (5,8) rectangle (6,9);
\draw[fill=gray] (6,8) rectangle (7,9);
\draw[fill=gray] (7,8) rectangle (8,9);
\draw[fill=gray] (8,8) rectangle (9,9);
\draw[fill=gray] (9,8) rectangle (10,9);
\draw[fill=gray] (10,8) rectangle (11,9);
\draw[fill=gray] (0,9) rectangle (1,10);
\draw[fill={rgb,255:red,0.0;green,0.0;blue,255.0}]  (1,9) rectangle (2,10);
\draw[fill=gray] (2,9) rectangle (3,10);
\draw[fill=gray] (3,9) rectangle (4,10);
\draw[fill=gray] (4,9) rectangle (5,10);
\draw[fill=gray] (5,9) rectangle (6,10);
\draw[fill=gray] (6,9) rectangle (7,10);
\draw[fill=gray] (7,9) rectangle (8,10);
\draw[fill=gray] (8,9) rectangle (9,10);
\draw[fill=gray] (9,9) rectangle (10,10);
\draw[fill=gray] (10,9) rectangle (11,10);
\draw[fill=gray] (0,10) rectangle (1,11);
\draw[fill=gray] (1,10) rectangle (2,11);
\draw[fill=gray] (2,10) rectangle (3,11);
\draw[fill=gray] (3,10) rectangle (4,11);
\draw[fill=gray] (4,10) rectangle (5,11);
\draw[fill=gray] (5,10) rectangle (6,11);
\draw[fill=gray] (6,10) rectangle (7,11);
\draw[fill=gray] (7,10) rectangle (8,11);
\draw[fill=gray] (8,10) rectangle (9,11);
\draw[fill=gray] (9,10) rectangle (10,11);
\draw[fill=gray] (10,10) rectangle (11,11);
\node [align=center] at(-0.500000,0.500000) {0};
\node [align=center] at(-0.500000,1.500000) {1};
\node [align=center] at(-0.500000,2.500000) {2};
\node [align=center] at(-0.500000,3.500000) {3};
\node [align=center] at(-0.500000,4.500000) {4};
\node [align=center] at(-0.500000,5.500000) {5};
\node [align=center] at(-0.500000,6.500000) {6};
\node [align=center] at(-0.500000,7.500000) {7};
\node [align=center] at(-0.500000,8.500000) {8};
\node [align=center] at(-0.500000,9.500000) {9};
\node [align=center] at(-0.500000,10.500000) {10};
\node [align=center] at(0.500000,-0.500000) {A};
\node [align=center] at(1.500000,-0.500000) {B};
\node [align=center] at(2.500000,-0.500000) {C};
\node [align=center] at(3.500000,-0.500000) {D};
\node [align=center] at(4.500000,-0.500000) {E};
\node [align=center] at(5.500000,-0.500000) {F};
\node [align=center] at(6.500000,-0.500000) {G};
\node [align=center] at(7.500000,-0.500000) {H};
\node [align=center] at(8.500000,-0.500000) {I};
\node [align=center] at(9.500000,-0.500000) {J};
\node [align=center] at(10.500000,-0.500000) {K};
\end{tikzpicture}
      }
    \end{minipage}
    \hfill
    \begin{minipage}{\mpwidth}
      \centering
      \adjustbox{width=\aboxwidth}{
        \begin{tikzpicture}
\draw (0,0) grid (11,11);
\draw[fill=gray] (0,0) rectangle (1,1);
\draw[fill=gray] (1,0) rectangle (2,1);
\draw[fill=gray] (2,0) rectangle (3,1);
\draw[fill=gray] (3,0) rectangle (4,1);
\draw[fill=gray] (4,0) rectangle (5,1);
\draw[fill=gray] (5,0) rectangle (6,1);
\draw[fill=gray] (6,0) rectangle (7,1);
\draw[fill=gray] (7,0) rectangle (8,1);
\draw[fill=gray] (8,0) rectangle (9,1);
\draw[fill=gray] (9,0) rectangle (10,1);
\draw[fill=gray] (10,0) rectangle (11,1);
\draw[fill=gray] (0,1) rectangle (1,2);
\draw[fill={rgb,255:red,93.0;green,93.0;blue,255.0}]  (4,1) rectangle (5,2);
\draw[fill={rgb,255:red,255.0;green,148.0;blue,42.0}]  (4.250000,1.250000) rectangle (4.750000,1.750000);
\draw[fill={rgb,255:red,93.0;green,93.0;blue,255.0}]  (5,1) rectangle (6,2);
\draw[fill={rgb,255:red,255.0;green,212.0;blue,170.0}]  (5.250000,1.250000) rectangle (5.750000,1.750000);
\draw[fill={rgb,255:red,93.0;green,93.0;blue,255.0}]  (6,1) rectangle (7,2);
\draw[fill={rgb,255:red,255.0;green,212.0;blue,170.0}]  (6.250000,1.250000) rectangle (6.750000,1.750000);
\draw[fill={rgb,255:red,93.0;green,93.0;blue,255.0}]  (7,1) rectangle (8,2);
\draw[fill={rgb,255:red,255.0;green,212.0;blue,170.0}]  (7.250000,1.250000) rectangle (7.750000,1.750000);
\draw[fill={rgb,255:red,93.0;green,93.0;blue,255.0}]  (8,1) rectangle (9,2);
\draw[fill={rgb,255:red,255.0;green,212.0;blue,170.0}]  (8.250000,1.250000) rectangle (8.750000,1.750000);
\draw[fill=pink] (9.500000,1.500000) circle (0.500000);
\draw[fill=gray] (10,1) rectangle (11,2);
\draw[fill=gray] (0,2) rectangle (1,3);
\draw[fill={rgb,255:red,93.0;green,93.0;blue,255.0}]  (4,2) rectangle (5,3);
\draw[fill=gray] (5,2) rectangle (6,3);
\draw[fill=gray] (6,2) rectangle (7,3);
\draw[fill=gray] (7,2) rectangle (8,3);
\draw[fill=gray] (8,2) rectangle (9,3);
\draw[fill={rgb,255:red,161.0;green,161.0;blue,255.0}]  (9,2) rectangle (10,3);
\draw[fill=gray] (10,2) rectangle (11,3);
\draw[fill=gray] (0,3) rectangle (1,4);
\draw[fill={rgb,255:red,93.0;green,93.0;blue,255.0}]  (4,3) rectangle (5,4);
\draw[fill={rgb,255:red,255.0;green,170.0;blue,85.0}]  (4.250000,3.250000) rectangle (4.750000,3.750000);
\draw[fill=gray] (5,3) rectangle (6,4);
\draw[fill={rgb,255:red,161.0;green,161.0;blue,255.0}]  (9,3) rectangle (10,4);
\draw[fill={rgb,255:red,255.0;green,218.0;blue,182.0}]  (9.250000,3.250000) rectangle (9.750000,3.750000);
\draw[fill=gray] (10,3) rectangle (11,4);
\draw[fill=gray] (0,4) rectangle (1,5);
\draw[fill={rgb,255:red,93.0;green,93.0;blue,255.0}]  (4,4) rectangle (5,5);
\draw[fill={rgb,255:red,255.0;green,191.0;blue,127.0}]  (4.250000,4.250000) rectangle (4.750000,4.750000);
\draw[fill=gray] (5,4) rectangle (6,5);
\draw[fill={rgb,255:red,161.0;green,161.0;blue,255.0}]  (9,4) rectangle (10,5);
\draw[fill=gray] (10,4) rectangle (11,5);
\draw[fill=gray] (0,5) rectangle (1,6);
\draw[fill={rgb,255:red,93.0;green,93.0;blue,255.0}]  (1,5) rectangle (2,6);
\draw[fill={rgb,255:red,255.0;green,212.0;blue,170.0}]  (1.250000,5.250000) rectangle (1.750000,5.750000);
\draw[fill={rgb,255:red,93.0;green,93.0;blue,255.0}]  (2,5) rectangle (3,6);
\draw[fill={rgb,255:red,255.0;green,170.0;blue,85.0}]  (2.250000,5.250000) rectangle (2.750000,5.750000);
\draw[fill={rgb,255:red,93.0;green,93.0;blue,255.0}]  (3,5) rectangle (4,6);
\draw[fill={rgb,255:red,93.0;green,93.0;blue,255.0}]  (4,5) rectangle (5,6);
\draw[fill={rgb,255:red,255.0;green,212.0;blue,170.0}]  (4.250000,5.250000) rectangle (4.750000,5.750000);
\draw[fill=gray] (5,5) rectangle (6,6);
\draw[fill={rgb,255:red,161.0;green,161.0;blue,255.0}]  (9,5) rectangle (10,6);
\draw[fill=gray] (10,5) rectangle (11,6);
\draw[fill=gray] (0,6) rectangle (1,7);
\draw[fill={rgb,255:red,93.0;green,93.0;blue,255.0}]  (1,6) rectangle (2,7);
\draw[fill=gray] (2,6) rectangle (3,7);
\draw[fill=gray] (3,6) rectangle (4,7);
\draw[fill=gray] (4,6) rectangle (5,7);
\draw[fill=gray] (5,6) rectangle (6,7);
\draw[fill={rgb,255:red,161.0;green,161.0;blue,255.0}]  (9,6) rectangle (10,7);
\draw[fill={rgb,255:red,255.0;green,182.0;blue,109.0}]  (9.250000,6.250000) rectangle (9.750000,6.750000);
\draw[fill=gray] (10,6) rectangle (11,7);
\draw[fill=gray] (0,7) rectangle (1,8);
\draw[fill={rgb,255:red,0.0;green,0.0;blue,255.0}]  (1,7) rectangle (2,8);
\draw[fill={rgb,255:red,255.0;green,228.0;blue,201.0}]  (1.250000,7.250000) rectangle (1.750000,7.750000);
\draw[fill={rgb,255:red,161.0;green,161.0;blue,255.0}]  (2,7) rectangle (3,8);
\draw[fill={rgb,255:red,161.0;green,161.0;blue,255.0}]  (3,7) rectangle (4,8);
\draw[fill={rgb,255:red,255.0;green,218.0;blue,182.0}]  (3.250000,7.250000) rectangle (3.750000,7.750000);
\draw[fill={rgb,255:red,161.0;green,161.0;blue,255.0}]  (4,7) rectangle (5,8);
\draw[fill={rgb,255:red,255.0;green,218.0;blue,182.0}]  (4.250000,7.250000) rectangle (4.750000,7.750000);
\draw[fill={rgb,255:red,161.0;green,161.0;blue,255.0}]  (5,7) rectangle (6,8);
\draw[fill={rgb,255:red,255.0;green,218.0;blue,182.0}]  (5.250000,7.250000) rectangle (5.750000,7.750000);
\draw[fill={rgb,255:red,161.0;green,161.0;blue,255.0}]  (6,7) rectangle (7,8);
\draw[fill={rgb,255:red,161.0;green,161.0;blue,255.0}]  (7,7) rectangle (8,8);
\draw[fill={rgb,255:red,161.0;green,161.0;blue,255.0}]  (8,7) rectangle (9,8);
\draw[fill={rgb,255:red,161.0;green,161.0;blue,255.0}]  (9,7) rectangle (10,8);
\draw[fill={rgb,255:red,255.0;green,218.0;blue,182.0}]  (9.250000,7.250000) rectangle (9.750000,7.750000);
\draw[fill=gray] (10,7) rectangle (11,8);
\draw[fill=gray] (0,8) rectangle (1,9);
\draw[fill={rgb,255:red,0.0;green,0.0;blue,255.0}]  (1,8) rectangle (2,9);
\draw[fill={rgb,255:red,255.0;green,241.0;blue,228.0}]  (1.250000,8.250000) rectangle (1.750000,8.750000);
\draw[fill=gray] (2,8) rectangle (3,9);
\draw[fill=gray] (3,8) rectangle (4,9);
\draw[fill=gray] (4,8) rectangle (5,9);
\draw[fill=gray] (5,8) rectangle (6,9);
\draw[fill=gray] (6,8) rectangle (7,9);
\draw[fill=gray] (7,8) rectangle (8,9);
\draw[fill=gray] (8,8) rectangle (9,9);
\draw[fill=gray] (9,8) rectangle (10,9);
\draw[fill=gray] (10,8) rectangle (11,9);
\draw[fill=gray] (0,9) rectangle (1,10);
\draw[fill={rgb,255:red,0.0;green,0.0;blue,255.0}]  (1,9) rectangle (2,10);
\draw[fill=gray] (2,9) rectangle (3,10);
\draw[fill=gray] (3,9) rectangle (4,10);
\draw[fill=gray] (4,9) rectangle (5,10);
\draw[fill=gray] (5,9) rectangle (6,10);
\draw[fill=gray] (6,9) rectangle (7,10);
\draw[fill=gray] (7,9) rectangle (8,10);
\draw[fill=gray] (8,9) rectangle (9,10);
\draw[fill=gray] (9,9) rectangle (10,10);
\draw[fill=gray] (10,9) rectangle (11,10);
\draw[fill=gray] (0,10) rectangle (1,11);
\draw[fill=gray] (1,10) rectangle (2,11);
\draw[fill=gray] (2,10) rectangle (3,11);
\draw[fill=gray] (3,10) rectangle (4,11);
\draw[fill=gray] (4,10) rectangle (5,11);
\draw[fill=gray] (5,10) rectangle (6,11);
\draw[fill=gray] (6,10) rectangle (7,11);
\draw[fill=gray] (7,10) rectangle (8,11);
\draw[fill=gray] (8,10) rectangle (9,11);
\draw[fill=gray] (9,10) rectangle (10,11);
\draw[fill=gray] (10,10) rectangle (11,11);
\node [align=center] at(-0.500000,0.500000) {0};
\node [align=center] at(-0.500000,1.500000) {1};
\node [align=center] at(-0.500000,2.500000) {2};
\node [align=center] at(-0.500000,3.500000) {3};
\node [align=center] at(-0.500000,4.500000) {4};
\node [align=center] at(-0.500000,5.500000) {5};
\node [align=center] at(-0.500000,6.500000) {6};
\node [align=center] at(-0.500000,7.500000) {7};
\node [align=center] at(-0.500000,8.500000) {8};
\node [align=center] at(-0.500000,9.500000) {9};
\node [align=center] at(-0.500000,10.500000) {10};
\node [align=center] at(0.500000,-0.500000) {A};
\node [align=center] at(1.500000,-0.500000) {B};
\node [align=center] at(2.500000,-0.500000) {C};
\node [align=center] at(3.500000,-0.500000) {D};
\node [align=center] at(4.500000,-0.500000) {E};
\node [align=center] at(5.500000,-0.500000) {F};
\node [align=center] at(6.500000,-0.500000) {G};
\node [align=center] at(7.500000,-0.500000) {H};
\node [align=center] at(8.500000,-0.500000) {I};
\node [align=center] at(9.500000,-0.500000) {J};
\node [align=center] at(10.500000,-0.500000) {K};
\end{tikzpicture}
      }
    \end{minipage}

    \centerline{\small $M_{3}$}
  \end{minipage}

  \hfill
  
  \begin{minipage}{\colwidth}
    \begin{minipage}{\mpwidth}
      \centering
      \adjustbox{width=\aboxwidth}{
        \input{\dir 3_MDP_policyStochHeatmap.tikz}
      }
    \end{minipage}
    \hfill
    \begin{minipage}{\mpwidth}
      \centering
      \adjustbox{width=\aboxwidth}{
        \begin{tikzpicture}
\draw (0,0) grid (11,11);
\draw[fill=gray] (0,0) rectangle (1,1);
\draw[fill=gray] (1,0) rectangle (2,1);
\draw[fill=gray] (2,0) rectangle (3,1);
\draw[fill=gray] (3,0) rectangle (4,1);
\draw[fill=gray] (4,0) rectangle (5,1);
\draw[fill=gray] (5,0) rectangle (6,1);
\draw[fill=gray] (6,0) rectangle (7,1);
\draw[fill=gray] (7,0) rectangle (8,1);
\draw[fill=gray] (8,0) rectangle (9,1);
\draw[fill=gray] (9,0) rectangle (10,1);
\draw[fill=gray] (10,0) rectangle (11,1);
\draw[fill=gray] (0,1) rectangle (1,2);
\draw[fill=pink] (9.500000,1.500000) circle (0.500000);
\draw[fill=gray] (10,1) rectangle (11,2);
\draw[fill=gray] (0,2) rectangle (1,3);
\draw[fill=gray] (5,2) rectangle (6,3);
\draw[fill=gray] (6,2) rectangle (7,3);
\draw[fill=gray] (7,2) rectangle (8,3);
\draw[fill=gray] (8,2) rectangle (9,3);
\draw[fill={rgb,255:red,0.0;green,0.0;blue,255.0}]  (9,2) rectangle (10,3);
\draw[fill={rgb,255:red,255.0;green,228.0;blue,201.0}]  (9.250000,2.250000) rectangle (9.750000,2.750000);
\draw[fill=gray] (10,2) rectangle (11,3);
\draw[fill=gray] (0,3) rectangle (1,4);
\draw[fill=gray] (5,3) rectangle (6,4);
\draw[fill={rgb,255:red,0.0;green,0.0;blue,255.0}]  (9,3) rectangle (10,4);
\draw[fill={rgb,255:red,255.0;green,187.0;blue,120.0}]  (9.250000,3.250000) rectangle (9.750000,3.750000);
\draw[fill=gray] (10,3) rectangle (11,4);
\draw[fill=gray] (0,4) rectangle (1,5);
\draw[fill={rgb,255:red,0.0;green,0.0;blue,255.0}]  (9,4) rectangle (10,5);
\draw[fill={rgb,255:red,255.0;green,187.0;blue,120.0}]  (9.250000,4.250000) rectangle (9.750000,4.750000);
\draw[fill=gray] (10,4) rectangle (11,5);
\draw[fill=gray] (0,5) rectangle (1,6);
\draw[fill=gray] (5,5) rectangle (6,6);
\draw[fill={rgb,255:red,0.0;green,0.0;blue,255.0}]  (9,5) rectangle (10,6);
\draw[fill={rgb,255:red,255.0;green,147.0;blue,40.0}]  (9.250000,5.250000) rectangle (9.750000,5.750000);
\draw[fill=gray] (10,5) rectangle (11,6);
\draw[fill=gray] (0,6) rectangle (1,7);
\draw[fill=gray] (2,6) rectangle (3,7);
\draw[fill=gray] (3,6) rectangle (4,7);
\draw[fill=gray] (4,6) rectangle (5,7);
\draw[fill=gray] (5,6) rectangle (6,7);
\draw[fill={rgb,255:red,0.0;green,0.0;blue,255.0}]  (9,6) rectangle (10,7);
\draw[fill={rgb,255:red,255.0;green,228.0;blue,201.0}]  (9.250000,6.250000) rectangle (9.750000,6.750000);
\draw[fill=gray] (10,6) rectangle (11,7);
\draw[fill=gray] (0,7) rectangle (1,8);
\draw[fill={rgb,255:red,0.0;green,0.0;blue,255.0}]  (1,7) rectangle (2,8);
\draw[fill={rgb,255:red,255.0;green,228.0;blue,201.0}]  (1.250000,7.250000) rectangle (1.750000,7.750000);
\draw[fill={rgb,255:red,0.0;green,0.0;blue,255.0}]  (2,7) rectangle (3,8);
\draw[fill={rgb,255:red,0.0;green,0.0;blue,255.0}]  (3,7) rectangle (4,8);
\draw[fill={rgb,255:red,255.0;green,201.0;blue,147.0}]  (3.250000,7.250000) rectangle (3.750000,7.750000);
\draw[fill={rgb,255:red,0.0;green,0.0;blue,255.0}]  (4,7) rectangle (5,8);
\draw[fill={rgb,255:red,255.0;green,187.0;blue,120.0}]  (4.250000,7.250000) rectangle (4.750000,7.750000);
\draw[fill={rgb,255:red,0.0;green,0.0;blue,255.0}]  (5,7) rectangle (6,8);
\draw[fill={rgb,255:red,255.0;green,174.0;blue,93.0}]  (5.250000,7.250000) rectangle (5.750000,7.750000);
\draw[fill={rgb,255:red,0.0;green,0.0;blue,255.0}]  (6,7) rectangle (7,8);
\draw[fill={rgb,255:red,255.0;green,228.0;blue,201.0}]  (6.250000,7.250000) rectangle (6.750000,7.750000);
\draw[fill={rgb,255:red,0.0;green,0.0;blue,255.0}]  (7,7) rectangle (8,8);
\draw[fill={rgb,255:red,255.0;green,214.0;blue,174.0}]  (7.250000,7.250000) rectangle (7.750000,7.750000);
\draw[fill={rgb,255:red,0.0;green,0.0;blue,255.0}]  (8,7) rectangle (9,8);
\draw[fill={rgb,255:red,255.0;green,241.0;blue,228.0}]  (8.250000,7.250000) rectangle (8.750000,7.750000);
\draw[fill={rgb,255:red,0.0;green,0.0;blue,255.0}]  (9,7) rectangle (10,8);
\draw[fill={rgb,255:red,255.0;green,228.0;blue,201.0}]  (9.250000,7.250000) rectangle (9.750000,7.750000);
\draw[fill=gray] (10,7) rectangle (11,8);
\draw[fill=gray] (0,8) rectangle (1,9);
\draw[fill={rgb,255:red,0.0;green,0.0;blue,255.0}]  (1,8) rectangle (2,9);
\draw[fill={rgb,255:red,255.0;green,241.0;blue,228.0}]  (1.250000,8.250000) rectangle (1.750000,8.750000);
\draw[fill=gray] (2,8) rectangle (3,9);
\draw[fill=gray] (3,8) rectangle (4,9);
\draw[fill=gray] (4,8) rectangle (5,9);
\draw[fill=gray] (5,8) rectangle (6,9);
\draw[fill=gray] (6,8) rectangle (7,9);
\draw[fill=gray] (7,8) rectangle (8,9);
\draw[fill=gray] (8,8) rectangle (9,9);
\draw[fill=gray] (9,8) rectangle (10,9);
\draw[fill=gray] (10,8) rectangle (11,9);
\draw[fill=gray] (0,9) rectangle (1,10);
\draw[fill={rgb,255:red,0.0;green,0.0;blue,255.0}]  (1,9) rectangle (2,10);
\draw[fill=gray] (2,9) rectangle (3,10);
\draw[fill=gray] (3,9) rectangle (4,10);
\draw[fill=gray] (4,9) rectangle (5,10);
\draw[fill=gray] (5,9) rectangle (6,10);
\draw[fill=gray] (6,9) rectangle (7,10);
\draw[fill=gray] (7,9) rectangle (8,10);
\draw[fill=gray] (8,9) rectangle (9,10);
\draw[fill=gray] (9,9) rectangle (10,10);
\draw[fill=gray] (10,9) rectangle (11,10);
\draw[fill=gray] (0,10) rectangle (1,11);
\draw[fill=gray] (1,10) rectangle (2,11);
\draw[fill=gray] (2,10) rectangle (3,11);
\draw[fill=gray] (3,10) rectangle (4,11);
\draw[fill=gray] (4,10) rectangle (5,11);
\draw[fill=gray] (5,10) rectangle (6,11);
\draw[fill=gray] (6,10) rectangle (7,11);
\draw[fill=gray] (7,10) rectangle (8,11);
\draw[fill=gray] (8,10) rectangle (9,11);
\draw[fill=gray] (9,10) rectangle (10,11);
\draw[fill=gray] (10,10) rectangle (11,11);
\node [align=center] at(-0.500000,0.500000) {0};
\node [align=center] at(-0.500000,1.500000) {1};
\node [align=center] at(-0.500000,2.500000) {2};
\node [align=center] at(-0.500000,3.500000) {3};
\node [align=center] at(-0.500000,4.500000) {4};
\node [align=center] at(-0.500000,5.500000) {5};
\node [align=center] at(-0.500000,6.500000) {6};
\node [align=center] at(-0.500000,7.500000) {7};
\node [align=center] at(-0.500000,8.500000) {8};
\node [align=center] at(-0.500000,9.500000) {9};
\node [align=center] at(-0.500000,10.500000) {10};
\node [align=center] at(0.500000,-0.500000) {A};
\node [align=center] at(1.500000,-0.500000) {B};
\node [align=center] at(2.500000,-0.500000) {C};
\node [align=center] at(3.500000,-0.500000) {D};
\node [align=center] at(4.500000,-0.500000) {E};
\node [align=center] at(5.500000,-0.500000) {F};
\node [align=center] at(6.500000,-0.500000) {G};
\node [align=center] at(7.500000,-0.500000) {H};
\node [align=center] at(8.500000,-0.500000) {I};
\node [align=center] at(9.500000,-0.500000) {J};
\node [align=center] at(10.500000,-0.500000) {K};
\end{tikzpicture}
      }
    \end{minipage}
    \hfill
    \begin{minipage}{\mpwidth}
      \centering
      \adjustbox{width=\aboxwidth}{
        \begin{tikzpicture}
\draw (0,0) grid (11,11);
\draw[fill=gray] (0,0) rectangle (1,1);
\draw[fill=gray] (1,0) rectangle (2,1);
\draw[fill=gray] (2,0) rectangle (3,1);
\draw[fill=gray] (3,0) rectangle (4,1);
\draw[fill=gray] (4,0) rectangle (5,1);
\draw[fill=gray] (5,0) rectangle (6,1);
\draw[fill=gray] (6,0) rectangle (7,1);
\draw[fill=gray] (7,0) rectangle (8,1);
\draw[fill=gray] (8,0) rectangle (9,1);
\draw[fill=gray] (9,0) rectangle (10,1);
\draw[fill=gray] (10,0) rectangle (11,1);
\draw[fill=gray] (0,1) rectangle (1,2);
\draw[fill=pink] (9.500000,1.500000) circle (0.500000);
\draw[fill=gray] (10,1) rectangle (11,2);
\draw[fill=gray] (0,2) rectangle (1,3);
\draw[fill=gray] (5,2) rectangle (6,3);
\draw[fill=gray] (6,2) rectangle (7,3);
\draw[fill=gray] (7,2) rectangle (8,3);
\draw[fill=gray] (8,2) rectangle (9,3);
\draw[fill={rgb,255:red,0.0;green,0.0;blue,255.0}]  (9,2) rectangle (10,3);
\draw[fill={rgb,255:red,255.0;green,214.0;blue,174.0}]  (9.250000,2.250000) rectangle (9.750000,2.750000);
\draw[fill=gray] (10,2) rectangle (11,3);
\draw[fill=gray] (0,3) rectangle (1,4);
\draw[fill=gray] (5,3) rectangle (6,4);
\draw[fill={rgb,255:red,0.0;green,0.0;blue,255.0}]  (9,3) rectangle (10,4);
\draw[fill={rgb,255:red,255.0;green,187.0;blue,120.0}]  (9.250000,3.250000) rectangle (9.750000,3.750000);
\draw[fill=gray] (10,3) rectangle (11,4);
\draw[fill=gray] (0,4) rectangle (1,5);
\draw[fill={rgb,255:red,0.0;green,0.0;blue,255.0}]  (9,4) rectangle (10,5);
\draw[fill={rgb,255:red,255.0;green,161.0;blue,67.0}]  (9.250000,4.250000) rectangle (9.750000,4.750000);
\draw[fill=gray] (10,4) rectangle (11,5);
\draw[fill=gray] (0,5) rectangle (1,6);
\draw[fill=gray] (5,5) rectangle (6,6);
\draw[fill={rgb,255:red,0.0;green,0.0;blue,255.0}]  (9,5) rectangle (10,6);
\draw[fill={rgb,255:red,255.0;green,147.0;blue,40.0}]  (9.250000,5.250000) rectangle (9.750000,5.750000);
\draw[fill=gray] (10,5) rectangle (11,6);
\draw[fill=gray] (0,6) rectangle (1,7);
\draw[fill=gray] (2,6) rectangle (3,7);
\draw[fill=gray] (3,6) rectangle (4,7);
\draw[fill=gray] (4,6) rectangle (5,7);
\draw[fill=gray] (5,6) rectangle (6,7);
\draw[fill={rgb,255:red,0.0;green,0.0;blue,255.0}]  (9,6) rectangle (10,7);
\draw[fill={rgb,255:red,255.0;green,174.0;blue,93.0}]  (9.250000,6.250000) rectangle (9.750000,6.750000);
\draw[fill=gray] (10,6) rectangle (11,7);
\draw[fill=gray] (0,7) rectangle (1,8);
\draw[fill={rgb,255:red,0.0;green,0.0;blue,255.0}]  (1,7) rectangle (2,8);
\draw[fill={rgb,255:red,255.0;green,228.0;blue,201.0}]  (1.250000,7.250000) rectangle (1.750000,7.750000);
\draw[fill={rgb,255:red,0.0;green,0.0;blue,255.0}]  (2,7) rectangle (3,8);
\draw[fill={rgb,255:red,0.0;green,0.0;blue,255.0}]  (3,7) rectangle (4,8);
\draw[fill={rgb,255:red,255.0;green,214.0;blue,174.0}]  (3.250000,7.250000) rectangle (3.750000,7.750000);
\draw[fill={rgb,255:red,0.0;green,0.0;blue,255.0}]  (4,7) rectangle (5,8);
\draw[fill={rgb,255:red,255.0;green,214.0;blue,174.0}]  (4.250000,7.250000) rectangle (4.750000,7.750000);
\draw[fill={rgb,255:red,0.0;green,0.0;blue,255.0}]  (5,7) rectangle (6,8);
\draw[fill={rgb,255:red,255.0;green,201.0;blue,147.0}]  (5.250000,7.250000) rectangle (5.750000,7.750000);
\draw[fill={rgb,255:red,0.0;green,0.0;blue,255.0}]  (6,7) rectangle (7,8);
\draw[fill={rgb,255:red,255.0;green,214.0;blue,174.0}]  (6.250000,7.250000) rectangle (6.750000,7.750000);
\draw[fill={rgb,255:red,0.0;green,0.0;blue,255.0}]  (7,7) rectangle (8,8);
\draw[fill={rgb,255:red,255.0;green,241.0;blue,228.0}]  (7.250000,7.250000) rectangle (7.750000,7.750000);
\draw[fill={rgb,255:red,0.0;green,0.0;blue,255.0}]  (8,7) rectangle (9,8);
\draw[fill={rgb,255:red,255.0;green,241.0;blue,228.0}]  (8.250000,7.250000) rectangle (8.750000,7.750000);
\draw[fill={rgb,255:red,0.0;green,0.0;blue,255.0}]  (9,7) rectangle (10,8);
\draw[fill={rgb,255:red,255.0;green,228.0;blue,201.0}]  (9.250000,7.250000) rectangle (9.750000,7.750000);
\draw[fill=gray] (10,7) rectangle (11,8);
\draw[fill=gray] (0,8) rectangle (1,9);
\draw[fill={rgb,255:red,0.0;green,0.0;blue,255.0}]  (1,8) rectangle (2,9);
\draw[fill={rgb,255:red,255.0;green,241.0;blue,228.0}]  (1.250000,8.250000) rectangle (1.750000,8.750000);
\draw[fill=gray] (2,8) rectangle (3,9);
\draw[fill=gray] (3,8) rectangle (4,9);
\draw[fill=gray] (4,8) rectangle (5,9);
\draw[fill=gray] (5,8) rectangle (6,9);
\draw[fill=gray] (6,8) rectangle (7,9);
\draw[fill=gray] (7,8) rectangle (8,9);
\draw[fill=gray] (8,8) rectangle (9,9);
\draw[fill=gray] (9,8) rectangle (10,9);
\draw[fill=gray] (10,8) rectangle (11,9);
\draw[fill=gray] (0,9) rectangle (1,10);
\draw[fill={rgb,255:red,0.0;green,0.0;blue,255.0}]  (1,9) rectangle (2,10);
\draw[fill=gray] (2,9) rectangle (3,10);
\draw[fill=gray] (3,9) rectangle (4,10);
\draw[fill=gray] (4,9) rectangle (5,10);
\draw[fill=gray] (5,9) rectangle (6,10);
\draw[fill=gray] (6,9) rectangle (7,10);
\draw[fill=gray] (7,9) rectangle (8,10);
\draw[fill=gray] (8,9) rectangle (9,10);
\draw[fill=gray] (9,9) rectangle (10,10);
\draw[fill=gray] (10,9) rectangle (11,10);
\draw[fill=gray] (0,10) rectangle (1,11);
\draw[fill=gray] (1,10) rectangle (2,11);
\draw[fill=gray] (2,10) rectangle (3,11);
\draw[fill=gray] (3,10) rectangle (4,11);
\draw[fill=gray] (4,10) rectangle (5,11);
\draw[fill=gray] (5,10) rectangle (6,11);
\draw[fill=gray] (6,10) rectangle (7,11);
\draw[fill=gray] (7,10) rectangle (8,11);
\draw[fill=gray] (8,10) rectangle (9,11);
\draw[fill=gray] (9,10) rectangle (10,11);
\draw[fill=gray] (10,10) rectangle (11,11);
\node [align=center] at(-0.500000,0.500000) {0};
\node [align=center] at(-0.500000,1.500000) {1};
\node [align=center] at(-0.500000,2.500000) {2};
\node [align=center] at(-0.500000,3.500000) {3};
\node [align=center] at(-0.500000,4.500000) {4};
\node [align=center] at(-0.500000,5.500000) {5};
\node [align=center] at(-0.500000,6.500000) {6};
\node [align=center] at(-0.500000,7.500000) {7};
\node [align=center] at(-0.500000,8.500000) {8};
\node [align=center] at(-0.500000,9.500000) {9};
\node [align=center] at(-0.500000,10.500000) {10};
\node [align=center] at(0.500000,-0.500000) {A};
\node [align=center] at(1.500000,-0.500000) {B};
\node [align=center] at(2.500000,-0.500000) {C};
\node [align=center] at(3.500000,-0.500000) {D};
\node [align=center] at(4.500000,-0.500000) {E};
\node [align=center] at(5.500000,-0.500000) {F};
\node [align=center] at(6.500000,-0.500000) {G};
\node [align=center] at(7.500000,-0.500000) {H};
\node [align=center] at(8.500000,-0.500000) {I};
\node [align=center] at(9.500000,-0.500000) {J};
\node [align=center] at(10.500000,-0.500000) {K};
\end{tikzpicture}
      }
    \end{minipage}

    \centerline{\small $M_{4}$}
  \end{minipage}

  \medskip

  \begin{minipage}{\colwidth}
    \begin{minipage}{\mpwidth}
      \centering
      \adjustbox{width=\aboxwidth}{
        \input{\dir 6_MDP_policyStochHeatmap.tikz}
      }
    \end{minipage}
    \hfill
    \begin{minipage}{\mpwidth}
      \centering
      \adjustbox{width=\aboxwidth}{
        \begin{tikzpicture}
\draw (0,0) grid (11,12);
\draw[fill=gray] (0,0) rectangle (1,1);
\draw[fill=gray] (1,0) rectangle (2,1);
\draw[fill=gray] (2,0) rectangle (3,1);
\draw[fill=gray] (3,0) rectangle (4,1);
\draw[fill=gray] (4,0) rectangle (5,1);
\draw[fill=gray] (5,0) rectangle (6,1);
\draw[fill=gray] (6,0) rectangle (7,1);
\draw[fill=gray] (7,0) rectangle (8,1);
\draw[fill=gray] (8,0) rectangle (9,1);
\draw[fill=gray] (9,0) rectangle (10,1);
\draw[fill=gray] (10,0) rectangle (11,1);
\draw[fill=gray] (0,1) rectangle (1,2);
\draw[fill={rgb,255:red,0.0;green,0.0;blue,255.0}]  (1,1) rectangle (2,2);
\draw[fill=gray] (2,1) rectangle (3,2);
\draw[fill=gray] (3,1) rectangle (4,2);
\draw[fill=gray] (4,1) rectangle (5,2);
\draw[fill=gray] (5,1) rectangle (6,2);
\draw[fill=gray] (6,1) rectangle (7,2);
\draw[fill=gray] (7,1) rectangle (8,2);
\draw[fill=gray] (8,1) rectangle (9,2);
\draw[fill=gray] (9,1) rectangle (10,2);
\draw[fill=gray] (10,1) rectangle (11,2);
\draw[fill=gray] (0,2) rectangle (1,3);
\draw[fill={rgb,255:red,0.0;green,0.0;blue,255.0}]  (1,2) rectangle (2,3);
\draw[fill={rgb,255:red,255.0;green,228.0;blue,201.0}]  (1.250000,2.250000) rectangle (1.750000,2.750000);
\draw[fill=gray] (2,2) rectangle (3,3);
\draw[fill=gray] (3,2) rectangle (4,3);
\draw[fill=gray] (4,2) rectangle (5,3);
\draw[fill=gray] (5,2) rectangle (6,3);
\draw[fill=gray] (6,2) rectangle (7,3);
\draw[fill=gray] (7,2) rectangle (8,3);
\draw[fill=gray] (8,2) rectangle (9,3);
\draw[fill=gray] (9,2) rectangle (10,3);
\draw[fill=gray] (10,2) rectangle (11,3);
\draw[fill=gray] (0,3) rectangle (1,4);
\draw[fill={rgb,255:red,0.0;green,0.0;blue,255.0}]  (1,3) rectangle (2,4);
\draw[fill={rgb,255:red,0.0;green,0.0;blue,255.0}]  (2,3) rectangle (3,4);
\draw[fill={rgb,255:red,255.0;green,241.0;blue,228.0}]  (2.250000,3.250000) rectangle (2.750000,3.750000);
\draw[fill={rgb,255:red,0.0;green,0.0;blue,255.0}]  (3,3) rectangle (4,4);
\draw[fill={rgb,255:red,255.0;green,187.0;blue,120.0}]  (3.250000,3.250000) rectangle (3.750000,3.750000);
\draw[fill={rgb,255:red,0.0;green,0.0;blue,255.0}]  (4,3) rectangle (5,4);
\draw[fill={rgb,255:red,255.0;green,201.0;blue,147.0}]  (4.250000,3.250000) rectangle (4.750000,3.750000);
\draw[fill={rgb,255:red,0.0;green,0.0;blue,255.0}]  (5,3) rectangle (6,4);
\draw[fill={rgb,255:red,255.0;green,187.0;blue,120.0}]  (5.250000,3.250000) rectangle (5.750000,3.750000);
\draw[fill={rgb,255:red,0.0;green,0.0;blue,255.0}]  (6,3) rectangle (7,4);
\draw[fill={rgb,255:red,255.0;green,228.0;blue,201.0}]  (6.250000,3.250000) rectangle (6.750000,3.750000);
\draw[fill={rgb,255:red,0.0;green,0.0;blue,255.0}]  (7,3) rectangle (8,4);
\draw[fill={rgb,255:red,0.0;green,0.0;blue,255.0}]  (8,3) rectangle (9,4);
\draw[fill={rgb,255:red,255.0;green,241.0;blue,228.0}]  (8.250000,3.250000) rectangle (8.750000,3.750000);
\draw[fill={rgb,255:red,0.0;green,0.0;blue,255.0}]  (9,3) rectangle (10,4);
\draw[fill={rgb,255:red,255.0;green,241.0;blue,228.0}]  (9.250000,3.250000) rectangle (9.750000,3.750000);
\draw[fill=gray] (10,3) rectangle (11,4);
\draw[fill=gray] (0,4) rectangle (1,5);
\draw[fill=gray] (2,4) rectangle (3,5);
\draw[fill=gray] (3,4) rectangle (4,5);
\draw[fill=gray] (4,4) rectangle (5,5);
\draw[fill=gray] (5,4) rectangle (6,5);
\draw[fill={rgb,255:red,0.0;green,0.0;blue,255.0}]  (9,4) rectangle (10,5);
\draw[fill={rgb,255:red,255.0;green,201.0;blue,147.0}]  (9.250000,4.250000) rectangle (9.750000,4.750000);
\draw[fill=gray] (10,4) rectangle (11,5);
\draw[fill=gray] (0,5) rectangle (1,6);
\draw[fill=gray] (5,5) rectangle (6,6);
\draw[fill={rgb,255:red,0.0;green,0.0;blue,255.0}]  (9,5) rectangle (10,6);
\draw[fill={rgb,255:red,255.0;green,201.0;blue,147.0}]  (9.250000,5.250000) rectangle (9.750000,5.750000);
\draw[fill=gray] (10,5) rectangle (11,6);
\draw[fill=gray] (0,6) rectangle (1,7);
\draw[fill=gray] (5,6) rectangle (6,7);
\draw[fill={rgb,255:red,0.0;green,0.0;blue,255.0}]  (9,6) rectangle (10,7);
\draw[fill={rgb,255:red,255.0;green,174.0;blue,93.0}]  (9.250000,6.250000) rectangle (9.750000,6.750000);
\draw[fill=gray] (10,6) rectangle (11,7);
\draw[fill=gray] (0,7) rectangle (1,8);
\draw[fill=gray] (5,7) rectangle (6,8);
\draw[fill={rgb,255:red,0.0;green,0.0;blue,255.0}]  (9,7) rectangle (10,8);
\draw[fill={rgb,255:red,255.0;green,187.0;blue,120.0}]  (9.250000,7.250000) rectangle (9.750000,7.750000);
\draw[fill=gray] (10,7) rectangle (11,8);
\draw[fill=gray] (0,8) rectangle (1,9);
\draw[fill=gray] (5,8) rectangle (6,9);
\draw[fill={rgb,255:red,0.0;green,0.0;blue,255.0}]  (9,8) rectangle (10,9);
\draw[fill={rgb,255:red,255.0;green,201.0;blue,147.0}]  (9.250000,8.250000) rectangle (9.750000,8.750000);
\draw[fill=gray] (10,8) rectangle (11,9);
\draw[fill=gray] (0,9) rectangle (1,10);
\draw[fill=gray] (5,9) rectangle (6,10);
\draw[fill={rgb,255:red,0.0;green,0.0;blue,255.0}]  (9,9) rectangle (10,10);
\draw[fill={rgb,255:red,255.0;green,214.0;blue,174.0}]  (9.250000,9.250000) rectangle (9.750000,9.750000);
\draw[fill=gray] (10,9) rectangle (11,10);
\draw[fill=gray] (0,10) rectangle (1,11);
\draw[fill=pink] (9.500000,10.500000) circle (0.500000);
\draw[fill=gray] (10,10) rectangle (11,11);
\draw[fill=gray] (0,11) rectangle (1,12);
\draw[fill=gray] (1,11) rectangle (2,12);
\draw[fill=gray] (2,11) rectangle (3,12);
\draw[fill=gray] (3,11) rectangle (4,12);
\draw[fill=gray] (4,11) rectangle (5,12);
\draw[fill=gray] (5,11) rectangle (6,12);
\draw[fill=gray] (6,11) rectangle (7,12);
\draw[fill=gray] (7,11) rectangle (8,12);
\draw[fill=gray] (8,11) rectangle (9,12);
\draw[fill=gray] (9,11) rectangle (10,12);
\draw[fill=gray] (10,11) rectangle (11,12);
\node [align=center] at(-0.500000,0.500000) {0};
\node [align=center] at(-0.500000,1.500000) {1};
\node [align=center] at(-0.500000,2.500000) {2};
\node [align=center] at(-0.500000,3.500000) {3};
\node [align=center] at(-0.500000,4.500000) {4};
\node [align=center] at(-0.500000,5.500000) {5};
\node [align=center] at(-0.500000,6.500000) {6};
\node [align=center] at(-0.500000,7.500000) {7};
\node [align=center] at(-0.500000,8.500000) {8};
\node [align=center] at(-0.500000,9.500000) {9};
\node [align=center] at(-0.500000,10.500000) {10};
\node [align=center] at(-0.500000,11.500000) {11};
\node [align=center] at(0.500000,-0.500000) {A};
\node [align=center] at(1.500000,-0.500000) {B};
\node [align=center] at(2.500000,-0.500000) {C};
\node [align=center] at(3.500000,-0.500000) {D};
\node [align=center] at(4.500000,-0.500000) {E};
\node [align=center] at(5.500000,-0.500000) {F};
\node [align=center] at(6.500000,-0.500000) {G};
\node [align=center] at(7.500000,-0.500000) {H};
\node [align=center] at(8.500000,-0.500000) {I};
\node [align=center] at(9.500000,-0.500000) {J};
\node [align=center] at(10.500000,-0.500000) {K};
\end{tikzpicture}
      }
    \end{minipage}
    \hfill
    \begin{minipage}{\mpwidth}
      \centering
      \adjustbox{width=\aboxwidth}{
        \begin{tikzpicture}
\draw (0,0) grid (11,12);
\draw[fill=gray] (0,0) rectangle (1,1);
\draw[fill=gray] (1,0) rectangle (2,1);
\draw[fill=gray] (2,0) rectangle (3,1);
\draw[fill=gray] (3,0) rectangle (4,1);
\draw[fill=gray] (4,0) rectangle (5,1);
\draw[fill=gray] (5,0) rectangle (6,1);
\draw[fill=gray] (6,0) rectangle (7,1);
\draw[fill=gray] (7,0) rectangle (8,1);
\draw[fill=gray] (8,0) rectangle (9,1);
\draw[fill=gray] (9,0) rectangle (10,1);
\draw[fill=gray] (10,0) rectangle (11,1);
\draw[fill=gray] (0,1) rectangle (1,2);
\draw[fill={rgb,255:red,0.0;green,0.0;blue,255.0}]  (1,1) rectangle (2,2);
\draw[fill=gray] (2,1) rectangle (3,2);
\draw[fill=gray] (3,1) rectangle (4,2);
\draw[fill=gray] (4,1) rectangle (5,2);
\draw[fill=gray] (5,1) rectangle (6,2);
\draw[fill=gray] (6,1) rectangle (7,2);
\draw[fill=gray] (7,1) rectangle (8,2);
\draw[fill=gray] (8,1) rectangle (9,2);
\draw[fill=gray] (9,1) rectangle (10,2);
\draw[fill=gray] (10,1) rectangle (11,2);
\draw[fill=gray] (0,2) rectangle (1,3);
\draw[fill={rgb,255:red,0.0;green,0.0;blue,255.0}]  (1,2) rectangle (2,3);
\draw[fill={rgb,255:red,255.0;green,228.0;blue,201.0}]  (1.250000,2.250000) rectangle (1.750000,2.750000);
\draw[fill=gray] (2,2) rectangle (3,3);
\draw[fill=gray] (3,2) rectangle (4,3);
\draw[fill=gray] (4,2) rectangle (5,3);
\draw[fill=gray] (5,2) rectangle (6,3);
\draw[fill=gray] (6,2) rectangle (7,3);
\draw[fill=gray] (7,2) rectangle (8,3);
\draw[fill=gray] (8,2) rectangle (9,3);
\draw[fill=gray] (9,2) rectangle (10,3);
\draw[fill=gray] (10,2) rectangle (11,3);
\draw[fill=gray] (0,3) rectangle (1,4);
\draw[fill={rgb,255:red,0.0;green,0.0;blue,255.0}]  (1,3) rectangle (2,4);
\draw[fill={rgb,255:red,134.0;green,134.0;blue,255.0}]  (2,3) rectangle (3,4);
\draw[fill={rgb,255:red,134.0;green,134.0;blue,255.0}]  (3,3) rectangle (4,4);
\draw[fill={rgb,255:red,255.0;green,226.0;blue,198.0}]  (3.250000,3.250000) rectangle (3.750000,3.750000);
\draw[fill={rgb,255:red,134.0;green,134.0;blue,255.0}]  (4,3) rectangle (5,4);
\draw[fill={rgb,255:red,255.0;green,198.0;blue,141.0}]  (4.250000,3.250000) rectangle (4.750000,3.750000);
\draw[fill={rgb,255:red,134.0;green,134.0;blue,255.0}]  (5,3) rectangle (6,4);
\draw[fill={rgb,255:red,255.0;green,170.0;blue,85.0}]  (5.250000,3.250000) rectangle (5.750000,3.750000);
\draw[fill={rgb,255:red,134.0;green,134.0;blue,255.0}]  (6,3) rectangle (7,4);
\draw[fill={rgb,255:red,134.0;green,134.0;blue,255.0}]  (7,3) rectangle (8,4);
\draw[fill={rgb,255:red,134.0;green,134.0;blue,255.0}]  (8,3) rectangle (9,4);
\draw[fill={rgb,255:red,134.0;green,134.0;blue,255.0}]  (9,3) rectangle (10,4);
\draw[fill=gray] (10,3) rectangle (11,4);
\draw[fill=gray] (0,4) rectangle (1,5);
\draw[fill={rgb,255:red,120.0;green,120.0;blue,255.0}]  (1,4) rectangle (2,5);
\draw[fill={rgb,255:red,255.0;green,229.0;blue,204.0}]  (1.250000,4.250000) rectangle (1.750000,4.750000);
\draw[fill=gray] (2,4) rectangle (3,5);
\draw[fill=gray] (3,4) rectangle (4,5);
\draw[fill=gray] (4,4) rectangle (5,5);
\draw[fill=gray] (5,4) rectangle (6,5);
\draw[fill={rgb,255:red,134.0;green,134.0;blue,255.0}]  (9,4) rectangle (10,5);
\draw[fill={rgb,255:red,255.0;green,170.0;blue,85.0}]  (9.250000,4.250000) rectangle (9.750000,4.750000);
\draw[fill=gray] (10,4) rectangle (11,5);
\draw[fill=gray] (0,5) rectangle (1,6);
\draw[fill={rgb,255:red,120.0;green,120.0;blue,255.0}]  (1,5) rectangle (2,6);
\draw[fill={rgb,255:red,255.0;green,229.0;blue,204.0}]  (1.250000,5.250000) rectangle (1.750000,5.750000);
\draw[fill={rgb,255:red,120.0;green,120.0;blue,255.0}]  (2,5) rectangle (3,6);
\draw[fill={rgb,255:red,255.0;green,178.0;blue,102.0}]  (2.250000,5.250000) rectangle (2.750000,5.750000);
\draw[fill={rgb,255:red,120.0;green,120.0;blue,255.0}]  (3,5) rectangle (4,6);
\draw[fill={rgb,255:red,120.0;green,120.0;blue,255.0}]  (4,5) rectangle (5,6);
\draw[fill={rgb,255:red,255.0;green,204.0;blue,153.0}]  (4.250000,5.250000) rectangle (4.750000,5.750000);
\draw[fill=gray] (5,5) rectangle (6,6);
\draw[fill={rgb,255:red,134.0;green,134.0;blue,255.0}]  (9,5) rectangle (10,6);
\draw[fill={rgb,255:red,255.0;green,170.0;blue,85.0}]  (9.250000,5.250000) rectangle (9.750000,5.750000);
\draw[fill=gray] (10,5) rectangle (11,6);
\draw[fill=gray] (0,6) rectangle (1,7);
\draw[fill={rgb,255:red,120.0;green,120.0;blue,255.0}]  (4,6) rectangle (5,7);
\draw[fill={rgb,255:red,255.0;green,178.0;blue,102.0}]  (4.250000,6.250000) rectangle (4.750000,6.750000);
\draw[fill=gray] (5,6) rectangle (6,7);
\draw[fill={rgb,255:red,134.0;green,134.0;blue,255.0}]  (9,6) rectangle (10,7);
\draw[fill={rgb,255:red,255.0;green,226.0;blue,198.0}]  (9.250000,6.250000) rectangle (9.750000,6.750000);
\draw[fill=gray] (10,6) rectangle (11,7);
\draw[fill=gray] (0,7) rectangle (1,8);
\draw[fill={rgb,255:red,120.0;green,120.0;blue,255.0}]  (4,7) rectangle (5,8);
\draw[fill=gray] (5,7) rectangle (6,8);
\draw[fill={rgb,255:red,134.0;green,134.0;blue,255.0}]  (9,7) rectangle (10,8);
\draw[fill={rgb,255:red,255.0;green,226.0;blue,198.0}]  (9.250000,7.250000) rectangle (9.750000,7.750000);
\draw[fill=gray] (10,7) rectangle (11,8);
\draw[fill=gray] (0,8) rectangle (1,9);
\draw[fill={rgb,255:red,120.0;green,120.0;blue,255.0}]  (4,8) rectangle (5,9);
\draw[fill=gray] (5,8) rectangle (6,9);
\draw[fill={rgb,255:red,134.0;green,134.0;blue,255.0}]  (9,8) rectangle (10,9);
\draw[fill={rgb,255:red,255.0;green,226.0;blue,198.0}]  (9.250000,8.250000) rectangle (9.750000,8.750000);
\draw[fill=gray] (10,8) rectangle (11,9);
\draw[fill=gray] (0,9) rectangle (1,10);
\draw[fill={rgb,255:red,120.0;green,120.0;blue,255.0}]  (4,9) rectangle (5,10);
\draw[fill={rgb,255:red,255.0;green,229.0;blue,204.0}]  (4.250000,9.250000) rectangle (4.750000,9.750000);
\draw[fill=gray] (5,9) rectangle (6,10);
\draw[fill={rgb,255:red,134.0;green,134.0;blue,255.0}]  (9,9) rectangle (10,10);
\draw[fill=gray] (10,9) rectangle (11,10);
\draw[fill=gray] (0,10) rectangle (1,11);
\draw[fill={rgb,255:red,120.0;green,120.0;blue,255.0}]  (4,10) rectangle (5,11);
\draw[fill={rgb,255:red,120.0;green,120.0;blue,255.0}]  (5,10) rectangle (6,11);
\draw[fill={rgb,255:red,255.0;green,229.0;blue,204.0}]  (5.250000,10.250000) rectangle (5.750000,10.750000);
\draw[fill={rgb,255:red,120.0;green,120.0;blue,255.0}]  (6,10) rectangle (7,11);
\draw[fill={rgb,255:red,255.0;green,153.0;blue,50.0}]  (6.250000,10.250000) rectangle (6.750000,10.750000);
\draw[fill={rgb,255:red,120.0;green,120.0;blue,255.0}]  (7,10) rectangle (8,11);
\draw[fill={rgb,255:red,255.0;green,204.0;blue,153.0}]  (7.250000,10.250000) rectangle (7.750000,10.750000);
\draw[fill={rgb,255:red,120.0;green,120.0;blue,255.0}]  (8,10) rectangle (9,11);
\draw[fill=pink] (9.500000,10.500000) circle (0.500000);
\draw[fill=gray] (10,10) rectangle (11,11);
\draw[fill=gray] (0,11) rectangle (1,12);
\draw[fill=gray] (1,11) rectangle (2,12);
\draw[fill=gray] (2,11) rectangle (3,12);
\draw[fill=gray] (3,11) rectangle (4,12);
\draw[fill=gray] (4,11) rectangle (5,12);
\draw[fill=gray] (5,11) rectangle (6,12);
\draw[fill=gray] (6,11) rectangle (7,12);
\draw[fill=gray] (7,11) rectangle (8,12);
\draw[fill=gray] (8,11) rectangle (9,12);
\draw[fill=gray] (9,11) rectangle (10,12);
\draw[fill=gray] (10,11) rectangle (11,12);
\node [align=center] at(-0.500000,0.500000) {0};
\node [align=center] at(-0.500000,1.500000) {1};
\node [align=center] at(-0.500000,2.500000) {2};
\node [align=center] at(-0.500000,3.500000) {3};
\node [align=center] at(-0.500000,4.500000) {4};
\node [align=center] at(-0.500000,5.500000) {5};
\node [align=center] at(-0.500000,6.500000) {6};
\node [align=center] at(-0.500000,7.500000) {7};
\node [align=center] at(-0.500000,8.500000) {8};
\node [align=center] at(-0.500000,9.500000) {9};
\node [align=center] at(-0.500000,10.500000) {10};
\node [align=center] at(-0.500000,11.500000) {11};
\node [align=center] at(0.500000,-0.500000) {A};
\node [align=center] at(1.500000,-0.500000) {B};
\node [align=center] at(2.500000,-0.500000) {C};
\node [align=center] at(3.500000,-0.500000) {D};
\node [align=center] at(4.500000,-0.500000) {E};
\node [align=center] at(5.500000,-0.500000) {F};
\node [align=center] at(6.500000,-0.500000) {G};
\node [align=center] at(7.500000,-0.500000) {H};
\node [align=center] at(8.500000,-0.500000) {I};
\node [align=center] at(9.500000,-0.500000) {J};
\node [align=center] at(10.500000,-0.500000) {K};
\end{tikzpicture}
      }
    \end{minipage}

    \centerline{\small $M_{5}$}
  \end{minipage}

  \hfill
  
  \begin{minipage}{\colwidth}
    \begin{minipage}{\mpwidth}
      \centering
      \adjustbox{width=\aboxwidth}{
        \input{\dir 7_MDP_policyStochHeatmap.tikz}
      }
    \end{minipage}
    \hfill
    \begin{minipage}{\mpwidth}
      \centering
      \adjustbox{width=\aboxwidth}{
        \input{\dir 7_MDP_policyBiasedHeatMap.tikz}
      }
    \end{minipage}
    \hfill
    \begin{minipage}{\mpwidth}
      \centering
      \adjustbox{width=\aboxwidth}{
        \input{\dir 7_OAMDP_policyStochHeatMap.tikz}
      }
    \end{minipage}

    \centerline{\small $M_{7}$}
  \end{minipage}

  \medskip

  \def\colwidth{.8\textwidth}

  \begin{minipage}{\colwidth}
    \begin{minipage}{\mpwidth}
      \centering
      \adjustbox{width=\aboxwidth}{
        \input{\dir 4_MDP_policyStochHeatmap.tikz}
      }
    \end{minipage}
    \hfill
    \begin{minipage}{\mpwidth}
      \centering
      \adjustbox{width=\aboxwidth}{
        \input{\dir 4_MDP_policyBiasedHeatMap.tikz}
      }
    \end{minipage}
    \hfill
    \begin{minipage}{\mpwidth}
      \centering
      \adjustbox{width=\aboxwidth}{
        \input{\dir 4_OAMDP_policyStochHeatMap.tikz}
      }
    \end{minipage}

    \centerline{\small $M_{6}$}
  \end{minipage}

\caption{[``Reflex'' Responses] Heatmaps showing, for each cell,
      (1) [in background] the probability of visit during a trajectory from dark blue ($P=1$) to white ($P=0$), and
      (2) [in the middle] the rate of $<150$\,ms response times from orange (max. rate: 50\% or more) to white (min. rate: 0\%).
      These heatmaps are provided for each maze and each policy, from left to right: stochastic MDP policy, biased MDP policy, and pOAMDP policy.
      \label{fig|heatmap|timeMaptime150}}

\end{figure}

\subsubsection{Questionnaire and ranking}

\begin{figure}
  \centering
  \includegraphics[width=.7\textwidth]{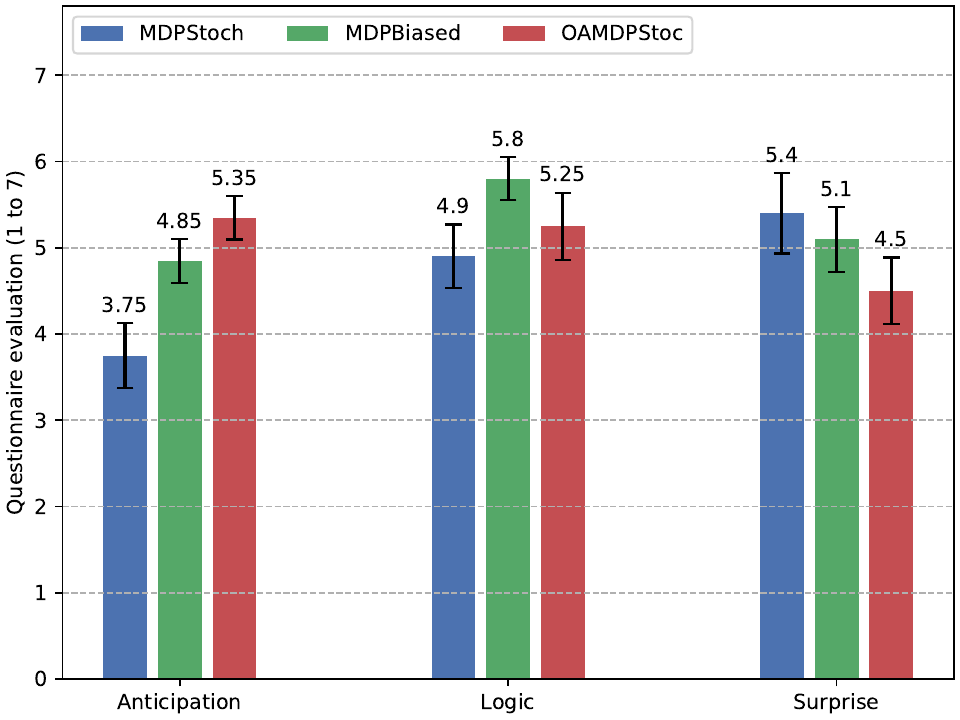}
		\caption{Graph showing questionnaire results (mean score and standard error) on a 7-point Lickert scale for each of the three policies and for the three dimensions assessed (Anticipation, Logic and Surprise).
			\label{fig|graphQuestionnaire}}
\end{figure}

After each sequence of mazes corresponding to a policy, participants were asked to rate this policy on a scale of 1 to 7 according to three dimensions: Anticipation, Logic and Surprise (see \Cref{sec|inVivoProtocol}).
The average scores obtained, as well as standard errors, are shown in \Cref{fig|graphQuestionnaire}.
Overlapping error bars (based on standard errors) indicate no differences, while non-overlapping error bars indicate significant differences.
Concerning Anticipation, the results seem to indicate that $\pi_\mdps$ is considered more difficult to anticipate than the other two policies, and that $\pi_\mdpb$ is tendentially a little more difficult to anticipate than $\piApred$.
Concerning Logic, the results seem to indicate no significant difference between the three policies, with $\pi_\mdpb$ tending to be judged slightly more logical than the other two.
Concerning Surprise, the results seem to indicate no significant difference between the three policies.

The results of the ranking of the three policies given at the end by the participants in terms of Anticipation and Logic are given by \Cref{tab|prefBis} (complete orderings).
The two rankings show very similar patterns.
For both Anticipation and Logic, as presented in \Cref{tab|prefBis} (score rankings),
\begin{enumerate*}
	\item $\pi_\mdps$, which most participants consider hard to predict and even random, is typically ranked last, sometimes second, and
	\item participants have a preference for $\piApred$ over $\pi_\mdpb$.
\end{enumerate*}

\begin{table}[ht!]
	\caption{Human preferences over policies,
		where A=$\pi_\mdps$, B=$\pi_\mdpb$, and C=$\pi_\pred$
		\label{tab|prefBis}
	}
	\centering
	\begin{subtable}[t]{0.48\columnwidth}
		\caption{Anticipation: Complete orderings}
		\label{tab|pref|order}
		\centering
		\begin{tabular}{lr}
			\toprule
			order & \#votes \\
			\midrule
			CBA & 8 \\
			BCA & 5 \\
			CAB & 3 \\
			BAC & 2 \\
			ABC & 1 \\
\bottomrule
		\end{tabular}
	\end{subtable}
	\hfill
	\begin{subtable}[t]{0.48\columnwidth}
		\caption{Anticipation: Score ranking}
		\label{tab|pref|rankBis2}
		\centering
		{\newcommand{\frax}[1]{$#1$}
			\begin{tabular}{lrrr}
				\toprule
				& A & B & C \\
				\midrule
				1st & \frax{1} & \frax{7} & \frax{11} \\
				2nd & \frax{5} & \frax{9} & \frax{5} \\
				3rd & \frax{14} & \frax{3} & \frax{3} \\
				\bottomrule
			\end{tabular}
}
	\end{subtable}

	\begin{subtable}[t]{0.48\columnwidth}
		\caption{Logic: Complete orderings}
		\label{tab|pref|order}
		\centering
		\begin{tabular}{lr}
			\toprule
			order & \#votes \\
			\midrule
			CBA & 8 \\
			BAC & 4 \\
			CAB & 3 \\
			BCA & 3 \\
			ABC & 2 \\
\bottomrule
		\end{tabular}
	\end{subtable}
	\hfill
	\begin{subtable}[t]{0.48\columnwidth}
		\caption{Logic: Score ranking}
		\label{tab|pref|rankBis}
		\centering
		{ \newcommand{\frax}[1]{$#1$}
			\begin{tabular}{lrrr}
				\toprule
				& A & B & C \\
				\midrule
				1st & \frax{2} & \frax{7} & \frax{11} \\
				2nd & \frax{7} & \frax{10} & \frax{3} \\
				3rd & \frax{11} & \frax{3} & \frax{6} \\
				\bottomrule
			\end{tabular}
		}
\end{subtable}
\end{table}

Participants often declare that the initial choice of $\piApred$ can be surprising.
This is especially the case in maze $M_6$, and  if the participants had worked with $\piApred$ after $\pi_\mdps$ and $\pi_\mdpb$.
However, despite those statements, humans still performed better with $\piApred$ (especially in maze $M_6$).

\subsection{Discussion}

There are few differences between the biased MDP policy and the pOAMDP policy  in terms of errors, response times, and human perception, in comparison with the stochastic MDP policy, whose behavior is much harder to predict.
The pOAMDP policies' response times are better on some mazes ($M_1$, $M_4$ and $M_6$).
The pOAMDP policy induces less errors on more complex mazes ($M_6$ and $M_7$),
while the biased MDP policy induces less errors on mazes $M_3$ and $M_5$, where the pOAMDP policy is not deterministic (having two possible trajectories).
Subjective feedback from human participants is consistent with observed performance.
The pOAMDP policy was preferred, probably because it seems easier to anticipate, while the biased MDP policy is considered as slightly more logical (as it always follows shortest paths).

The pOAMDP policy is more efficient in complexe mazes such as $M_6$ which is interesting if we want to consider more realistic scenarios. However in less complex mazes, we do not observer significative differences between the pOAMDP policy and the biased policy. Considering more mazes such as  $M_6$ in the experiment could improve the results and better emphasize the differences between the biased policy and the pOAMDP policy.
The biased condition was added to this study to be able to compare the pOAMDP policy not only to a stochastic policy but also to a policy that could be easier to anticipate for the participants. We were expecting the differences between the pOAMDP policy and the biased policy to be less important than the differences between the stochastic MDP policy and the pOAMDP policy. However the bias used for the maze was learned really fast by the participants and  facilitated their predictions to the point where in certain mazes, there was not any differences between the two.A better model of a human observer would thus account for the possible biases of the agent, \eg, the bias being a hidden state variable (a {\em type} in the OAMDP sense) that the observer could try to infer.
This being said, a scenario that could better match the human participants' model of the agent could be if the agent's state were described not only by its location $(x,y)$, but also by a direction $d\in \{North, South, East, West\}$, and action set $\cA=\{forward, turnRight, turnLeft\}$, so that, without modifying the reward function, minimizing trajectory lengths will lead to prefer straight lines whenever possible.

\section{Conclusion}
\label{sec|conclusion}

We have introduced a new formalism, predictable observer-aware MDPs (pOAMDPs), that allows deriving policies whose next actions or next states are more predictable, and proposed accounting not only for discounted problems, but also for stochastic shortest-path problems (which requires ensuring that valid solution policies can be found).
With the objective of minimizing the number of prediction errors along a trajectory in an undiscounted setting, and assuming rational observer predictions, we derived two reward functions, respectively for action and state predictability and demonstrated that they both induce valid stochastic shortest-path problems, \ie, the solution predictable policies reach terminal states with probability 1.
A notable property is that the solving complexity of pOAMDPs is comparable to MDPs, thus much less than OAMDPs.
In some cases, the resulting policies select counter-intuitive actions early on to increase predictability later on.
The interpretation of generated policies shows significant reductions in the expected number of errors when using pOAMDP solutions on some scenarios (up to fourfold), and also benefits in using biased MDP policies, which prefer following walls.
Results of the experiment with human participants are consistent with these observations.

As illustrated by some benchmark problems, the proposed performance criterion can lead to poor policies in terms of the original performance criterion (here used only for the observer predictions).
This can be addressed in various ways as, for instance, by linearly combining both reward functions, or, using {\em constrained MDPs} \cite{Altman99,TreEtAl-ijcai17}, by minimizing the prediction error while constraining the value of the original criterion.

On another note, considering goal-oriented problems as we did would of course also be relevant for Miura and Zilberstein's OAMDPs, first to determine which of their scenarios result in valid SSPs.
Then, to handle SSPs with traps, \ie, subsets of (non-terminal) states that cannot be escaped, an interesting direction would be to extend our work to generalized SSPs \cite{KolMauWelGef-icaps11,TreEtAl-uai17}.

Finally, we had to depart from Miura and Zilberstein's original formalism and their static types \cite{pmlr-v161-miura21a}, but an important perspective is to generalize both formalisms, making for a more unified theory of observer-aware sequential decision-making.
We believe that a key point to achieve this is to restrict the observer's observability of states and actions so that the target variable, whether static or dynamic, can be a state variable, even for action predictability.
What is more, this partial observability would also allow covering more real-world scenarios.
In this setting, we envision looking at the continuity properties of the optimal value function to possibly propose bounding approximators and derive point-based solvers (as was done for POMDPs and related models \cite{SmiSim-uai05,SpaVla05,KurHsuLee-rss08,PinGorThr06,ShaPinKap-aamas13,DibAmaBufCha-jair16,HorBosPec-aaai17,HorBos-aaai19}).

\bibliographystyle{ios1}

\end{document}